\documentclass[10pt,twocolumn,letterpaper]{article}

\usepackage{iccv, times, epsfig, xspace, enumitem}
\usepackage{graphicx, amsmath, amssymb, caption, subcaption, multirow, overpic, textpos, pifont, adjustbox}
\usepackage[table]{xcolor}
\usepackage[british, english, american]{babel}
\usepackage[utf8x]{inputenc}
\makeatletter
\@namedef{ver@everyshi.sty}{}
\makeatother
\usepackage{pgf, tikz, pgfplots}
\usepackage{makecell}
\usepackage[pagebackref=true, breaklinks=true, letterpaper=true, colorlinks,
 citecolor=citecolor, linkcolor=linkcolor, bookmarks=false]{hyperref}

\definecolor{citecolor}{HTML}{0071BC}
\definecolor{linkcolor}{HTML}{ED1C24}
\definecolor{acceptcolor}{HTML}{74C219}
\definecolor{rejectcolor}{HTML}{DE1616}
\definecolor{qcolor}{HTML}{536872}
\definecolor{demphcolor}{RGB}{100,100,100}

\newlength\savewidth
\newcommand{\tablestyle}[2]{\setlength{\tabcolsep}{#1}\renewcommand{\arraystretch}{#2}\centering\footnotesize}
\newenvironment{enumeratecard}{\begin{enumerate}[leftmargin=4mm, itemsep=0pt, topsep=3pt]}{\end{enumerate}}
\renewcommand{\paragraph}[1]{\vspace{1.25mm}\noindent\textbf{#1}}
\newcolumntype{x}[1]{>{\centering\arraybackslash}p{#1pt}}
\newcolumntype{L}[1]{>{\raggedright\let\newline\\\arraybackslash\hspace{0pt}}m{#1}}
\newcommand{\app}{\raise.17ex\hbox{$\scriptstyle\sim$}}
\newcommand{\x}{{\times}}

\newcommand{\sad}{\mbox{SA-1B}\xspace}
\newcommand{\sam}{SAM\xspace}
\newcommand{\fig}[1]{Fig.~\ref{#1}}
\newcommand{\supp}{appendix\xspace}
\newcommand{\cmark}{\textcolor{acceptcolor}{\ding{51}}}
\newcommand{\xmark}{\textcolor{rejectcolor}{\ding{55}}}

\newcommand{\qtxt}[1]{{\color{qcolor}{\textit{#1}}}}
\newcommand{\demph}[1]{\textcolor{demphcolor}{#1}}
\newcommand{\mypm}[1]{{\scriptsize{{\demph{{\kern.4ex$\pm$\kern.1ex#1}}}}}}
\newcommand{\lr}{\emph{lr}\xspace}
\newcommand{\ld}{\emph{ld}\xspace}
\newcommand{\wtd}{\emph{wd}\xspace}
\newcommand{\drp}{\emph{dp}\xspace}
\newcommand{\expnum}[2]{{#1}\mathrm{e}^{#2}}
\setlist[enumerate]{itemsep=-0.5mm,partopsep=0pt}

\iccvfinalcopy
\begin{document}
\title{\vspace{-3mm}\LARGE Segment Anything\vspace{-6mm}}
\author{
{\normalsize Alexander Kirillov$^{1,2,4}$ \quad Eric Mintun$^{2}$ \quad Nikhila Ravi$^{1,2}$
 \quad Hanzi Mao$^{2}$ \quad Chloe Rolland$^{3}$ \quad Laura Gustafson$^{3}$}\\[0mm]
{\normalsize Tete Xiao$^{3}$ \hspace{4.65mm} Spencer Whitehead \hspace{4.65mm} Alexander C. Berg \hspace{4.65mm}
Wan-Yen Lo \hspace{4.65mm} Piotr Doll\'{a}r$^{4}$ \hspace{4.65mm} Ross Girshick$^{4}$}\\[0mm]
 {\small$^1$project lead \qquad $^2$joint first author \qquad $^3$equal contribution \qquad $^4$directional lead}\\[1mm]
{Meta AI Research, FAIR}\vspace{-4mm}}

%##################################################################################################
\twocolumn[{
\maketitle\centering
\captionsetup{type=figure}
\includegraphics[width=0.99\textwidth]{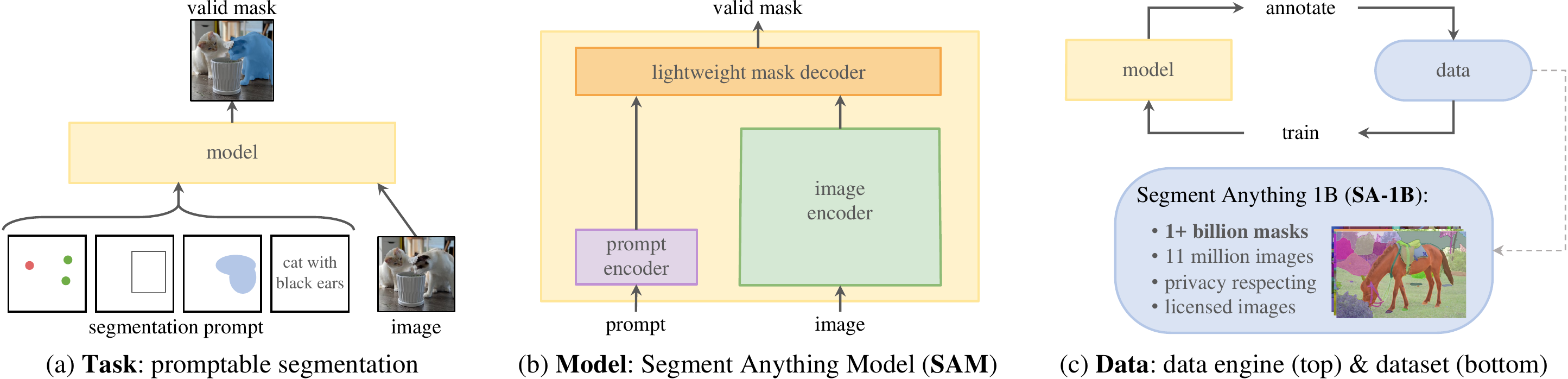}\vspace{-2mm}
\captionof{figure}{We aim to build a foundation model for segmentation by introducing three interconnected components: a promptable segmentation \emph{task}, a segmentation \emph{model} (\sam) that powers data annotation and enables zero-shot transfer to a range of tasks via prompt engineering, and a \emph{data} engine for collecting \sad, our dataset of over 1 billion masks.}
\label{fig:teaser}\vspace{5mm}
}]
%##################################################################################################

\maketitle

%%%%%%%%%%%%%%%%%%%%%%%%%%%%%%%%%%%%%%%%%%%%%%%%%%%%%%%%%%%%%%%%%%%%%%%%%%%%%%%%%%%%%%%%%%%%%%%%%%%
\begin{abstract}\vspace{-3mm}
We introduce the Segment Anything (SA) project: a new task, model, and dataset for image segmentation. Using our efficient model in a data collection loop, we built the largest segmentation dataset to date (by far), with over 1 \textbf{billion} masks on 11M licensed and privacy respecting images. The model is designed and trained to be promptable, so it can transfer zero-shot to new image distributions and tasks. We evaluate its capabilities on numerous tasks and find that its zero-shot performance is impressive -- often competitive with or even superior to prior fully supervised results. We are releasing the Segment Anything Model (\sam) and corresponding dataset (\sad) of 1B masks and 11M images at \href{https://segment-anything.com}{https://segment-anything.com} to foster research into foundation models for computer vision.
\vspace{-3mm}
\end{abstract}

%%%%%%%%%%%%%%%%%%%%%%%%%%%%%%%%%%%%%%%%%%%%%%%%%%%%%%%%%%%%%%%%%%%%%%%%%%%%%%%%%%%%%%%%%%%%%%%%%%%
\section{Introduction}\label{sec:intro}

Large language models pre-trained on web-scale datasets are revolutionizing NLP with strong zero-shot and few-shot generalization~\cite{Brown2020}. These ``foundation models''~\cite{bommasani2021opportunities} can generalize to tasks and data distributions beyond those seen during training. This capability is often implemented with \emph{prompt engineering} in which hand-crafted text is used to prompt the language model to generate a valid textual response for the task at hand. When scaled and trained with abundant text corpora from the web, these models' zero and few-shot performance compares surprisingly well to (even matching in some cases) fine-tuned models~\cite{Brown2020,chowdhery2022palm}. Empirical trends show this behavior improving with model scale, dataset size, and total training compute~\cite{kaplan2020scaling,Brown2020,chowdhery2022palm,hoffmann2022training}.

Foundation models have also been explored in computer vision, albeit to a lesser extent. Perhaps the most prominent illustration aligns paired text and images from the web. For example, CLIP~\cite{Radford2021} and ALIGN~\cite{jia2021scaling} use contrastive learning to train text and image encoders that align the two modalities. Once trained, engineered text prompts enable zero-shot generalization to novel visual concepts and data distributions. Such encoders also compose effectively with other modules to enable downstream tasks, such as image generation (\eg, DALL·E~\cite{Ramesh2021}). While much progress has been made on vision and language encoders, computer vision includes a wide range of problems beyond this scope, and for many of these, abundant training data does not exist.

In this work, our goal is to build \emph{a foundation model for image segmentation}. That is, we seek to develop a promptable model and pre-train it on a broad dataset using a task that enables powerful generalization. With this model, we aim to solve a range of downstream segmentation problems on new data distributions using prompt engineering.

The success of this plan hinges on three components: \textbf{task}, \textbf{model}, and \textbf{data}. To develop them, we address the following questions about image segmentation:
\begin{enumerate}[itemsep=-0.6mm]
\item What \textbf{task} will enable zero-shot generalization?
\item What is the corresponding \textbf{model} architecture?
\item What \textbf{data} can power this task and model?
\end{enumerate}
These questions are entangled and require a comprehensive solution. We start by defining a \emph{promptable segmentation} \textbf{task} that is general enough to provide a powerful pre-training objective and to enable a wide range of downstream applications. This task requires a \textbf{model} that supports flexible prompting and can output segmentation masks in real-time when prompted to allow for interactive use. To train our model, we need a diverse, large-scale source of \textbf{data}. Unfortunately, there is no web-scale data source for segmentation; to address this, we build a ``data engine'', \ie, we iterate between using our efficient model to assist in data collection and using the newly collected data to improve the model. We introduce each interconnected component next, followed by the dataset we created and the experiments that demonstrate the effectiveness of our approach.

\paragraph{Task (\S\ref{sec:task}).} In NLP and more recently computer vision, foundation models are a promising development that can perform zero-shot and few-shot learning for new datasets and tasks often by using ``prompting'' techniques. Inspired by this line of work, we propose the \emph{promptable segmentation task}, where the goal is to return a \emph{valid} segmentation mask given any segmentation \emph{prompt} (see~\fig{fig:teaser}{\color{linkcolor}a}). A prompt simply specifies what to segment in an image, \eg, a prompt can include spatial or text information identifying an object. The requirement of a valid output mask means that even when a prompt is ambiguous and could refer to multiple objects (for example, a point on a shirt may indicate either the shirt or the person wearing it), the output should be a reasonable mask for at least one of those objects. We use the promptable segmentation task as both a pre-training objective and to solve general downstream segmentation tasks via prompt engineering.

\paragraph{Model (\S\ref{sec:model}).} The promptable segmentation task and the goal of real-world use impose constraints on the model architecture. In particular, the model must support \emph{flexible prompts}, needs to compute masks in amortized \emph{real-time} to allow interactive use, and must be \emph{ambiguity-aware}. Surprisingly, we find that a simple design satisfies all three constraints: a powerful image encoder computes an image embedding, a prompt encoder embeds prompts, and then the two information sources are combined in a lightweight mask decoder that predicts segmentation masks. We refer to this model as the Segment Anything Model, or \sam (see~\fig{fig:teaser}{\color{linkcolor}b}). By separating \sam into an image encoder and a fast prompt encoder / mask decoder, the same image embedding can be reused (and its cost amortized) with different prompts. Given an image embedding, the prompt encoder and mask decoder predict a mask from a prompt in $\app$50ms in a web browser. We focus on point, box, and mask prompts, and also present initial results with free-form text prompts. To make \sam ambiguity-aware, we design it to predict multiple masks for a single prompt allowing \sam to naturally handle ambiguity, such as the shirt \vs person example.

\paragraph{Data engine (\S\ref{sec:engine}).} To achieve strong generalization to new data distributions, we found it necessary to train \sam on a large and diverse set of masks, beyond any segmentation dataset that already exists. While a typical approach for foundation models is to obtain data online~\cite{Radford2021}, masks are not naturally abundant and thus we need an alternative strategy. Our solution is to build a ``data engine'', \ie, we co-develop our model with model-in-the-loop dataset annotation (see~\fig{fig:teaser}{\color{linkcolor}c}). Our data engine has three stages: \emph{assisted-manual}, \emph{semi-automatic}, and \emph{fully automatic}. In the first stage, \sam assists annotators in annotating masks, similar to a classic interactive segmentation setup. In the second stage, \sam can automatically generate masks for a subset of objects by prompting it with likely object locations and annotators focus on annotating the remaining objects, helping increase mask diversity. In the final stage, we prompt \sam with a regular grid of foreground points, yielding on average \app100 high-quality masks per image.

\paragraph{Dataset (\S\ref{sec:dataset}).} Our final dataset, \sad, includes more than \emph{1B} masks from \emph{11M} licensed and privacy-preserving images (see \fig{fig:sa1bvisuals}). \sad, collected fully automatically using the final stage of our data engine, has 400$\x$ more masks than any existing segmentation dataset~\cite{Lin2014,Gupta2019,Zhou2019,OpenImages}, and as we verify extensively, the masks are of high quality and diversity. Beyond its use in training \sam to be robust and general, we hope \sad becomes a valuable resource for research aiming to build new foundation models.

\paragraph{Responsible AI (\S\ref{sec:rai}).} We study and report on potential fairness concerns and biases when using \sad and \sam. Images in \sad span a geographically and economically diverse set of countries and we found that \sam performs similarly across different groups of people. Together, we hope this will make our work more equitable for real-world use cases. We provide model and dataset cards in the \supp.

\paragraph{Experiments (\S\ref{sec:eval}).} We extensively evaluate \sam. First, using a diverse new suite of 23 segmentation datasets, we find that \sam produces high-quality masks from a single foreground point, often only slightly below that of the manually annotated ground truth. Second, we find consistently strong quantitative and qualitative results on a variety of downstream tasks under a zero-shot transfer protocol using prompt engineering, including edge detection, object proposal generation, instance segmentation, and a preliminary exploration of text-to-mask prediction. These results suggest that \sam can be used out-of-the-box with prompt engineering to solve a variety of tasks involving object and image distributions beyond \sam's training data. Nevertheless, room for improvement remains, as we discuss in \S\ref{sec:disc}.

\paragraph{Release.} We are releasing the \sad dataset for research purposes and making \sam available under a permissive open license (Apache 2.0) at \href{https://segment-anything.com}{https://segment-anything.com}. We also showcase \sam's capabilities with an \href{https://segment-anything.com/demo}{online demo}.

%##################################################################################################
\newcommand{\incvisualsrow}[6]{
\begin{subfigure}{\linewidth}\begin{adjustbox}{width=\textwidth}{
\def\arraystretch{0.2}\setlength\tabcolsep{#1pt}\begin{tabular}{ccccc}
\raisebox{1.2\normalbaselineskip}[0pt][0pt]{\rotatebox{90}{\tiny #2}} &
\includegraphics[height=2cm]{figs/sa1b_examples/#3.jpg} &
\includegraphics[height=2cm]{figs/sa1b_examples/#4.jpg} &
\includegraphics[height=2cm]{figs/sa1b_examples/#5.jpg} &
\includegraphics[height=2cm]{figs/sa1b_examples/#6.jpg} \\
\end{tabular}}\end{adjustbox}\end{subfigure}}
%##################################################################################################

%##################################################################################################
\begin{figure*}[!t]
\centering
\incvisualsrow{0.5}{$<$50 masks}{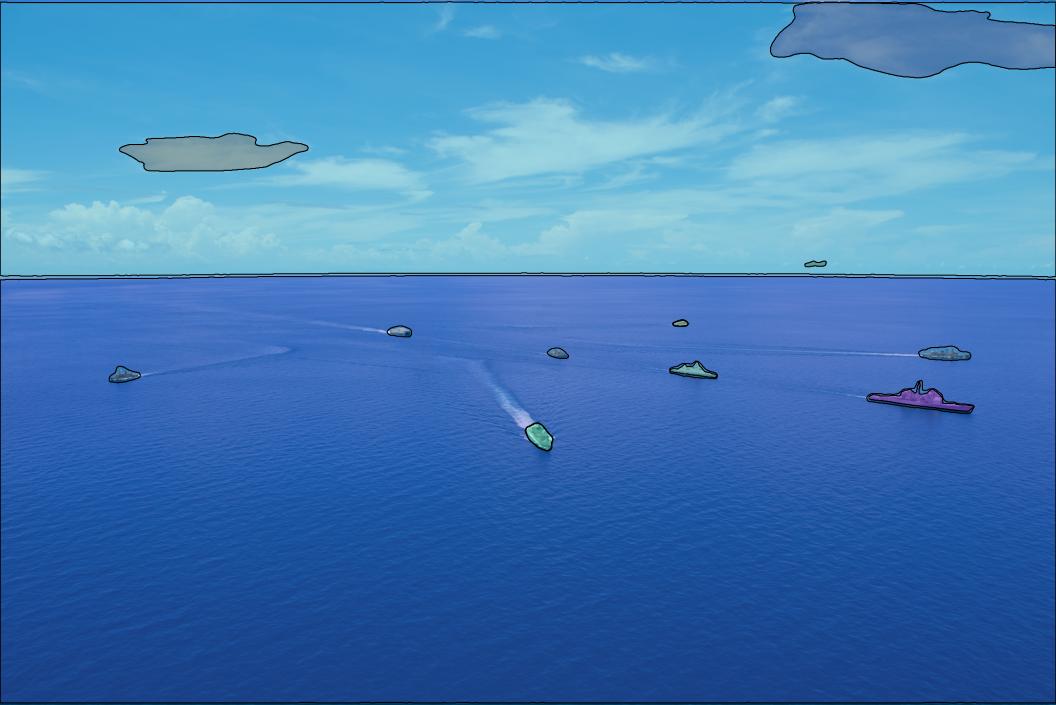}{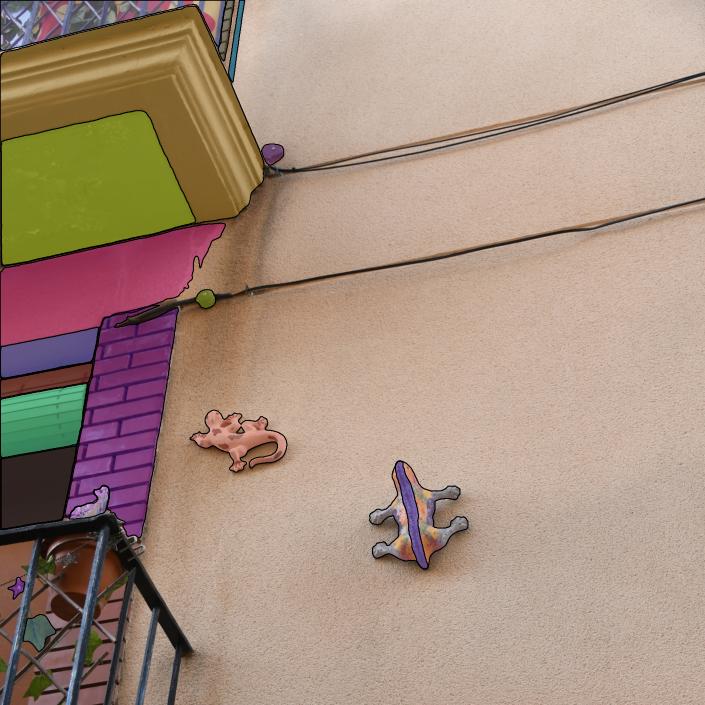}{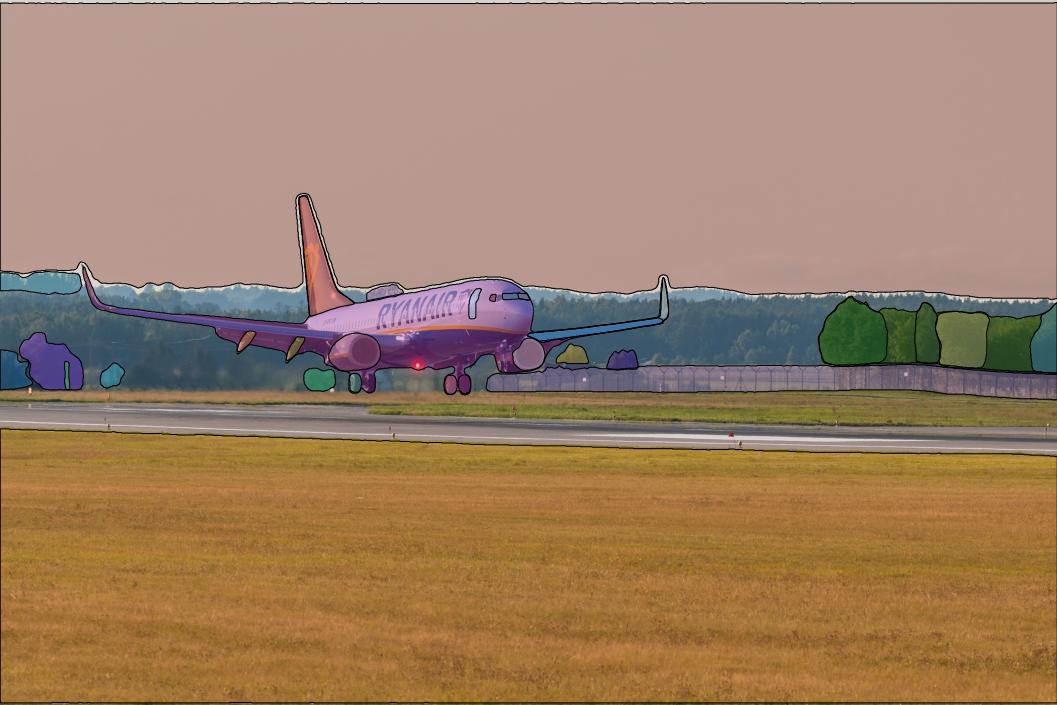}{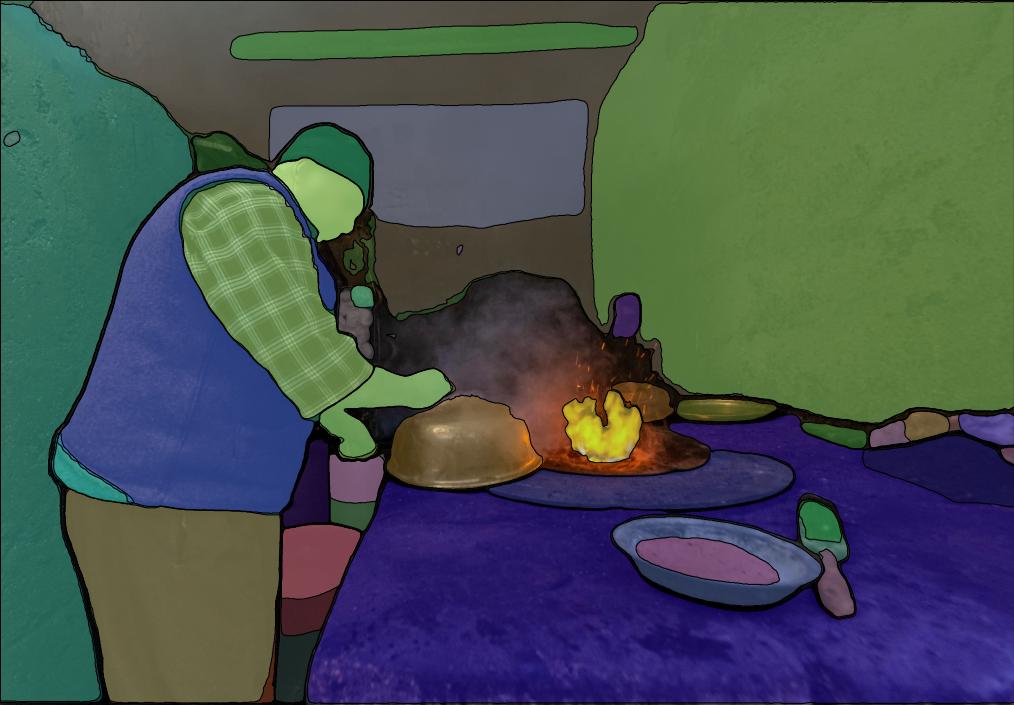}
\incvisualsrow{1.0}{50-100 masks}{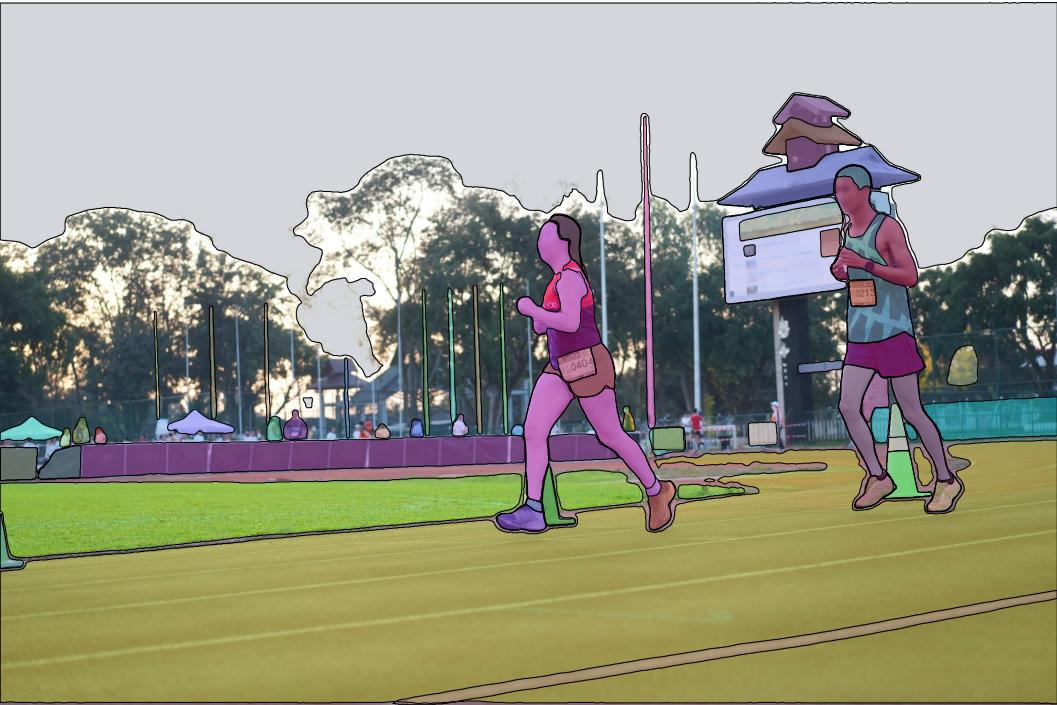}{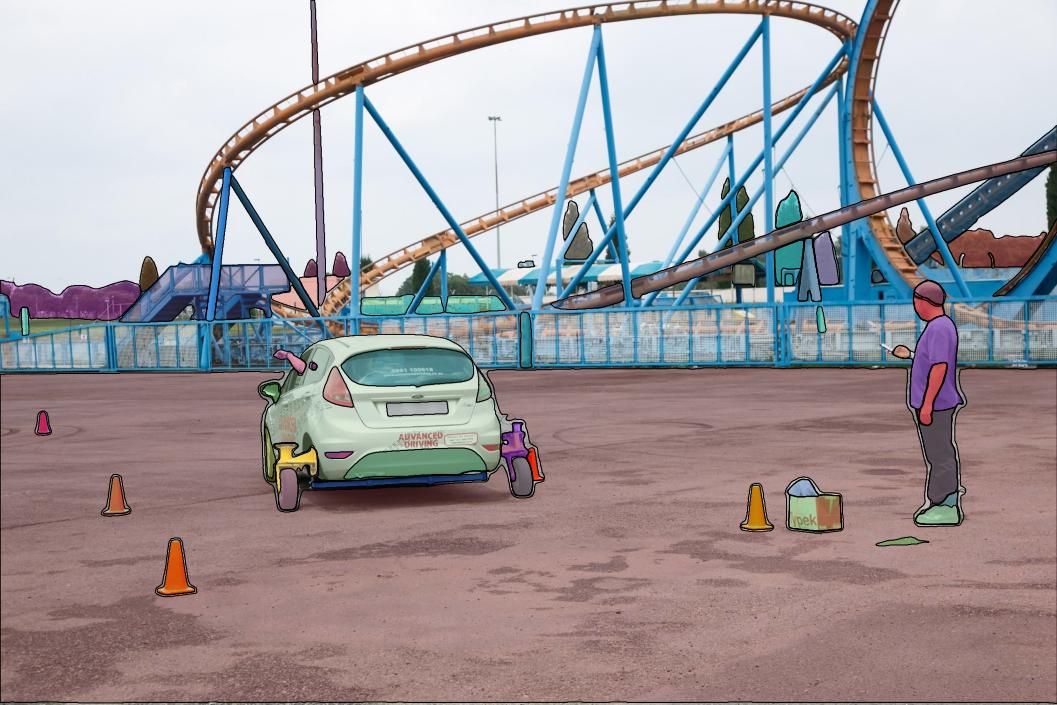}{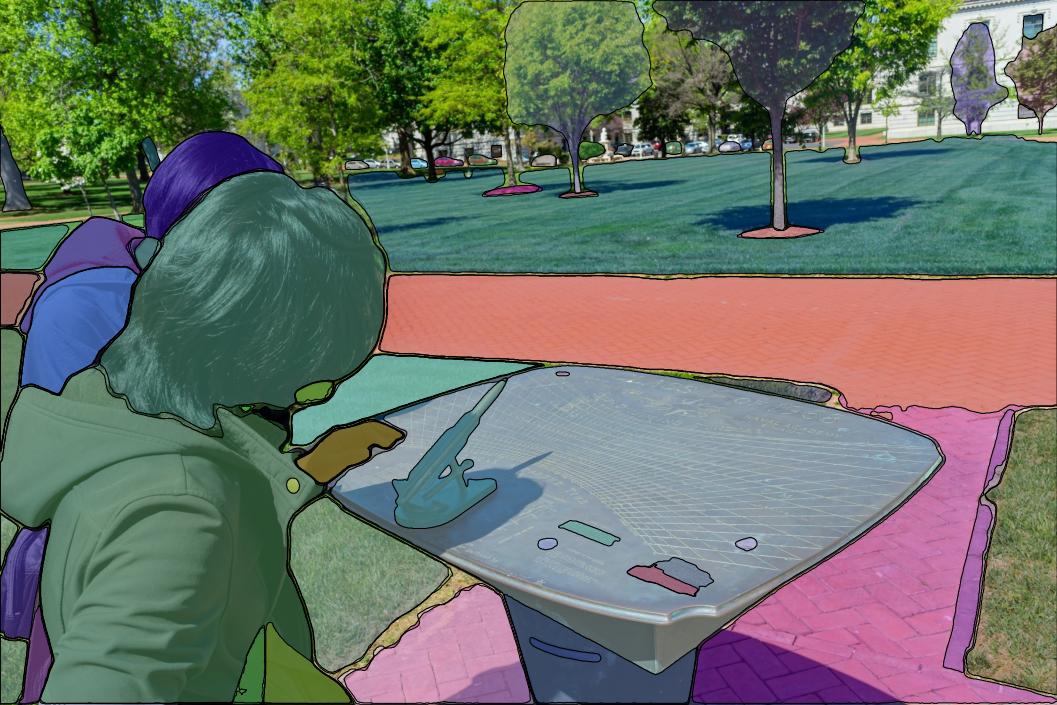}{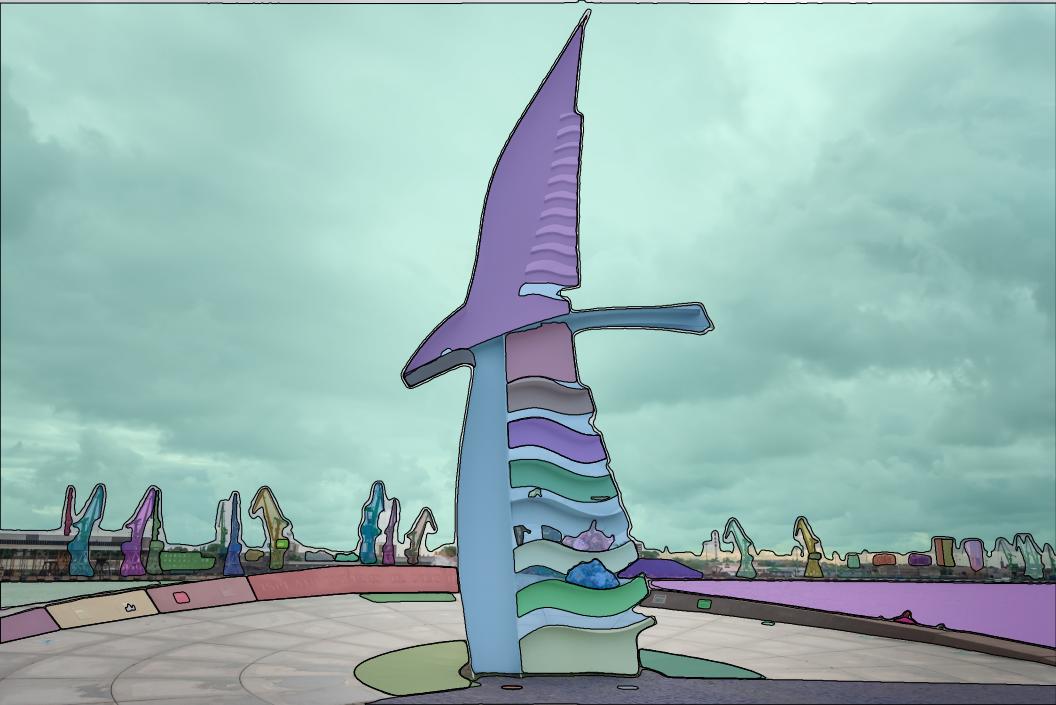}
\incvisualsrow{0.8}{100-200 masks}{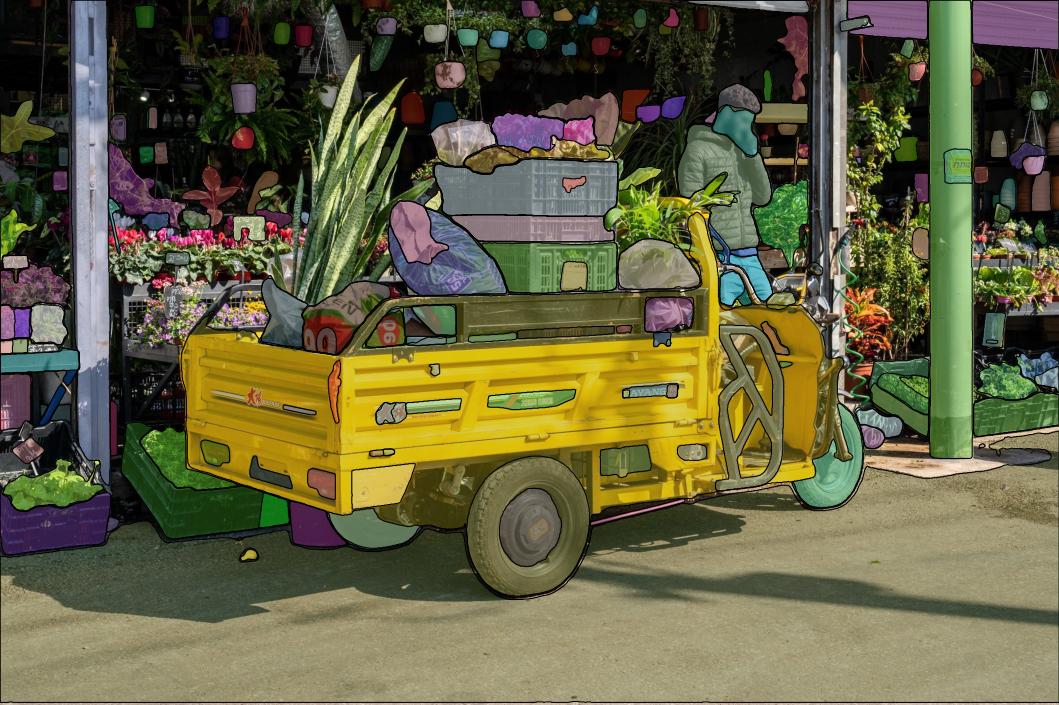}{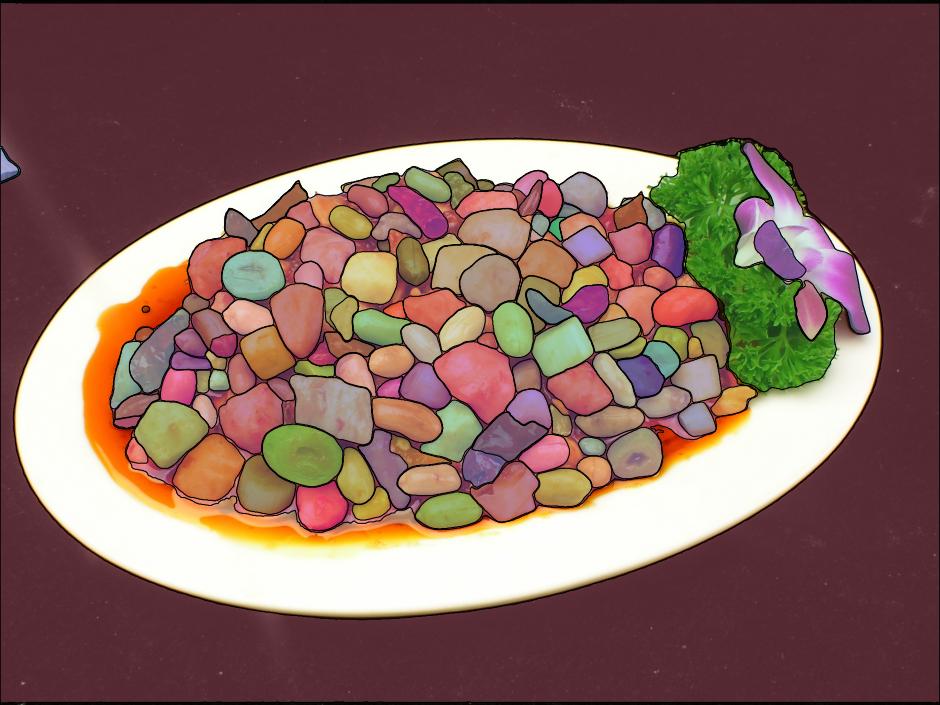}{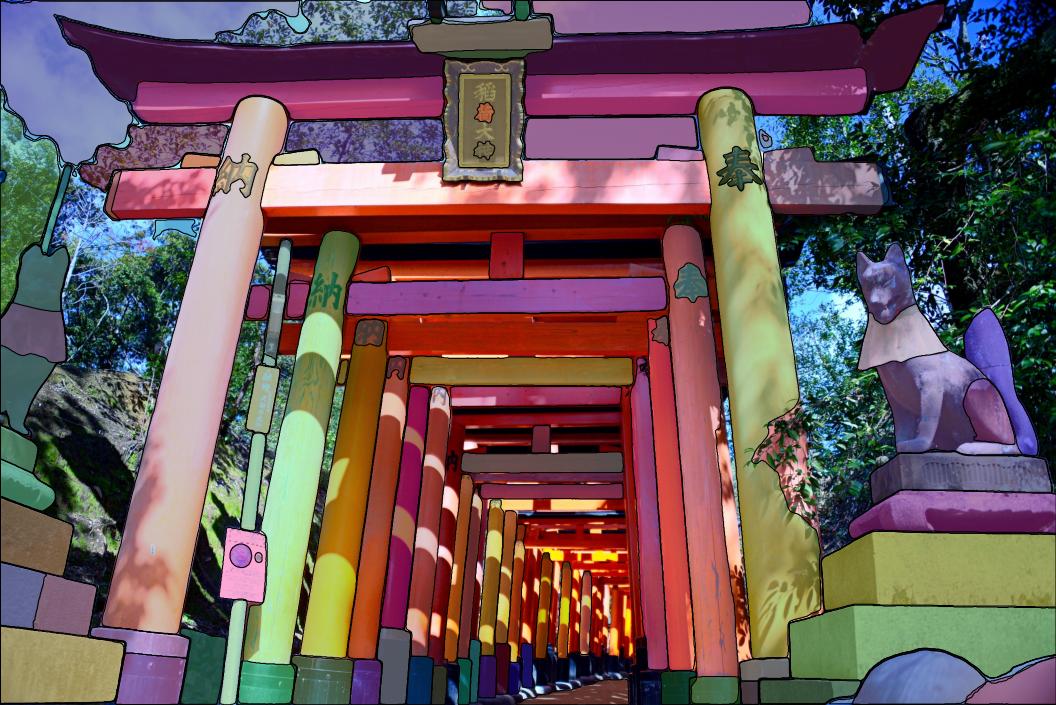}{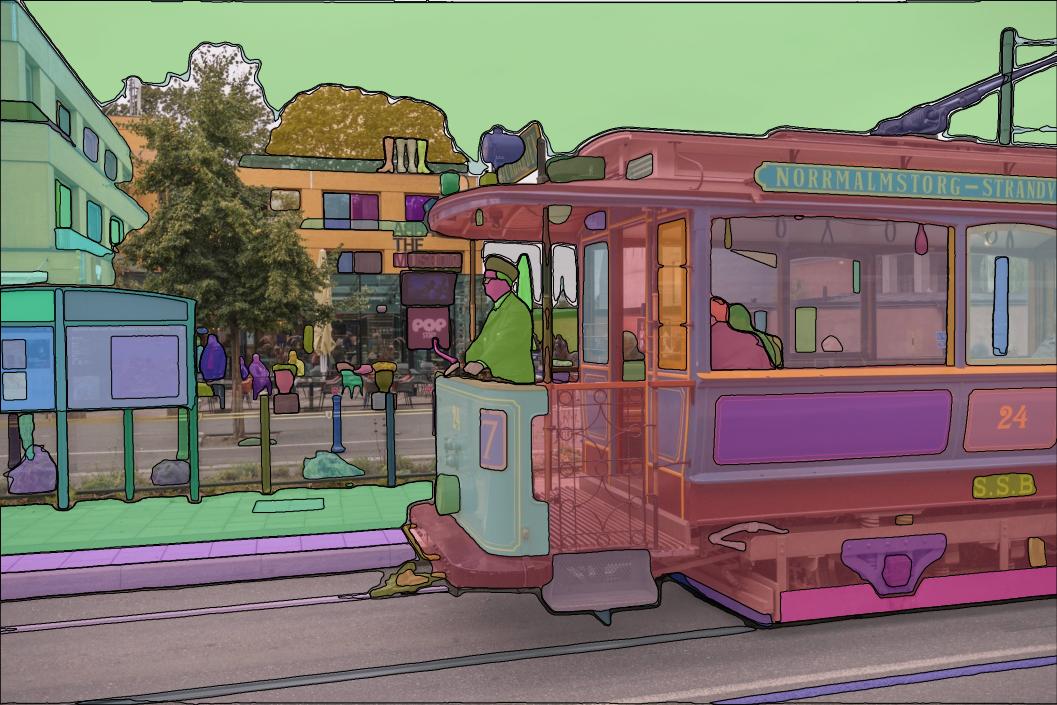}
\incvisualsrow{0.8}{200-300 masks}{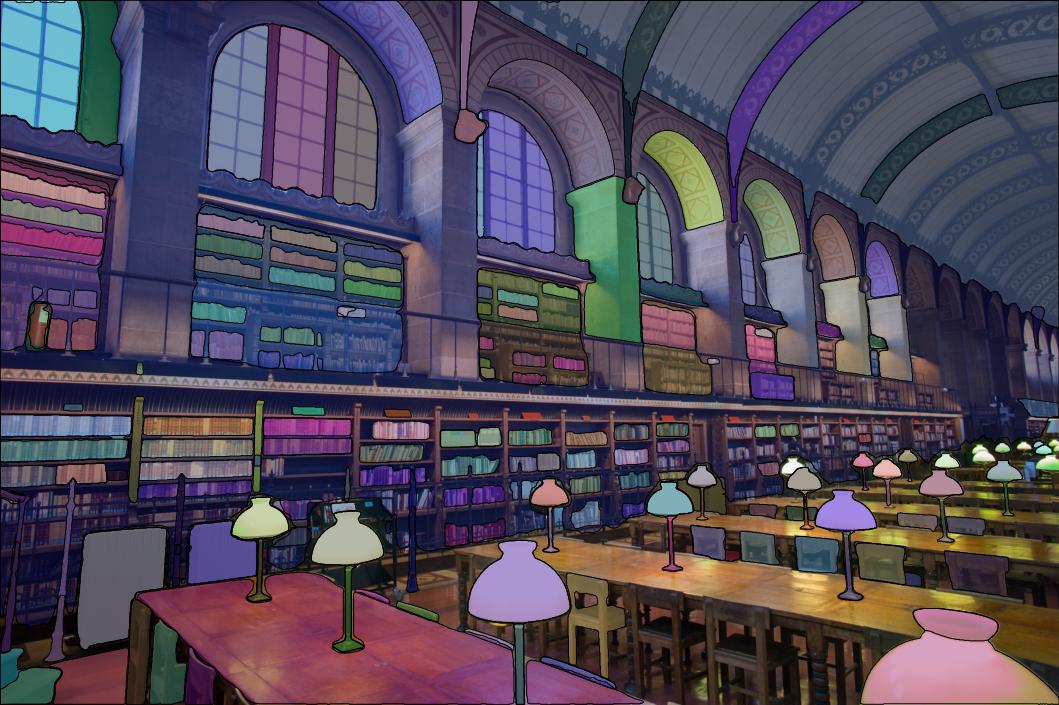}{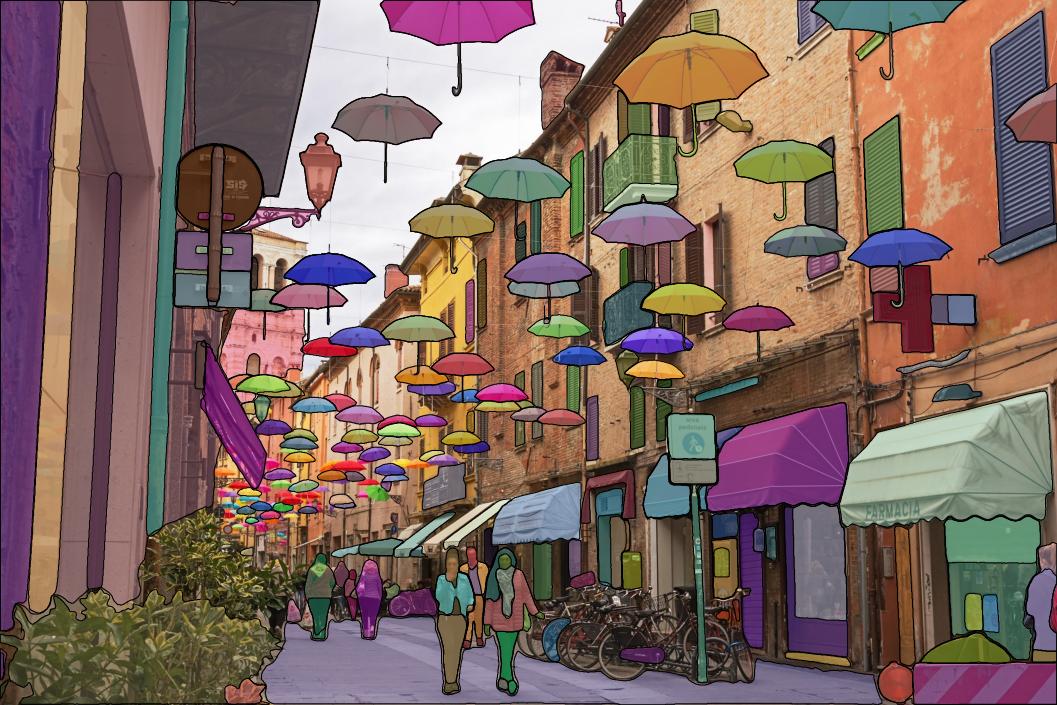}{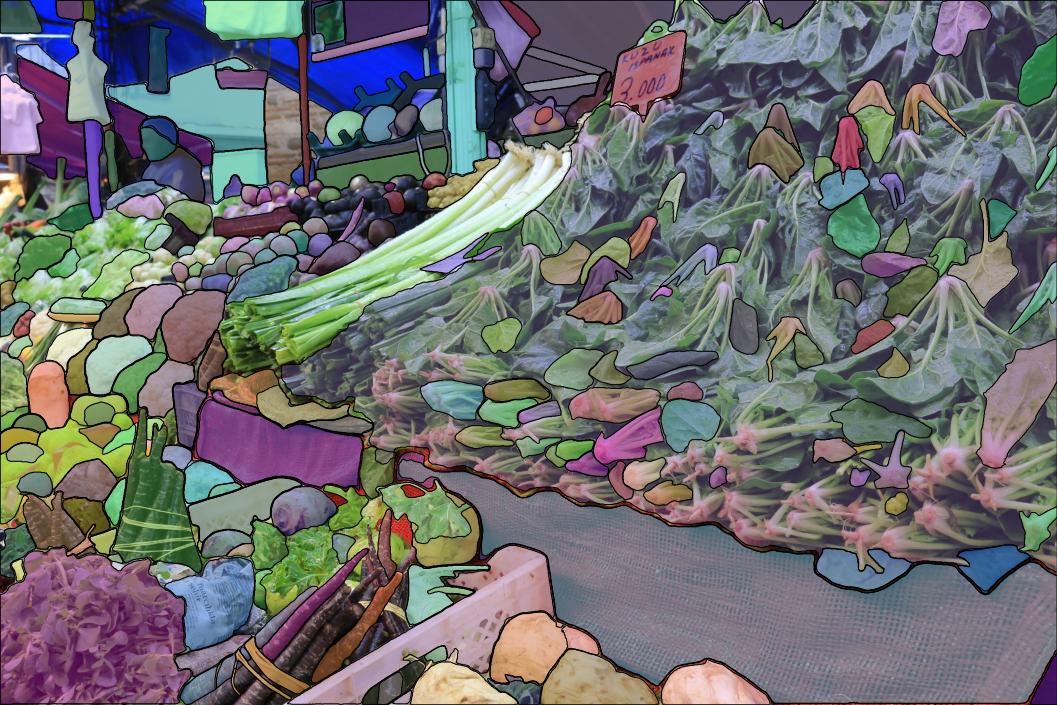}{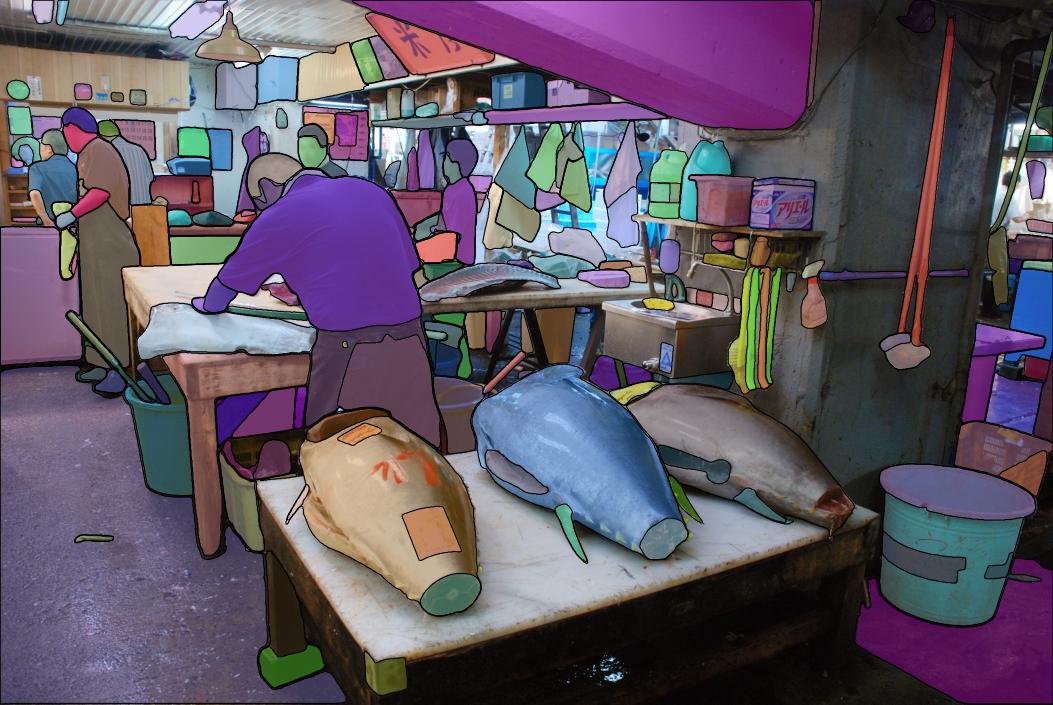}
\incvisualsrow{0.6}{300-400 masks}{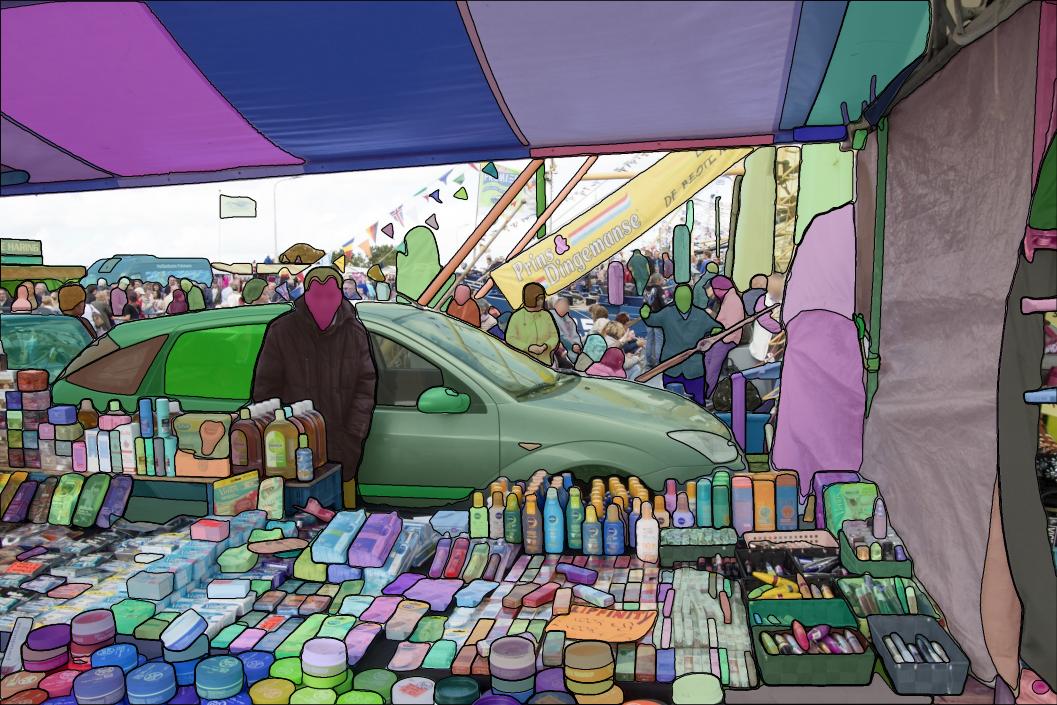}{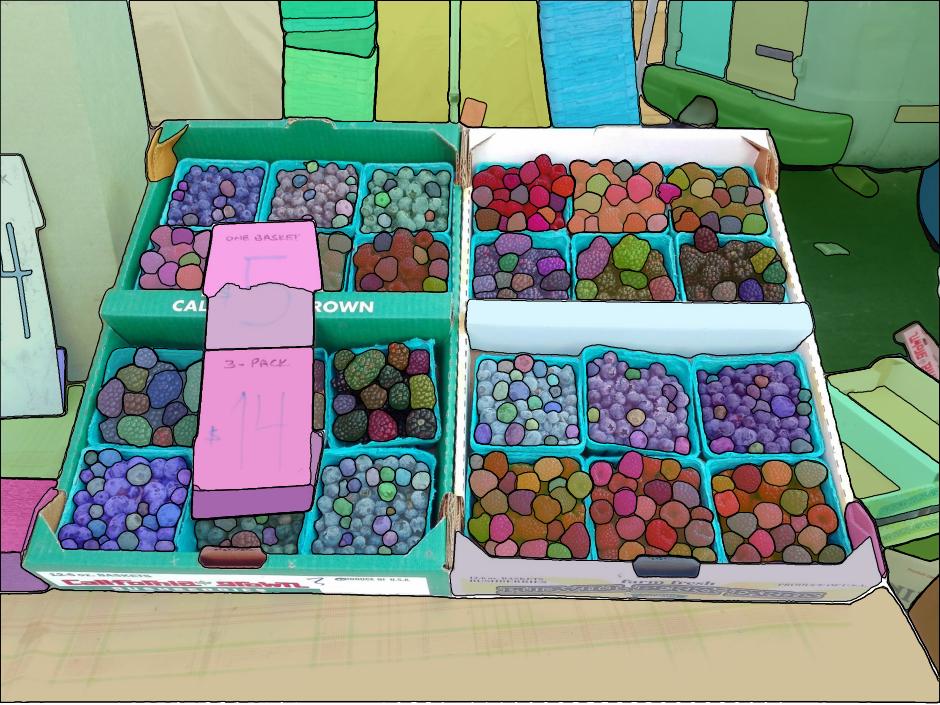}{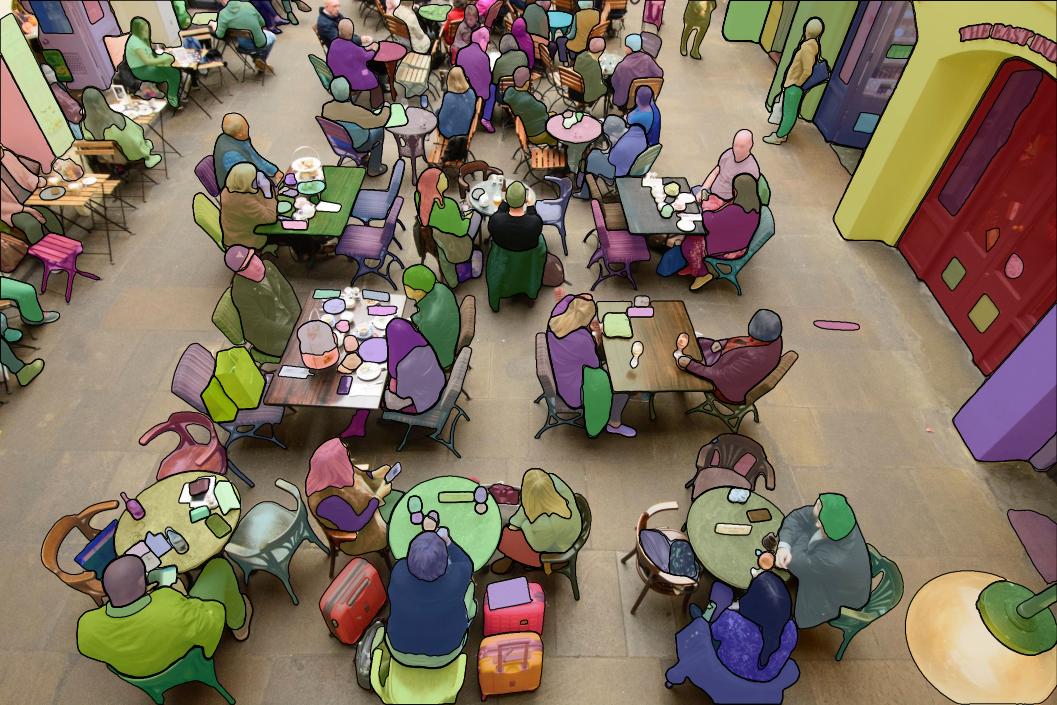}{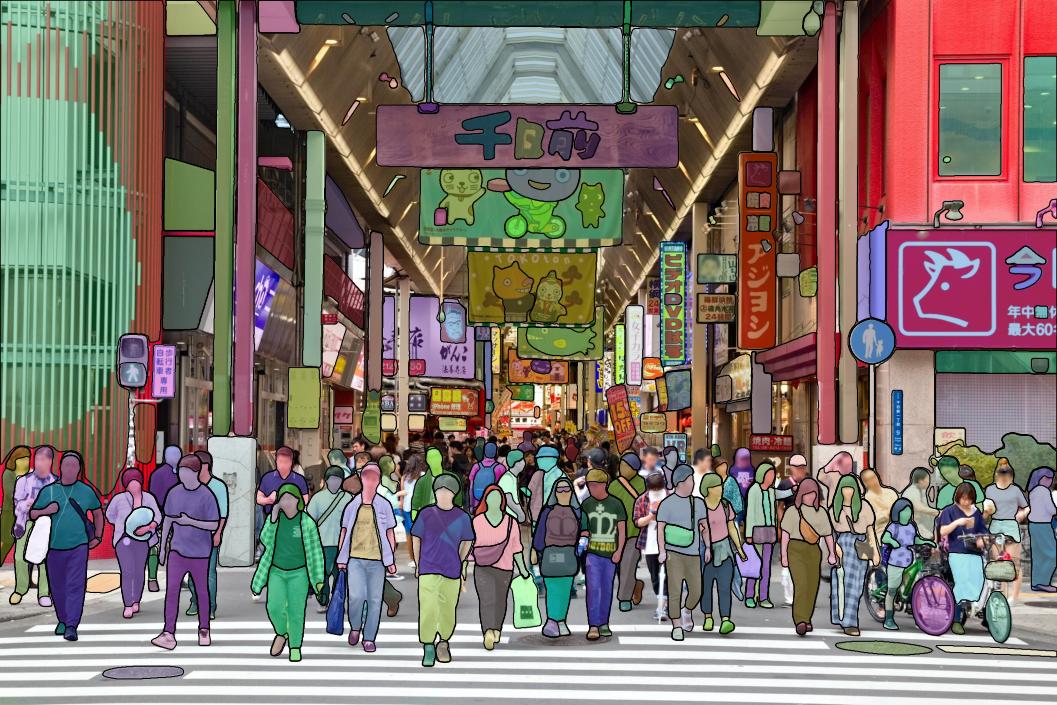}
\incvisualsrow{0.7}{400-500 masks}{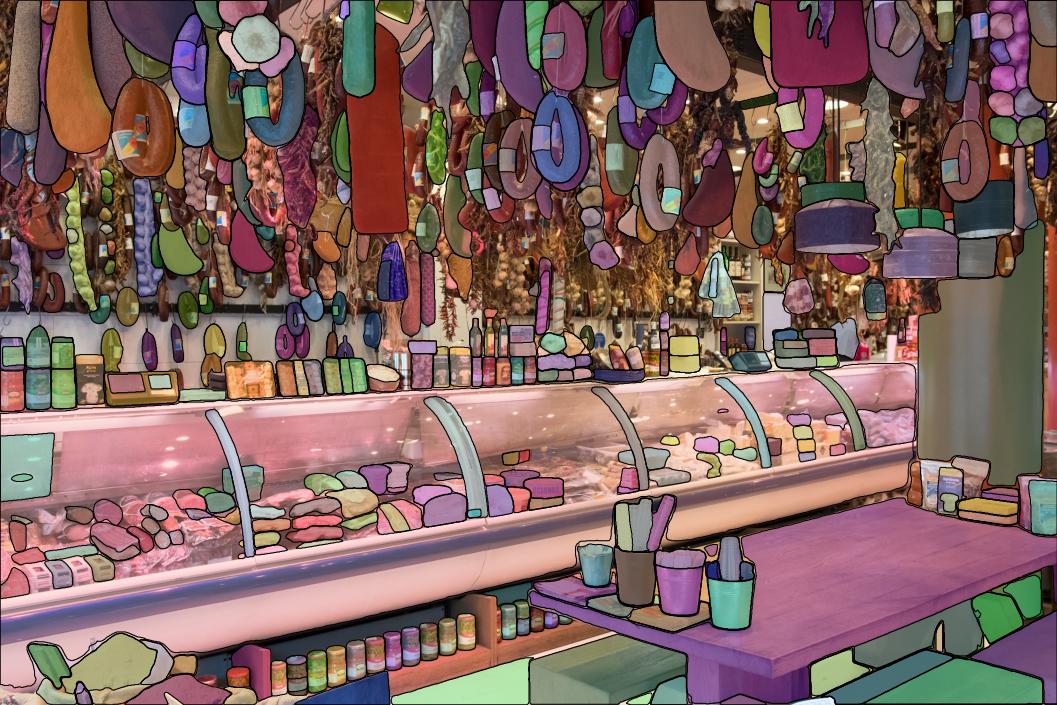}{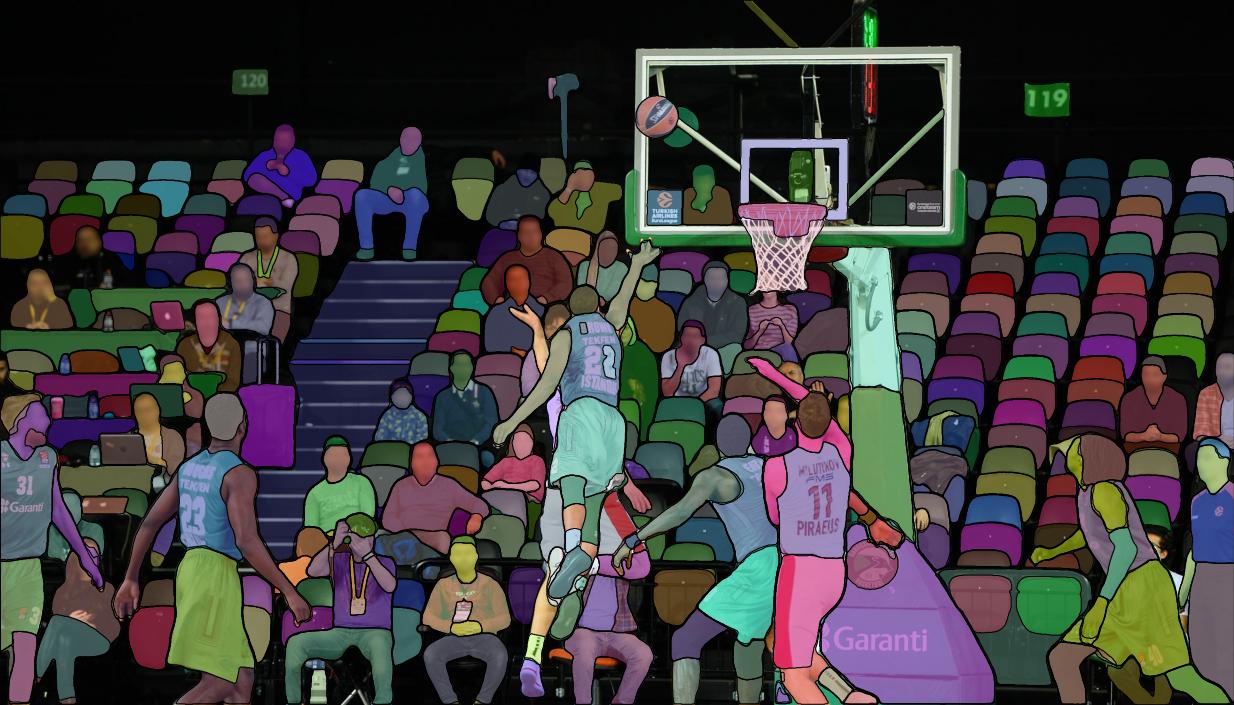}{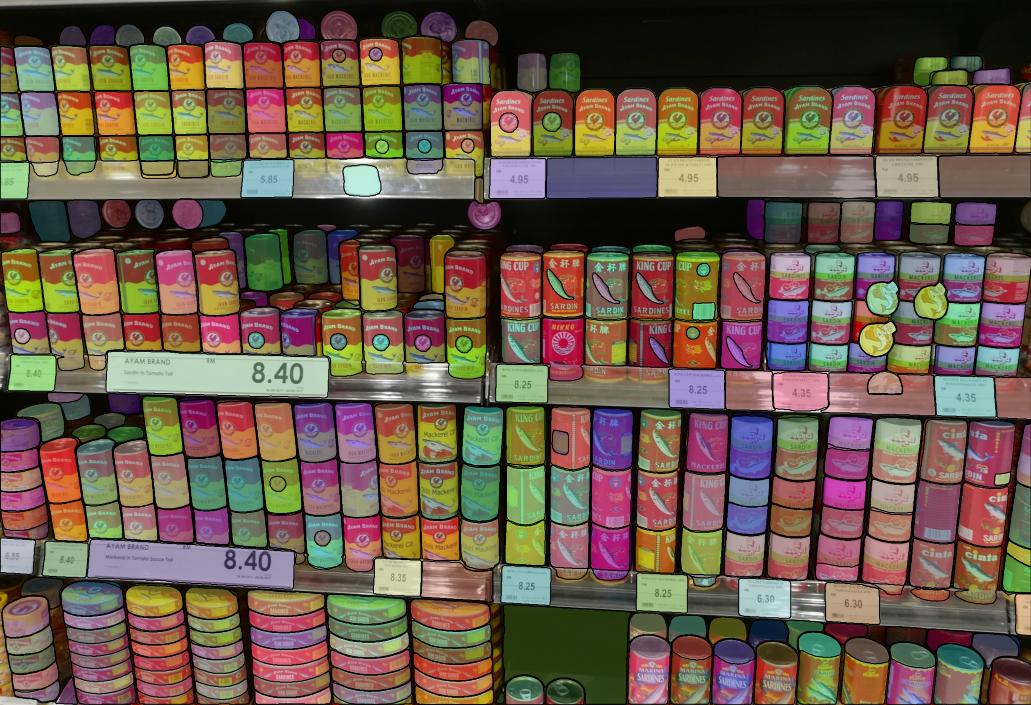}{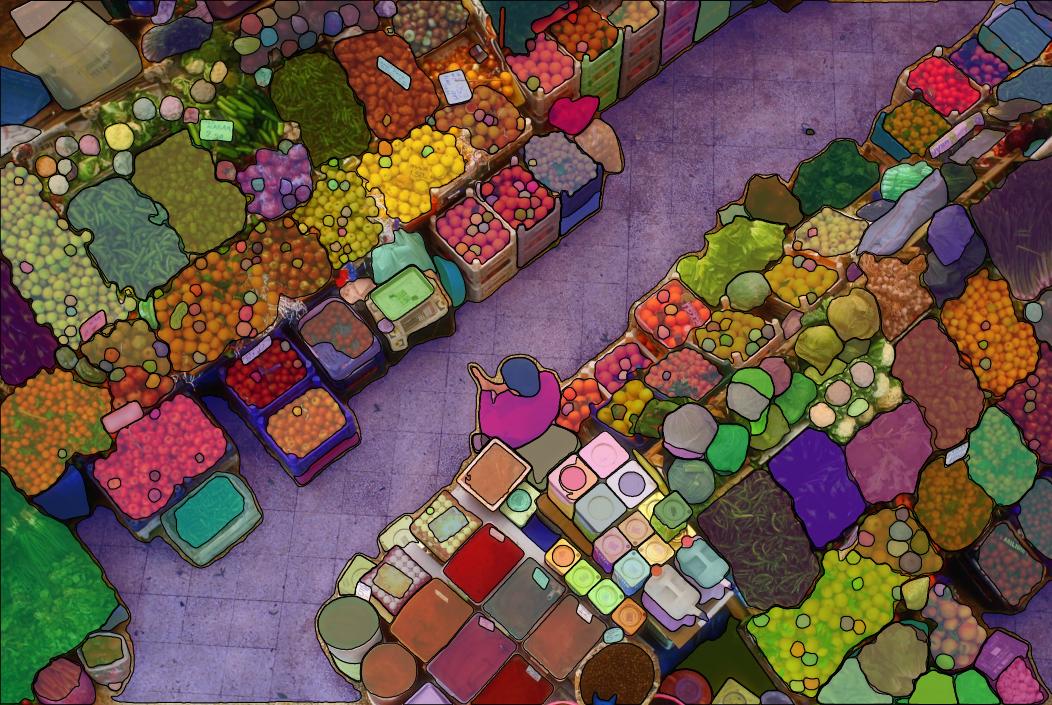}
\incvisualsrow{0.3}{$>$500 masks}{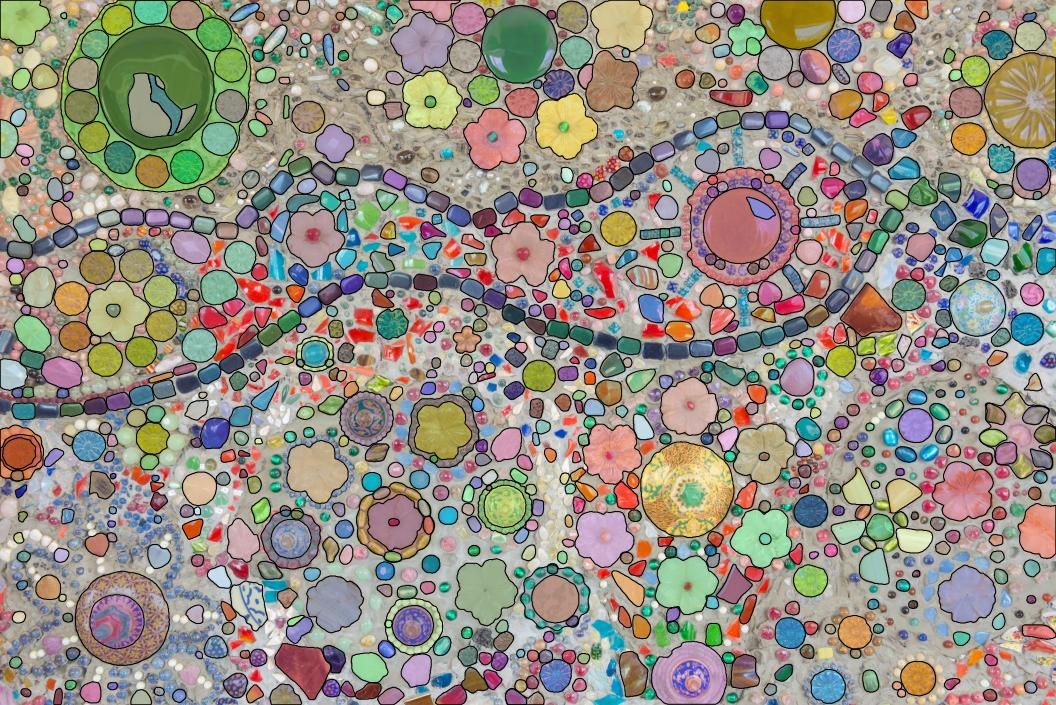}{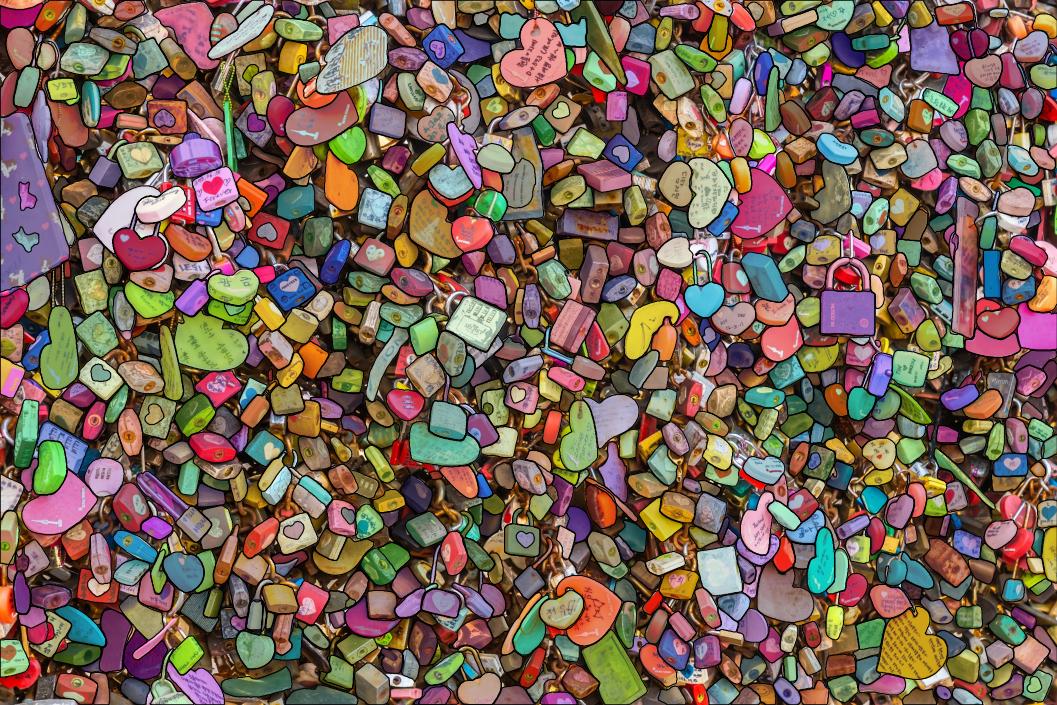}{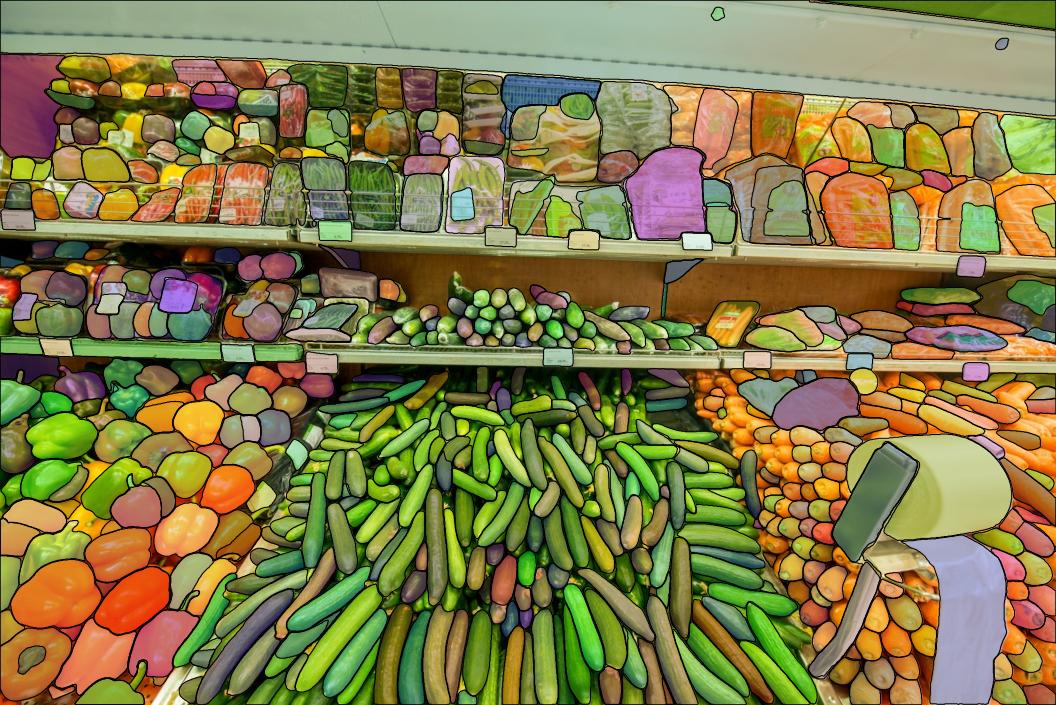}{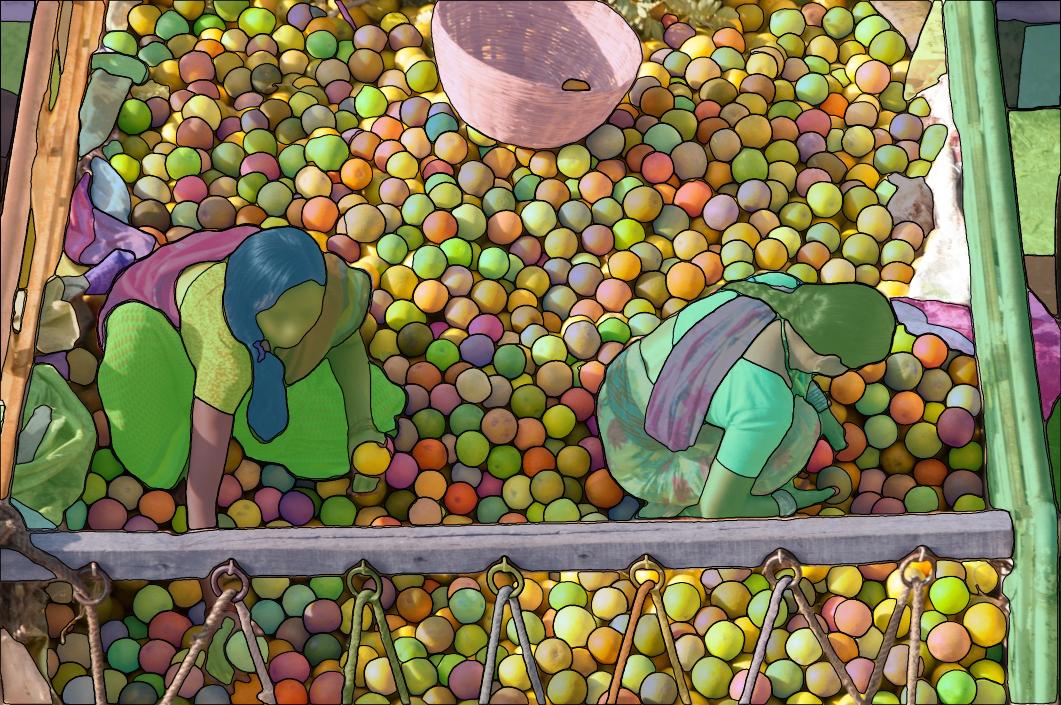}
\caption{Example images with overlaid masks from our newly introduced dataset, \textbf{\sad}. \sad contains 11M diverse, high-resolution, licensed, and privacy protecting images and 1.1B high-quality segmentation masks. These masks were annotated \emph{fully automatically} by \sam, and as we verify by human ratings and numerous experiments, are of high quality and diversity. We group images by number of masks per image for visualization (there are \app100 masks per image on average).}
\label{fig:sa1bvisuals}
\end{figure*}
%##################################################################################################

%%%%%%%%%%%%%%%%%%%%%%%%%%%%%%%%%%%%%%%%%%%%%%%%%%%%%%%%%%%%%%%%%%%%%%%%%%%%%%%%%%%%%%%%%%%%%%%%%%%
\section{Segment Anything Task}\label{sec:task}

We take inspiration from NLP, where the next token prediction task is used for foundation model pre-training \emph{and} to solve diverse downstream tasks via prompt engineering~\cite{Brown2020}. To build a foundation model for segmentation, we aim to define a task with analogous capabilities.

\begin{figure}[t]\centering
\includegraphics[width=.97\linewidth]{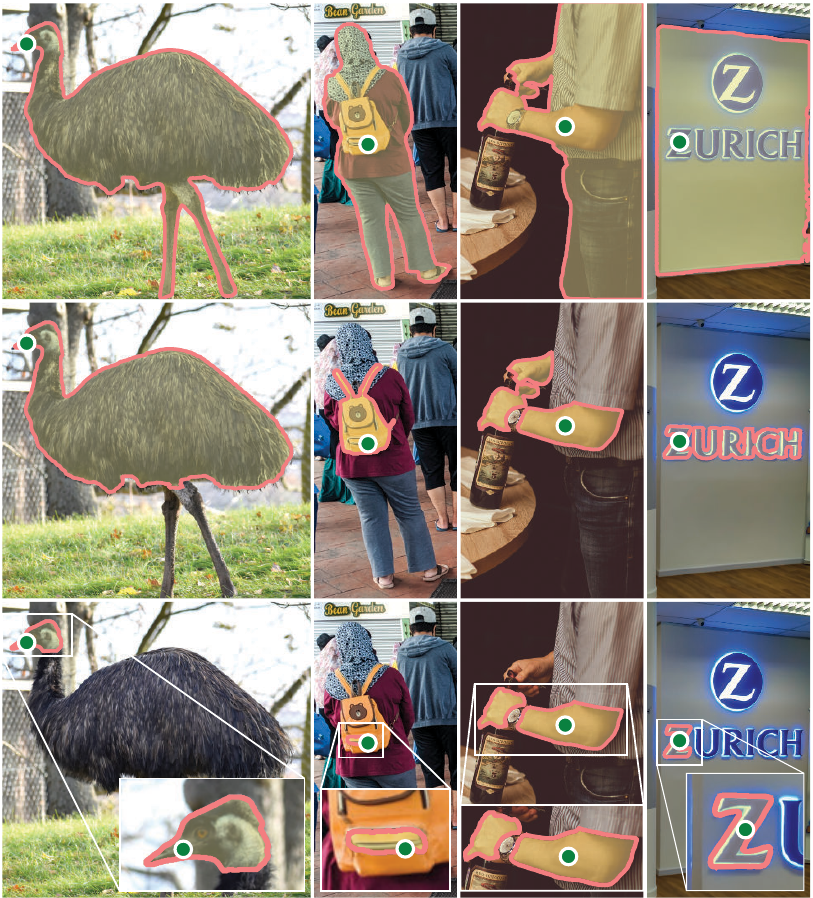}\vspace{-1mm}
\caption{Each column shows 3 valid masks generated by \sam from a single ambiguous point prompt (green circle).}
\label{fig:ambiguity_examples}\vspace{-1mm}
\end{figure}

\paragraph{Task.} We start by translating the idea of a prompt from NLP to segmentation, where a prompt can be a set of foreground / background points, a rough box or mask, free-form text, or, in general, any information indicating what to segment in an image. The \emph{promptable segmentation task}, then, is to return a \emph{valid} segmentation mask given any \emph{prompt}. The requirement of a ``valid'' mask simply means that even when a prompt is \emph{ambiguous} and could refer to multiple objects (\eg, recall the shirt \vs person example, and see \fig{fig:ambiguity_examples}), the output should be a reasonable mask for at least \emph{one} of those objects. This requirement is similar to expecting a language model to output a coherent response to an ambiguous prompt. We choose this task because it leads to a natural pre-training algorithm \emph{and} a general method for zero-shot transfer to downstream segmentation tasks via prompting.

\paragraph{Pre-training.} The promptable segmentation task suggests a natural pre-training algorithm that simulates a sequence of prompts (\eg, points, boxes, masks) for each training sample and compares the model's mask predictions against the ground truth. We adapt this method from interactive segmentation~\cite{xu2016deep,mahadevan2018iteratively}, although unlike interactive segmentation whose aim is to eventually predict a valid mask after enough user input, our aim is to always predict a \emph{valid mask} for \emph{any prompt} even when the prompt is \emph{ambiguous}. This ensures that a pre-trained model is effective in use cases that involve ambiguity, including automatic annotation as required by our data engine \S\ref{sec:engine}. We note that performing well at this task is challenging and requires specialized modeling and training loss choices, which we discuss in \S\ref{sec:model}.

\paragraph{Zero-shot transfer.} Intuitively, our pre-training task endows the model with the ability to respond appropriately to any prompt at inference time, and thus downstream tasks can be solved by engineering appropriate prompts. For example, if one has a bounding box detector for cats, cat instance segmentation can be solved by providing the detector's box output as a prompt to our model. In general, a wide array of practical segmentation tasks can be cast as prompting. In addition to automatic dataset labeling, we explore five diverse example tasks in our experiments in \S\ref{sec:eval}.

%##################################################################################################
%##################################################################################################

\paragraph{Related tasks.} Segmentation is a broad field: there's interactive segmentation~\cite{kass1988snakes,xu2016deep}, edge detection~\cite{arbelaez2010contour}, super pixelization~\cite{ren2003learning}, object proposal generation~\cite{alexe2010object}, foreground segmentation~\cite{stauffer1999adaptive}, semantic segmentation~\cite{shotton2006textonboost}, instance segmentation~\cite{Lin2014}, panoptic segmentation~\cite{kirillov2019panoptic}, \etc. The goal of our promptable segmentation task is to produce a broadly capable model that can adapt to \emph{many} (though not all) existing and \emph{new} segmentation tasks via prompt engineering. This capability is a form of task generalization~\cite{da2012learning}. Note that this is different than previous work on multi-task segmentation systems. In a multi-task system, a single model performs a \emph{fixed} set of tasks, \eg, joint semantic, instance, and panoptic segmentation~\cite{zhang2021knet,cheng2022masked,jain2022oneformer}, but the training and test tasks are the same. An important distinction in our work is that a model trained for promptable segmentation can perform a new, different task at inference time by acting as a \emph{component} in a larger system, \eg, to perform instance segmentation, a promptable segmentation model is \emph{combined} with an existing object detector.

\paragraph{Discussion.} Prompting and composition are powerful tools that enable a single model to be used in extensible ways, potentially to accomplish tasks unknown at the time of model design. This approach is analogous to how other foundation models are used, \eg, how CLIP~\cite{Radford2021} is the text-image alignment component of the DALL$\cdot$E~\cite{Ramesh2021} image generation system. We anticipate that composable system design, powered by techniques such as prompt engineering, will enable a wider variety of applications than systems trained specifically for a fixed set of tasks. It's also interesting to compare promptable and interactive segmentation through the lens of composition: while interactive segmentation models are designed with human users in mind, a model trained for promptable segmentation can also be composed into a larger algorithmic system as we will demonstrate.

%##################################################################################################
\begin{figure*}[t]\centering
\includegraphics[width=0.99\linewidth]{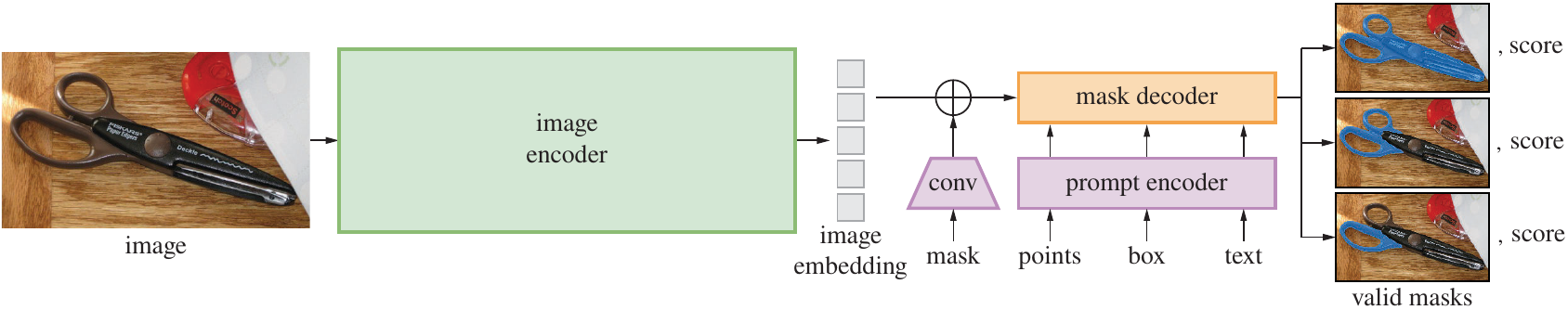}\vspace{-1mm}
\caption{Segment Anything Model (\sam) overview. A heavyweight image encoder outputs an image embedding that can then be efficiently queried by a variety of input prompts to produce object masks at amortized real-time speed. For ambiguous prompts corresponding to more than one object, \sam can output multiple valid masks and associated confidence scores.}
\label{fig:model_diagram}
\vspace{-1mm}
\end{figure*}
%##################################################################################################

%%%%%%%%%%%%%%%%%%%%%%%%%%%%%%%%%%%%%%%%%%%%%%%%%%%%%%%%%%%%%%%%%%%%%%%%%%%%%%%%%%%%%%%%%%%%%%%%%%%
\section{Segment Anything Model}\label{sec:model}

We next describe the Segment Anything Model (\sam) for promptable segmentation. \sam has three components, illustrated in~\fig{fig:model_diagram}: an image encoder, a flexible prompt encoder, and a fast mask decoder. We build on Transformer vision models~\cite{Carion2020,Dosovitskiy2021,cheng2021per,li2022exploring} with specific tradeoffs for (amortized) real-time performance. We describe these components at a high-level here, with details in \S\ref{app:model}.

\paragraph{Image encoder.} Motivated by scalability and powerful pre-training methods, we use an MAE~\cite{he2022masked} pre-trained Vision Transformer (ViT)~\cite{Dosovitskiy2021} minimally adapted to process high resolution inputs~\cite{li2022exploring}. The image encoder runs once per image and can be applied prior to prompting the model.

\paragraph{Prompt encoder.} We consider two sets of prompts: \emph{sparse} (points, boxes, text) and \emph{dense} (masks). We represent points and boxes by positional encodings~\cite{tancik2020fourier} summed with learned embeddings for each prompt type and free-form text with an off-the-shelf text encoder from CLIP~\cite{Radford2021}. Dense prompts (\ie, masks) are embedded using convolutions and summed element-wise with the image embedding.

\paragraph{Mask decoder.} The mask decoder efficiently maps the image embedding, prompt embeddings, and an output token to a mask. This design, inspired by~\cite{Carion2020,cheng2021per}, employs a modification of a Transformer decoder block~\cite{Vaswani2017} followed by a dynamic mask prediction head. Our modified decoder block uses prompt self-attention and cross-attention in two directions (prompt-to-image embedding and vice-versa) to update \emph{all} embeddings. After running two blocks, we upsample the image embedding and an MLP maps the output token to a dynamic linear classifier, which then computes the mask foreground probability at each image location.

\paragraph{Resolving ambiguity.} With one output, the model will average multiple valid masks if given an ambiguous prompt. To address this, we modify the model to predict multiple output masks for a single prompt (see \fig{fig:ambiguity_examples}). We found 3 mask outputs is sufficient to address most common cases (nested masks are often at most three deep: whole, part, and subpart). During training, we backprop only the minimum loss~\cite{charpiat2008automatic,guzman2012multiple,li2018interactive} over masks. To rank masks, the model predicts a confidence score (\ie, estimated IoU) for each mask.

\paragraph{Efficiency.} The overall model design is largely motivated by efficiency. Given a precomputed image embedding, the prompt encoder and mask decoder run in a web browser, on CPU, in \app50ms. This runtime performance enables seamless, real-time interactive prompting of our model.

\paragraph{Losses and training.} We supervise mask prediction with the linear combination of focal loss~\cite{Lin2017a} and dice loss~\cite{milletari2016v} used in~\cite{Carion2020}. We train for the promptable segmentation task using a mixture of geometric prompts (for text prompts see \S\ref{sec:eval:text_to_mask}). Following~\cite{sofiiuk2022reviving,forte2020getting}, we simulate an interactive setup by randomly sampling prompts in 11 rounds per mask, allowing \sam to integrate seamlessly into our data engine.

%%%%%%%%%%%%%%%%%%%%%%%%%%%%%%%%%%%%%%%%%%%%%%%%%%%%%%%%%%%%%%%%%%%%%%%%%%%%%%%%%%%%%%%%%%%%%%%%%%%
\section{Segment Anything Data Engine}\label{sec:engine}

As segmentation masks are not abundant on the internet, we built a data engine to enable the collection of our 1.1B mask dataset, \sad. The data engine has three stages: (1) a model-assisted manual annotation stage, (2) a semi-automatic stage with a mix of automatically predicted masks and model-assisted annotation, and (3) a fully automatic stage in which our model generates masks without annotator input. We go into details of each next.

\paragraph{Assisted-manual stage.} In the first stage, resembling classic interactive segmentation, a team of professional annotators labeled masks by clicking foreground / background object points using a browser-based interactive segmentation tool powered by \sam. Masks could be refined using pixel-precise ``brush'' and ``eraser'' tools. Our model-assisted annotation runs in real-time directly inside a browser (using precomputed image embeddings) enabling a truly interactive experience. We did not impose semantic constraints for labeling objects, and annotators freely labeled both ``stuff'' and ``things''~\cite{adelson2001seeing}. We suggested annotators label objects they could name or describe, but did not collect these names or descriptions. Annotators were asked to label objects in order of prominence and were encouraged to proceed to the next image once a mask took over 30 seconds to annotate.

At the start of this stage, \sam was trained using common public segmentation datasets. After sufficient data annotation, \sam was retrained using only newly annotated masks. As more masks were collected, the image encoder was scaled from ViT-B to ViT-H and other architectural details evolved; in total we retrained our model 6 times. Average annotation time per mask decreased from 34 to 14 seconds as the model improved. We note that 14 seconds is 6.5$\x$ faster than mask annotation for COCO~\cite{Lin2014} and only 2$\x$ slower than bounding-box labeling with extreme points~\cite{papadopoulos2017extreme,maninis2018deep}. As \sam improved, the average number of masks per image increased from 20 to 44 masks. Overall, we collected 4.3M masks from 120k images in this stage.

\paragraph{Semi-automatic stage.} In this stage, we aimed to increase the \emph{diversity} of masks in order to improve our model's ability to segment anything. To focus annotators on less prominent objects, we first automatically detected confident masks. Then we presented annotators with images prefilled with these masks and asked them to annotate any additional unannotated objects. To detect confident masks, we trained a bounding box detector~\cite{Ren2015} on all first stage masks using a generic ``object'' category. During this stage we collected an additional 5.9M masks in 180k images (for a total of 10.2M masks). As in the first stage, we periodically retrained our model on newly collected data (5 times). Average annotation time per mask went back up to 34 seconds (excluding the automatic masks) as these objects were more challenging to label. The average number of masks per image went from 44 to 72 masks (including the automatic masks).

\paragraph{Fully automatic stage.} In the final stage, annotation was \emph{fully automatic}. This was feasible due to two major enhancements to our model. First, at the start of this stage, we had collected enough masks to greatly improve the model, including the diverse masks from the previous stage. Second, by this stage we had developed the ambiguity-aware model, which allowed us to predict valid masks even in ambiguous cases. Specifically, we prompted the model with a 32$\x$32 regular grid of points and for each point predicted a set of masks that may correspond to valid objects. With the ambiguity-aware model, if a point lies on a part or subpart, our model will return the subpart, part, and whole object. The IoU prediction module of our model is used to select \emph{confident} masks; moreover, we identified and selected only \emph{stable} masks (we consider a mask stable if thresholding the probability map at $0.5-\delta$ and $0.5+\delta$ results in similar masks). Finally, after selecting the confident and stable masks, we applied non-maximal suppression (NMS) to filter duplicates. To further improve the quality of smaller masks, we also processed multiple overlapping zoomed-in image crops. For further details of this stage, see \S\ref{app:dataset_generation}. We applied fully automatic mask generation to all 11M images in our dataset, producing a total of 1.1B high-quality masks. We describe and analyze the resulting dataset, \sad, next.

%##################################################################################################
\begin{figure}[t]\centering
\includegraphics[width=.99\linewidth]{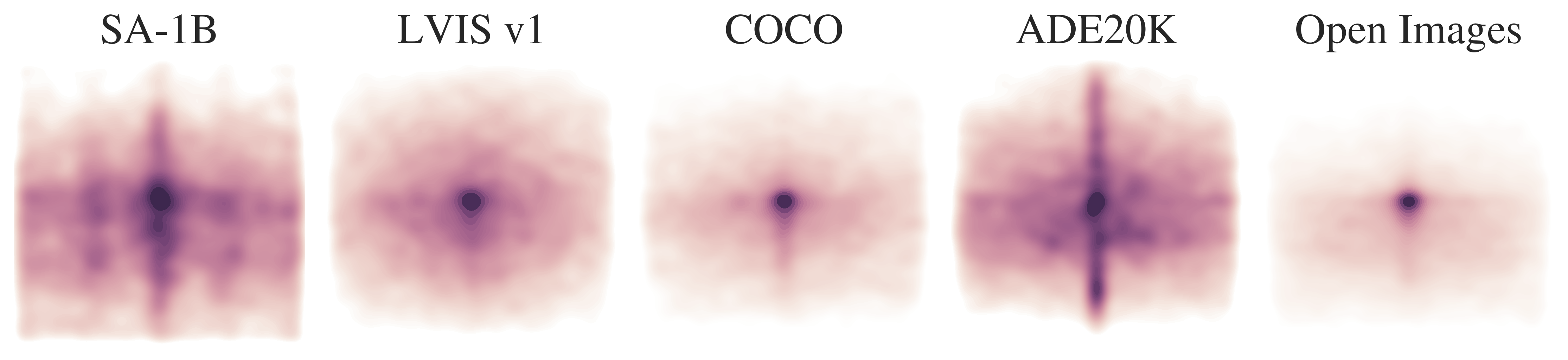}\vspace{-3mm}
\caption{Image-size normalized mask center distributions.}
\label{fig:analysis:center_distribution}\vspace{-2mm}
\end{figure}
%##################################################################################################

%##################################################################################################
\begin{figure*}[t]\centering\vspace{-1mm}
\includegraphics[width=0.88\linewidth]{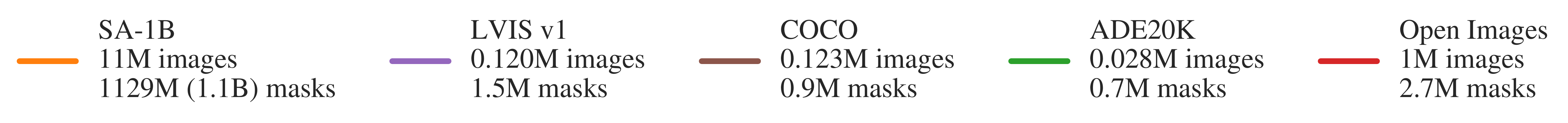}\\[-2mm]
\includegraphics[width=0.32\linewidth]{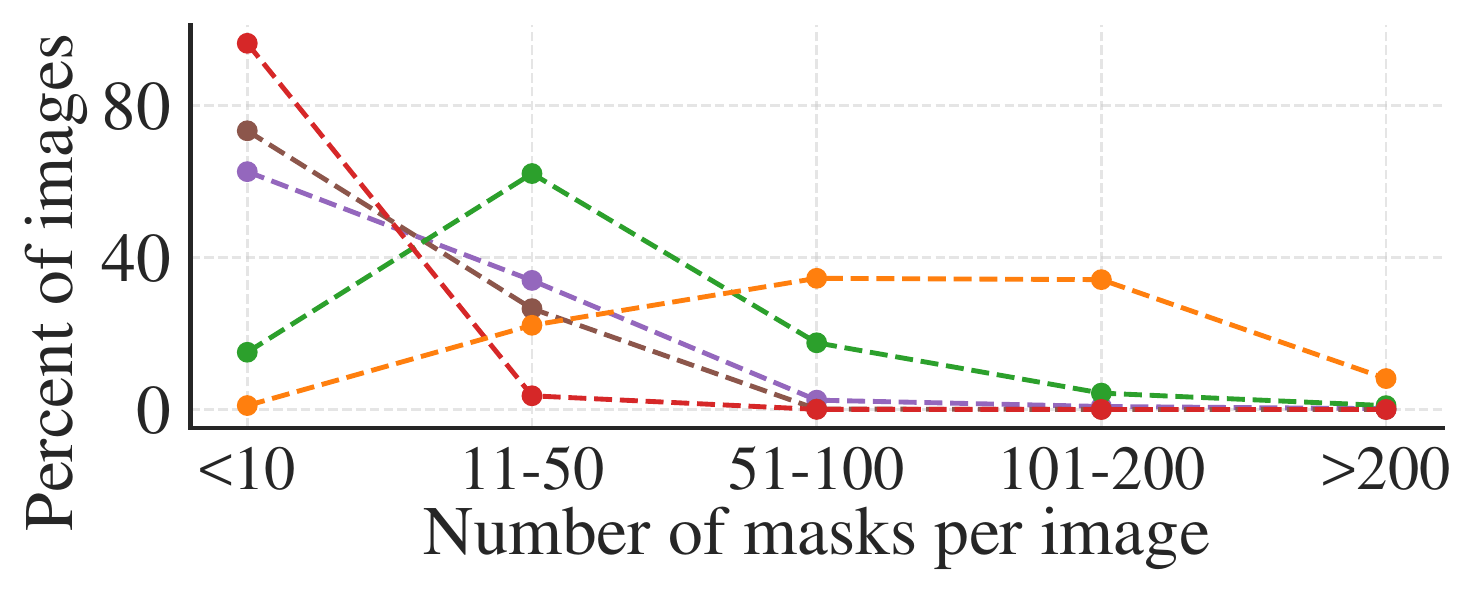}
\includegraphics[width=0.32\linewidth]{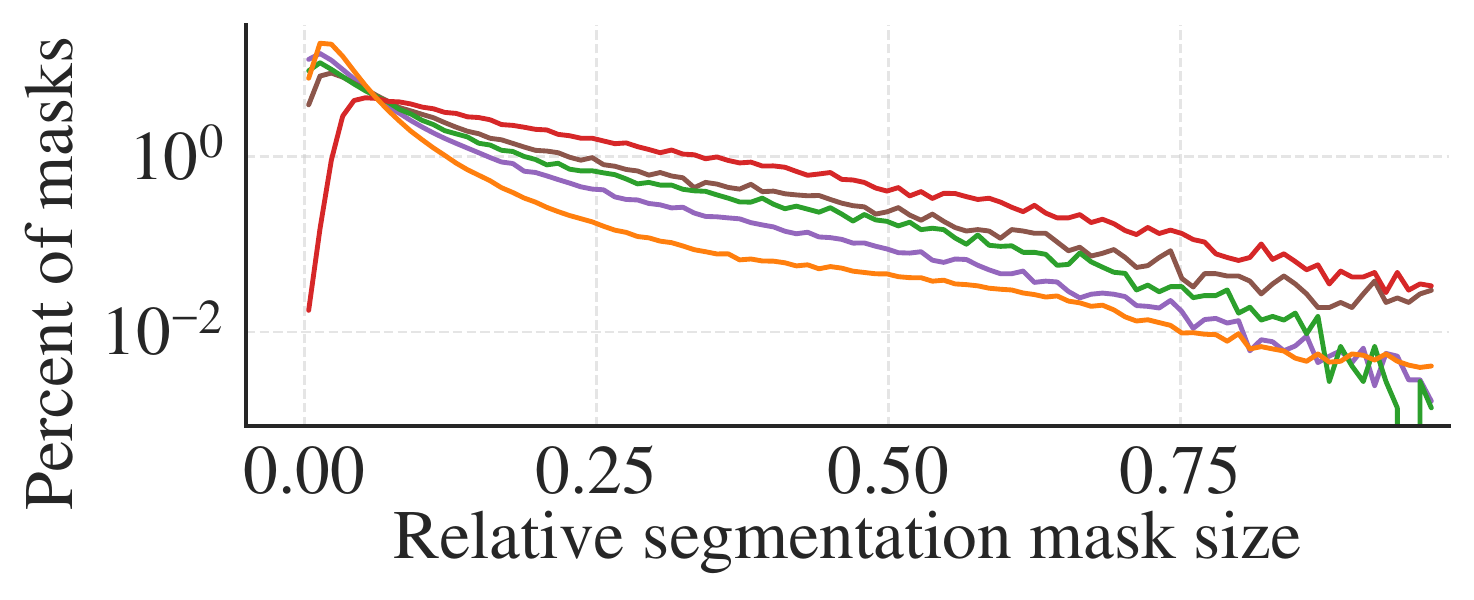}
\includegraphics[width=0.32\linewidth]{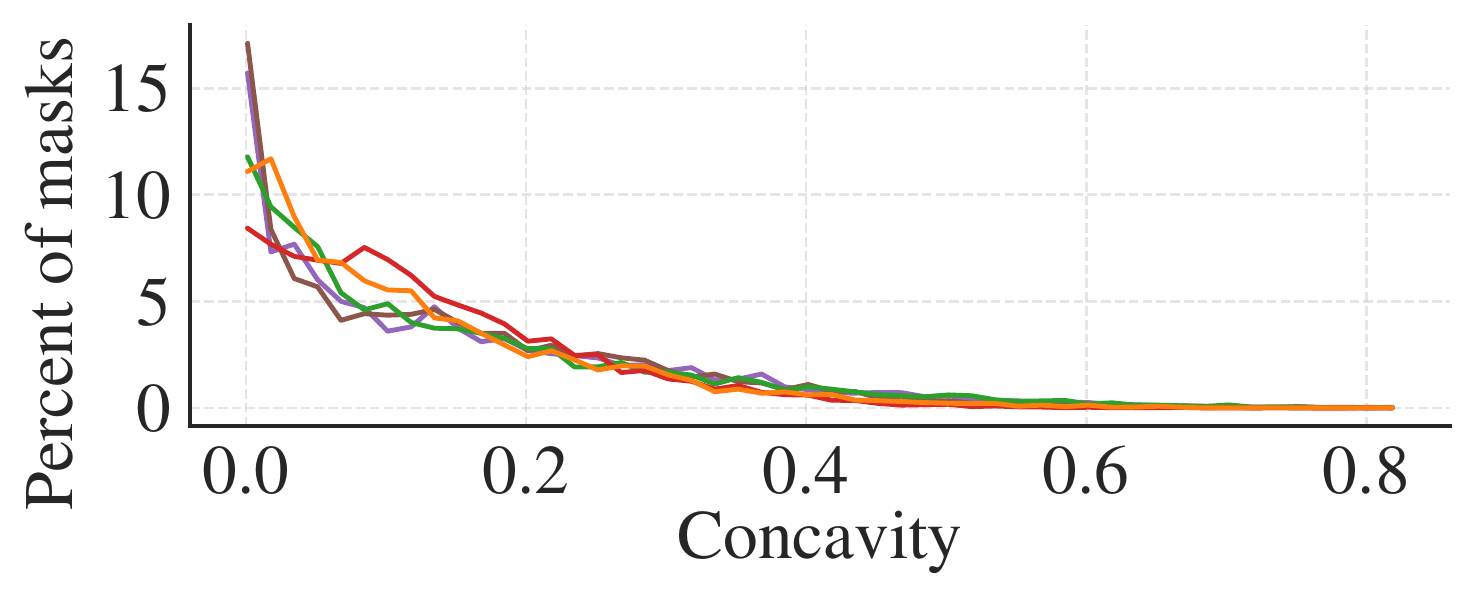}
\vspace{-3mm}
\caption{Dataset mask properties. The legend references the number of images and masks in each dataset. Note, that \sad has 11$\x$ more images and 400$\x$ more masks than the largest existing segmentation dataset Open Images~\cite{OpenImages}.}
\label{fig:analysis}\vspace{-3mm}
\end{figure*}
%##################################################################################################

%%%%%%%%%%%%%%%%%%%%%%%%%%%%%%%%%%%%%%%%%%%%%%%%%%%%%%%%%%%%%%%%%%%%%%%%%%%%%%%%%%%%%%%%%%%%%%%%%%%
\section{Segment Anything Dataset}\label{sec:dataset}

Our dataset, \sad, consists of 11M diverse, high-resolution, licensed, and privacy protecting images and 1.1B high-quality segmentation masks collected with our data engine. We compare \sad with existing datasets and analyze mask quality and properties. We are releasing \sad to aid future development of foundation models for computer vision. We note that \sad will be released under a favorable license agreement for certain research uses and with protections for researchers.

\paragraph{Images}. We licensed a new set of 11M images from a provider that works directly with photographers. These images are high resolution (3300$\x$4950 pixels on average), and the resulting data size can present accessibility and storage challenges. Therefore, we are releasing downsampled images with their shortest side set to 1500 pixels. Even after downsampling, our images are significantly higher resolution than many existing vision datasets (\eg, COCO~\cite{Lin2014} images are \app480$\x$640 pixels). Note that most models today operate on much lower resolution inputs. Faces and vehicle license plates have been blurred in the released images.

\paragraph{Masks}. Our data engine produced 1.1B masks, 99.1\% of which were generated fully automatically. Therefore, the quality of the automatic masks is centrally important. We compare them directly to professional annotations and look at how various mask properties compare to prominent segmentation datasets. Our main conclusion, as borne out in the analysis below and the experiments in \S\ref{sec:eval}, is that our automatic masks are high quality and effective for training models. Motivated by these findings, \sad \emph{only includes automatically generated masks.}

\paragraph{Mask quality.} To estimate mask quality, we randomly sampled 500 images ($\app$50k masks) and asked our professional annotators to improve the quality of all masks in these images. Annotators did so using our model and pixel-precise ``brush'' and ``eraser'' editing tools. This procedure resulted in pairs of automatically predicted and professionally corrected masks. We computed IoU between each pair and found that 94\% of pairs have greater than 90\% IoU (and 97\% of pairs have greater than 75\% IoU). For comparison, prior work estimates inter-annotator consistency at 85-91\% IoU~\cite{Gupta2019,OpenImages}. Our experiments in \S\ref{sec:eval} confirm by human ratings that mask quality is high relative to a variety of datasets and that training our model on automatic masks is nearly as good as using all masks produced by the data engine.

%##################################################################################################
\begin{figure*}[t]\centering
\includegraphics[width=0.44\linewidth]{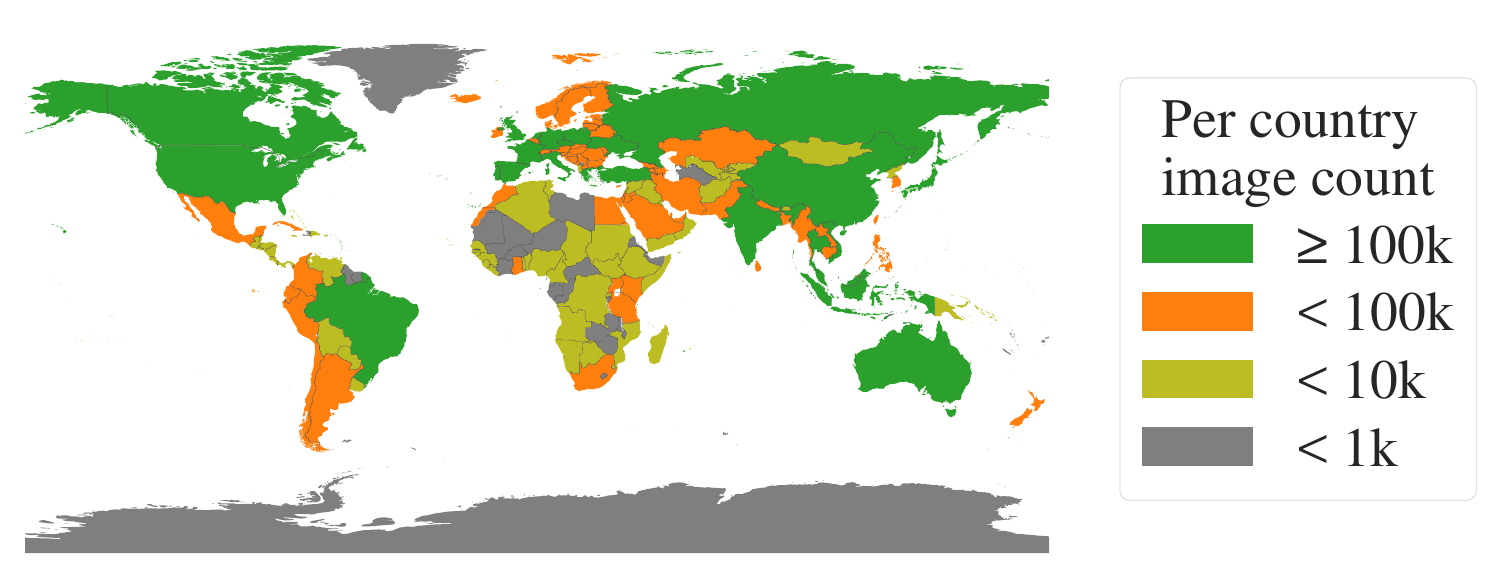}\hfill
\includegraphics[width=0.54\linewidth]{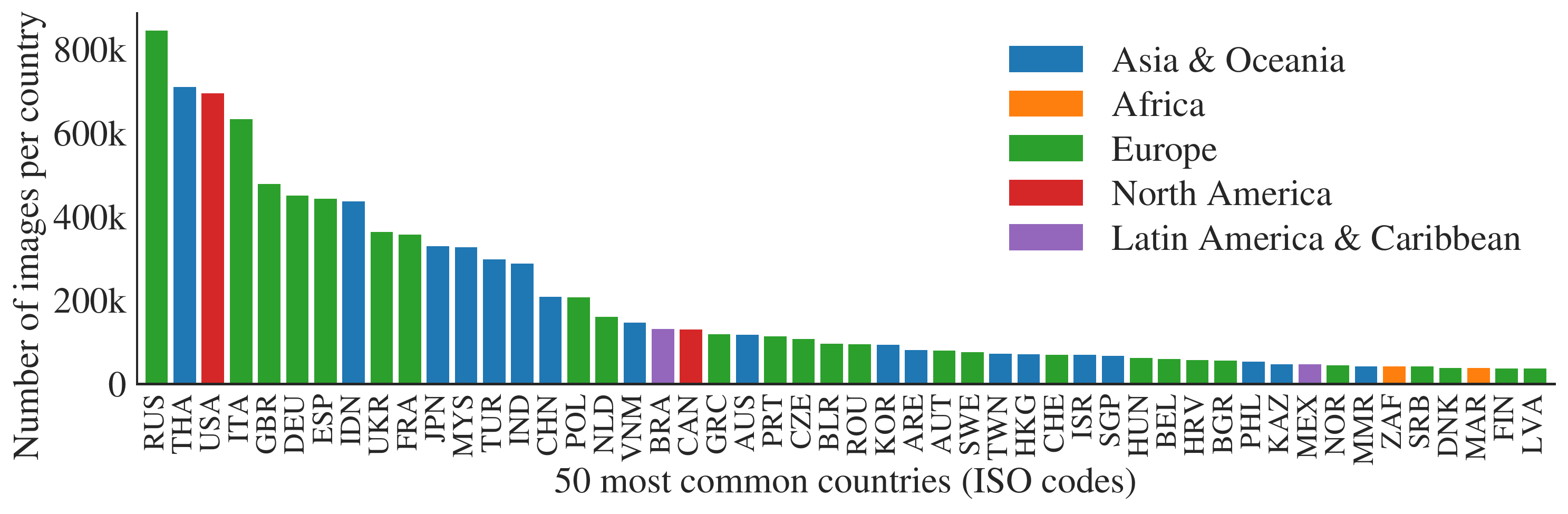}
\vspace{-3mm}
\caption{Estimated geographic distribution of \sad images. Most of the world's countries have more than 1000 images in \sad, and the three countries with the most images are from different parts of the world.}
\label{fig:geo_distribution}\vspace{-4mm}
\end{figure*}
%##################################################################################################

\paragraph{Mask properties.} In \fig{fig:analysis:center_distribution} we plot the spatial distribution of object centers in \sad compared to the largest existing segmentation datasets. Common photographer biases are present in all datasets. We observe that \sad has greater coverage of image corners compared to LVIS v1~\cite{Gupta2019} and ADE20K~\cite{Zhou2019}, the two most similarly distributed datasets, while COCO~\cite{Lin2014} and Open Images V5~\cite{OpenImages} have a more prominent center bias. In \fig{fig:analysis} (legend) we compare these datasets by size. \sad has 11$\x$ more images and 400$\x$ more masks than the second largest, Open Images. On average, it has 36$\x$ more masks per image than Open Images. The closest dataset in this respect, ADE20K, still has 3.5$\x$ fewer masks per image. \fig{fig:analysis} (left) plots the masks-per-image distribution. Next, we look at image-relative mask size (square root of the mask area divided by image area) in \fig{fig:analysis} (middle). As expected, since our dataset has more masks per image, it also tends to include a greater percentage of small and medium relative-size masks. Finally, to analyze shape complexity, we look at mask concavity (1 minus mask area divided by area of mask's convex hull) in \fig{fig:analysis} (right). Since shape complexity is correlated with mask size, we control for the datasets' mask size distributions by first performing stratified sampling from binned mask sizes. We observe that the concavity distribution of our masks is broadly similar to that of other datasets.

%%%%%%%%%%%%%%%%%%%%%%%%%%%%%%%%%%%%%%%%%%%%%%%%%%%%%%%%%%%%%%%%%%%%%%%%%%%%%%%%%%%%%%%%%%%%%%%%%%%
\section{Segment Anything RAI Analysis}\label{sec:rai}

We next perform a Responsible AI (RAI) analysis of our work by investigating potential fairness concerns and biases when using \sad and \sam. We focus on the geographic and income distribution of \sad and fairness of \sam across protected attributes of people. We also provide dataset, data annotation, and model cards in \S\ref{app:cards}.

%##################################################################################################
\begin{table}[t]\centering
\resizebox{\linewidth}{!}{
\tablestyle{3pt}{1.2}\begin{tabular}{@{}lr|rr|rrr@{}}
& & \multicolumn{2}{c|}{\sad} & \multicolumn{3}{c}{\% images} \\
& \hspace{-9mm} \# countries & \#imgs & \#masks & \sad & COCO & O.I. \\
\hline
\scriptsize Africa & 54 & 300k & 28M & 2.8\% & 3.0\% & 1.7\% \\
\scriptsize Asia \& Oceania & 70 & 3.9M & 423M & 36.2\% & 11.4\% & 14.3\% \\
\scriptsize Europe & 47 & 5.4M & 540M & 49.8\% & 34.2\% & 36.2\% \\
\scriptsize Latin America \& Carib. & 42 & 380k & 36M & 3.5\% & 3.1\% & 5.0\% \\
\scriptsize North America & 4 & 830k & 80M & 7.7\% & 48.3\% & 42.8\% \\
\hline
\scriptsize high income countries & 81 & 5.8M & 598M & 54.0\% & 89.1\% & 87.5\% \\
\scriptsize middle income countries & 108 & 4.9M & 499M & 45.0\% & 10.5\% & 12.0\% \\
\scriptsize low income countries & 28 & 100k & 9.4M & 0.9\% & 0.4\% & 0.5\% \\
\end{tabular}}\vspace{-2mm}
\caption{Comparison of geographic and income representation. \sad has higher representation in Europe and Asia \& Oceania as well as middle income countries. Images from Africa, Latin America \& Caribbean, as well as low income countries, are underrepresented in all datasets.}
\label{tab:region}\vspace{0mm}
\end{table}
%##################################################################################################

\paragraph{Geographic and income representation.} We infer the country images were photographed in using standard methods (see \S\ref{app:rai}). In \fig{fig:geo_distribution} we visualize the per-country image counts in \sad (left) and the 50 countries with the most images (right). We note that the top-three countries are from different parts of the world. Next, in Table~\ref{tab:region} we compare the geographic and income representation of \sad, COCO~\cite{Lin2014}, and Open Images~\cite{OpenImages}. \sad has a substantially higher percentage of images in Europe and Asia \& Oceania as well as in middle income countries. All datasets underrepresent Africa as well as low income countries. We note that in \sad, all regions, including Africa, have at least 28 million masks, 10$\x$ more than the \emph{total} number of masks of any previous dataset. Finally, we observe that the average number of masks per image (not shown) is fairly consistent across region and income (94-108 per image).

\paragraph{Fairness in segmenting people.} We investigate potential fairness concerns across perceived gender presentation, perceived age group, and perceived skin tone by measuring the performance discrepancy of \sam between groups. We use the More Inclusive Annotations for People (MIAP)~\cite{schumann2021step} dataset for gender presentation and age and a proprietary dataset for skin tone (see \S\ref{app:rai}). Our evaluation uses simulated interactive segmentation with random sampling of 1 and 3 points (see \S\ref{app:experimental_design}). Table \ref{tab:rai_person} (top left) shows results for perceived gender presentation. We note that females have been shown to be underrepresented in detection and segmentation datasets~\cite{zhao2017}, but observe that \sam performs similarly across groups. We repeat the analysis for perceived age in Table~\ref{tab:rai_person} (bottom left), noting that those who are perceived to be younger and older have been shown to be underrepresented in large-scale datasets~\cite{yang2020towards}. \sam performs best on those who are perceived older (although the confidence interval is large). Finally, we repeat the analysis for perceived skin tone in Table~\ref{tab:rai_person} (right), noting that those with lighter apparent skin tones have been shown to be overrepresented and those with darker skin tones underrepresented in large-scale datasets~\cite{yang2020towards}. As MIAP does not contain perceived skin tone annotations, we use a proprietary dataset that contains annotations for the perceived Fitzpatrick skin type~\cite{fitzpatrick1988}, which ranges from 1 (lightest skin tone) to 6 (darkest skin tone). While the means vary somewhat, we do not find a significant difference across groups. We believe our findings stem from the nature of the task, and acknowledge biases may arise when \sam is used as a component in larger systems. Finally, in \S\ref{app:rai} we extend the analysis to segmenting clothing where we find an indication of bias across perceived gender presentation.

%##################################################################################################
\begin{table}[t]\centering
\resizebox{!}{19mm}{
\tablestyle{4pt}{1.1}\begin{tabular}{@{}lcc@{}}
{} & \multicolumn{2}{c}{mIoU at} \\
{} & 1 point & 3 points \\
\hline
\multicolumn{3}{@{}l}{\emph{perceived gender presentation}} \\
feminine & 54.4\mypm{1.7} & 90.4\mypm{0.6} \\
masculine & 55.7\mypm{1.7} & 90.1\mypm{0.6} \\
\hline
\multicolumn{3}{@{}l}{\emph{perceived age group}} \\
older & 62.9\mypm{6.7} & 92.6\mypm{1.3} \\
middle & 54.5\mypm{1.3} & 90.2\mypm{0.5} \\
young & 54.2\mypm{2.2} & 91.2\mypm{0.7} \\
\end{tabular}}\hspace{5mm}
\resizebox{!}{19mm}{
\tablestyle{4pt}{1.1}\begin{tabular}{@{}lcc@{}}
{} & \multicolumn{2}{c}{mIoU at} \\
{} & 1 point & 3 points \\
\hline
\multicolumn{3}{@{}l}{\emph{perceived skin tone}} \\
1 & 52.9\mypm{2.2} & 91.0\mypm{0.9} \\
2 & 51.5\mypm{1.4} & 91.1\mypm{0.5} \\
3 & 52.2\mypm{1.9} & 91.4\mypm{0.7} \\
4 & 51.5\mypm{2.7} & 91.7\mypm{1.0} \\
5 & 52.4\mypm{4.2} & 92.5\mypm{1.4} \\
6 & 56.7\mypm{6.3} & 91.2\mypm{2.4} \\
\end{tabular}}
\vspace{-2mm}
\caption{\sam's performance segmenting people across perceived gender presentation, age group, and skin tone. 95\% confidence intervals are shown. Within each grouping, all confidence intervals overlap except older \vs middle.}
\label{tab:rai_person}\vspace{-3mm}
\end{table}
%##################################################################################################

%%%%%%%%%%%%%%%%%%%%%%%%%%%%%%%%%%%%%%%%%%%%%%%%%%%%%%%%%%%%%%%%%%%%%%%%%%%%%%%%%%%%%%%%%%%%%%%%%%%
\section{Zero-Shot Transfer Experiments}\label{sec:eval}

In this section, we present \emph{zero-shot transfer} experiments with \sam, the Segment Anything Model. We consider five tasks, four of which differ significantly from the promptable segmentation task used to train \sam. These experiments evaluate \sam on datasets and tasks that were not seen during training (our usage of ``zero-shot transfer'' follows its usage in CLIP~\cite{Radford2021}). The datasets may include novel image distributions, such as underwater or ego-centric images (\eg \fig{fig:benchmark_examples}) that, to our knowledge, do not appear in \sad.

Our experiments begin by testing the core goal of promptable segmentation: producing a valid mask from any prompt. We emphasize the challenging scenario of a \emph{single} foreground point prompt, since it is more likely to be ambiguous than other more specific prompts. Next, we present a sequence of experiments that traverse low, mid, and high-level image understanding and roughly parallel the historical development of the field. Specifically, we prompt \sam to (1) perform edge detection, (2) segment everything, \ie object proposal generation, (3) segment detected objects, \ie instance segmentation, and (4), as a proof-of-concept, to segment objects from free-form text. These four tasks differ significantly from the promptable segmentation task that \sam was trained on and are implemented via prompt engineering. Our experiments conclude with an ablation study.

\paragraph{Implementation.} Unless otherwise specified: (1) \sam uses an MAE~\cite{he2022masked} pre-trained ViT-H~\cite{Dosovitskiy2021} image encoder and (2) \sam was trained on \sad, noting that this dataset includes only automatically generated masks from the final stage of our data engine. For all other model and training details, such as hyperparameters, refer to \S\ref{app:model}.

\subsection{Zero-Shot Single Point Valid Mask Evaluation}\label{sec:eval:single_point}\vspace{-2mm}

\paragraph{Task.} We evaluate segmenting an object from a \emph{single} foreground point. This task is ill-posed as one point can refer to multiple objects. Ground truth masks in most datasets do not enumerate \emph{all} possible masks, which can make automatic metrics unreliable. Therefore, we supplement the standard mIoU metric (\ie, the mean of all IoUs between predicted and ground truth masks) with a human study in which annotators rate mask quality from 1 (nonsense) to 10 (pixel-perfect). See \S\ref{app:benchmark}, \S\ref{app:human_study}, and \S\ref{app:annotation_guidelines} for additional details.

By default, we sample points from the ``center'' of ground truth masks (at a maximal value of the mask's interior distance transform), following the standard evaluation protocol in interactive segmentation~\cite{sofiiuk2022reviving}. Since \sam is capable of predicting multiple masks, we evaluate only the model's most confident mask by default. The baselines are all single-mask methods. We compare mainly to RITM~\cite{sofiiuk2022reviving}, a strong interactive segmenter that performs best on our benchmark compared to other strong baselines~\cite{liu2022simpleclick,chen2022focalclick}.

%##################################################################################################
\begin{figure*}[t]\centering
\tablestyle{0.9pt}{0.6}\fontsize{6}{7}\selectfont
\begin{tabular}{cccccccc}
ADE20K~\cite{Zhou2019} & BBBC038v1~\cite{cells} & Cityscapes~\cite{Cordts2016} & DOORS~\cite{doors} & DRAM~\cite{DRAM} & EgoHOS~\cite{egoHos} & GTEA~\cite{gtea1,gtea2} & Hypersim~\cite{hypersim} \\
\includegraphics[height=1.43cm]{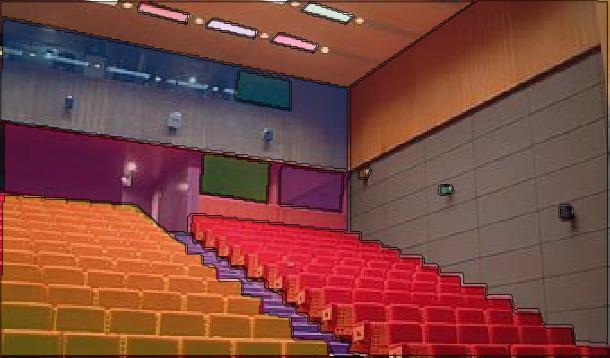} &
\includegraphics[height=1.43cm]{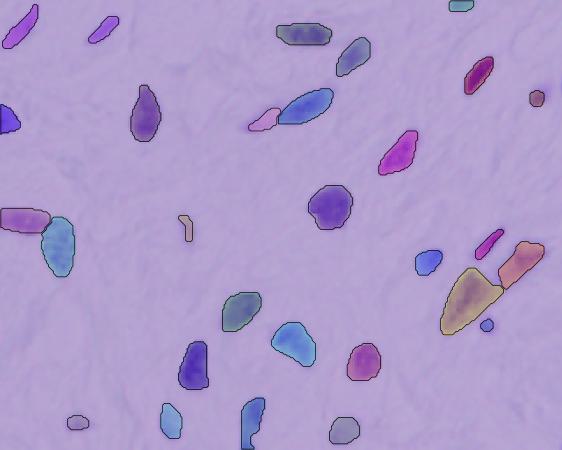} &
\includegraphics[height=1.43cm]{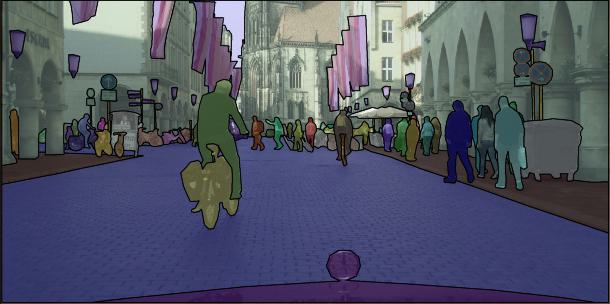} &
\includegraphics[height=1.43cm]{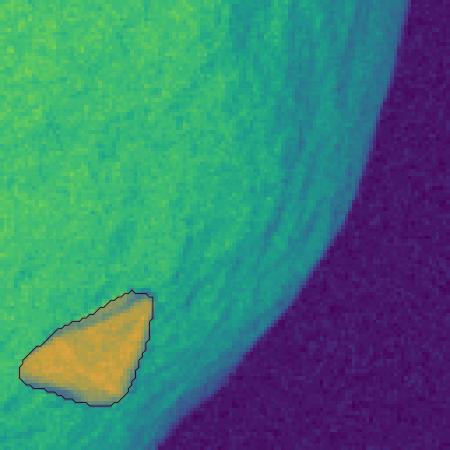} &
\includegraphics[height=1.43cm]{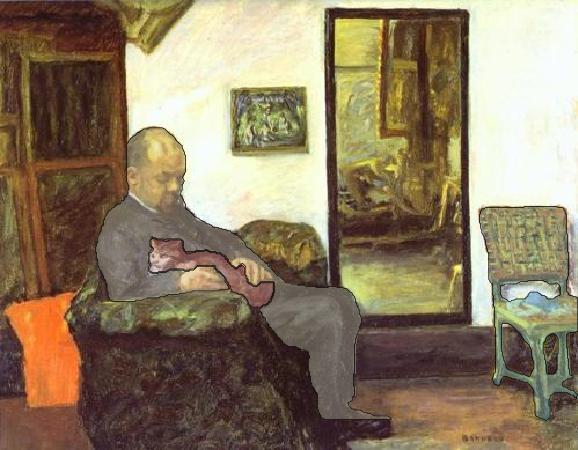} &
\includegraphics[height=1.43cm]{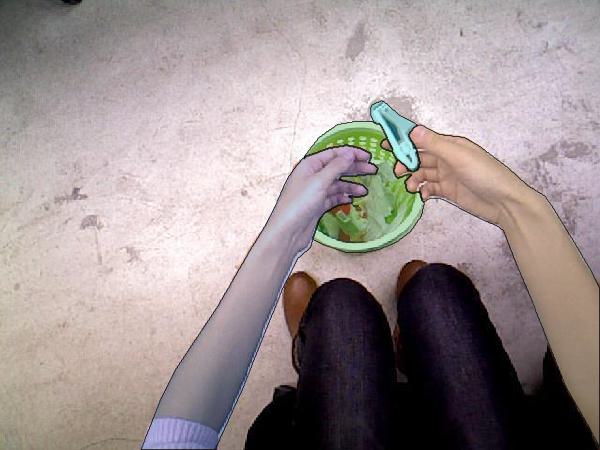} &
\includegraphics[height=1.43cm]{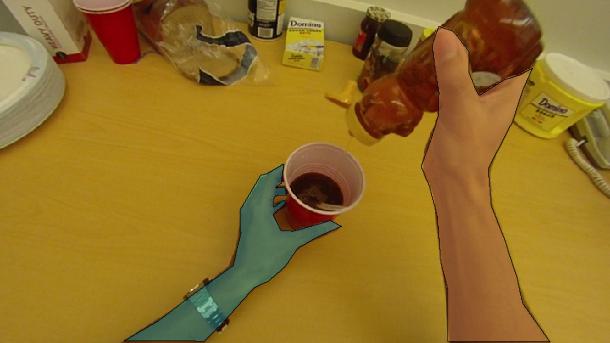} &
\includegraphics[height=1.43cm]{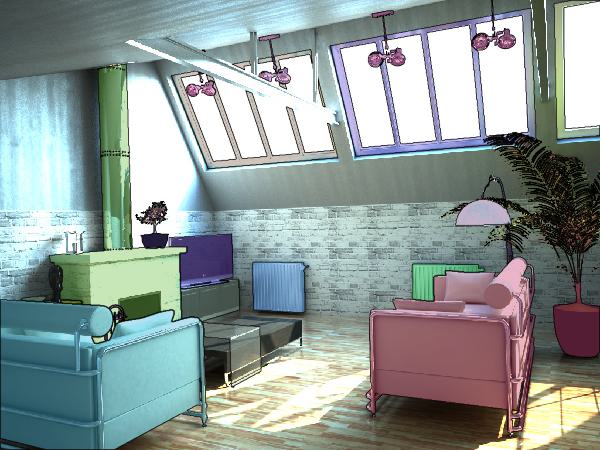}
\end{tabular}\\[0mm]
\tablestyle{1pt}{0.6}\fontsize{6}{7}\selectfont
\begin{tabular}{cccccccc}
IBD~\cite{chen20223D} & iShape~\cite{iShape} & LVIS~\cite{Gupta2019} & NDD20~\cite{ndd20} & NDISPark~\cite{ndis1,ndis2} & OVIS~\cite{ovis} & PPDLS~\cite{plants} & Plittersdorf~\cite{haucke2022socrates}\\
\includegraphics[height=1.64cm]{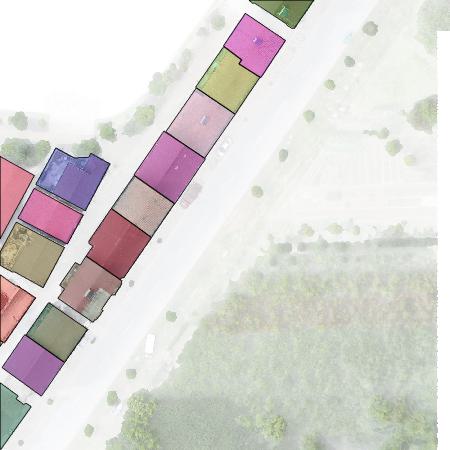} &
\includegraphics[height=1.64cm]{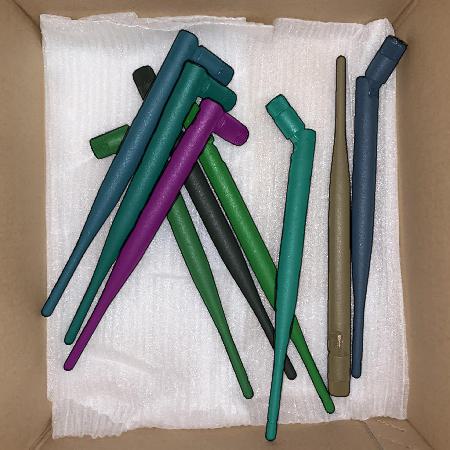} &
\includegraphics[height=1.64cm]{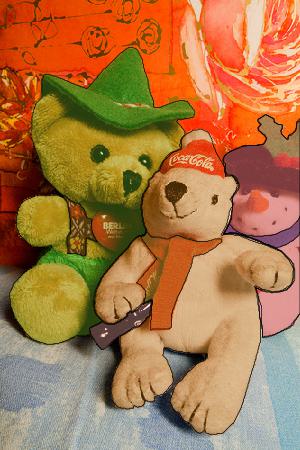} &
\includegraphics[height=1.64cm]{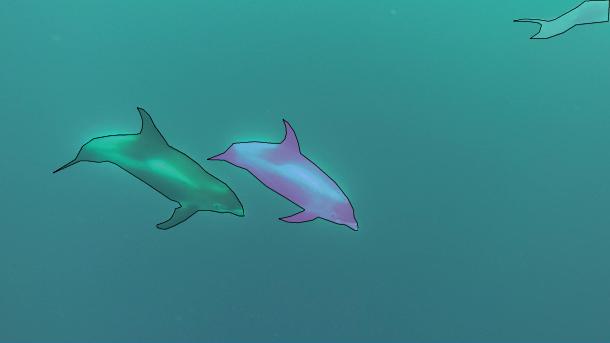} &
\includegraphics[height=1.64cm]{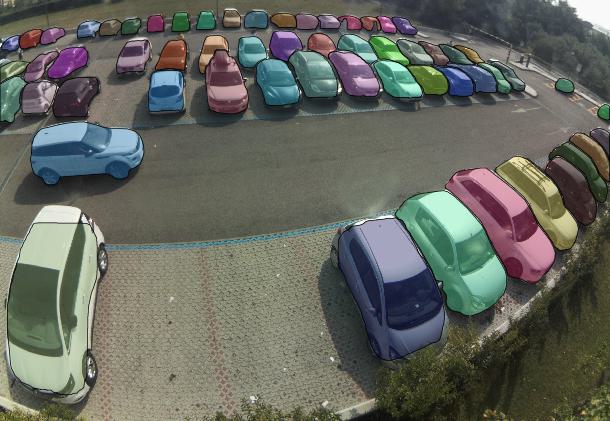} &
\includegraphics[height=1.64cm]{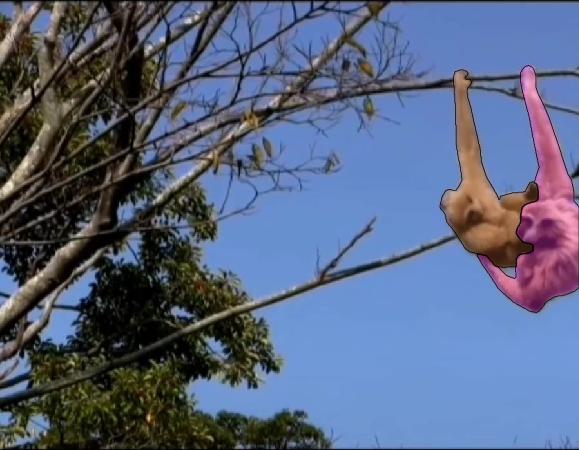} &
\includegraphics[height=1.64cm]{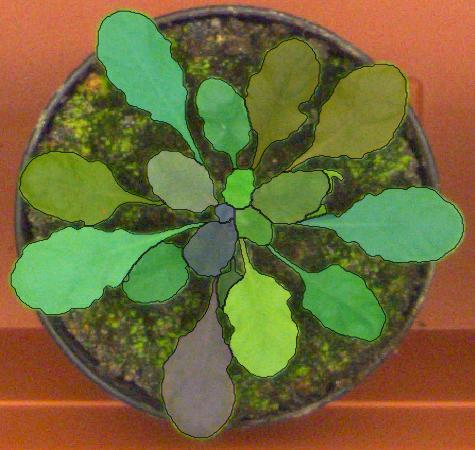} &
\includegraphics[height=1.64cm]{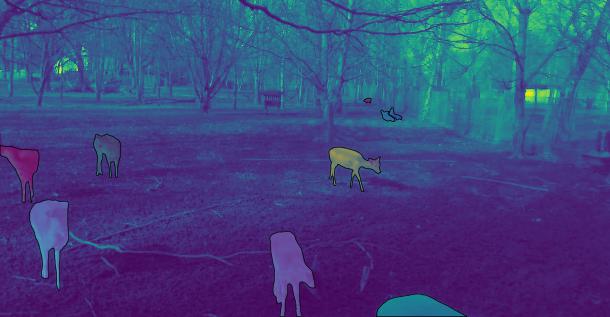}
\end{tabular}\\[0mm]
\tablestyle{1pt}{0.6}\fontsize{6}{7}\selectfont
\begin{tabular}{ccccccc}
STREETS~\cite{streets} & TimberSeg~\cite{timberSeg} & TrashCan~\cite{hong2020trashcan} & VISOR~\cite{VISOR,EpicKitchens} & WoodScape~\cite{yogamani2019woodscape} & PIDRay~\cite{wang2021towards} & ZeroWaste-f~\cite{zerowaste} \\
\includegraphics[height=1.57cm]{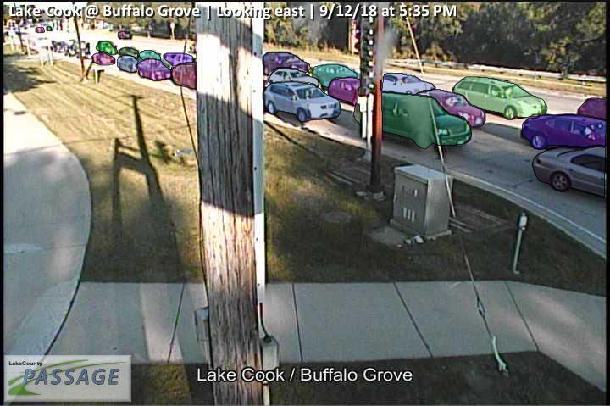} &
\includegraphics[height=1.57cm]{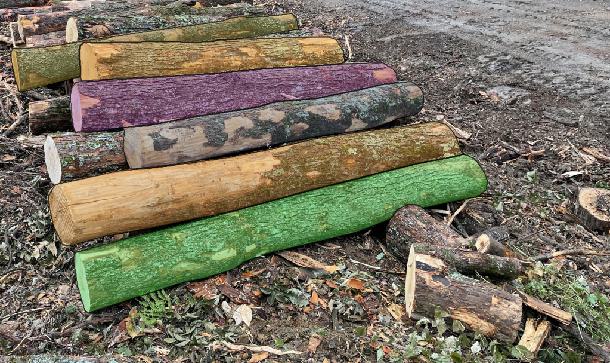} &
\includegraphics[height=1.57cm]{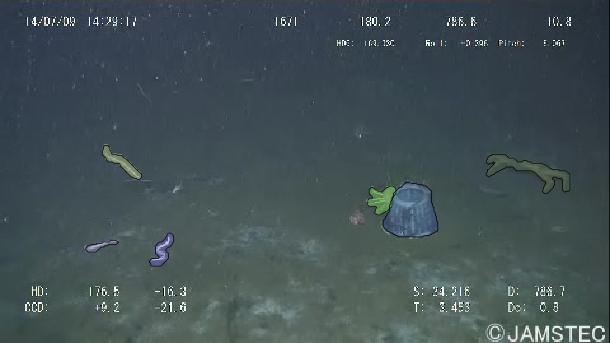} &
\includegraphics[height=1.57cm]{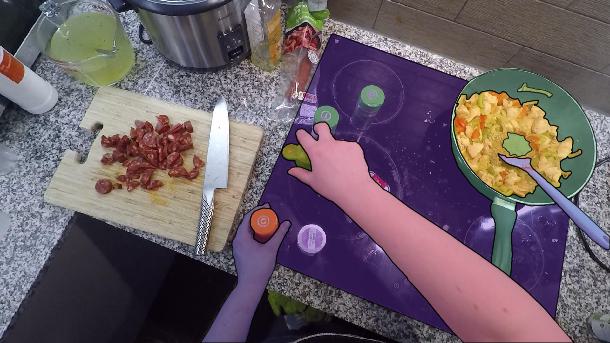} &
\includegraphics[height=1.57cm]{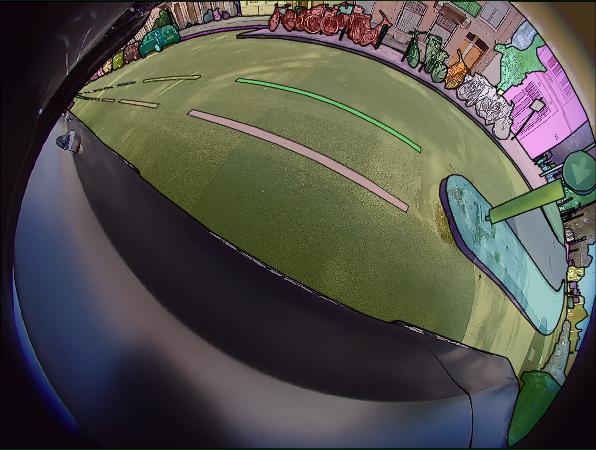} &
\includegraphics[height=1.57cm]{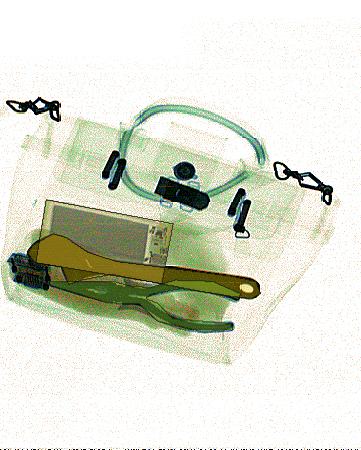} &
\includegraphics[height=1.57cm]{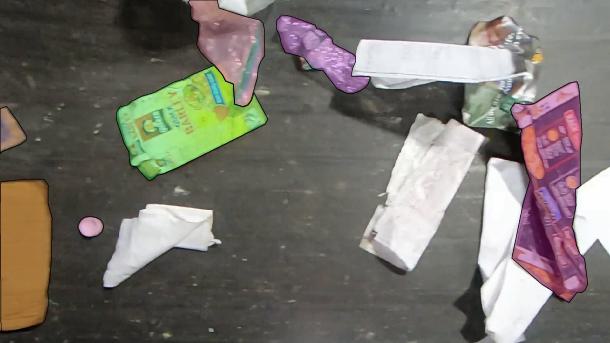}
\end{tabular}\vspace{-2mm}
\caption{Samples from the 23 diverse segmentation datasets used to evaluate \sam's zero-shot transfer capabilities.}
\label{fig:benchmark_examples}
\end{figure*}
%##################################################################################################

\begin{figure*}[t]\centering
\begin{tabular}[b]{@{}c@{}}
\resizebox{.49\linewidth}{!}{\input{figs/becnhmark_ritm_comparison.pgf}}\\[-1mm]
\footnotesize (a) \sam \vs RITM~\cite{sofiiuk2022reviving} on 23 datasets
\end{tabular}\hfill
\begin{tabular}[b]{@{}c@{}}
\includegraphics[width=.49\linewidth]{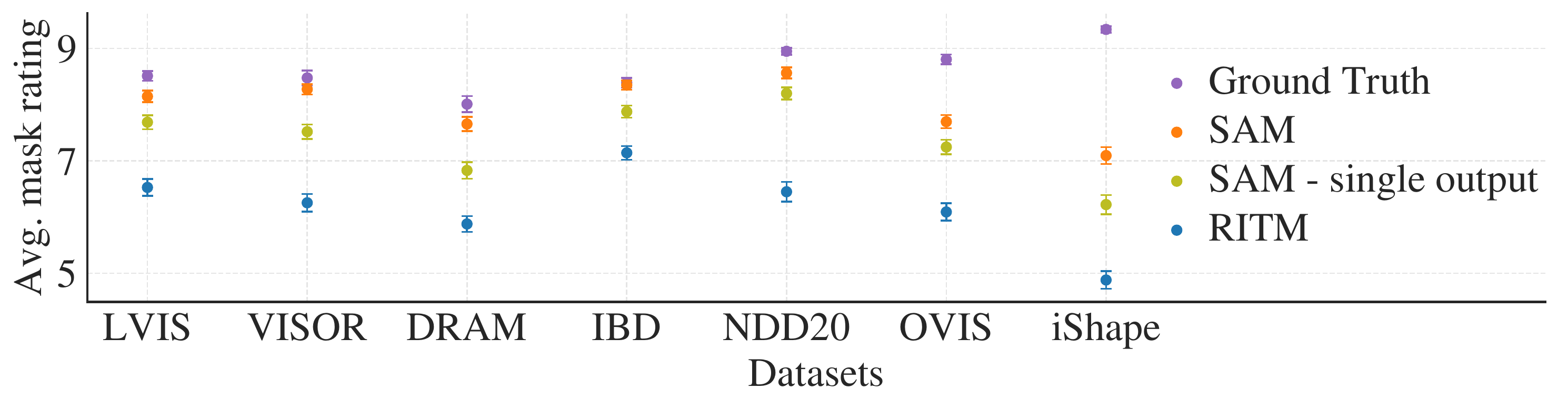}\\[-2mm]
\footnotesize (b) Mask quality ratings by human annotators\\[2mm]
\begin{tabular}[b]{@{}c@{}c@{}}
\includegraphics[width=0.24\linewidth,trim={0 2mm 0 0},clip]{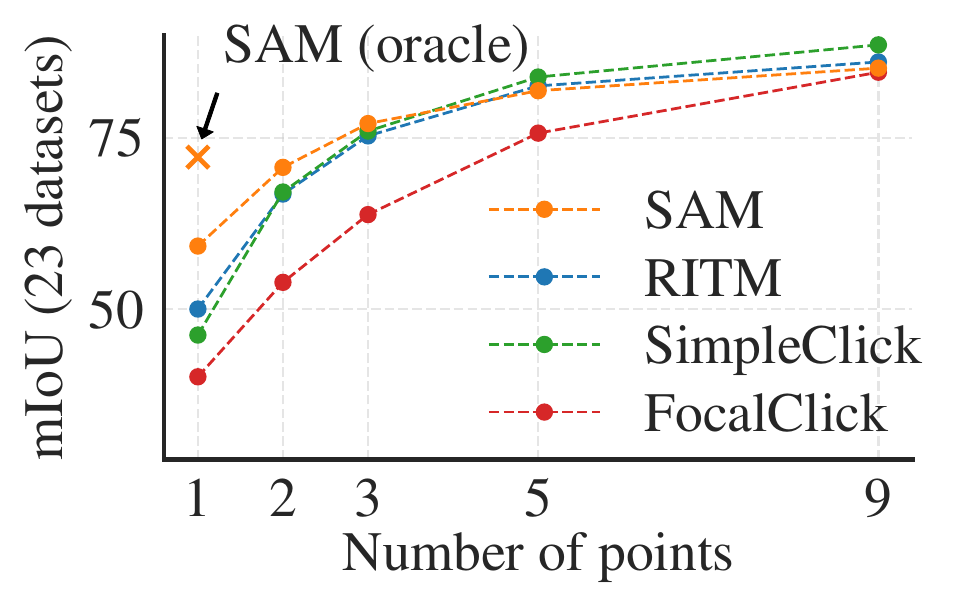} &
\includegraphics[width=0.24\linewidth,trim={0 2mm 0 0},clip]{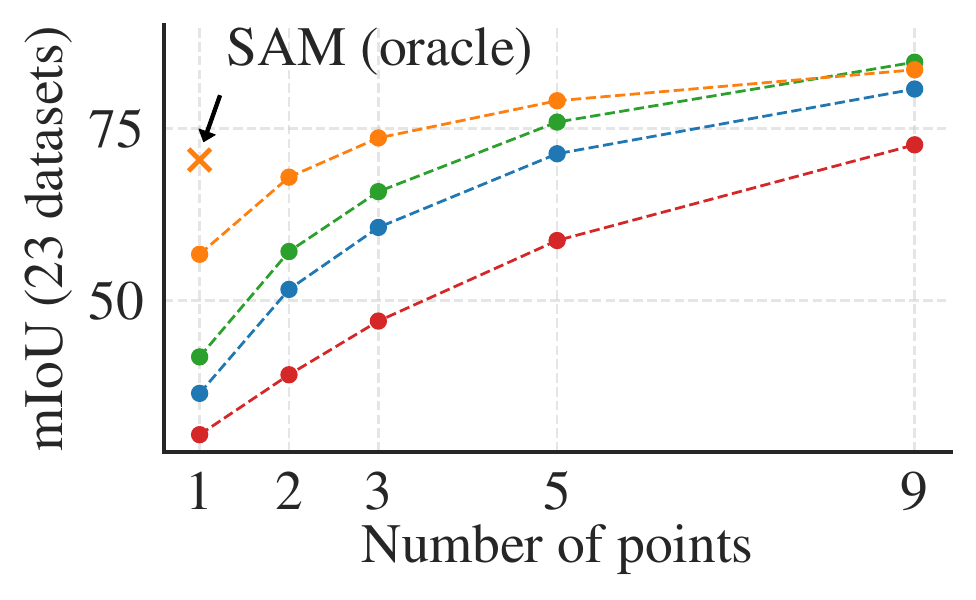}\\[-1mm]
\footnotesize (c) Center points (default) & \footnotesize (d) Random points
\end{tabular}
\end{tabular}\vspace{-2mm}
\caption{Point to mask evaluation on 23 datasets. (a) Mean IoU of \sam and the strongest single point segmenter, RITM~\cite{sofiiuk2022reviving}. Due to ambiguity, a single mask may not match ground truth; circles show ``oracle'' results of the most relevant of \sam's 3 predictions. (b) Per-dataset comparison of mask quality ratings by annotators from 1 (worst) to 10 (best). All methods use the ground truth mask center as the prompt. (c, d) mIoU with varying number of points. \sam significantly outperforms prior interactive segmenters with 1 point and is on par with more points. Low absolute mIoU at 1 point is the result of ambiguity.}
\label{fig:benchmark_exps}\vspace{-2mm}
\end{figure*}

\paragraph{Datasets.} We use a newly compiled suite of 23 datasets with diverse image distributions. \fig{fig:benchmark_examples} lists the datasets and shows a sample from each one (see appendix Table~\ref{app:tab:datasets_all} for more details). We use all 23 datasets for mIoU evaluation. For the human study, we use the subset listed in \fig{fig:benchmark_exps}{\color{linkcolor}b} (due to the resource requirements of such studies). This subset includes both datasets for which \sam outperforms and underperforms RITM according to automatic metrics.

%##################################################################################################
%##################################################################################################

\paragraph{Results.} First, we look at automatic evaluation on the full suite of 23 datasets using mIoU. We compare per-dataset results in \fig{fig:benchmark_exps}{\color{linkcolor}a} against RITM. \sam yields higher results on 16 of the 23 datasets, by as much as $\app$47 IoU. We also present an ``oracle'' result, in which the most relevant of \sam's 3 masks is selected by comparing them to the ground truth, rather than selecting the most confident mask. This reveals the impact of ambiguity on automatic evaluation. In particular, with the oracle to perform ambiguity resolution, \sam outperforms RITM on \emph{all} datasets.

Results of the human study are presented in \fig{fig:benchmark_exps}{\color{linkcolor}b}. Error bars are 95\% confidence intervals for mean mask ratings (all differences are significant; see \S\ref{app:human_study} for details). We observe that the annotators consistently rate the quality of \sam's masks substantially higher than the strongest baseline, RITM. An ablated, ``ambiguity-unaware'' version of \sam with a single output mask has consistently lower ratings, though still higher than RITM. \sam's mean ratings fall between 7 and 9, which corresponds to the qualitative rating guideline: ``\emph{A high score (7-9): The object is identifiable and errors are small and rare (\eg, missing a small, heavily obscured disconnected component, ...).}'' These results indicate that \sam has learned to segment valid masks from a single point. Note that for datasets like DRAM and IBD, where \sam is worse on automatic metrics, \emph{it receives consistently higher ratings in the human study}.

\fig{fig:benchmark_exps}{\color{linkcolor}c} shows additional baselines, SimpleClick~\cite{liu2022simpleclick} and FocalClick~\cite{chen2022focalclick}, which obtain lower single point performance than RITM and \sam. As the number of points increases from 1 to 9, we observe that the gap between methods decreases. This is expected as the task becomes easier; also, \sam is not optimized for the very high IoU regime. Finally, in \fig{fig:benchmark_exps}{\color{linkcolor}d} we replace the default center point sampling with random point sampling. We observe that the gap between \sam and the baselines grows and \sam is able to achieve comparable results under either sampling method.

\subsection{Zero-Shot Edge Detection}\label{sec:eval:edge}

\paragraph{Approach.} We evaluate \sam on the classic low-level task of edge detection using BSDS500~\cite{martin2001database,arbelaez2010contour}. We use a simplified version of our automatic mask generation pipeline. Specifically, we prompt \sam with a 16$\x$16 regular grid of foreground points resulting in 768 predicted masks (3 per point). Redundant masks are removed by NMS. Then, edge maps are computed using Sobel filtering of unthresholded mask probability maps and standard lightweight postprocessing, including edge NMS (see \S\ref{app:edges} for details).

%##################################################################################################
\begin{figure}[t]\centering
\tablestyle{1pt}{0.8}\begin{tabular}{ccc}
image & ground truth & \sam \\
\includegraphics[width=0.325\linewidth]{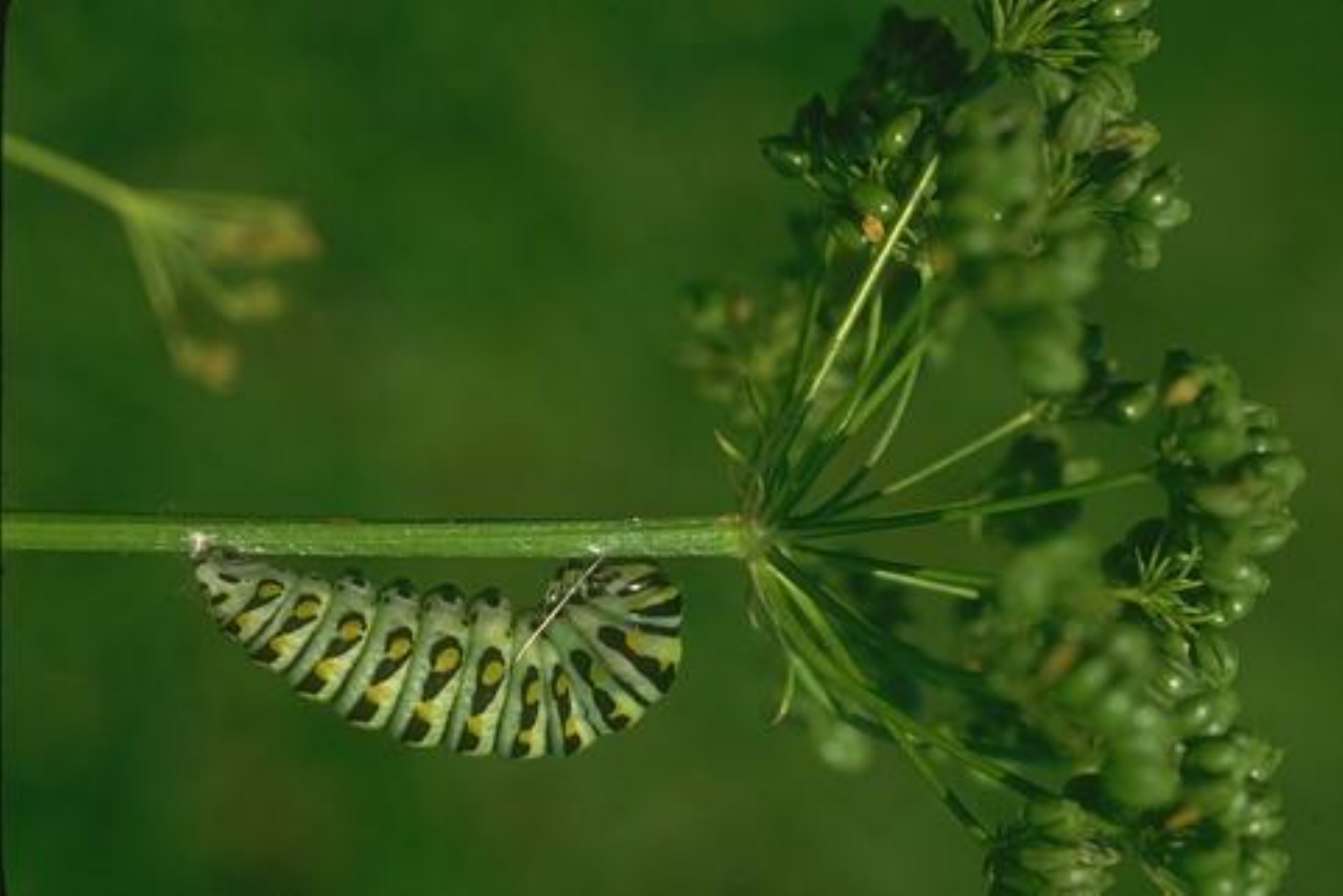} &
\includegraphics[width=0.325\linewidth]{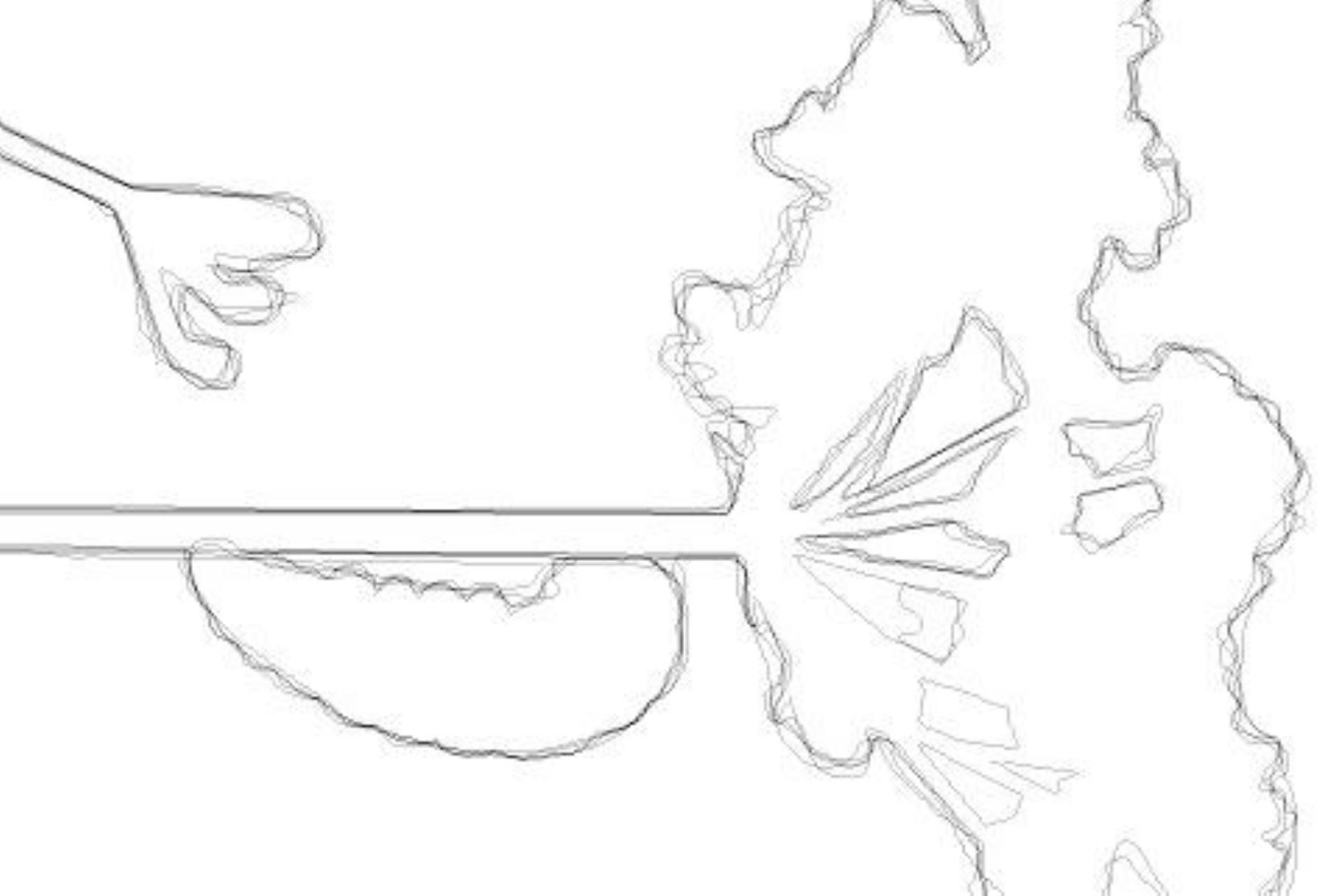} &
\includegraphics[width=0.325\linewidth]{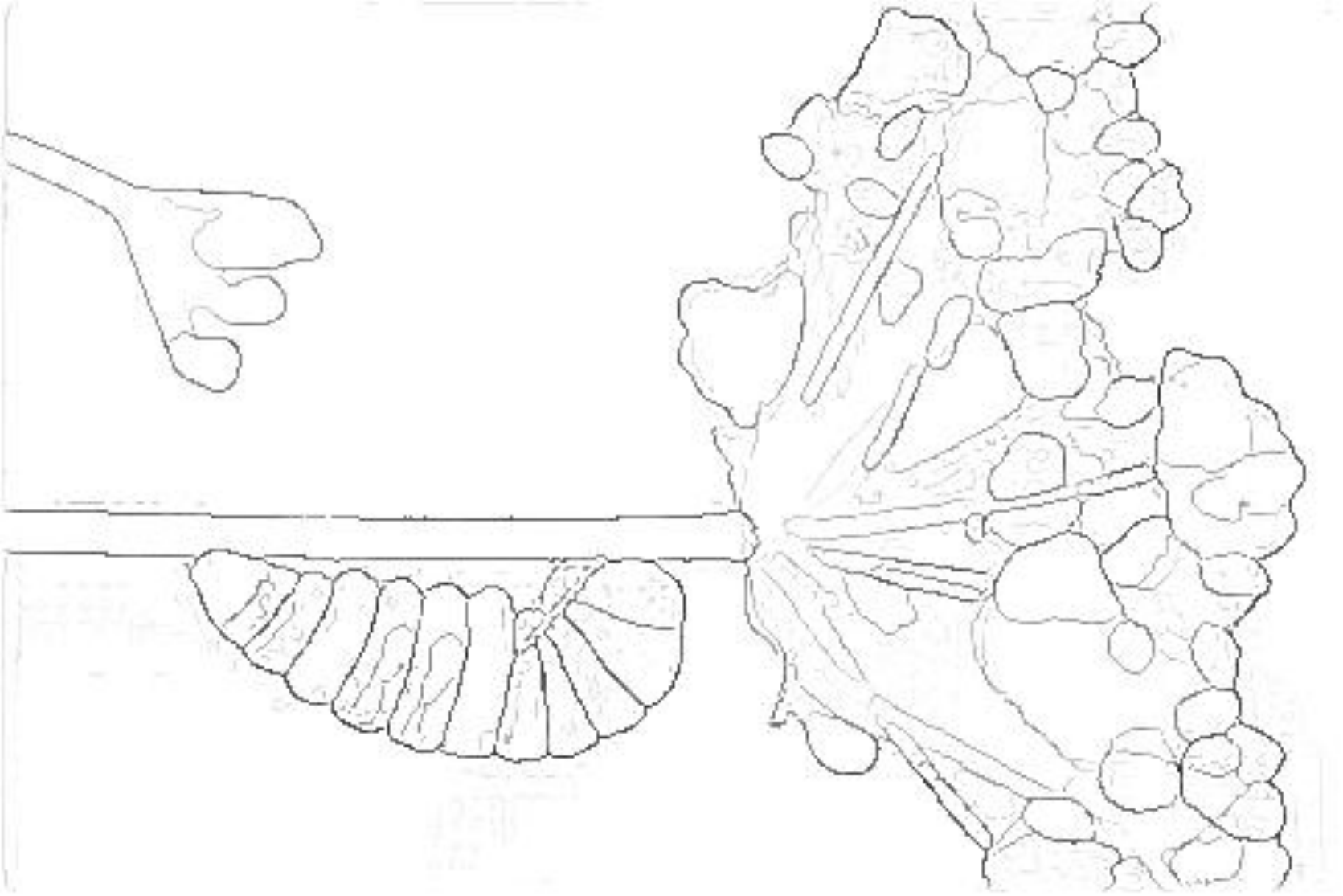} \\
\includegraphics[width=0.325\linewidth]{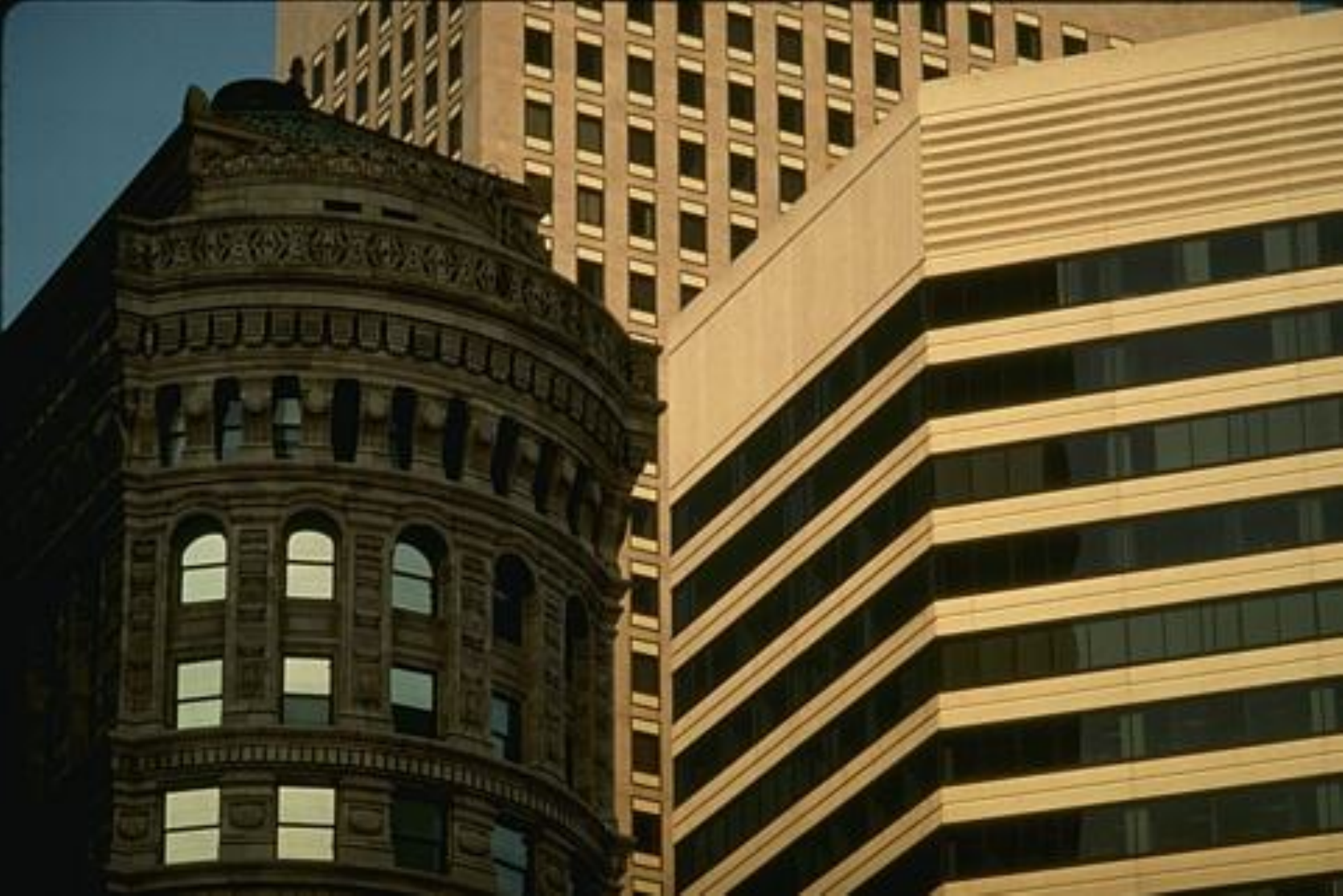} &
\includegraphics[width=0.325\linewidth]{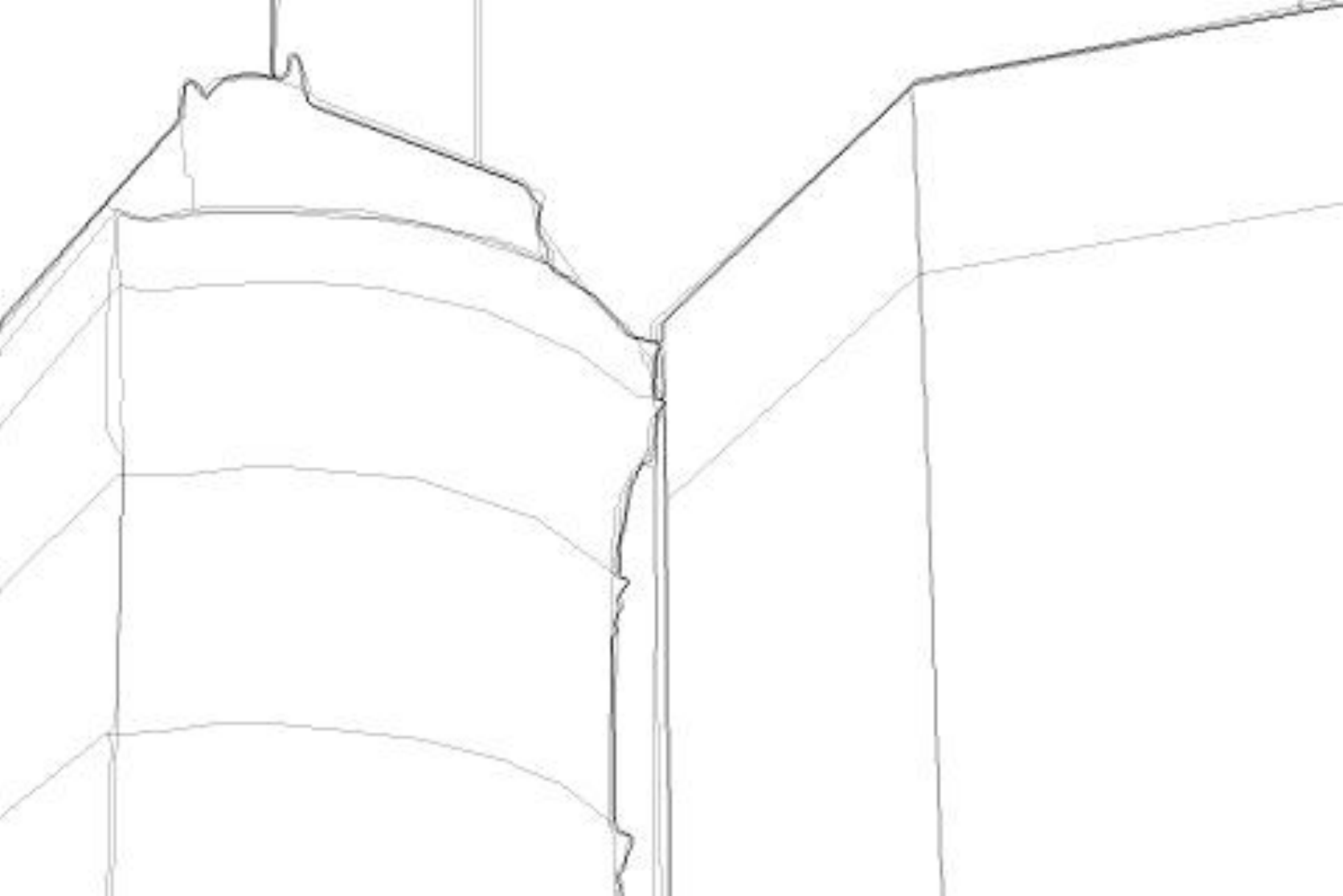} &
\includegraphics[width=0.325\linewidth]{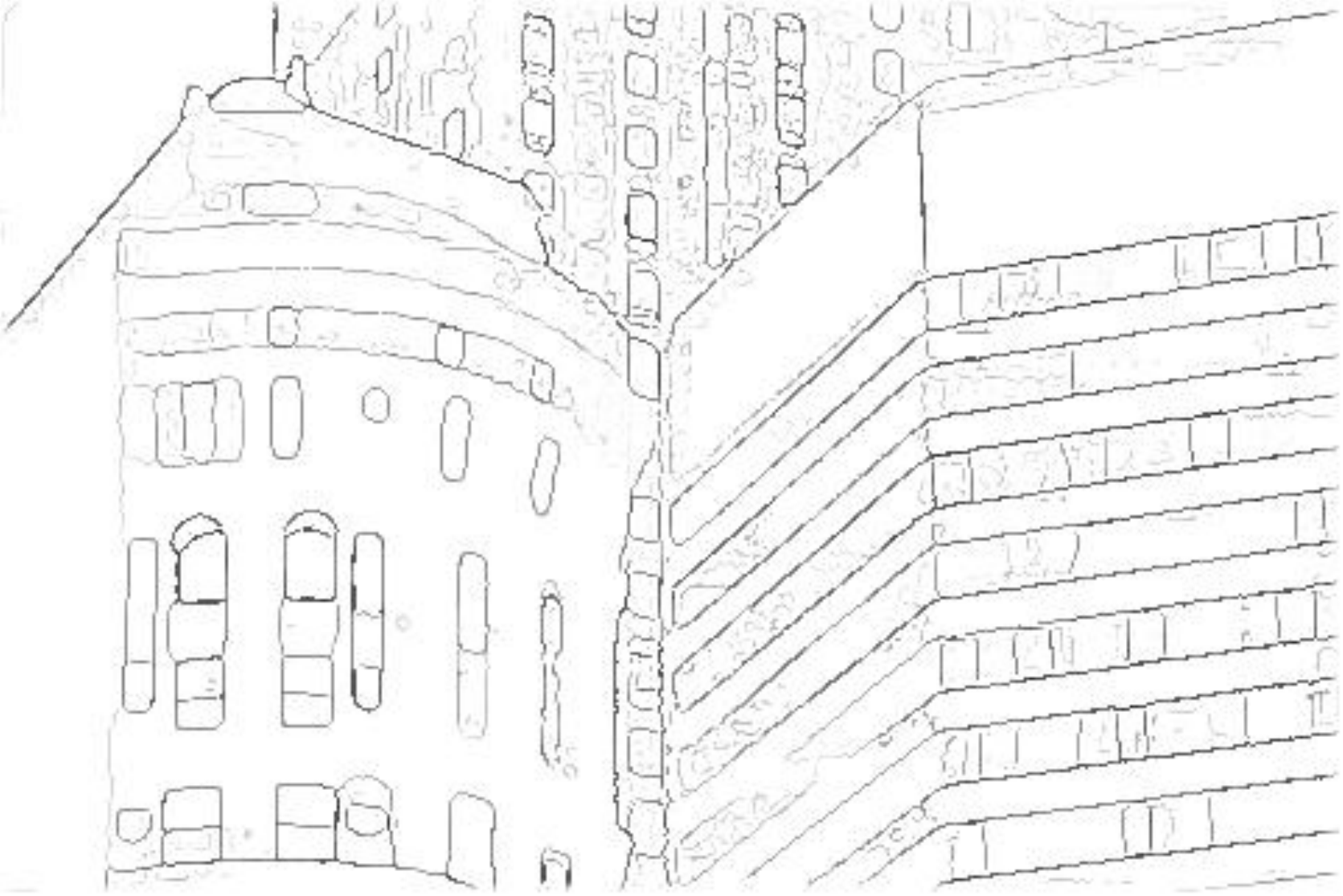} \\
\end{tabular}\vspace{-3mm}
\caption{Zero-shot edge prediction on BSDS500. \sam was not trained to predict edge maps nor did it have access to BSDS images or annotations during training.}
\label{fig:edges}
\end{figure}
%##################################################################################################

%##################################################################################################
\begin{table}[t]
\centering
\tablestyle{9pt}{1.1}
\footnotesize
\begin{tabular}{@{}lc|x{20}x{20}x{20}|x{9}}
method & year & ODS & OIS & AP & R50 \\
\hline
HED~\cite{xie2015holistically} & 2015& .788 & .808 & .840 & .923 \\
EDETR~\cite{pu2022edter} & 2022 & .840 & .858 & .896 & .930 \\
\multicolumn{6}{@{}l}{\emph{zero-shot transfer methods:}} \\
Sobel filter & 1968 & .539 & - & -& - \\
Canny~\cite{canny1986computational} & 1986 & .600 & .640 & .580 & - \\
Felz-Hutt~\cite{felzenszwalb2004efficient} & 2004 & .610 & .640 & .560 & - \\
\sam & 2023 &.768 &.786 & .794 & .928 \\
\end{tabular}\vspace{-2mm}
\caption{Zero-shot transfer to edge detection on BSDS500.}
\label{tab:edges}
\end{table}
%##################################################################################################

\paragraph{Results.} We visualize representative edge maps in \fig{fig:edges} (see \fig{fig:more_edges} for more). Qualitatively, we observe that even though \sam was not trained for edge detection, it produces reasonable edge maps. Compared to the ground truth, \sam predicts more edges, including sensible ones that are not annotated in BSDS500. This bias is reflected quantitatively in Table~\ref{tab:edges}: recall at 50\% precision (R50) is high, at the cost of precision. \sam naturally lags behind state-of-the-art methods that learn the biases of BSDS500, \ie, which edges to suppress. Nevertheless, \sam performs well compared to pioneering deep learning methods such as HED~\cite{xie2015holistically} (also trained on BSDS500) and significantly better than prior, though admittedly outdated, zero-shot transfer methods.

\subsection{Zero-Shot Object Proposals}\label{subsec:proposals}

\paragraph{Approach.} Next, we evaluate \sam on the mid-level task of object proposal generation~\cite{alexe2010object,Sande2011}. This task has played an important role in object detection research, serving as an intermediate step in pioneering systems (\eg,~\cite{Sande2011,Girshick2014,Ren2015}). To generate object proposals, we run a slightly modified version of our automatic mask generation pipeline and output the masks as proposals (see \S\ref{app:proposals} for details).

We compute the standard average recall (AR) metric on LVIS v1~\cite{Gupta2019}. We focus on LVIS because its large number of categories presents a challenging test. We compare to a \emph{strong} baseline implemented as a ViTDet~\cite{li2022exploring} detector (with cascade Mask R-CNN~\cite{He2017,Cai2018} ViT-H). We note that this ``baseline'' corresponds to the ``Detector Masquerading as Proposal generator'' (DMP) method~\cite{chavali2016object} that was shown to game AR, making it a truly demanding comparison.

%##################################################################################################
\begin{table}[t]
\centering
\tablestyle{2.8pt}{1.1}
\footnotesize
\begin{tabular}{@{}lx{20}|x{20}x{20}x{20}|x{20}x{20}x{12}}
\multirow{2}{*}{} & \multicolumn{7}{c}{mask AR@1000}\\
method & all & small & med. & large & freq. & com. & rare \\
\hline
ViTDet-H~\cite{li2022exploring} & 63.0 & 51.7 & 80.8 & 87.0 & 63.1 & 63.3 & 58.3 \\
\multicolumn{8}{@{}l}{\emph{zero-shot transfer methods:}} \\
\sam\ -- single out. & 54.9 & 42.8 & 76.7 & 74.4 & 54.7 & 59.8 & 62.0 \\
\sam & 59.3 & 45.5 & 81.6 & 86.9 & 59.1 & 63.9 & 65.8 \\
\end{tabular}
\vspace{-2mm}
\caption{Object proposal generation on LVIS v1. \sam is applied zero-shot, \ie it was not trained for object proposal generation nor did it access LVIS images or annotations.}
\label{tab:proposals}
\end{table}
%##################################################################################################

\paragraph{Results.} In Table~\ref{tab:proposals} we see unsurprisingly that using the detections from ViTDet-H as object proposals (\ie, the DMP method~\cite{chavali2016object} that games AR) performs the best overall. However, \sam does remarkably well on several metrics. Notably, it outperforms ViTDet-H on medium and large objects, as well as rare and common objects. In fact, \sam only underperforms ViTDet-H on small objects and frequent objects, where ViTDet-H can easily learn LVIS-specific annotation biases since it was trained on LVIS, unlike \sam. We also compare against an ablated ambiguity-unaware version of \sam (``single out.''), which performs significantly worse than \sam on all AR metrics.

\subsection{Zero-Shot Instance Segmentation}\label{sec:eval:instseg}

\paragraph{Approach.} Moving to higher-level vision, we use \sam as the segmentation module of an instance segmenter. The implementation is simple: we run a object detector (the ViTDet used before) and prompt \sam with its output boxes. This illustrates \emph{composing} \sam in a larger system.

\begin{figure}[t]\centering
\includegraphics[width=.99\linewidth]{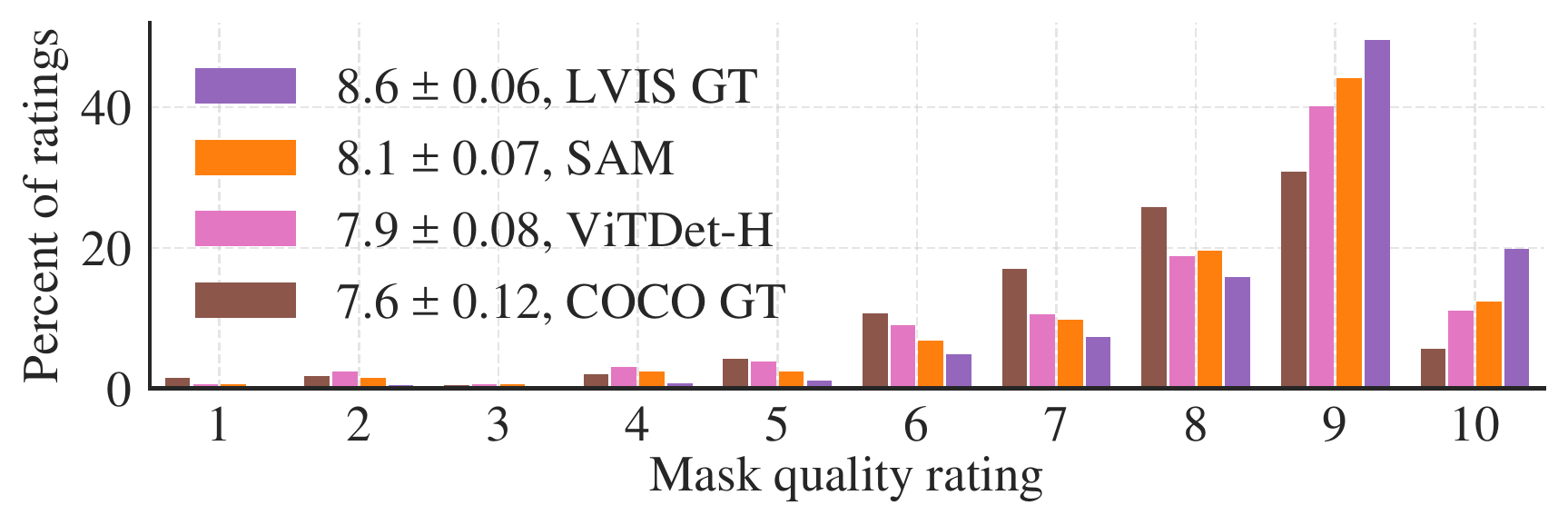}
\vspace{-3mm}
\caption{Mask quality rating distribution from our human study for ViTDet and \sam, both applied to LVIS ground truth boxes. We also report LVIS and COCO ground truth quality. The legend shows rating means and 95\% confidence intervals. Despite its lower AP (Table~\ref{tab:instance_segmentation}), \sam has higher ratings than ViTDet, suggesting that ViTDet exploits biases in the COCO and LVIS training data.}
\label{fig:humanstudy:inst}\vspace{-3mm}
\end{figure}

\paragraph{Results.} We compare the masks predicted by \sam and ViTDet on COCO and LVIS in Table~\ref{tab:instance_segmentation}. Looking at the mask AP metric we observe gaps on both datasets, where \sam is reasonably close, though certainly behind ViTDet. By visualizing outputs, we observed that \sam masks are often qualitatively better than those of ViTDet, with crisper boundaries (see \S\ref{app:instseg} and \fig{fig:instanceseg}). To investigate this observation, we conducted an additional human study asking annotators to rate the ViTDet masks and \sam masks on the 1 to 10 quality scale used before. In \fig{fig:humanstudy:inst} we observe that \sam consistently outperforms ViTDet in the human study.

%##################################################################################################
\begin{table}[t]
\centering
\tablestyle{2.2pt}{1.1}
\footnotesize
\begin{tabular}{@{}lx{19}x{19}x{19}x{19}|x{19}x{19}x{19}x{12}}
 \multirow{2}{*}{} & \multicolumn{4}{c}{COCO~\cite{Lin2014}} & \multicolumn{4}{c}{LVIS v1~\cite{Gupta2019}}\\
method & AP & AP\textsuperscript{S} & AP\textsuperscript{M} & AP\textsuperscript{L} & AP & AP\textsuperscript{S} & AP\textsuperscript{M} & AP\textsuperscript{L} \\
\hline
ViTDet-H~\cite{li2022exploring} & 51.0 & 32.0 & 54.3 & 68.9 & 46.6 & 35.0 & 58.0 & 66.3\\
\multicolumn{9}{@{}l}{\emph{zero-shot transfer methods (segmentation module only):}} \\
\sam & 46.5 & 30.8 & 51.0 & 61.7 & 44.7 & 32.5 & 57.6 & 65.5\\
\end{tabular}
\vspace{-2mm}
\caption{Instance segmentation results. \sam is prompted with ViTDet boxes to do zero-shot segmentation. The fully-supervised ViTDet outperforms \sam, but the gap shrinks on the higher-quality LVIS masks. Interestingly, \sam outperforms ViTDet according to human ratings (see \fig{fig:humanstudy:inst}).}
\label{tab:instance_segmentation}\vspace{-1mm}
\end{table}
%##################################################################################################

%##################################################################################################
%##################################################################################################

We hypothesize that on COCO, where the mask AP gap is larger and the ground truth quality is relatively low (as borne out by the human study), ViTDet learns the specific biases of COCO masks. \sam, being a zero-shot method, is unable to exploit these (generally undesirable) biases. The LVIS dataset has higher quality ground truth, but there are still specific idiosyncrasies (\eg, masks do not contain holes, they are simple polygons by construction) and biases for modal \vs amodal masks. Again, \sam is not trained to learn these biases, while ViTDet can exploit them.

\subsection{Zero-Shot Text-to-Mask}\label{sec:eval:text_to_mask}

\paragraph{Approach.} Finally, we consider an even higher-level task: segmenting objects from free-form text. This experiment is a proof-of-concept of \sam's ability to process text prompts. While we used the exact same \sam in all prior experiments, for this one \sam's training procedure is modified to make it text-aware, but in a way that does not require new text annotations. Specifically, for each manually collected mask with area larger than $\textrm{100}^\textrm{2}$ we extract the CLIP \emph{image} embedding. Then, during training, we prompt \sam with the extracted CLIP image embeddings as its first interaction. The key observation here is that because CLIP's \emph{image} embeddings are trained to align with its \emph{text} embeddings, we can train with image embeddings, but use text embeddings for inference. That is, at inference time we run text through CLIP's text encoder and then give the resulting text embedding as a prompt to \sam (see \S\ref{app:text_to_mask} for details).

%##################################################################################################
\begin{figure}[t]\centering
\tablestyle{1pt}{0.8}
\begin{tabular}{c c}\begin{tikzpicture}
\draw (0, 0) node[inner sep=0] {\includegraphics[width=0.49\linewidth]{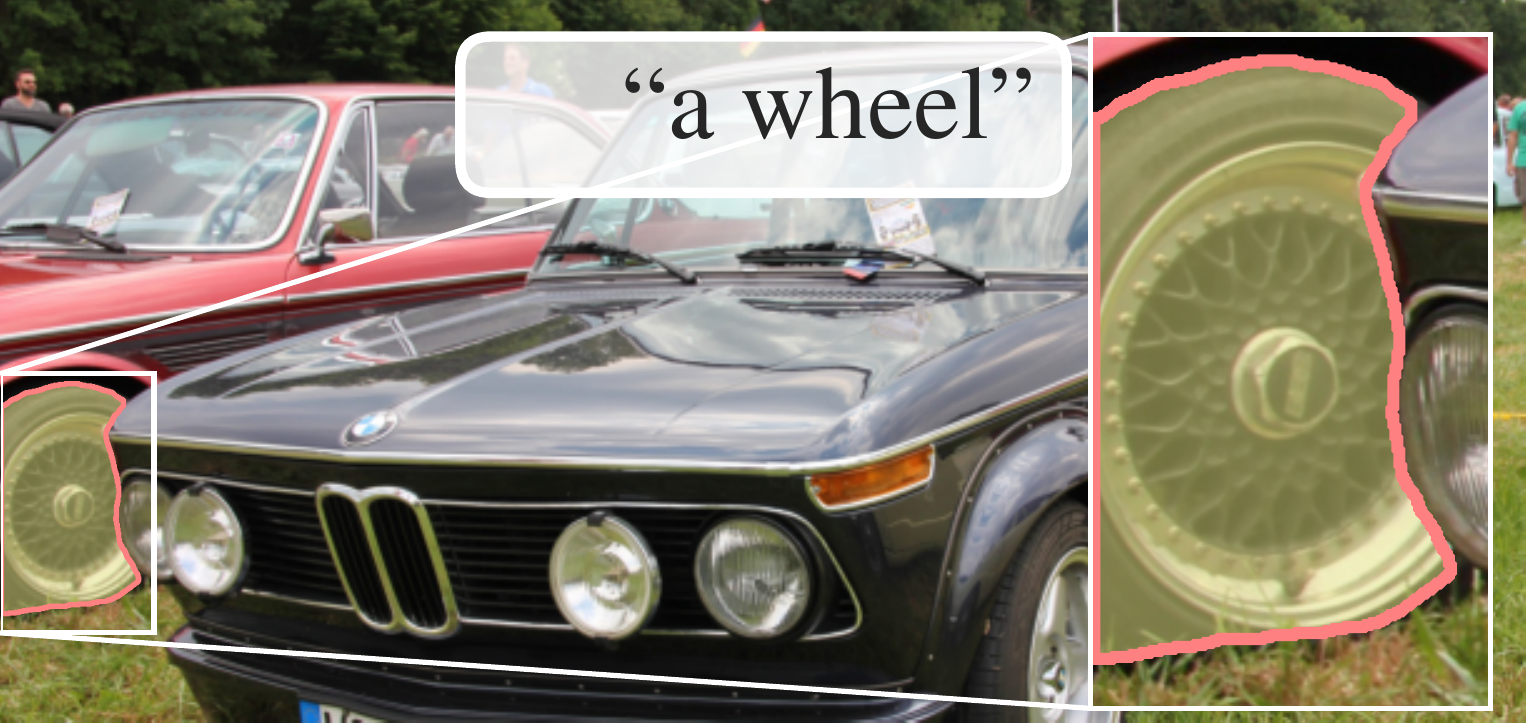}};
\draw (-0.6, 0.68) node {\cmark};
\end{tikzpicture} &
\begin{tikzpicture}
\draw (0, 0) node[inner sep=0] {\includegraphics[width=0.49\linewidth]{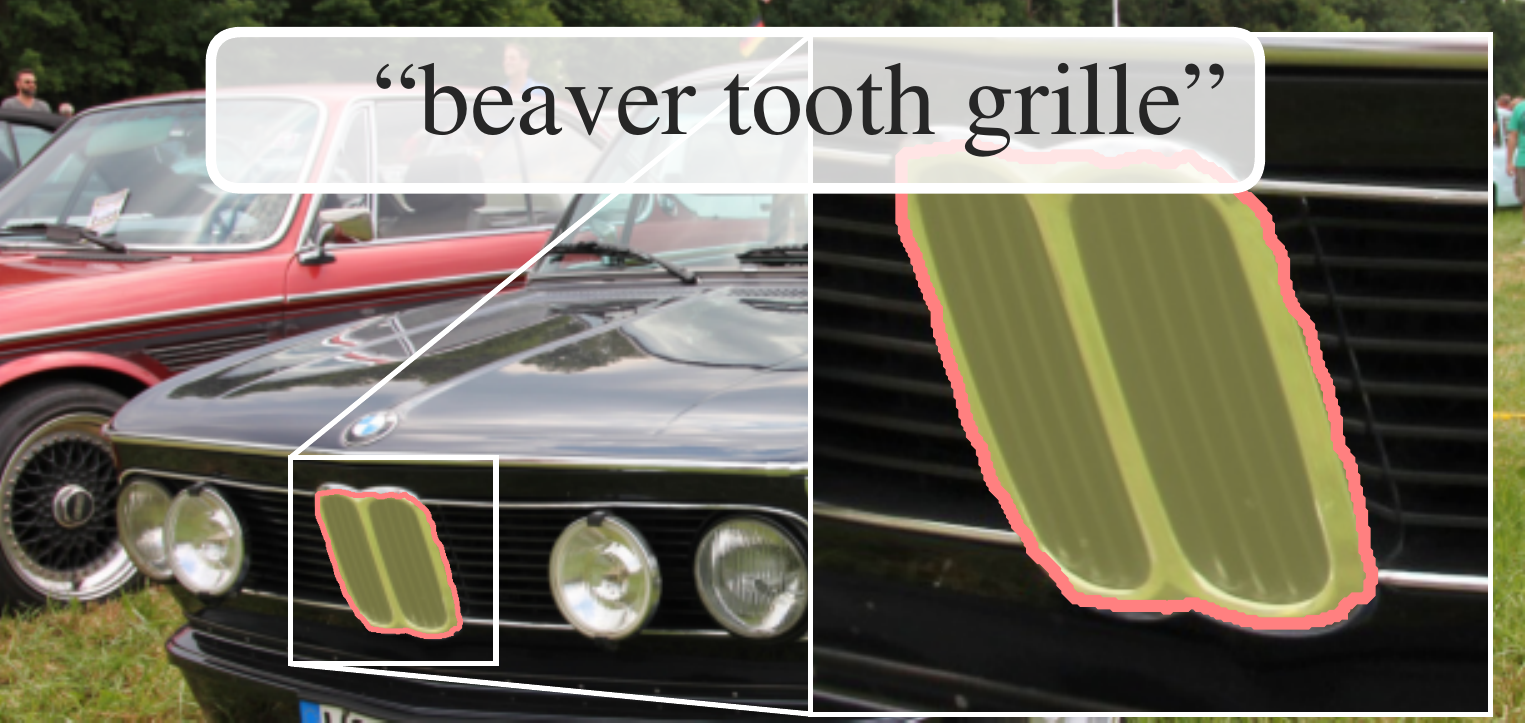}};
\draw (-1.25, 0.68) node {\cmark};
\end{tikzpicture}\\
\begin{tikzpicture}
\draw (0, 0) node[inner sep=0] {\includegraphics[width=0.49\linewidth]{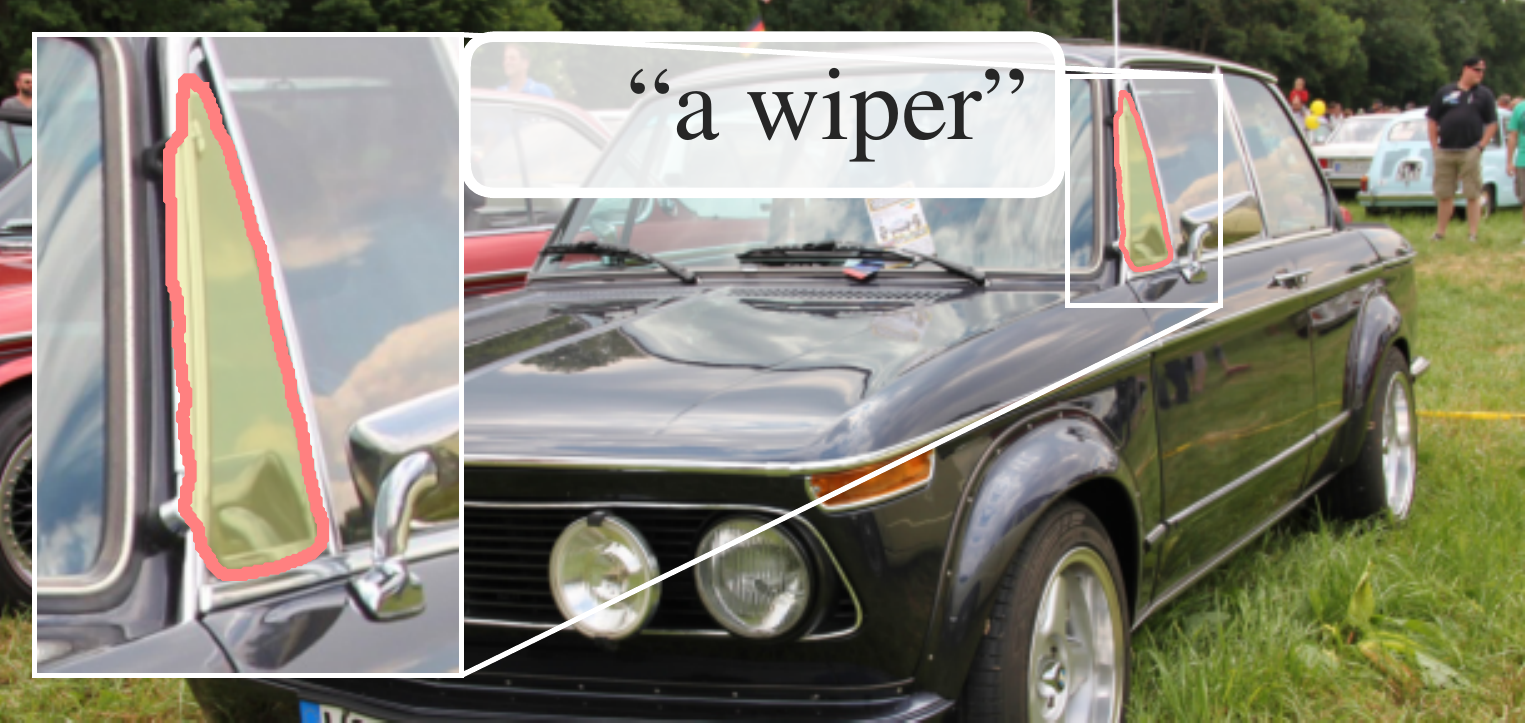}};
\draw (-0.55, 0.67) node {\xmark};
\end{tikzpicture} &
\begin{tikzpicture}
\draw (0, 0) node[inner sep=0] {\includegraphics[width=0.49\linewidth]{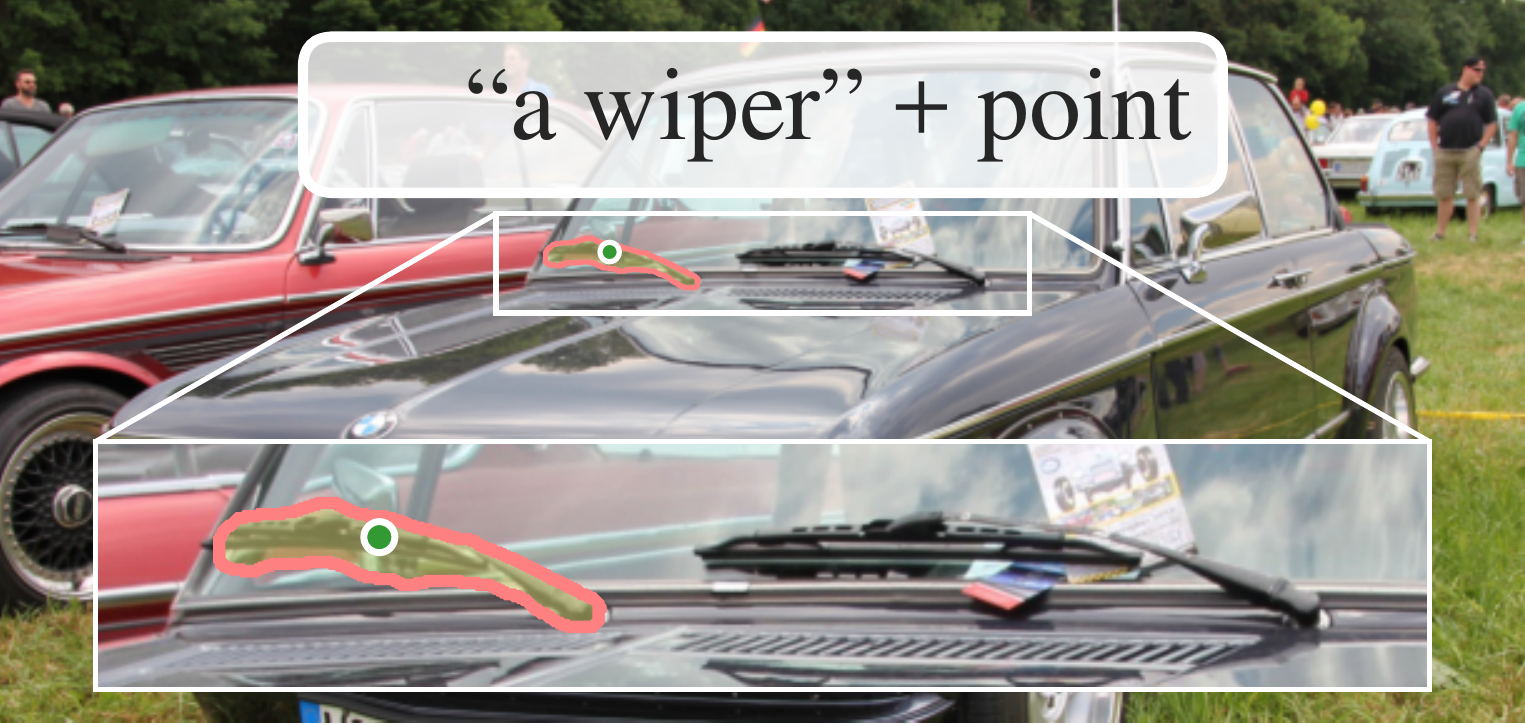}};
\draw (-0.97, 0.68) node {\cmark};
\end{tikzpicture}\\
\begin{tikzpicture}
\draw (0, 0) node[inner sep=0] {\includegraphics[width=0.49\linewidth]{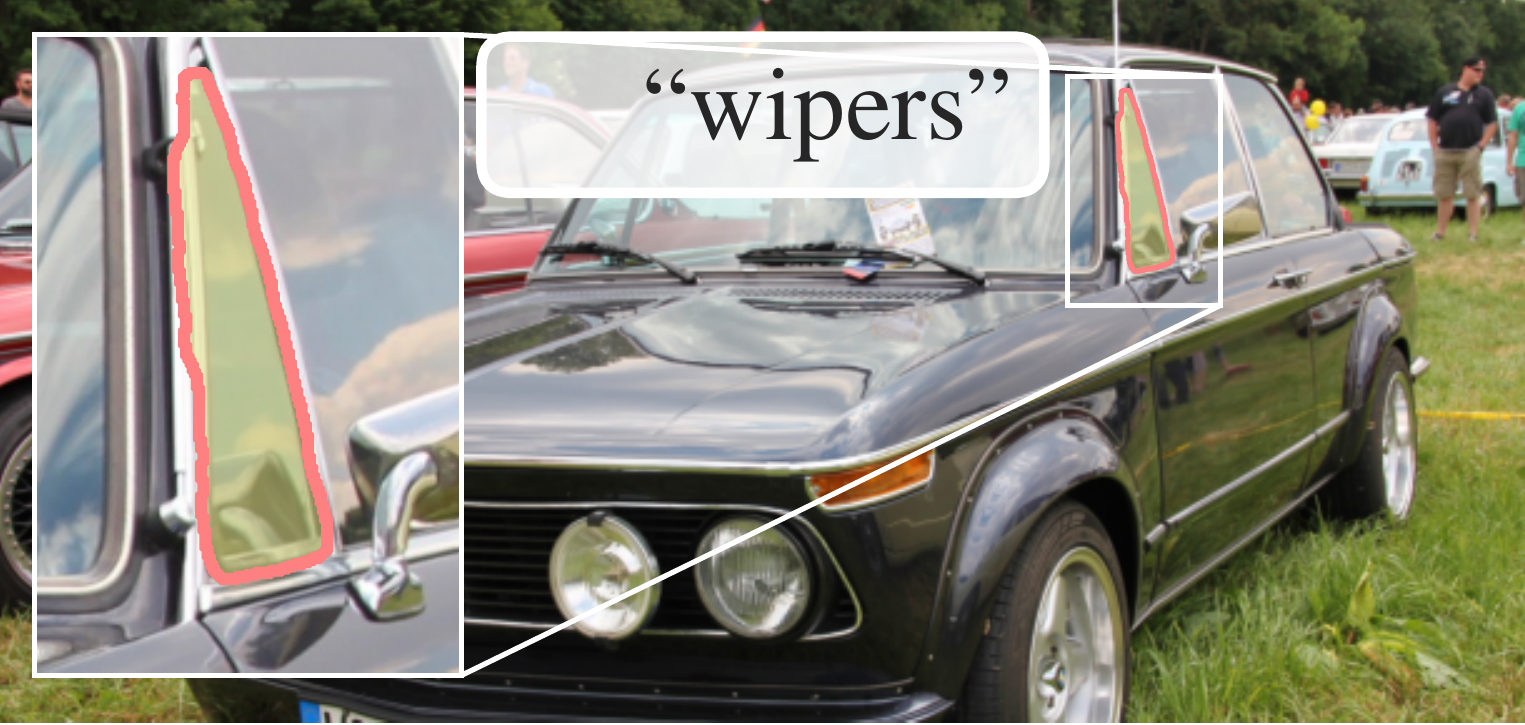}};
\draw (-0.50, 0.67) node {\xmark};
\end{tikzpicture} &
\begin{tikzpicture}
\draw (0, 0) node[inner sep=0] {\includegraphics[width=0.49\linewidth]{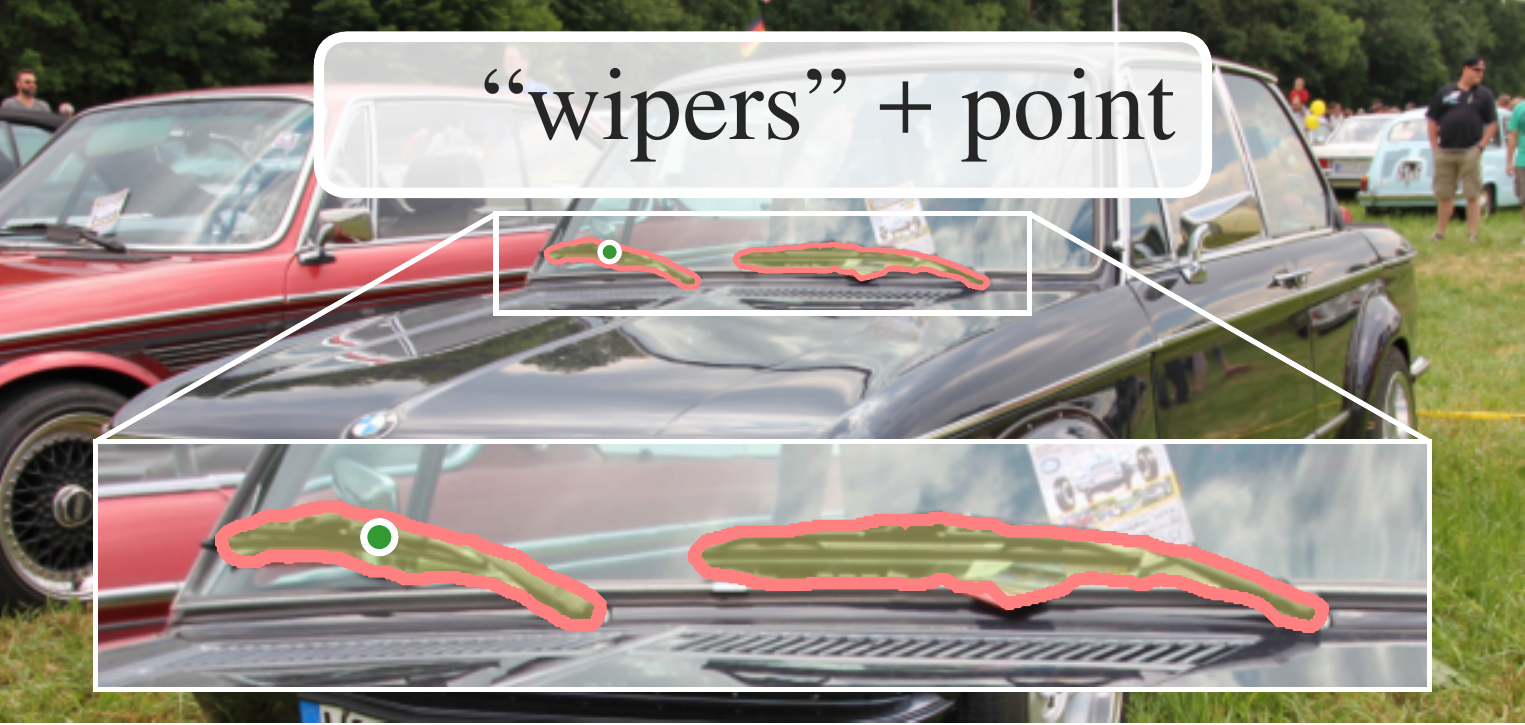}};
\draw (-0.95, 0.68) node {\cmark};
\end{tikzpicture}\\
\end{tabular}
\vspace{-4mm}
\caption{Zero-shot text-to-mask. \sam can work with simple and nuanced text prompts. When \sam fails to make a correct prediction, an additional point prompt can help.}
\label{fig:textprompts}\vspace{-5mm}
\end{figure}
%##################################################################################################
% %##################################################################################################
% \begin{figure}[t]\centering
% \tablestyle{1pt}{0.8}
% \begin{tabular}{c c}\begin{tikzpicture}
% \includegraphics[width=0.49\linewidth]{figs/text_to_mask/a_wheel.pdf}&
% \includegraphics[width=0.49\linewidth]{figs/text_to_mask/beaver_tooth_grille.pdf}\\
% \includegraphics[width=0.49\linewidth]{figs/text_to_mask/a_wiper.pdf}&
% \includegraphics[width=0.49\linewidth]{figs/text_to_mask/a_wiper_click.pdf}\\
% \includegraphics[width=0.49\linewidth]{figs/text_to_mask/wipers.pdf}&
% \includegraphics[width=0.49\linewidth]{figs/text_to_mask/wipers_click.pdf}
% \end{tabular}
% \vspace{-4mm}
% \caption{Zero-shot text-to-mask. \sam can work with simple and nuanced text prompts. When \sam fails to make a correct prediction, an additional point prompt can help.}
% \label{fig:textprompts}\vspace{-5mm}
% \end{figure}
% %##################################################################################################

\paragraph{Results.} We show qualitative results in \fig{fig:textprompts}. \sam can segment objects based on simple text prompts like ``a wheel'' as well as phrases like ``beaver tooth grille''. When \sam fails to pick the right object from a text prompt only, an additional point often fixes the prediction, similar to~\cite{ding2020phraseclick}.

%##################################################################################################
\begin{figure*}[t]\centering
\includegraphics[width=0.32\linewidth]{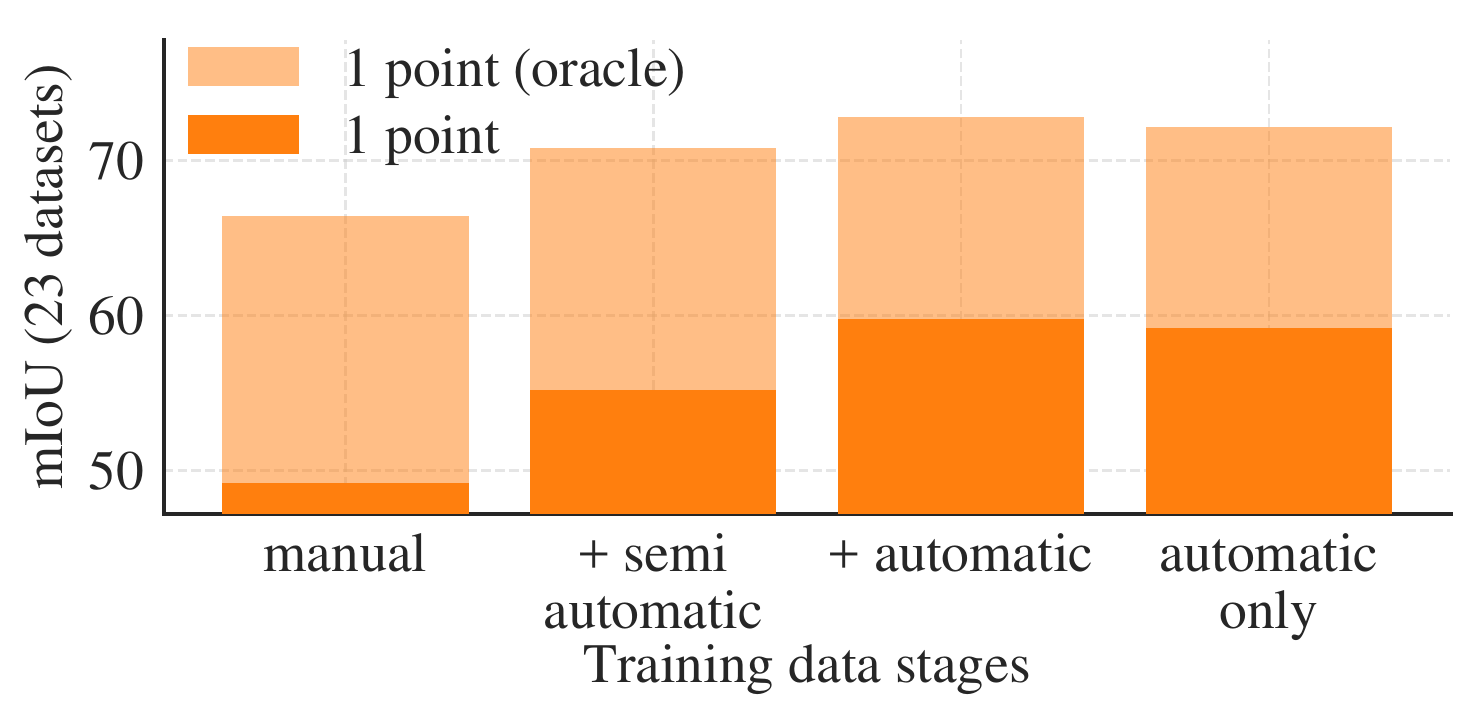}\hfill
\includegraphics[width=0.32\linewidth]{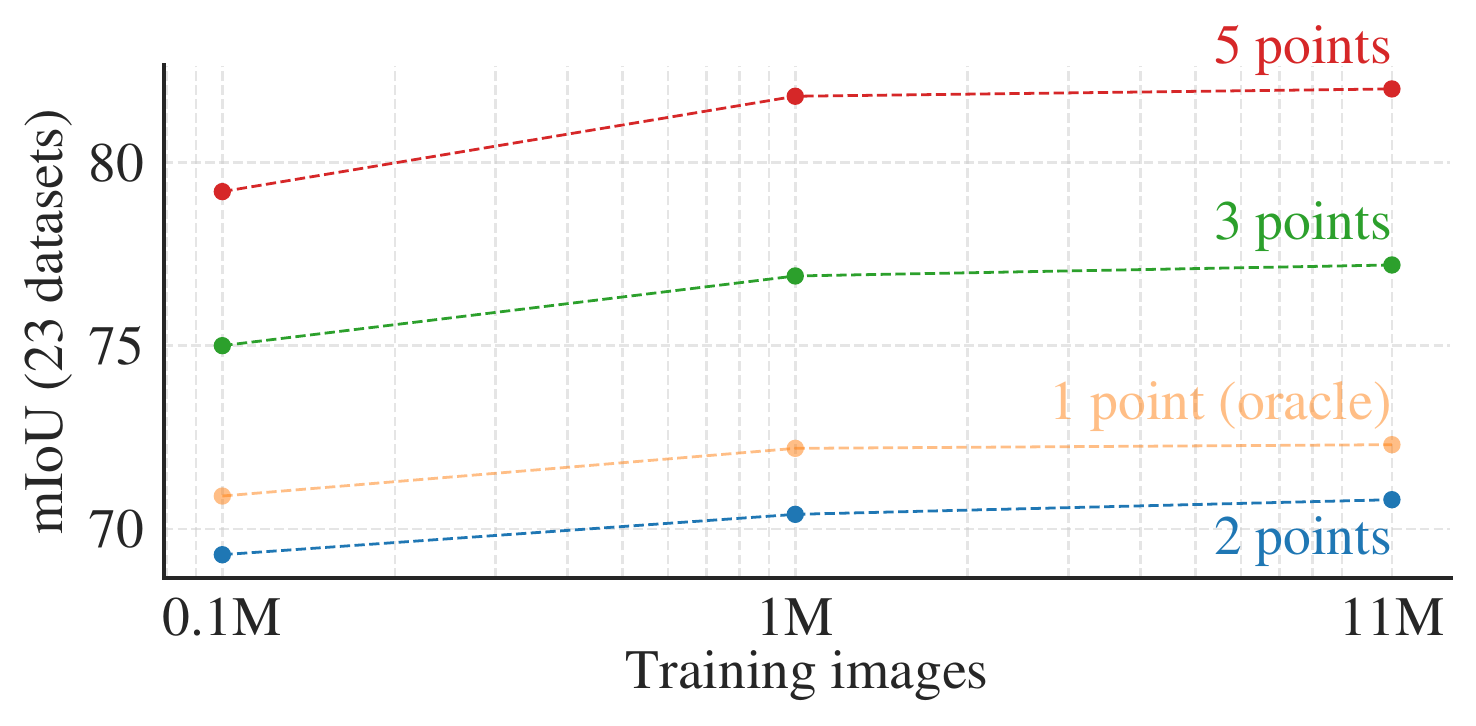}\hfill
\includegraphics[width=0.32\linewidth]{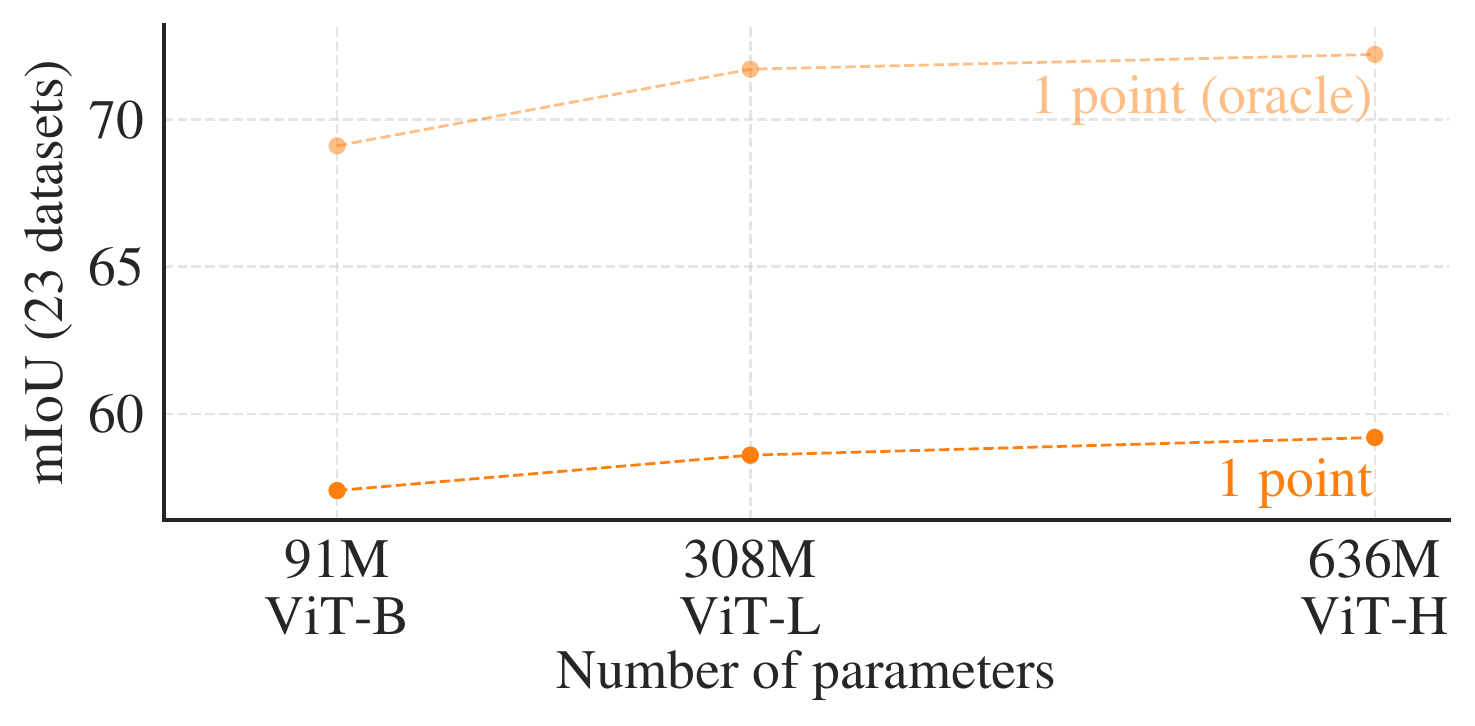}\vspace{-2mm}
\caption{Ablation studies of our data engine stages, image encoder scaling, and training data scaling. (Left) Each data engine stage leads to improvements on our 23 dataset suite, and training with only the automatic data (our default) yields similar results to using data from all three stages. (Middle) \sam trained with \app10\% of \sad and full \sad is comparable. We train with all 11M images by default, but using 1M images is a reasonable practical setting. (Right) Scaling \sam's image encoder shows meaningful, yet saturating gains. Nevertheless, smaller image encoders may be preferred in certain settings.}
\label{fig:ablations}\vspace{-3mm}
\end{figure*}
%##################################################################################################

\subsection{Ablations}\label{sec:eval:ablations}\vspace{-1mm}

We perform several ablations on our 23 dataset suite with the single center point prompt protocol. Recall that a single point may be ambiguous and that ambiguity may not be represented in the ground truth, which contains only a single mask per point. Since \sam is operating in a zero-shot transfer setting there can be systematic biases between \sam's top-ranked mask \vs the masks resulting from data annotation guidelines. We therefore additionally report the best mask with respect to the ground truth (``oracle'').

\fig{fig:ablations} (left) plots \sam's performance when trained on cumulative data from the data engine stages. We observe that each stage increases mIoU. When training with all three stages, the automatic masks vastly outnumber the manual and semi-automatic masks. To address this, we found that oversampling the manual and semi-automatic masks during training by 10$\x$ gave best results. This setup complicates training. We therefore tested a fourth setup that uses only the automatically generated masks. With this data, \sam performs only marginally lower than using all data (\app0.5 mIoU). Therefore, by default we use only the automatically generated masks to simplify the training setup.

In \fig{fig:ablations} (middle) we look at the impact of data volume. The full \sad contains 11M images, which we uniformly subsample to 1M and 0.1M for this ablation. At 0.1M images, we observe a large mIoU decline under all settings. However, with 1M images, about 10\% of the full dataset, we observe results comparable to using the full dataset. This data regime, which still includes approximately 100M masks, may be a practical setting for many use cases.

Finally, \fig{fig:ablations} (right) shows results with ViT-B, ViT-L, and ViT-H image encoders. ViT-H improves substantially over ViT-B, but has only marginal gains over ViT-L. Further image encoder scaling does not appear fruitful at this time.

%%%%%%%%%%%%%%%%%%%%%%%%%%%%%%%%%%%%%%%%%%%%%%%%%%%%%%%%%%%%%%%%%%%%%%%%%%%%%%%%%%%%%%%%%%%%%%%%%%%
\section{Discussion}\label{sec:disc}\vspace{-1mm}

\paragraph{Foundation models.} Pre-trained models have been adapted to downstream tasks since the early days of machine learning~\cite{thrun1995learning}. This paradigm has become increasingly important in recent years with a growing emphasis on scale, and such models have recently been (re-)branded as ``foundation models'': \ie models that are ``trained on broad data at scale and are adaptable to a wide range of downstream tasks''~\cite{bommasani2021opportunities}. Our work correlates well with this definition, though we note that a foundation model for image segmentation is an inherently limited scope, since it represents an important, yet fractional, subset of computer vision. We also contrast one aspect of our approach with~\cite{bommasani2021opportunities}, which emphasizes the role of \emph{self-supervised} learning in foundation models. While our model is initialized with a self-supervised technique (MAE~\cite{he2022masked}), the vast majority of its capabilities come from large-scale \emph{supervised} training. In cases where data engines can scale available annotations, like ours, supervised training provides an effective solution.

\paragraph{Compositionality.} Pre-trained models can power new capabilities even beyond ones imagined at the moment of training. One prominent example is how CLIP~\cite{Radford2021} is used as a \emph{component} in larger systems, such as DALL$\cdot$E~\cite{Ramesh2021}. Our goal is to make this kind of composition straightforward with \sam. We aim to achieve this by requiring \sam to predict a valid mask for a wide range of segmentation prompts. The effect is to create a reliable interface between \sam and other components. For example, MCC~\cite{wu2023multiview} can easily use \sam to segment an object of interest and achieve strong generalization to unseen objects for 3D reconstruction from a single RGB-D image. In another example, \sam can be prompted with gaze points detected by a wearable device, enabling new applications. Thanks to \sam's ability to generalize to new domains like ego-centric images, such systems work without need for additional training.

\paragraph{Limitations.} While \sam performs well in general, it is not perfect. It can miss fine structures, hallucinates small disconnected components at times, and does not produce boundaries as crisply as more computationally intensive methods that ``zoom-in'', \eg~\cite{chen2022focalclick}. In general, we expect dedicated interactive segmentation methods to outperform \sam when many points are provided, \eg~\cite{liu2022simpleclick}. Unlike these methods, \sam is designed for generality and breadth of use rather than high IoU interactive segmentation. Moreover, \sam can process prompts in real-time, but nevertheless \sam's overall performance is not real-time when using a heavy image encoder. Our foray into the text-to-mask task is exploratory and not entirely robust, although we believe it can be improved with more effort. While \sam can perform many tasks, it is unclear how to design simple prompts that implement semantic and panoptic segmentation. Finally, there are domain-specific tools, such as~\cite{berg2019}, that we expect to outperform \sam in their respective domains.

\paragraph{Conclusion.} The Segment Anything project is an attempt to lift image segmentation into the era of foundation models. Our principal contributions are a new task (promptable segmentation), model (\sam), and dataset (\sad) that make this leap possible. Whether \sam achieves the status of a foundation model remains to be seen by how it is used in the community, but regardless we expect the perspective of this work, the release of over 1B masks, and our promptable segmentation model will help pave the path ahead.

\paragraph{Acknowledgments.} We would like to thank Aaron Adcock and Jitendra Malik for helpful discussion. We thank Vaibhav Aggarwal and Yanghao Li for help with scaling the model. We thank Cheng-Yang Fu, Jiabo Hu, and Robert Kuo for help with data annotation platform. We thank Allen Goodman and Bram Wasti for help in optimizing web-version of our model. Finally, we thank Morteza Behrooz, Ashley Gabriel, Ahuva Goldstand, Sumanth Gurram, Somya Jain, Devansh Kukreja, Joshua Lane, Lilian Luong, Mallika Malhotra, William Ngan, Omkar Parkhi, Nikhil Raina, Dirk Rowe, Neil Sejoor, Vanessa Stark, Bala Varadarajan, and Zachary Winstrom for their help in making the demo, dataset viewer, and other assets and tooling.

%%%%%%%%%%%%%%%%%%%%%%%%%%%%%%%%%%%%%%%%%%%%%%%%%%%%%%%%%%%%%%%%%%%%%%%%%%%%%%%%%%%%%%%%%%%%%%%%%%%
{\footnotesize\linespread{.975}\selectfont\bibliographystyle{ieee_fullname}\bibliography{segany}}

%%%%%%%%%%%%%%%%%%%%%%%%%%%%%%%%%%%%%%%%%%%%%%%%%%%%%%%%%%%%%%%%%%%%%%%%%%%%%%%%%%%%%%%%%%%%%%%%%%%
\appendix
\section*{Appendix}
\label{appendix}
\paragraph{Table of contents:}
\begin{itemize}[itemsep=-1pt,topsep=-1pt]
\item \S\ref{app:model}: Segment Anything Model and Task Details
\item \S\ref{app:dataset_generation}: Automatic Mask Generation Details
\item \S\ref{app:rai}: RAI Additional Details
\item \S\ref{app:experimental_design}: Experiment Implementation Details
\item \S\ref{app:human_study}: Human Study Experimental Design
\item \S\ref{app:cards}: Dataset, Annotation, and Model Cards
\item \S\ref{app:annotation_guidelines}: Annotation Guidelines
\end{itemize}

%%%%%%%%%%%%%%%%%%%%%%%%%%%%%%%%%%%%%%%%%%%%%%%%%%%%%%%%%%%%%%%%%%%%%%%%%%%%%%%%%%%%%%%%%%%%%%%%%%%
\section{Segment Anything Model and Task Details}\label{app:model}

\paragraph{Image encoder.} In general, the image encoder can be any network that outputs a $C \x H \x W$ image embedding. Motivated by scalability and access to strong pre-training, we use an MAE~\cite{he2022masked} pre-trained Vision Transformer (ViT)~\cite{Dosovitskiy2021} with minimal adaptations to process high resolution inputs, specifically a ViT-H/16 with 14$\x$14 windowed attention and four equally-spaced global attention blocks, following~\cite{li2022exploring}. The image encoder's output is a 16$\x$ downscaled embedding of the input image. Since our runtime goal is to process each prompt in real-time, we can afford a high number of image encoder FLOPs because they are computed only once per image, \emph{not} per prompt.

Following standard practices (\eg,~\cite{Ghiasi2021}), we use an input resolution of 1024$\x$1024 obtained by rescaling the image and padding the shorter side. The image embedding is therefore 64$\x$64. To reduce the channel dimension, following~\cite{li2022exploring}, we use a 1$\x$1 convolution to get to 256 channels, followed by a 3$\x$3 convolution also with 256 channels. Each convolution is followed by a layer normalization~\cite{Ba2016}.

\paragraph{Prompt encoder.} Sparse prompts are mapped to 256-dimensional vectorial embeddings as follows. A point is represented as the sum of a positional encoding~\cite{tancik2020fourier} of the point's location and one of two learned embeddings that indicate if the point is either in the foreground or background. A box is represented by an embedding pair: (1) the positional encoding of its top-left corner summed with a learned embedding representing ``top-left corner'' and (2) the same structure but using a learned embedding indicating ``bottom-right corner''. Finally, to represent free-form text we use the text encoder from CLIP~\cite{Radford2021} (any text encoder is possible in general). We focus on geometric prompts for the remainder of this section and discuss text prompts in depth in \S\ref{app:text_to_mask}.

Dense prompts (\ie, masks) have a spatial correspondence with the image. We input masks at a 4$\x$ lower resolution than the input image, then downscale an additional 4$\x$ using two 2$\x$2, stride-2 convolutions with output channels 4 and 16, respectively. A final 1$\x$1 convolution maps the channel dimension to 256. Each layer is separated by GELU activations~\cite{Hendrycks2016} and layer normalization. The mask and image embedding are then added element-wise. If there is no mask prompt, a learned embedding representing ``no mask'' is added to each image embedding location.

%##################################################################################################
\begin{figure}[t]\centering
\includegraphics[width=0.99\linewidth]{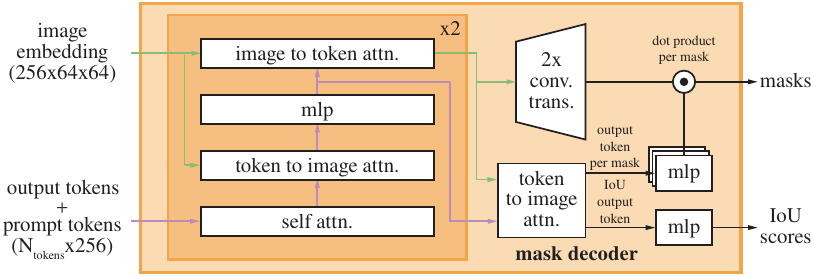}
\vspace{-2mm}
\caption{Details of the lightweight mask decoder. A two-layer decoder updates both the image embedding and prompt tokens via cross-attention. Then the image embedding is upscaled, from which the updated output tokens are used to dynamically predict masks. (Not illustrated for figure clarity: At every attention layer, positional encodings are added to the image embedding, and the entire original prompt token (including position encoding) is re-added to the token queries and keys.)}
\label{fig:model_diagram_zoomin}\vspace{-4mm}
\end{figure}
%##################################################################################################

\paragraph{Lightweight mask decoder.} This module efficiently maps the image embedding and a set of prompt embeddings to an output mask. To combine these inputs, we take inspiration from Transformer segmentation models~\cite{Carion2020,cheng2021per} and modify a standard Transformer decoder~\cite{Vaswani2017}. Before applying our decoder, we first insert into the set of prompt embeddings a learned output token embedding that will be used at the decoder's output, analogous to the \texttt{[class]} token in~\cite{Dosovitskiy2021}. For simplicity, we refer to these embeddings (\emph{not} including the image embedding) collectively as ``tokens''.

Our decoder design is shown in \fig{fig:model_diagram_zoomin}. Each decoder layer performs 4 steps: (1) self-attention on the tokens, (2) cross-attention from tokens (as queries) to the image embedding, (3) a point-wise MLP updates each token, and (4) cross-attention from the image embedding (as queries) to tokens. This last step updates the image embedding with prompt information. During cross-attention, the image embedding is treated as a set of 64$^\textrm{2}$ 256-dimensional vectors. Each self/cross-attention and MLP has a residual connection~\cite{He2016}, layer normalization, and a dropout~\cite{Srivastava2014} of 0.1 at training. The next decoder layer takes the updated tokens and the updated image embedding from the previous layer. We use a two-layer decoder.

To ensure the decoder has access to critical geometric information the positional encodings are added to the image embedding whenever they participate in an attention layer. Additionally, the \emph{entire} original prompt tokens (including their positional encodings) are re-added to the updated tokens whenever they participate in an attention layer. This allows for a strong dependence on both the prompt token's geometric location and type.

After running the decoder, we upsample the updated image embedding by 4$\x$ with two transposed convolutional layers (now it's downscaled 4$\x$ relative to the input image). Then, the tokens attend once more to the image embedding and we pass the updated output token embedding to a small 3-layer MLP that outputs a vector matching the channel dimension of the upscaled image embedding. Finally, we predict a mask with a spatially point-wise product between the upscaled image embedding and the MLP's output.

The transformer uses an embedding dimension of 256. The transformer MLP blocks have a large internal dimension of 2048, but the MLP is applied only to the prompt tokens for which there are relatively few (rarely greater than 20). However, in cross-attention layers where we have a 64$\x$64 image embedding, we reduce the channel dimension of the queries, keys, and values by 2$\x$ to 128 for computational efficiency. All attention layers use 8 heads.

The transposed convolutions used to upscale the output image embedding are 2$\x$2, stride 2 with output channel dimensions of 64 and 32 and have GELU activations. They are separated by layer normalization.

\paragraph{Making the model ambiguity-aware.} As described, a single input prompt may be ambiguous in the sense that it corresponds to multiple valid masks, and the model will learn to average over these masks. We eliminate this problem with a simple modification: instead of predicting a single mask, we use a small number of output tokens and predict multiple masks simultaneously. By default we predict three masks, since we observe that three layers (whole, part, and subpart) are often enough to describe nested masks. During training, we compute the loss (described shortly) between the ground truth and each of the predicted masks, but only backpropagate from the lowest loss. This is a common technique used for models with multiple outputs~\cite{charpiat2008automatic,guzman2012multiple,li2018interactive}. For use in applications, we'd like to rank predicted masks, so we add a small head (operating on an additional output token) that estimates the IoU between each predicted mask and the object it covers.

Ambiguity is much rarer with multiple prompts and the three output masks will usually become similar. To minimize computation of degenerate losses at training and ensure the single unambiguous mask receives a regular gradient signal, we only predict a single mask when more than one prompt is given. This is accomplished by adding a fourth output token for an additional mask prediction. This fourth mask is never returned for a single prompt and is the only mask returned for multiple prompts.

\paragraph{Losses.} We supervise mask prediction with a linear combination of focal loss~\cite{Lin2017a} and dice loss~\cite{milletari2016v} in a 20:1 ratio of focal loss to dice loss, following~\cite{cheng2021per,Carion2020}. Unlike~\cite{cheng2021per,Carion2020}, we observe that auxiliary deep supervision after each decoder layer is unhelpful. The IoU prediction head is trained with mean-square-error loss between the IoU prediction and the predicted mask's IoU with the ground truth mask. It is added to the mask loss with a constant scaling factor of 1.0.

\paragraph{Training algorithm.} Following recent approaches~\cite{sofiiuk2022reviving,forte2020getting}, we simulate an interactive segmentation setup during training. First, with equal probability either a foreground point or bounding box is selected randomly for the target mask. Points are sampled uniformly from the ground truth mask. Boxes are taken as the ground truth mask's bounding box, with random noise added in each coordinate with standard deviation equal to 10\% of the box sidelength, to a maximum of 20 pixels. This noise profile is a reasonable compromise between applications like instance segmentation, which produce a tight box around the target object, and interactive segmentation, where a user may draw a loose box.

After making a prediction from this first prompt, subsequent points are selected uniformly from the error region between the previous mask prediction and the ground truth mask. Each new point is foreground or background if the error region is a false negative or false positive, respectively. We also supply the mask prediction from the previous iteration as an additional prompt to our model. To provide the next iteration with maximal information, we supply the unthresholded mask logits instead of the binarized mask. When multiple masks are returned, the mask passed to the next iteration and used to sample the next point is the one with the highest predicted IoU.

We find diminishing returns after 8 iteratively sampled points (we have tested up to 16). Additionally, to encourage the model to benefit from the supplied mask, we also use two more iterations where no additional points are sampled. One of these iterations is randomly inserted among the 8 iteratively sampled points, and the other is always at the end. This gives 11 total iterations: one sampled initial input prompt, 8 iteratively sampled points, and two iterations where no new external information is supplied to the model so it can learn to refine its own mask predictions. We note that using a relatively large number of iterations is possible because our lightweight mask decoder requires less than 1\% of the image encoder's compute and, therefore, each iteration adds only a small overhead. This is unlike previous interactive methods that perform only one or a few interactive steps per optimizer update~\cite{mahadevan2018iteratively,bredell2018iterative,forte2020getting,sofiiuk2022reviving}.

\paragraph{Training recipe.} We use the AdamW~\cite{Loshchilov2019} optimizer ($\beta_1=0.9$, $\beta_2=0.999$) and a linear learning rate warmup~\cite{Goyal2017} for 250 iterations and a step-wise learning rate decay schedule. The initial learning rate (\lr), after warmup, is $\expnum{8}{-4}$. We train for 90k iterations ($\app$2 \sad epochs) and decrease the \lr by a factor of 10 at 60k iterations and again at 86666 iterations. The batch size is 256 images. To regularize \sam, we set weight decay (\wtd) to 0.1 and apply drop path~\cite{Huang2016deep} (\drp) with a rate of 0.4. We use a layer-wise learning rate decay~\cite{Bao2021} (\ld) of 0.8. No data augmentation is applied. We initialize \sam from an MAE~\cite{he2022masked} pre-trained ViT-H. We distribute training across 256 GPUs, due to the large image encoder and 1024$\x$1024 input size. To limit GPU memory usage, we train with up to 64 randomly sampled masks per GPU. Additionally, we find that lightly filtering \sad masks to discard any that cover more than 90\% of the image qualitatively improves results.

For ablations and others variations on training (\eg, text-to-mask \S\ref{app:text_to_mask}), we deviate from the default recipe above as follows. When training with data from the first and second data engine stages only, we augment the input with large-scale jitter~\cite{Ghiasi2021} with a scale range of [0.1, 2.0]. Intuitively, data augmentation may be helpful when training data is more limited. To train ViT-B and ViT-L, we use 180k iterations with batch size 128 distributed across 128 GPUs. We set \lr = $\expnum{8}{-4}$/$\expnum{4}{-4}$, \ld = 0.6/0.8, \wtd = 0.1, and \drp = 0.6/0.4 for ViT-B/L, respectively.

%%%%%%%%%%%%%%%%%%%%%%%%%%%%%%%%%%%%%%%%%%%%%%%%%%%%%%%%%%%%%%%%%%%%%%%%%%%%%%%%%%%%%%%%%%%%%%%%%%%
\section{Automatic Mask Generation Details}\label{app:dataset_generation}

Here we discuss details of the data engine's fully automatic stage that was used to generate the released \sad.

\paragraph{Cropping.} Masks were generated from a regular grid of 32$\x$32 points on the full image and 20 additional zoomed-in image crops arising from 2$\x$2 and 4$\x$4 partially overlapping windows using 16$\x$16 and 8$\x$8 regular point grids, respectively. The original high-resolution images were used for cropping (this was the only time we used them). We removed masks that touch the inner boundaries of the crops. We applied standard greedy box-based NMS (boxes were used for efficiency) in two phases: first within each crop and second across crops. When applying NMS within a crop, we used the model's predicted IoU to rank masks. When applying NMS across crops, we ranked masks from most zoomed-in (\ie, from a 4$\x$4 crop) to least zoomed-in (\ie, the original image), based on their source crop. In both cases, we used an NMS threshold of 0.7.

\paragraph{Filtering.} We used three filters to increase mask quality. First, to keep only \emph{confident} masks we filtered by the model's predicted IoU score at a threshold of 88.0. Second, to keep only \emph{stable} masks we compared two binary masks resulting from the same underlying soft mask by thresholding it at different values. We kept the prediction (\ie, the binary mask resulting from thresholding logits at 0) only if the IoU between its pair of -1 and +1 thresholded masks was equal to or greater than 95.0. Third, we noticed that occasionally an automatic mask would cover the entire image. These masks were generally uninteresting, and we filtered them by removing masks that covered 95\% or more of an image. All filtering thresholds were selected to achieve both a large number of masks and high mask quality as judged by professional annotators using the method described in \S\ref{sec:dataset}.

\paragraph{Postprocessing.} We observed two error types that are easily mitigated with postprocessing. First, an estimated 4\% of masks include small, spurious components. To address these, we removed connected components with area less than 100 pixels (including removing entire masks if the largest component is below this threshold). Second, another estimated 4\% of masks include small, spurious holes. To address these, we filled holes with area less than 100 pixels. Holes were identified as components of inverted masks.

\paragraph{Automatic mask generation model.} We trained a special version of \sam for fully automatic mask generation that sacrifices some inference speed for improved mask generation properties. We note the differences between our default \sam and the one used for data generation here: it was trained on manual and semi-automatic data only, it was trained for longer (177656 iterations instead of 90k) with large-scale jitter data augmentation~\cite{Ghiasi2021}, simulated interactive training used only point and mask prompts (no boxes) and sampled only 4 points per mask during training (reducing from our default of 9 to 4 sped up training iterations and had no impact on 1-point performance, though it would harm mIoU if evaluating with more points), and finally the mask decoder used 3 layers instead of 2.

\paragraph{\sad examples.} We show \sad samples in \fig{fig:sa1bvisuals}. For more examples, please see our \href{https://www.segment-anything.com/dataset/index.html}{dataset explorer}.

%%%%%%%%%%%%%%%%%%%%%%%%%%%%%%%%%%%%%%%%%%%%%%%%%%%%%%%%%%%%%%%%%%%%%%%%%%%%%%%%%%%%%%%%%%%%%%%%%%%
\section{RAI Additional Details}\label{app:rai}

\paragraph{Inferring geographic information for \sad.} While the images in \sad are not geo-tagged, each image has a caption describing its contents and where it was taken. We infer approximate image geo-locations from these captions using an Elmo-based named entity recognition model~\cite{peters2017semi}. Each extracted location entity is mapped to every matching country, province, and city. Captions are mapped to a single country by first considering the matching countries, then provinces, and finally cities. We note that there are ambiguities and potential for biases with this method (\eg, ``Georgia'' may refer to the country or the US state). As such, we use the extracted locations to analyze the dataset as a whole, but do not release the inferred locations. The captions will not be released publicly as required by the image provider.

\paragraph{Inferring geographic information for COCO and Open Images.} The COCO~\cite{Lin2014} and Open Images~\cite{OpenImages} datasets do not provide geo-locations. Following~\cite{De2019}, we retrieve geographic metadata using the Flickr API. We retrieved locations for 24\% of the COCO training set (19,562 images) and for Open Images we retrieved 18\% of the training set (493,517 images, after only considering images with masks). We note that the geographic information is approximate, and the sample of images with this information may not fully match the full dataset distribution.

\paragraph{Inferring income information.} We use each image's inferred country to look up its income level using the levels defined by The World Bank~\cite{worldbank2022}. We collapse the upper-middle and lower-middle levels into a single middle level.

\paragraph{Fairness in segmenting people.} To investigate \sam's fairness at segmenting people we use the More Inclusive Annotations for People (MIAP)~\cite{schumann2021step} test set annotations for Open Images~\cite{OpenImages}, which allows us to compare \sam's performance across perceived gender presentation and perceived age group. MIAP provides box annotations, while we need ground truth masks for this analysis. To get ground truth masks, we select each person-category mask from Open Images if its corresponding bounding box is within a 1\% margin (based on relative box side lengths) of an annotated bounding box in MIAP, resulting in 3.9k masks.

%##################################################################################################
\begin{table}[t]\centering
\resizebox{!}{12mm}{
\tablestyle{4pt}{1.1}\begin{tabular}{@{}lcc@{}}
{} & \multicolumn{2}{c}{mIoU at} \\
{} & 1 point & 3 points \\
\hline
\multicolumn{3}{@{}l}{\emph{perceived gender presentation}} \\
feminine & 76.3\mypm{1.1} & 90.7\mypm{0.5} \\
masculine & 81.0\mypm{1.2} & 92.3\mypm{0.4} \\
{} & {} & {} \\
\end{tabular}}\hspace{4mm}
\resizebox{!}{12mm}{
\tablestyle{4pt}{1.1}\begin{tabular}{@{}lcc@{}}
{} & \multicolumn{2}{c}{mIoU at} \\
{} & 1 point & 3 points \\
\hline
\multicolumn{3}{@{}l}{\emph{perceived age group}} \\
older & 81.9\mypm{3.8} & 92.8\mypm{1.6} \\
middle & 78.2\mypm{0.8} & 91.3\mypm{0.3} \\
young & 77.3\mypm{2.7} & 91.5\mypm{0.9} \\
\end{tabular}}
\vspace{-2mm}
\caption{\sam's performance segmenting clothing across perceived gender presentation and age group. The intervals for perceived gender are disjoint, with mIoU for masculine being higher. Confidence intervals for age group overlap.}
\label{app:tab:rai_clothing}\vspace{-2mm}
\end{table}
%##################################################################################################

\paragraph{Fairness in segmenting clothing.} We extend our analysis from \S\ref{sec:rai} to clothing segmentation. We look at \sam's performance on clothing relative to the attributes of those wearing the clothes. We use all 6.5k ground truth masks from Open Images that have a category under the clothing superclass and reside within a person box from MIAP. In Table~\ref{app:tab:rai_clothing} we compare performance across perceived gender presentation and age group. We find that \sam is better at segmenting clothing on those who present predominantly masculine, with disjoint 95\% confidence intervals. The gap closes when moving from 1 to 3 point evaluation. Differences for perceived age group are not significant. Our results indicate there is a bias when segmenting clothing across perceived gender presentation with a one point prompt, and we encourage users of \sam to be mindful of this limitation.

%%%%%%%%%%%%%%%%%%%%%%%%%%%%%%%%%%%%%%%%%%%%%%%%%%%%%%%%%%%%%%%%%%%%%%%%%%%%%%%%%%%%%%%%%%%%%%%%%%%
\section{Experiment Implementation Details}\label{app:experimental_design}

\subsection{Zero-Shot Single Point Valid Mask Evaluation}\label{app:benchmark}

%##################################################################################################
\begin{table*}[!htbp]
\vspace{5mm}
\resizebox{\textwidth}{!}{
\centering
\tablestyle{4pt}{1.8}
\begin{tabular}{L{3.5cm} r | l | L{7cm}| l | L{3.2cm} | l l}
dataset &
\makecell[l]{abbreviation\\ \& link} &
\makecell[l]{image\\ type} &
description &
\makecell[l]{mask\\ type} &
source split &
\makecell[l]{\# images \\ sampled} &
\makecell[l]{\# masks \\ sampled} \\
\hline
Plant Phenotyping Datasets Leaf Segmentation~\cite{plants} &
\href{https://www.plant-phenotyping.org/datasets-home}{PPDLS} &
Plants &
Leaf segmentation for images of tobacco and ara plants. &
Instance &
N/A &
182 &
2347 \\
\hline
BBBC038v1 from Broad Bioimage Benchmark Collection~\cite{cells} &
\href{https://bbbc.broadinstitute.org/BBBC038}{BBBC038v1} &
Microscopy &
Biological images of cells in a variety of settings testing robustness in nuclei segmentation. &
Instance &
Train &
227 &
10506 \\
\hline
Dataset fOr bOuldeRs Segmentation~\cite{doors} &
\href{https://zenodo.org/record/7107409\#.ZAzNnOzMJ47}{DOORS} &
Boulders &
Segmentation masks of single boulders positioned on the surface of a spherical mesh. &
Instance &
DS1 &
10000 &
10000 \\
\hline
TimberSeg 1.0~\cite{timberSeg} &
\href{https://data.mendeley.com/datasets/y5npsm3gkj}{TimberSeg} &
Logs &
Segmentation masks of individual logs in piles of timber in various environments and conditions. Images are taken from an operator's point-of-view. &
Instance &
N/A &
220 &
2487 \\
\hline
Northumberland Dolphin Dataset 2020~\cite{ndd20} &
\href{https://doi.org/10.25405/data.ncl.c.4982342}{NDD20} &
Underwater &
Segmentation masks of two different dolphin species in images taken above and under water. &
Instance &
N/A &
4402 &
6100 \\
\hline
Large Vocabulary Instance Segmentation~\cite{Gupta2019} &
\href{https://www.lvisdataset.org/}{LVIS} &
\makecell[l]{Scenes} &
Additional annotations for the COCO~\cite{Lin2014} dataset to enable the study of long-tailed object detection and segmentation. &
Instance &
Validation (v0.5) &
945 &
9642 \\
\hline
STREETS~\cite{streets} &
\href{https://databank.illinois.edu/datasets/IDB-3671567}{STREETS} &
\makecell[l]{Traffic \\ camera} &
Segmentation masks of cars in traffic camera footage. &
Instance &
N/A &
819 &
9854 \\
\hline
ZeroWaste-f~\cite{zerowaste} &
\href{http://ai.bu.edu/zerowaste/}{ZeroWaste-f} &
Recycling &
Segmentation masks in cluttered scenes of deformed recycling waste. &
Instance &
Train &
2947 &
6155 \\
\hline
iShape~\cite{iShape} &
\href{https://ishape.github.io/}{iShape} &
\makecell[l]{Irregular \\ shapes} &
Segmentation masks of irregular shapes like antennas, logs, fences, and hangers. &
Instance &
Validation &
754 &
9742 \\
\hline
ADE20K~\cite{Zhou2019} &
\href{https://groups.csail.mit.edu/vision/datasets/ADE20K/}{ADE20K} &
Scenes &
Object and part segmentation masks for images from SUN~\cite{sun2010} and Places~\cite{zhou2017places} datasets. &
Instance &
Validation &
302 &
10128 \\
\hline
Occluded Video Instance Segmentation~\cite{ovis} &
\href{http://songbai.site/ovis/}{OVIS} &
Occlusions &
Instance segmentation masks in videos, focusing on objects that are occluded. &
Instance &
Train &
2044 &
10011 \\
\hline
Hypersim~\cite{hypersim} &
\href{https://github.com/apple/ml-hypersim}{Hypersim} &
Simulation &
Photorealistic synthetic dataset of indoor scenes with instance masks. &
Instance &
Evermotion archinteriors volumes 1-55 excluding 20,25,40,49 &
338 &
9445 \\
\hline
Night and Day Instance Segmented Park~\cite{ndis1, ndis2} &
\href{https://zenodo.org/record/6560823\#.ZAzLlezMJ46}{NDISPark} &
Parking lots &
Images of parking lots from video footage taken at day and night during different weather conditions and camera angles for vehicle segmentation. &
Instance &
Train &
111 &
2577 \\
\hline
EPIC-KITCHENS VISOR~\cite{VISOR, EpicKitchens} &
\href{https://epic-kitchens.github.io/VISOR/}{VISOR} &
Egocentric &
Segmentation masks for hands and active objects in ego-centric video from the cooking dataset EPIC-KITCHENS~\cite{EpicKitchens}. &
Instance &
Validation &
1864 &
10141 \\
\hline
Plittersdorf dataset~\cite{haucke2022socrates} &
\href{https://timm.haucke.xyz/datasets/plittersdorf}{Plittersdorf} &
\makecell[l]{Stereo \\ images} &
Segmentation masks of wildlife in images taken with the SOCRATES stereo camera trap. &
Instance &
Train, validation, test &
187 &
546 \\
\hline
Egocentric Hand-Object Segmentation~\cite{egoHos} &
\href{https://github.com/owenzlz/EgoHOS}{EgoHOS} &
Egocentric &
Fine-grained egocentric hand-object segmentation dataset. Dataset contains mask annotations for existing datasets. &
Instance &
Train (including only Ego4D~\cite{Ego4D2022CVPR} and THU-READ~\cite{THU-READ1, THU-READ2}) &
2940 &
9961 \\
\hline
InstanceBuilding 2D~\cite{chen20223D} &
\href{https://californiachen.github.io/datasets/InstanceBuilding}{IBD} &
Drones &
High-resolution drone UAV images annotated with roof instance segmentation masks. &
Instance &
Train (2D annotations) &
467 &
11953 \\
\hline
WoodScape~\cite{yogamani2019woodscape} &
\href{https://woodscape.valeo.com/home}{WoodScape} &
\makecell[l]{Fisheye \\driving} &
Fisheye driving dataset with segmentation masks. Images are taken from four surround-view cameras. &
Instance &
Set 1 &
107 &
10266 \\
\hline
Cityscapes~\cite{Cordts2016} &
\href{https://www.cityscapes-dataset.com/}{Cityscapes} &
Driving &
Stereo video of street scenes with segmentation masks. &
Panoptic & Validation &
293 &
9973 \\
\hline
PIDray~\cite{wang2021towards} &
\href{https://github.com/bywang2018/security-dataset}{PIDRay} &
X-ray &
Segmentation masks of prohibited items in X-ray images of baggage. &
Instance &
Test (hard) &
3733 &
8892 \\
\hline
Diverse Realism in Art Movements~\cite{DRAM} &
\href{https://faculty.runi.ac.il/arik/site/artseg/Dram-Dataset.html}{DRAM} &
Paintings &
Domain adaptation dataset for semantic segmentation of art paintings. &
Semantic &
Test &
718 &
1179 \\
\hline
TrashCan~\cite{hong2020trashcan} &
\href{https://conservancy.umn.edu/handle/11299/214865}{TrashCan} &
Underwater &
Segmentation masks of trash in images taken by underwater ROVs. Images are sourced from the J-EDI~\cite{lucas2014gelatinous} dataset. &
Instance &
Train (instance task) &
5936 &
9540 \\
\hline
Georgia Tech Egocentric Activity Datasets~\cite{gtea1, gtea2} &
\href{https://cbs.ic.gatech.edu/fpv/}{GTEA} &
Egocentric &
Videos are composed of four different subjects performing seven types of daily activities with segmentation masks of hands. &
Instance &
Train (segmenting hands task) &
652 &
1208 \\
\end{tabular}}
\caption{Segmentation datasets used to evaluate zero-shot segmentation with point prompts. The 23 datasets cover a broad range of domains; see column ``image type''. To make our evaluation efficient, we subsample datasets that have more than 15k masks. Specifically, we randomly sampled images so that the total number of masks in the images is \app10k.}
\label{app:tab:datasets_all}\vspace{5mm}
\end{table*}
%##################################################################################################

\paragraph{Datasets.} We built a new segmentation benchmark to evaluate the zero-shot transfer capabilities of our model using a suite of 23 diverse segmentation datasets from prior work. A description of each dataset is given in Table~\ref{app:tab:datasets_all}. For examples, see main text \fig{fig:benchmark_examples}. This suite covers a range of domains including egocentric~\cite{gtea1,VISOR,egoHos}, microscopy~\cite{cells}, X-ray~\cite{wang2021towards}, underwater~\cite{hong2020trashcan,ndd20}, aerial~\cite{chen20223D}, simulation~\cite{hypersim}, driving~\cite{Cordts2016}, and painting~\cite{DRAM} images. For efficient evaluation we subsampled datasets with more than 15k masks. Specifically, we randomly picked images so that the total number of masks in the sampled images was \app10k. We blurred faces of people in all the datasets.

\paragraph{Point sampling.} Our default point sampling follows standard practice in interactive segmentation~\cite{xu2016deep,li2018interactive,sofiiuk2022reviving}. The first point is chosen deterministically as the point farthest from the object boundary. Each subsequent point is the farthest from the boundary of the error region between ground truth and the previous prediction. Some experiments (where specified) use a more challenging sampling strategy in which the first point is a \emph{random} point, rather than a deterministically selected ``center'' point. Each subsequent point is selected as described above. This setting better reflects use cases in which the first point is not reliably near the center of the mask, such as prompting from eye gaze.

\paragraph{Evaluation.} We measure IoU between a prediction after $N$ point prompts and a ground truth mask, where $N=\{1,2,3,5,9\}$ and points are sampled iteratively with either of the strategies described above. The per-dataset mIoU is the per-mask IoU averaged across all objects in the dataset. Finally, we report the top-line metric by averaging the per-dataset mIoUs across all 23 datasets. Our evaluation differs from the standard interactive segmentation evaluation protocol which measures the average number of points needed to achieve $X$\% IoU, with up to 20 points. We focus on predictions after just one, or possibly a few points, since many of our use cases involve a single or very few prompts. Given our application focus, which requires real-time prompt processing, we expect the best interactive segmentation models to outperform \sam when using a large number of points.

\paragraph{Baselines.} We use three recent strong interactive baselines: RITM~\cite{sofiiuk2022reviving}, FocalClick~\cite{chen2022focalclick}, and SimpleClick~\cite{liu2022simpleclick}. For each, we use the largest models trained on the broadest datasets publicly released by the authors. For RITM, we use \mbox{\tt HRNet32 IT-M} trained on the combination of COCO~\cite{Lin2014} and LVIS~\cite{Gupta2019} introduced by the authors. For FocalClick, we use {\tt SegFormerB3-S2} trained on a ``combined dataset'' that includes 8 different segmentation datasets~\cite{chen2022focalclick}. For SimpleClick, we use {\tt ViT-H448} trained on a combination of COCO and LVIS. We follow the suggested default strategies for data pre-processing (\ie, data augmentations or image resizing) and do not change or adapt any parameters for our evaluation. In our experiments, we observe that RITM outperforms other baselines on our 23 dataset suite with 1 point evaluation. Therefore, we use RITM as the default baseline. When evaluating with more points we report results for all baselines.

\paragraph{Single point ambiguity and oracle evaluation.} In addition to IoU after $N$ points prompts, we report \sam's ``oracle'' performance at 1 point by evaluating the predicted mask that best matches ground truth from amongst \sam's three predictions (rather than using the one that \sam itself ranks first, as we do by default). This protocol addresses possible single point prompt ambiguity by relaxing the requirement to guess the one right mask among several valid objects.

%##################################################################################################
\begin{figure*}[t]\footnotesize\centering
\tablestyle{2pt}{1.1}
\begin{tabular}{@{}cccccc@{}}\centering
image & ground truth & \sam & image & ground truth & \sam \\
\includegraphics[width=0.16\linewidth]{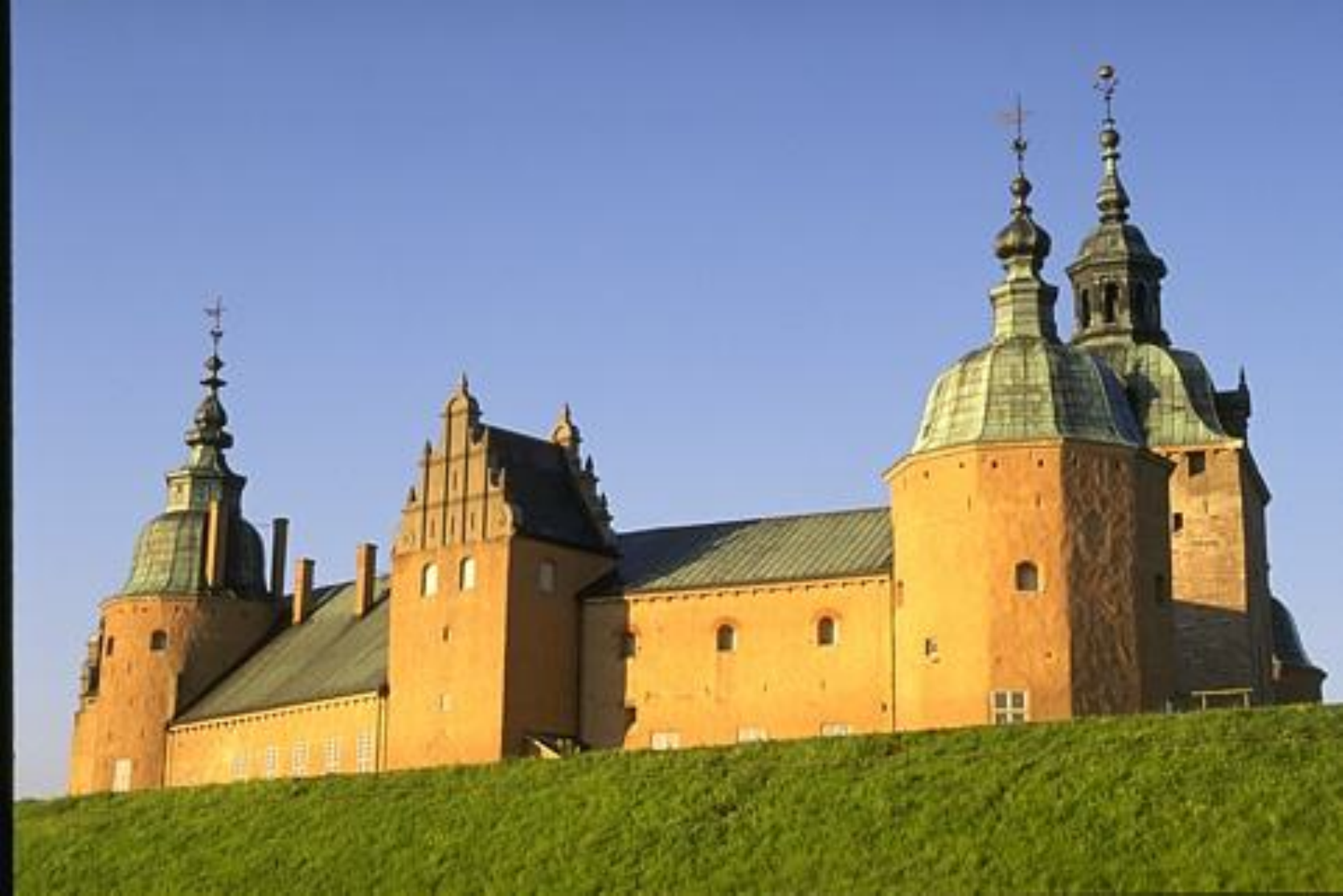} &
\includegraphics[width=0.16\linewidth]{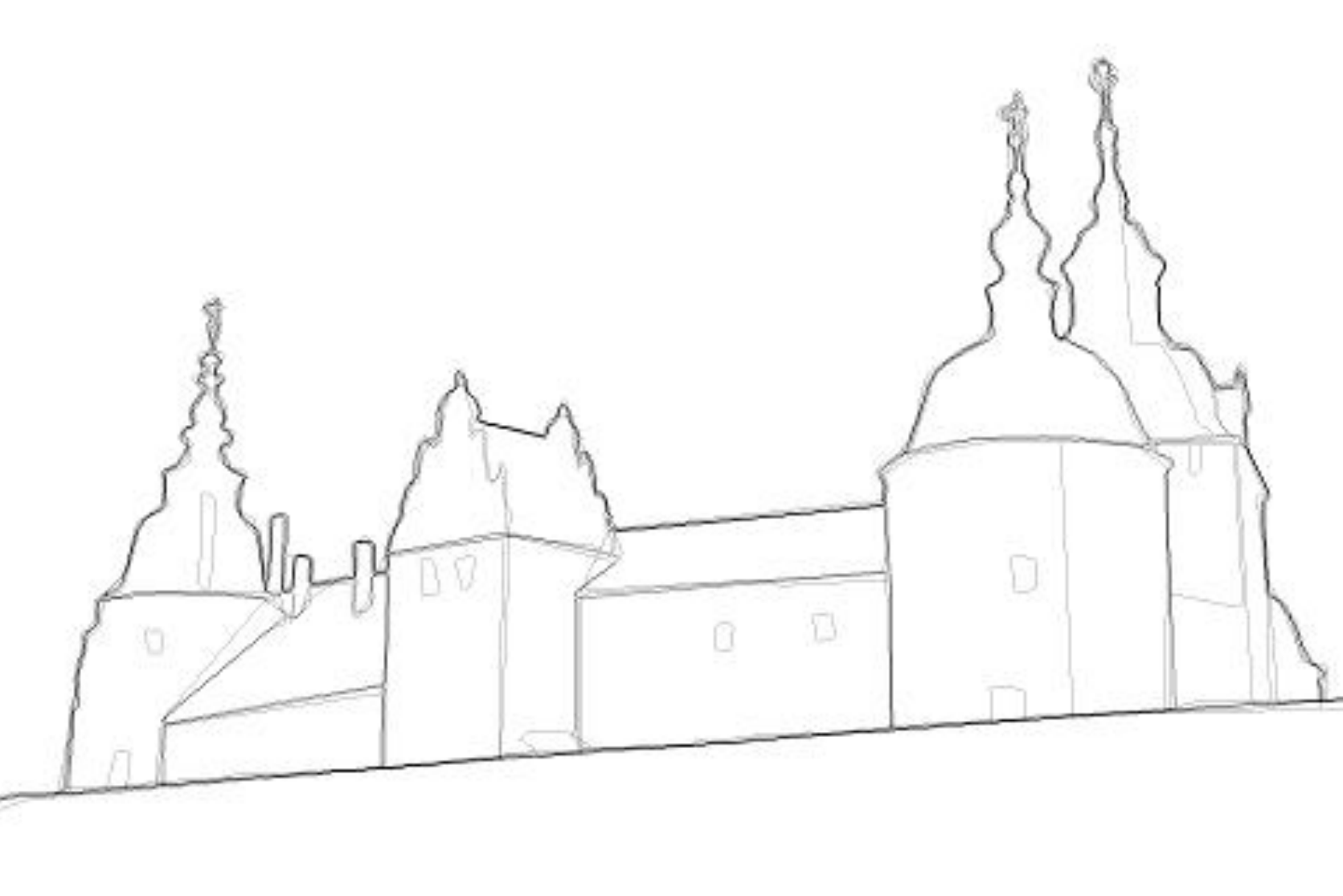} &
\includegraphics[width=0.16\linewidth]{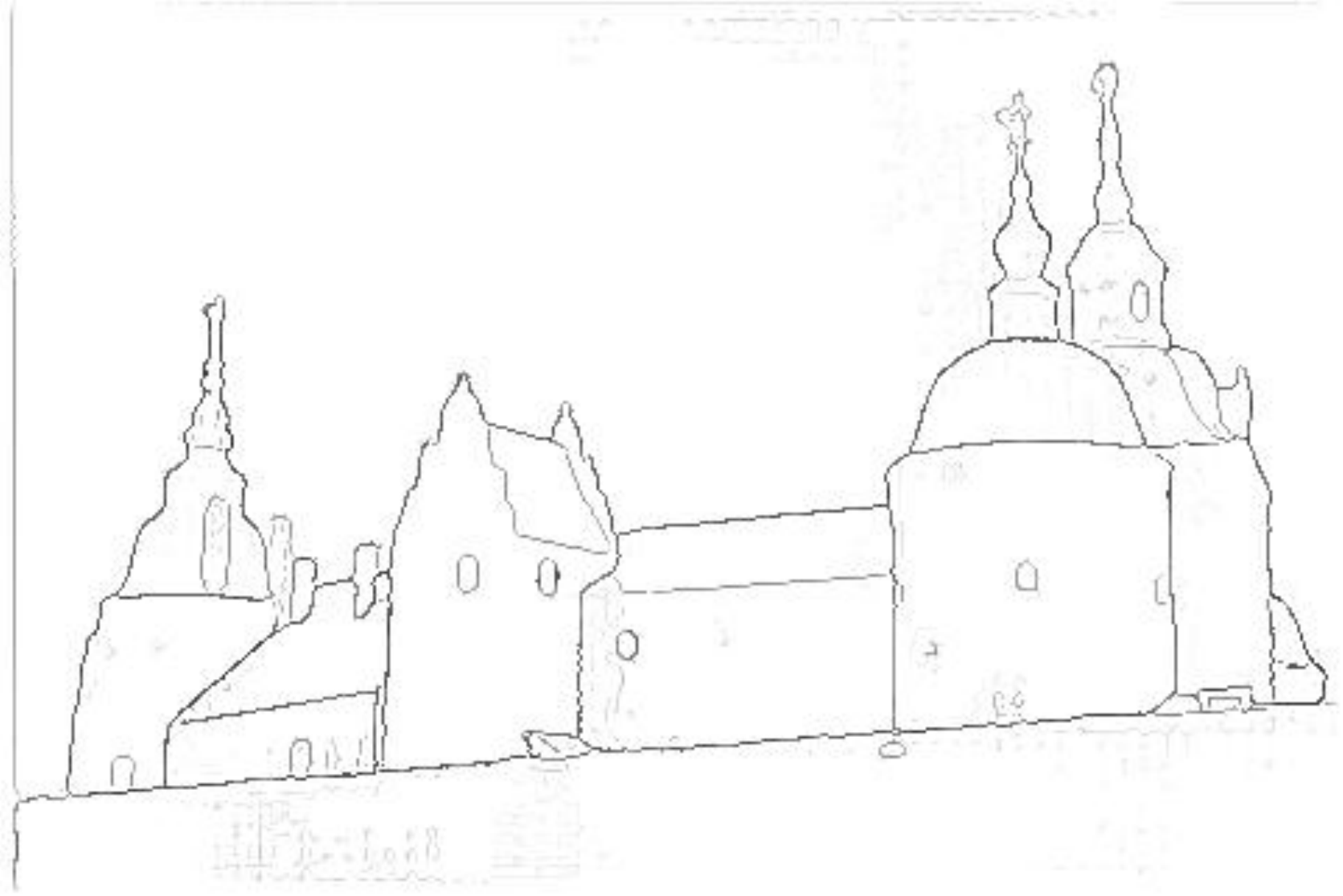} &
\includegraphics[width=0.16\linewidth]{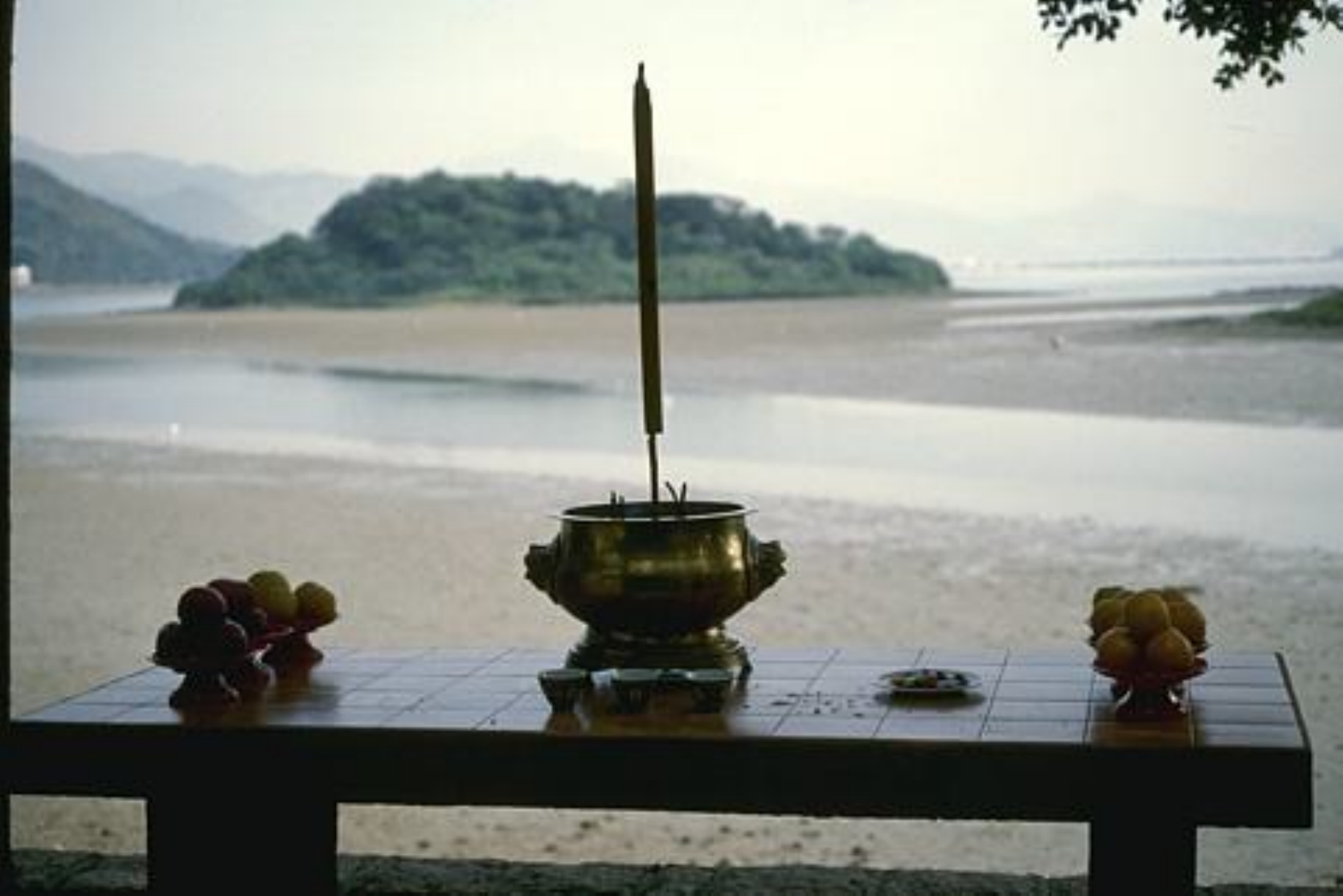} &
\includegraphics[width=0.16\linewidth]{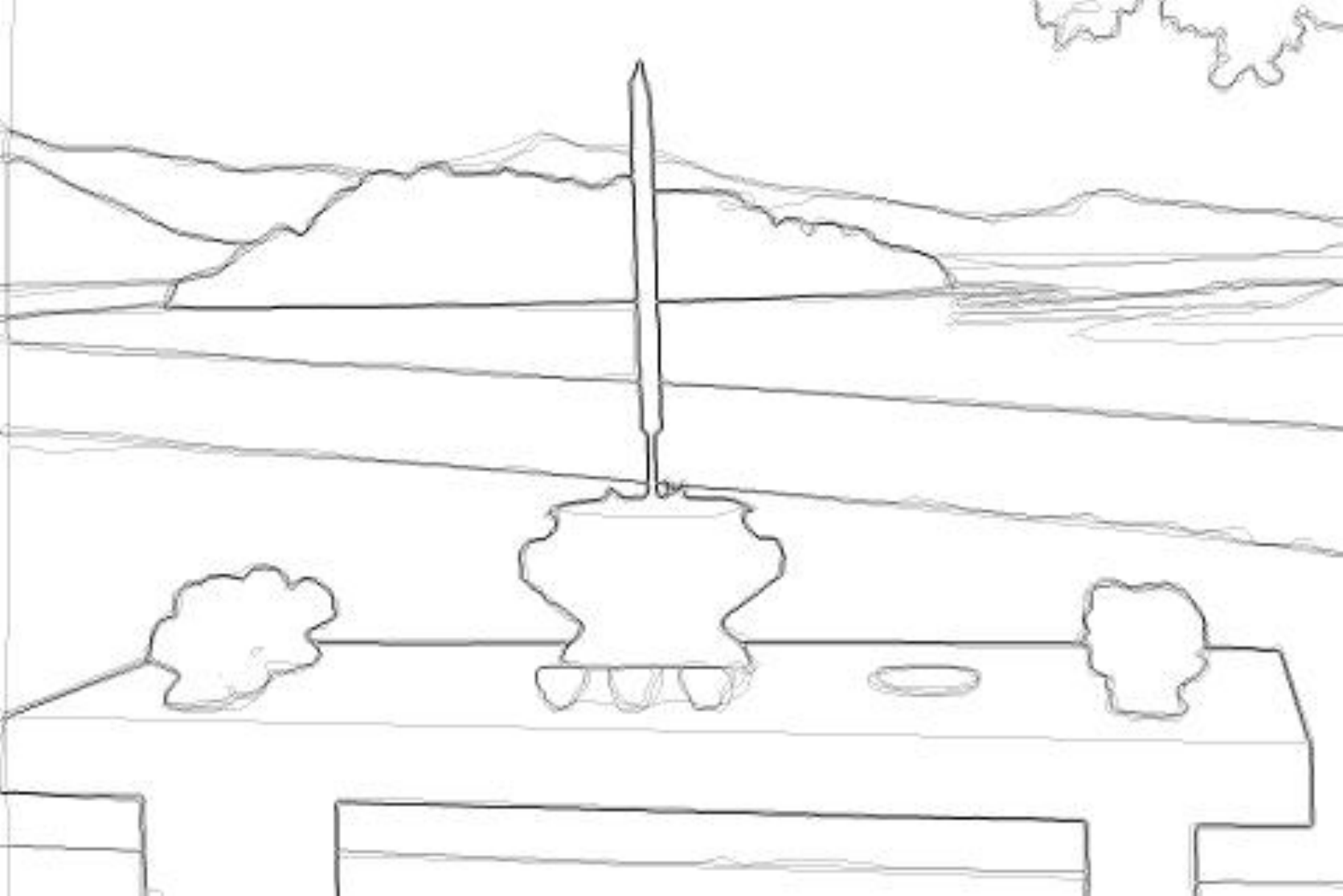} &
\includegraphics[width=0.15\linewidth]{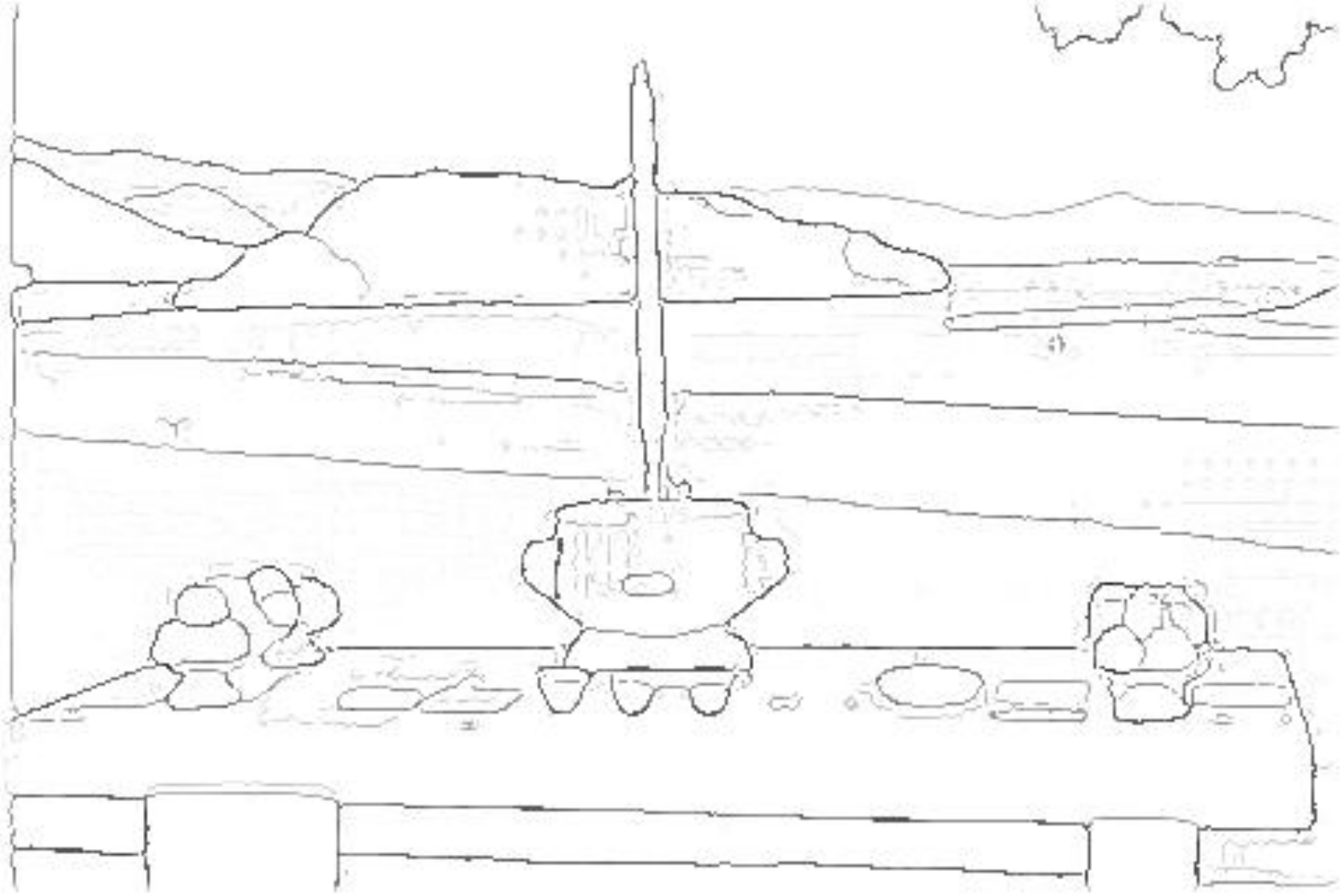}\\
\includegraphics[width=0.16\linewidth]{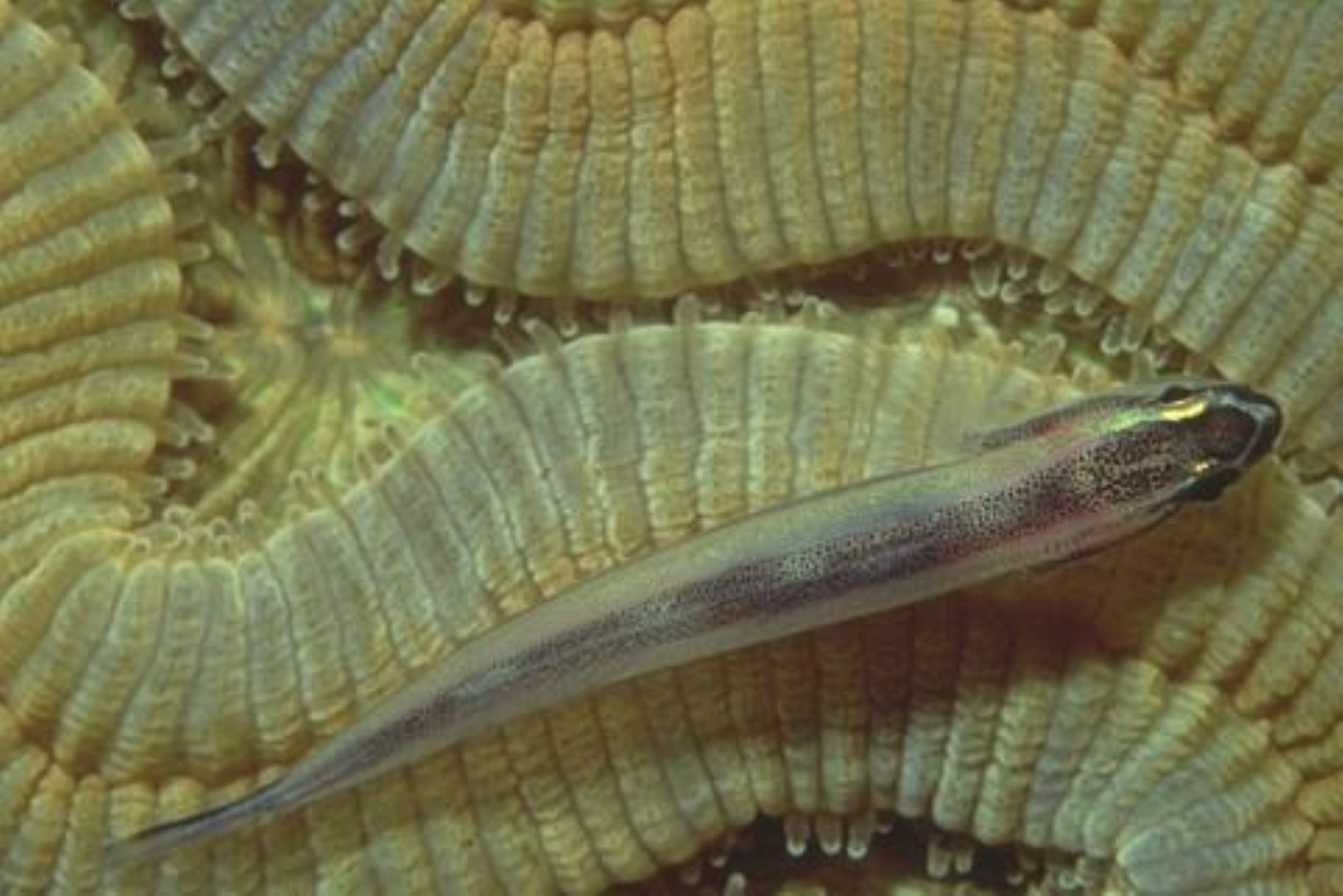} &
\includegraphics[width=0.16\linewidth]{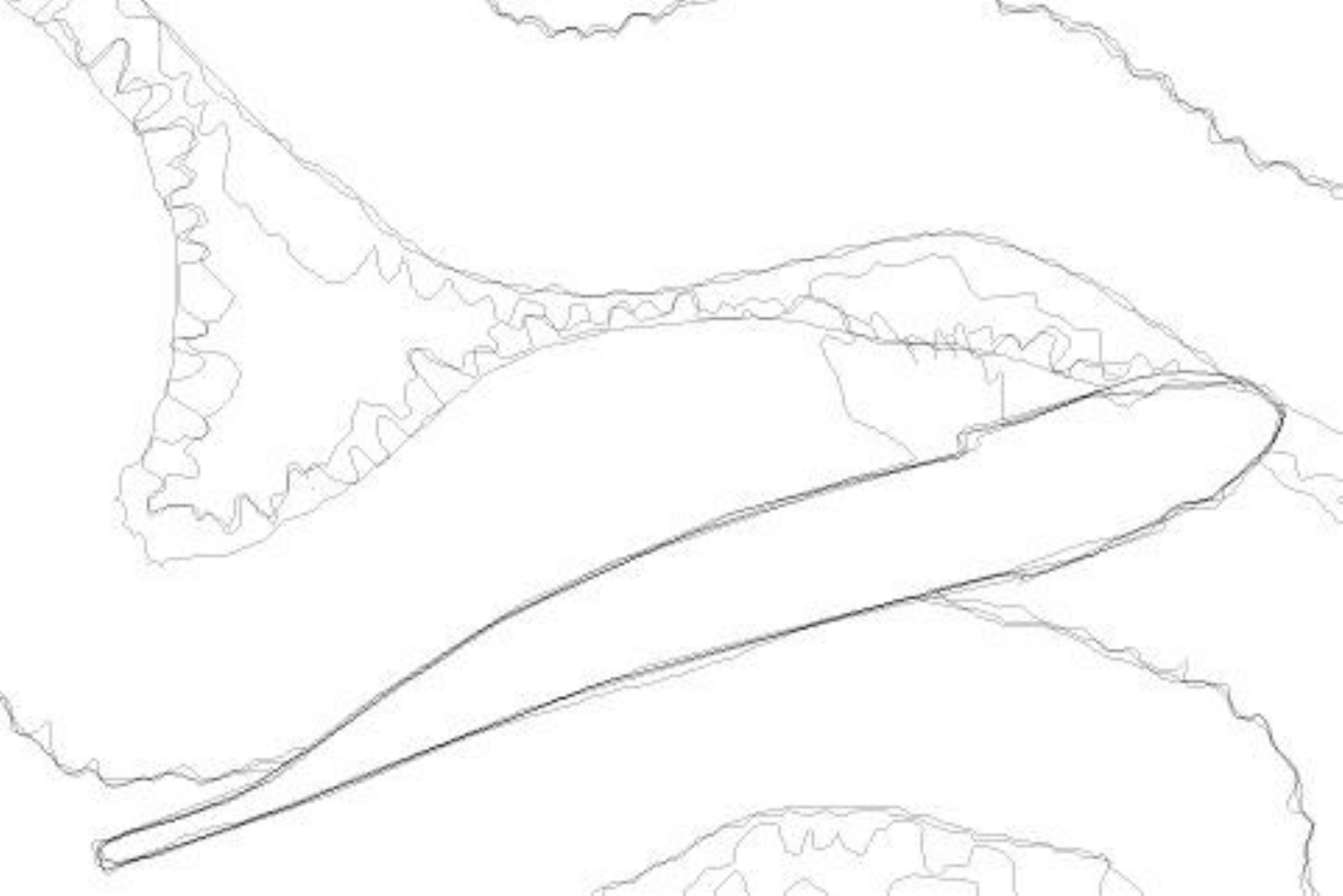} &
\includegraphics[width=0.16\linewidth]{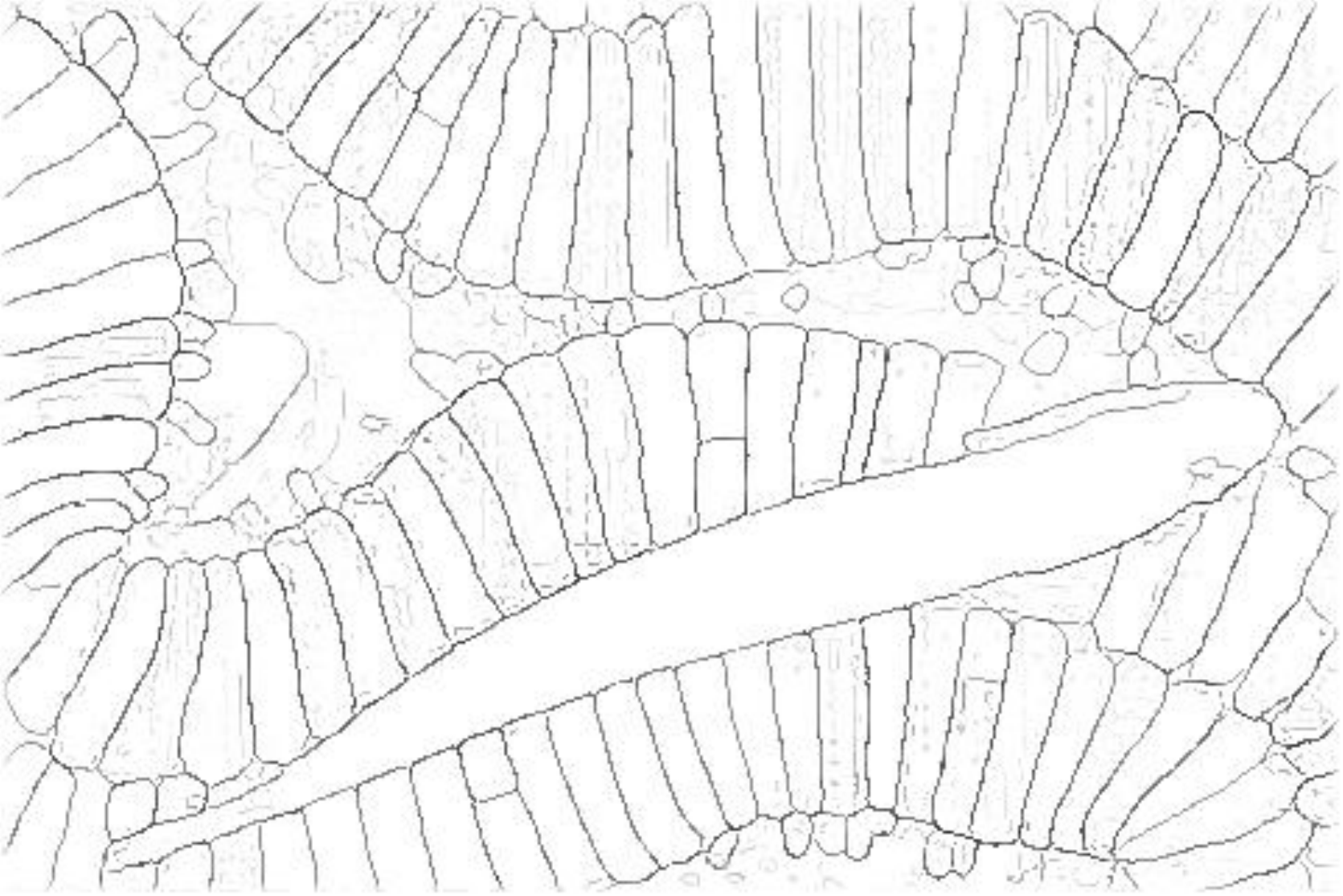} &
\includegraphics[width=0.16\linewidth]{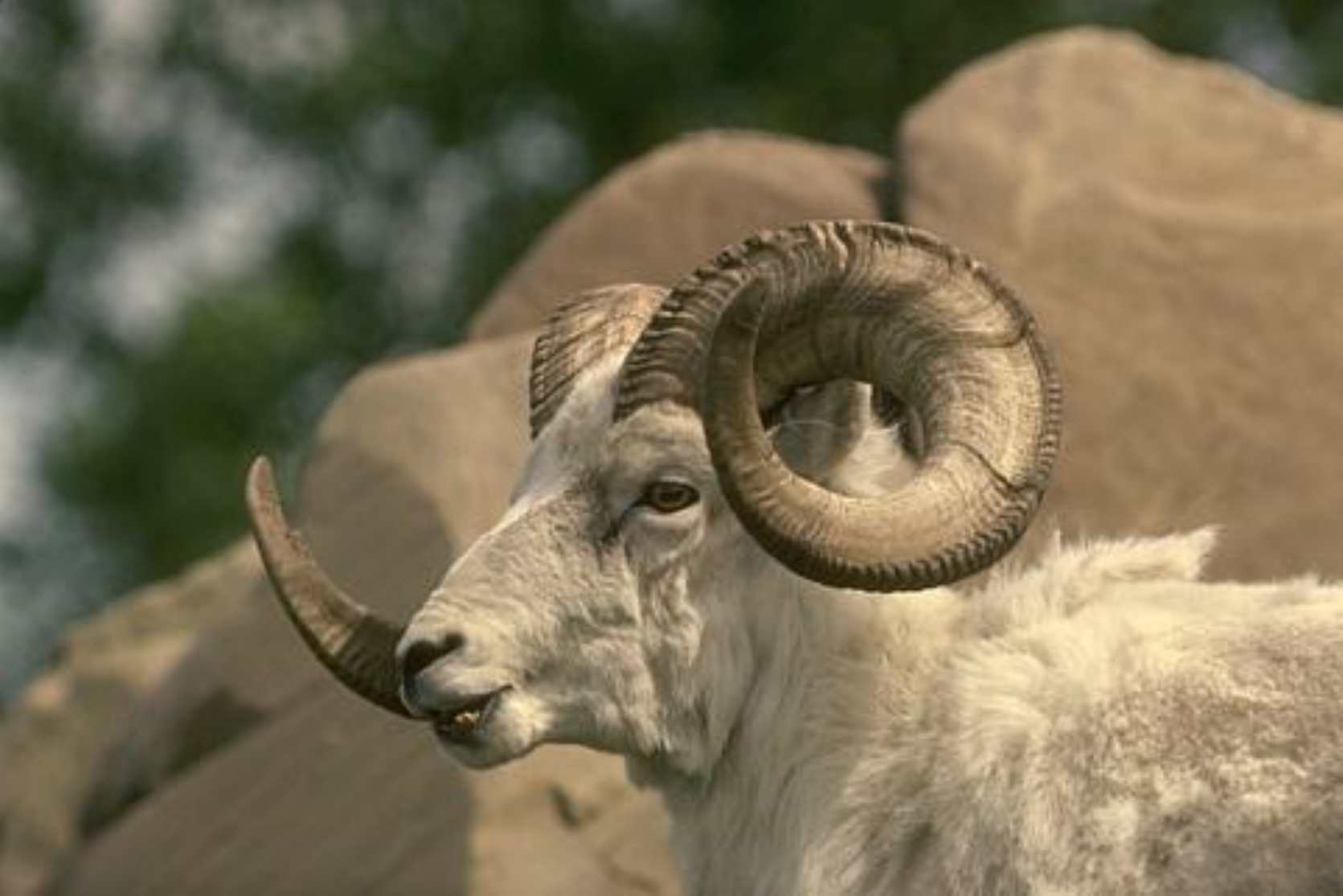} &
\includegraphics[width=0.16\linewidth]{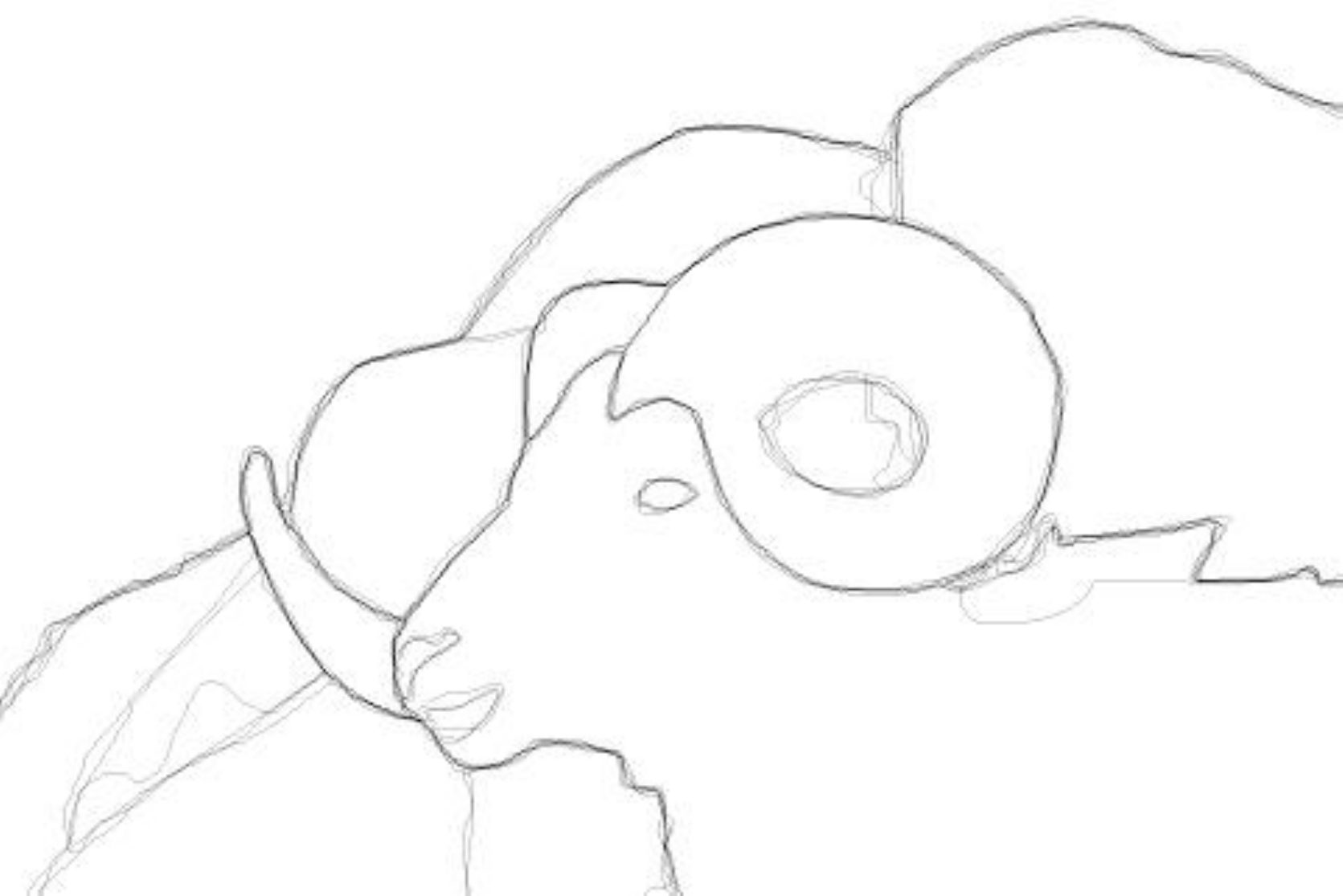} &
\includegraphics[width=0.16\linewidth]{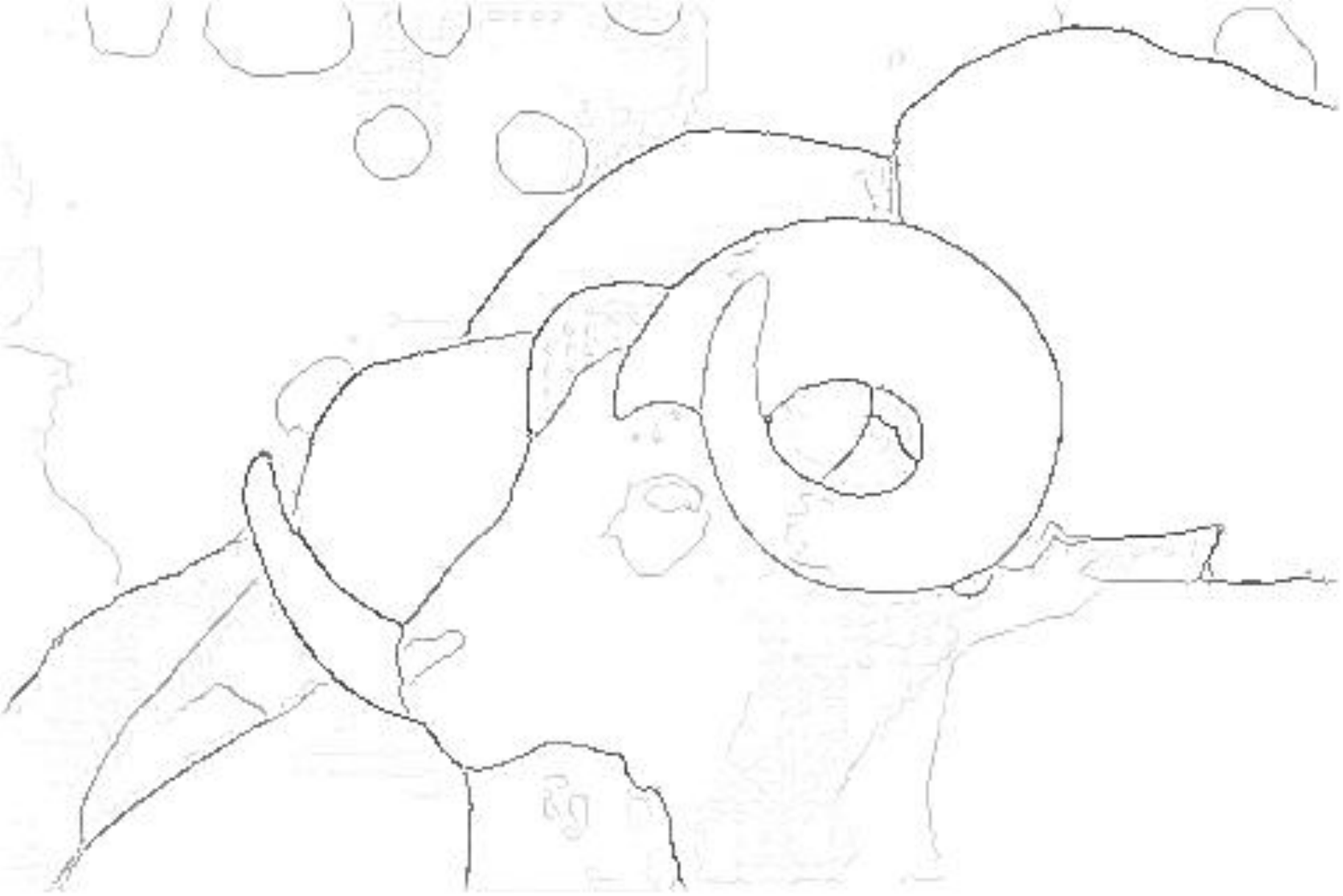}\\
\end{tabular}\vspace{-2mm}
\caption{Additional visualizations of zero-shot edge predictions on BSDS500. Recall that \sam was not trained to predict edge maps and did not have access to BSDS images and annotations during training.}
\label{fig:more_edges}\vspace{-2mm}
\end{figure*}
%##################################################################################################

\subsection{Zero-Shot Edge Detection}\label{app:edges}

\paragraph{Dataset and metrics.} We perform zero-shot edge detection experiments on BSDS500~\cite{martin2001database,arbelaez2010contour}. The ground truth for each image comes from the manual annotations of five different subjects. We report results on the 200 image test subset using the four standard metrics for edge detection~\cite{arbelaez2010contour,dollar2014fast}: optimal dataset scale (ODS), optimal image scale (OIS), average precision (AP), and recall at 50\% precision (R50).

\paragraph{Method.} For zero-shot transfer, we use a simplified version of our automatic mask generation pipeline. We prompt \sam with a 16$\x$16 regular grid of foreground points, which yields 768 predicted masks (three per point). We do not filter by predicted IoU or stability. Redundant masks are removed by NMS. Then we apply a Sobel filter to the remaining masks' unthresholded probability maps and set values to zero if they do not intersect with the outer boundary pixels of a mask. Finally, we take a pixel-wise max over all the predictions, linearly normalize the result to \mbox{[0,1]}, and apply edge NMS~\cite{canny1986computational} to thin the edges.

\paragraph{Visualizations.} In \fig{fig:more_edges}, we show additional examples of zero-shot edge predictions from \sam. These qualitative examples further illustrate how \sam tends to output sensible edge maps, despite not being trained for edge detection. We see that the edges can align well with the human annotations. Although, as previously mentioned, since \sam is not trained for edge detection it does not learn the biases of the BSDS500 dataset and often outputs more edges than are present in the ground truth annotations.

\subsection{Zero-Shot Object Proposals}\label{app:proposals}

\paragraph{Dataset and metrics.} We report the standard average recall (AR) metric for masks at 1000 proposals on the LVIS v1 validation set~\cite{Gupta2019}. Since LVIS has high-quality masks for 1203 object classes, it provides a challenging test for object proposal generation. We focus on AR@1000 due to the open-world nature of our model, which will likely produce many valid masks outside even the 1203 classes in LVIS. To measure performance on frequent, common, and rare categories, we use AR@1000 but measured against a ground truth set containing just the corresponding LVIS categories.

\paragraph{Baseline.} We use cascade ViTDet-H as a baseline, the strongest model from~\cite{li2022exploring} by AP on LVIS. As noted in the main text, an object detector trained in-domain can ``game'' AR~\cite{chavali2016object} and is expected to be a stronger baseline than other models that focus on open-world proposals or segmentation~\cite{kim2022learning,wang2022open}. To produce 1000 proposals, we disable score thresholding in the three cascade stages and as raise the maximum number of predictions per stage to 1000.

\paragraph{Method.} We use a modified version of \sam's automatic mask generation pipeline for zero-shot transfer. First, to make inference time comparable to that of ViTDet we do not process image crops. Second, we remove filtering by predicted IoU and stability. This leaves two tunable parameters to get \app1000 masks per image: the input point grid and the NMS threshold duplicate mask suppression. We choose a 64$\x$64 point grid and an NMS threshold of 0.9, which produces \app900 masks per image on average. At evaluation, if greater than 1000 masks have been proposed in an image, they are ranked by the average of their confidence and stability scores, then truncated to the top 1000 proposals.

We hypothesize that \sam's ability to output multiple masks is especially valuable for this task, since recall should benefit from proposals generated at multiple scales from a single input point. To test this, we compare to an ablated version \sam that only outputs a single mask instead of three (\sam\ - single-output). Since this model produces fewer masks, we further increase the number of points sampled and NMS threshold to 128$\x$128 and 0.95, respectively, obtaining \app950 masks per image on average. Additionally, single-output \sam does not produce the IoU score used to rank masks for NMS in the automatic mask generation pipeline, so instead masks are ranked randomly. Testing suggests this has similar performance to more sophisticated methods of ranking masks, such as using the max logit value of the mask as a proxy for model confidence.

%##################################################################################################
\begin{figure*}[t]\footnotesize\centering
\tablestyle{2pt}{1}
\begin{tabular}[t]{@{}cccccc@{}}\centering
ground truth & ViTDet & \sam & ground truth & ViTDet & \sam\\
\includegraphics[width=0.16\linewidth,trim={5mm 50mm 5mm 10mm},clip]{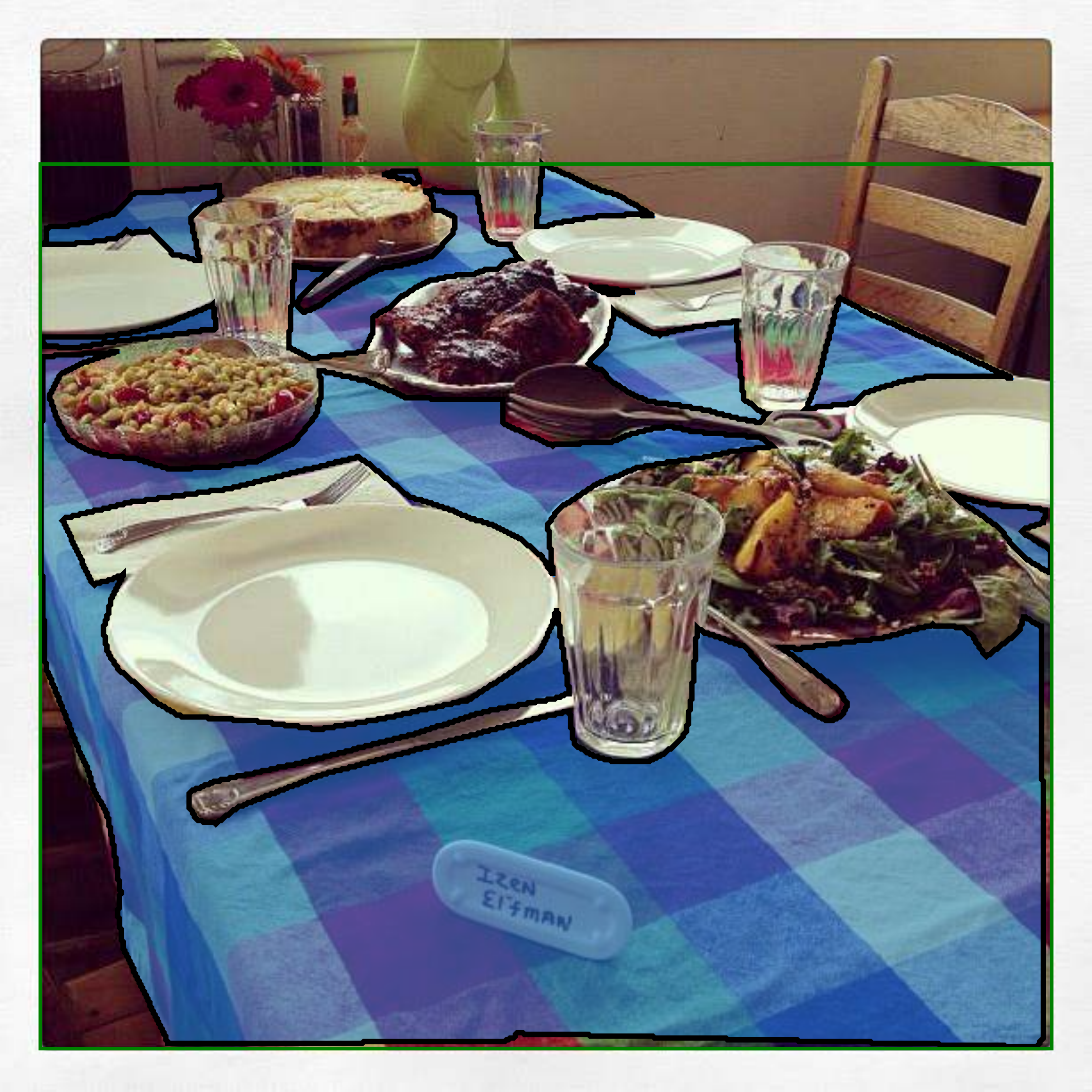} &
\includegraphics[width=0.16\linewidth,trim={5mm 50mm 5mm 10mm},clip]{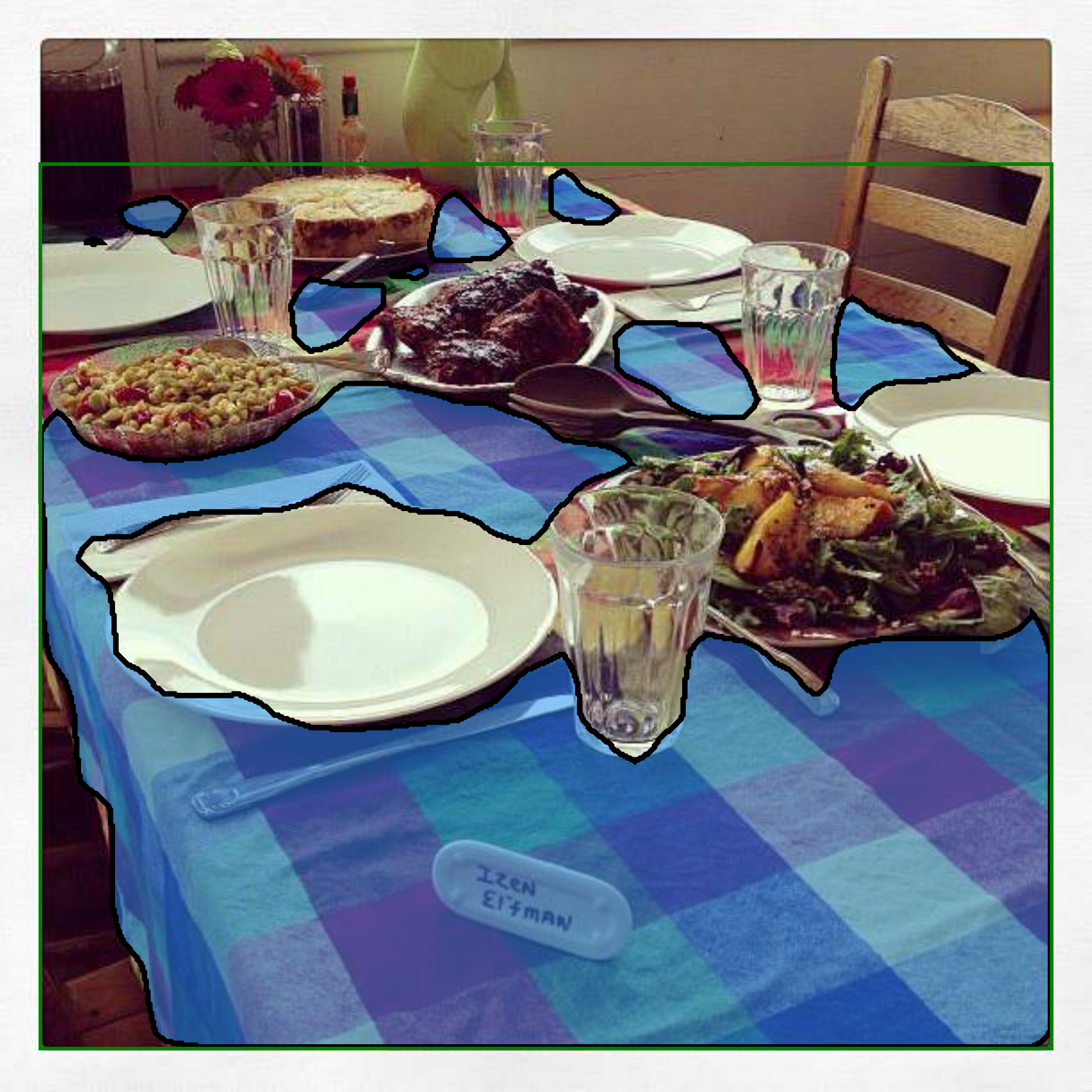} &
\includegraphics[width=0.16\linewidth,trim={5mm 50mm 5mm 10mm},clip]{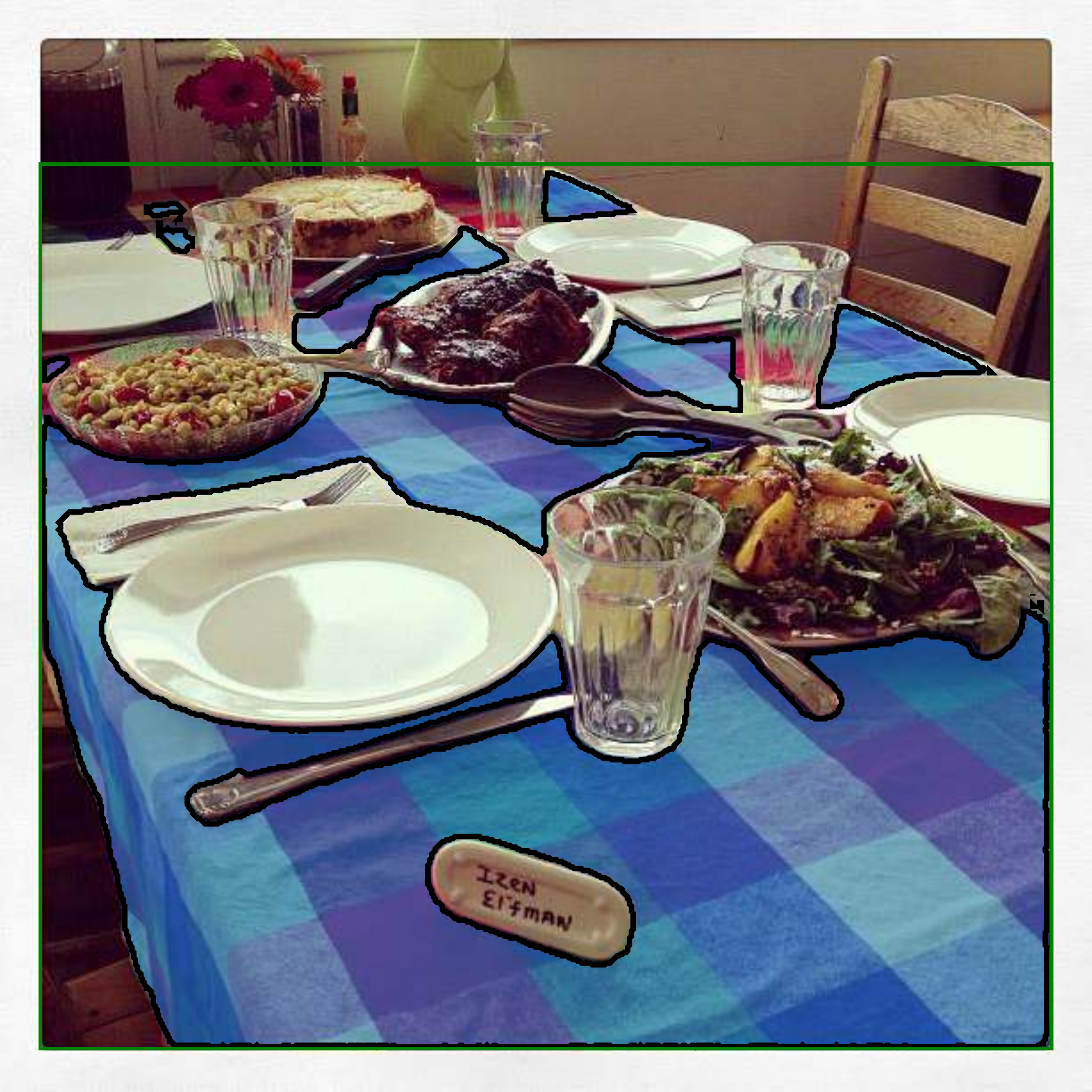} &
\includegraphics[width=0.16\linewidth]{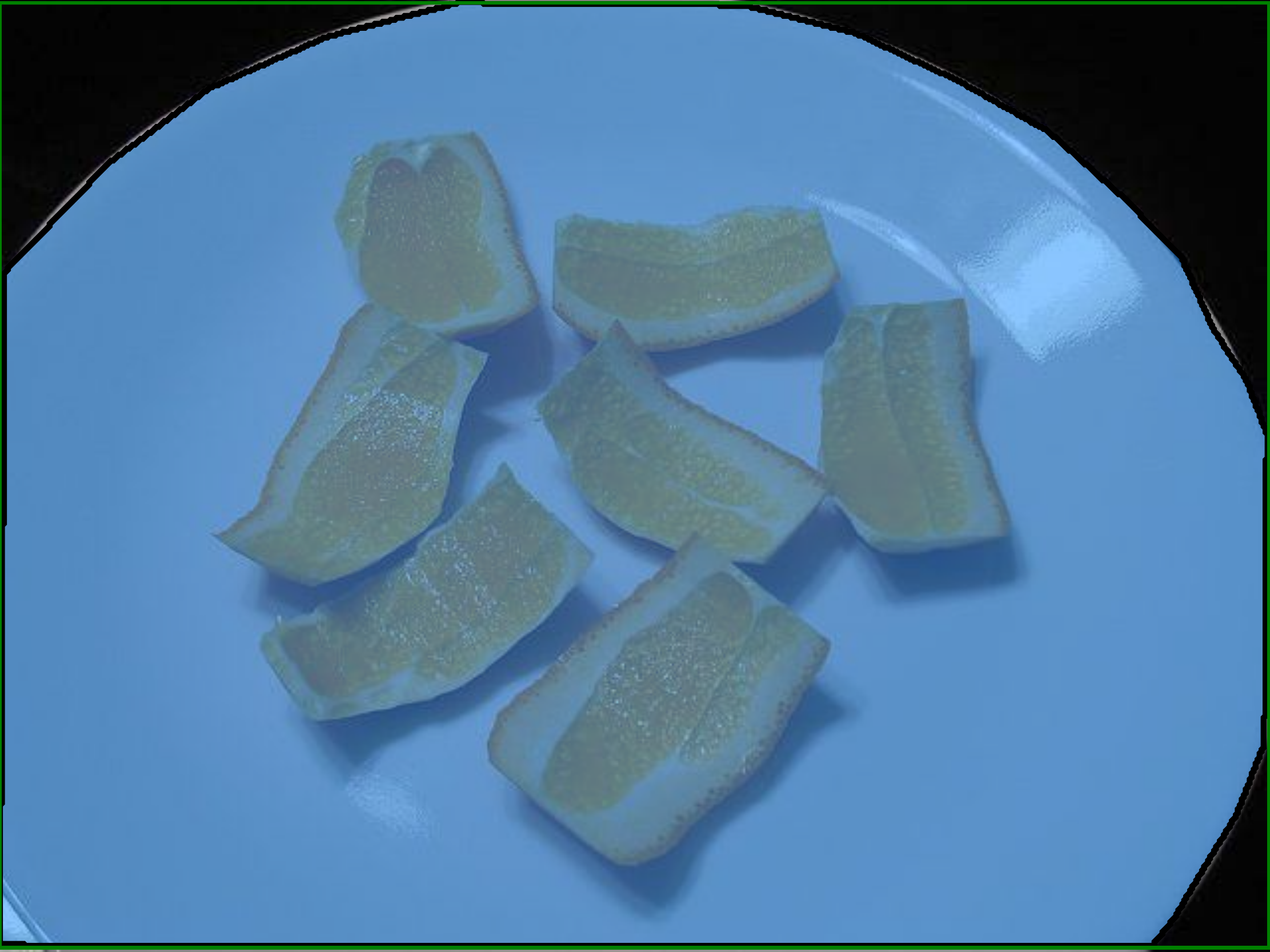} &
\includegraphics[width=0.16\linewidth]{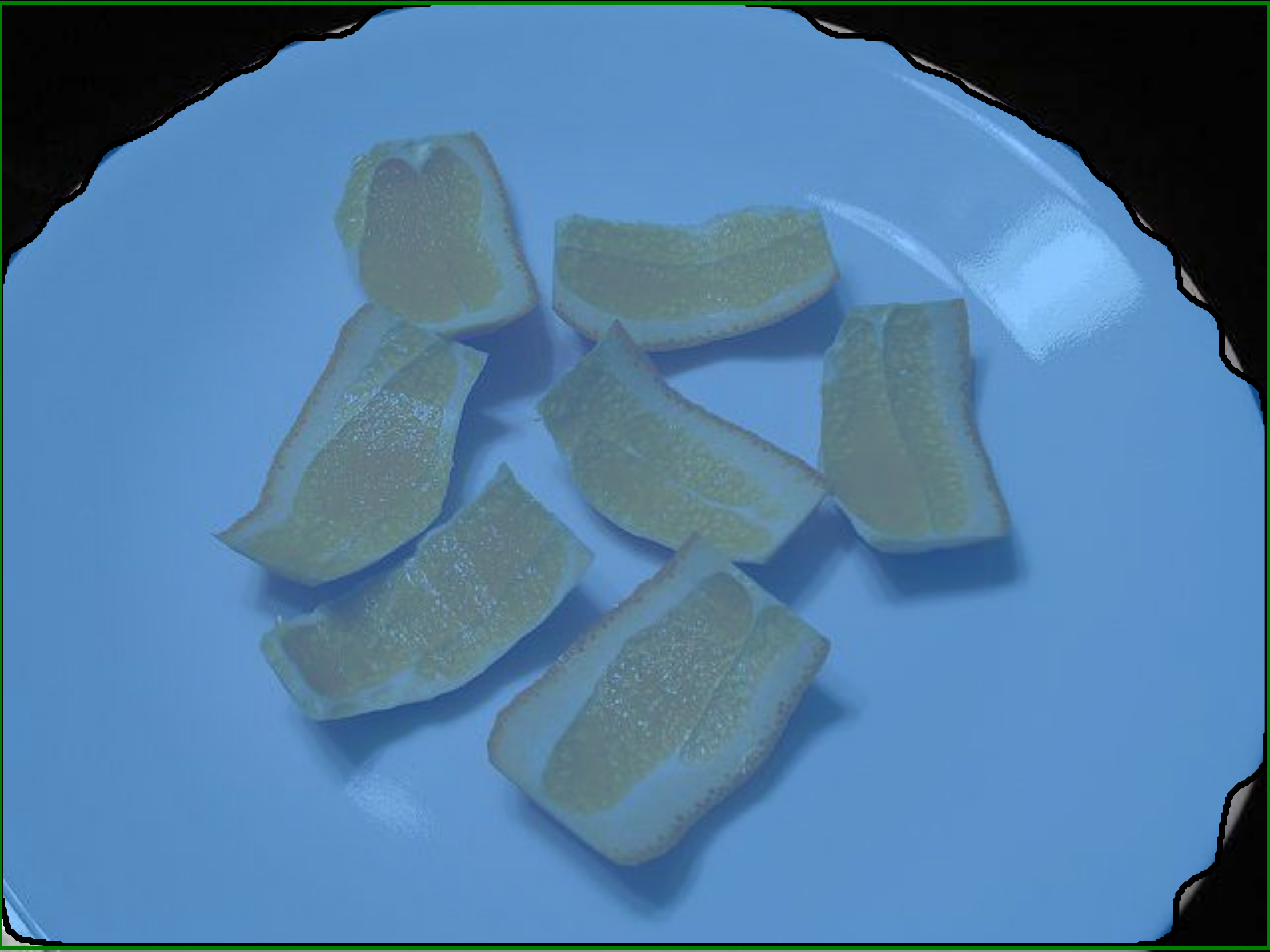} &
\includegraphics[width=0.16\linewidth]{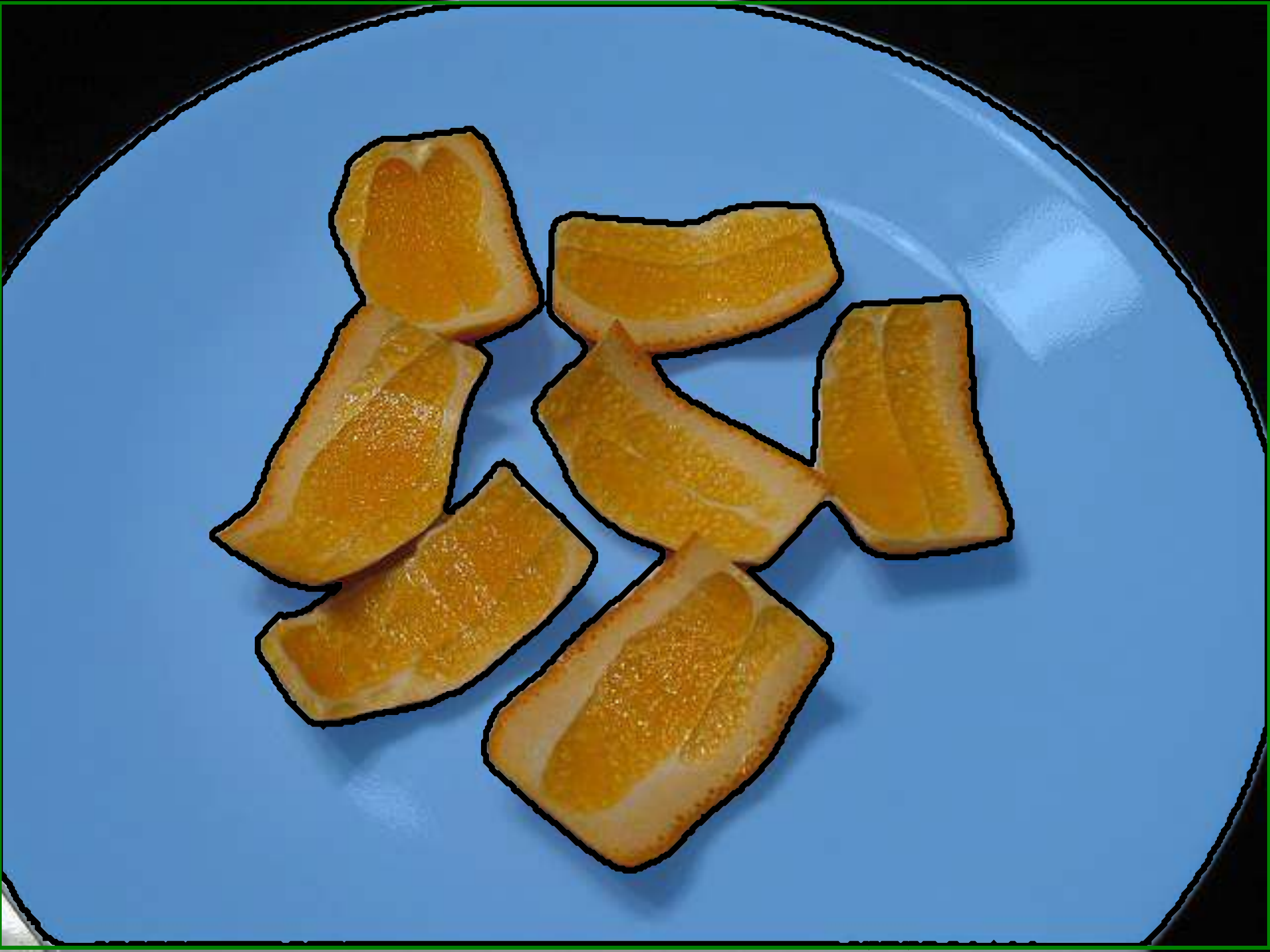}\\
\includegraphics[width=0.16\linewidth]{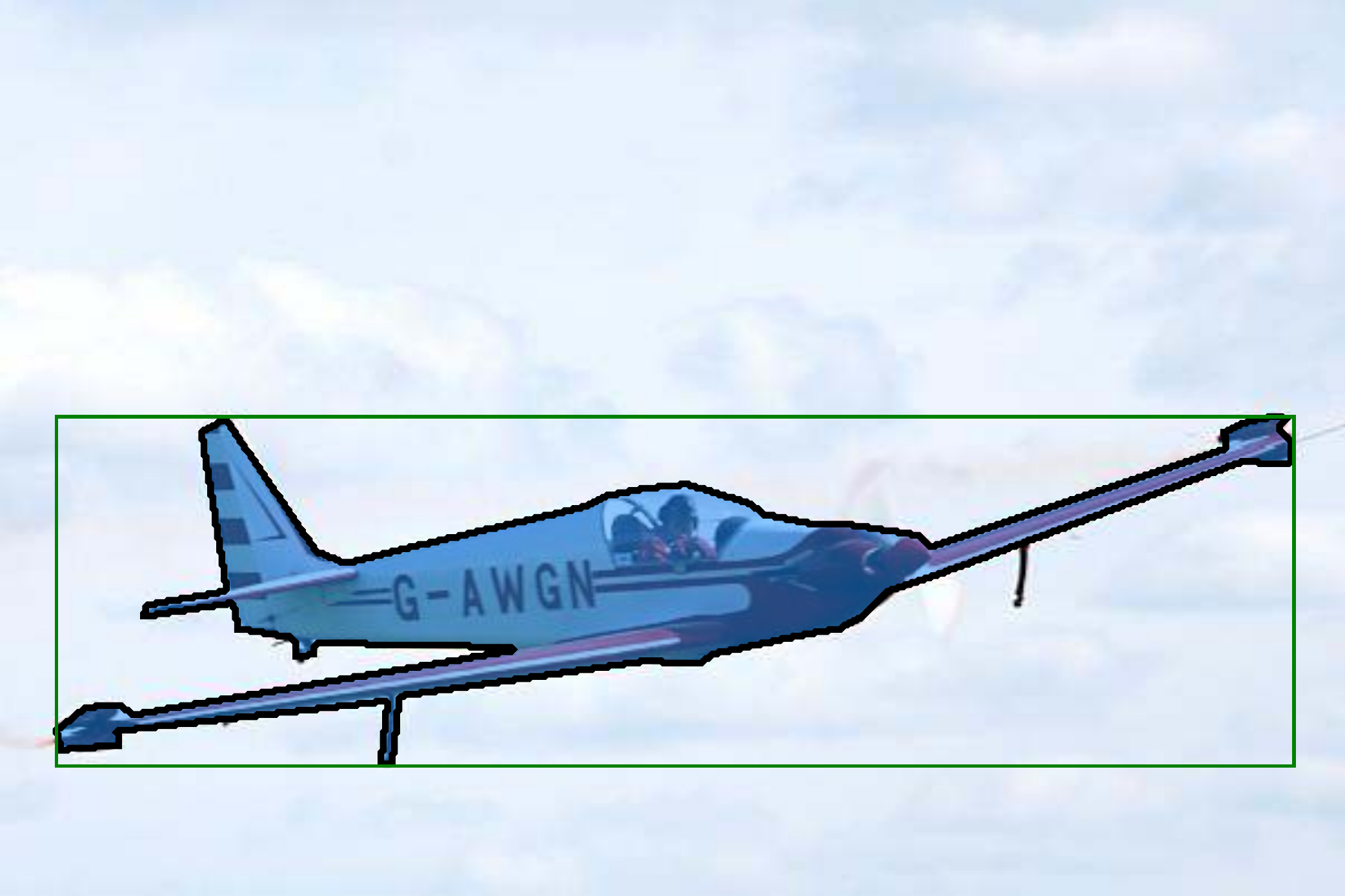} &
\includegraphics[width=0.16\linewidth]{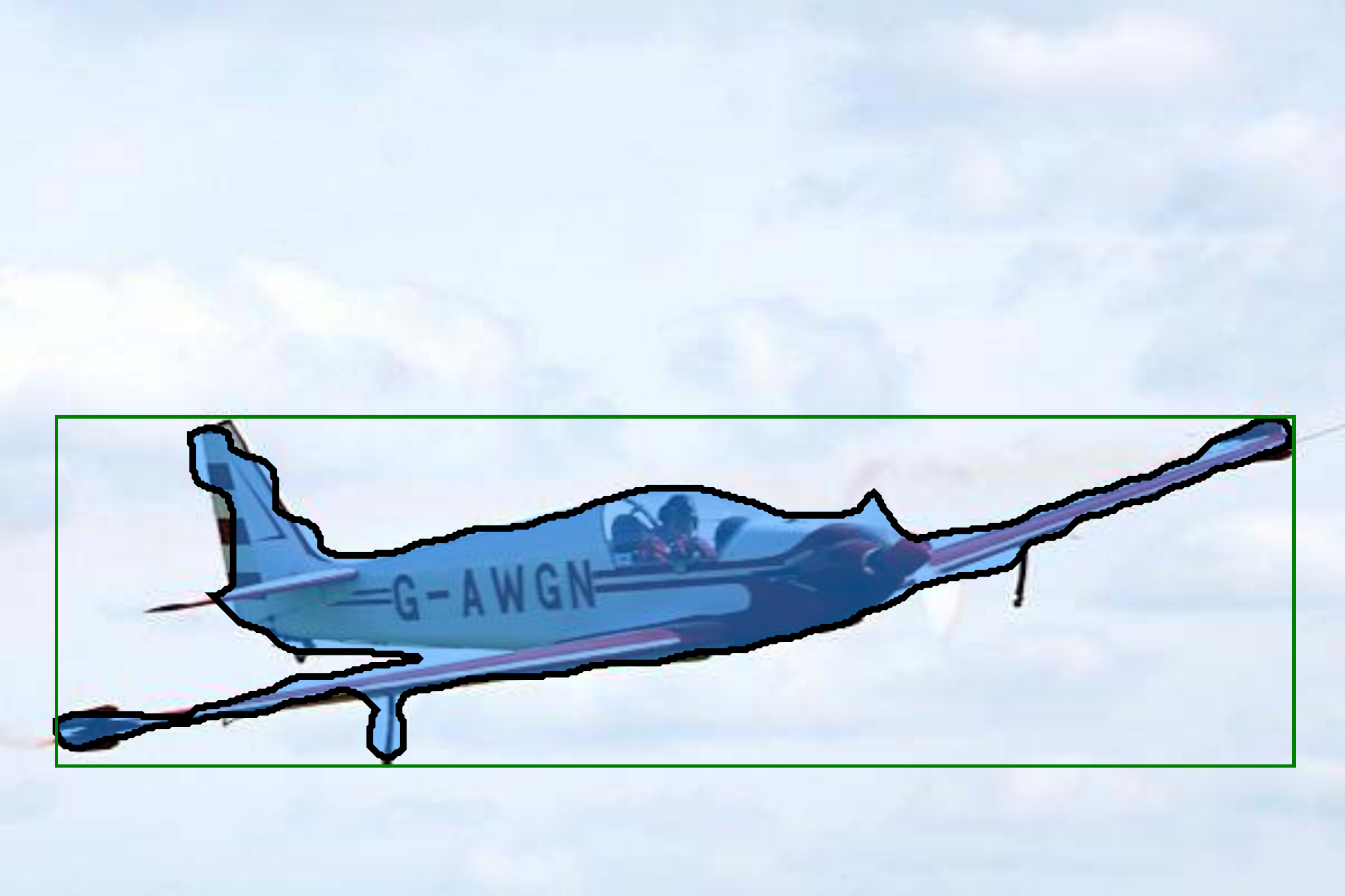} &
\includegraphics[width=0.16\linewidth]{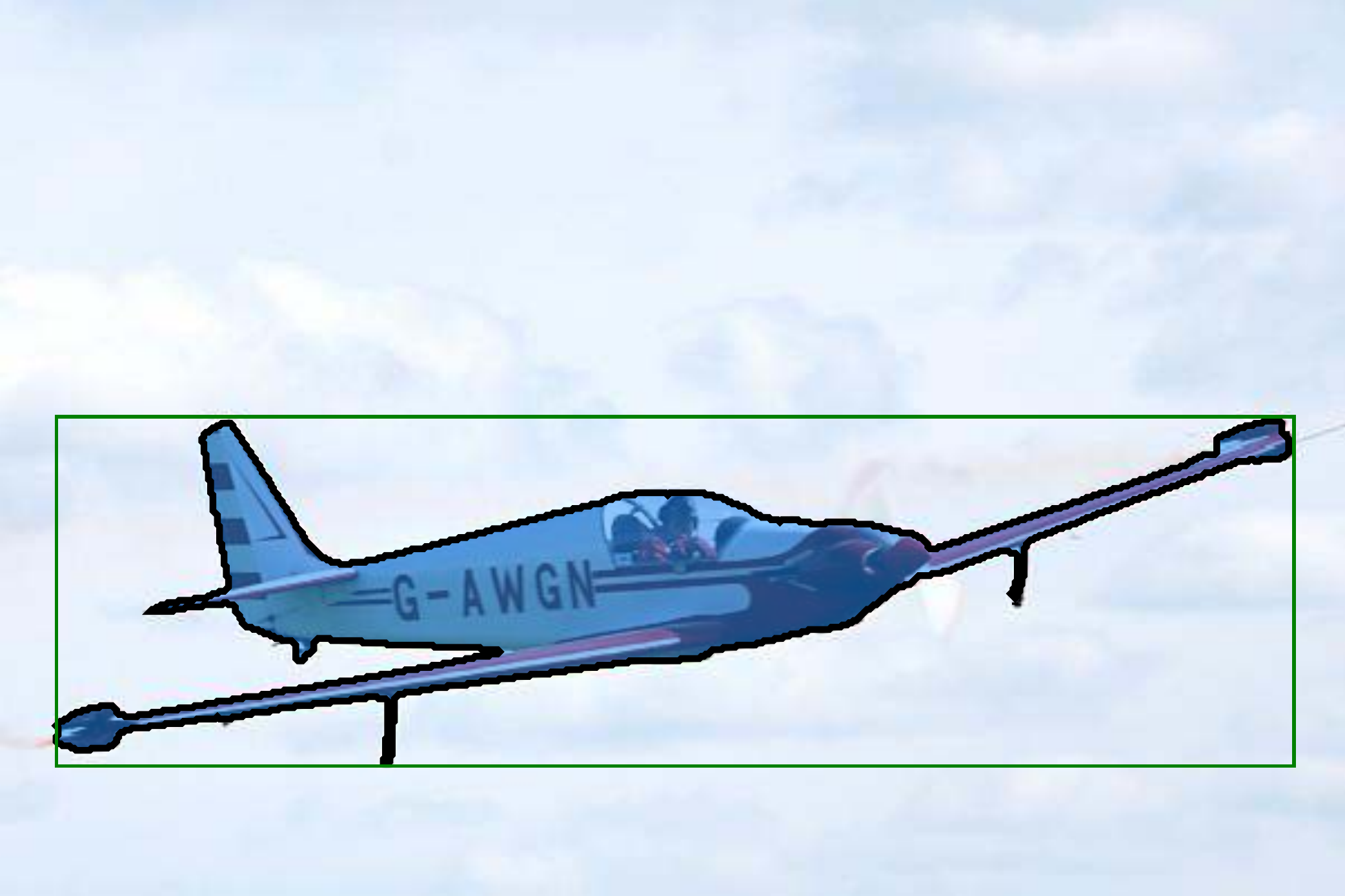} &
\includegraphics[width=0.16\linewidth]{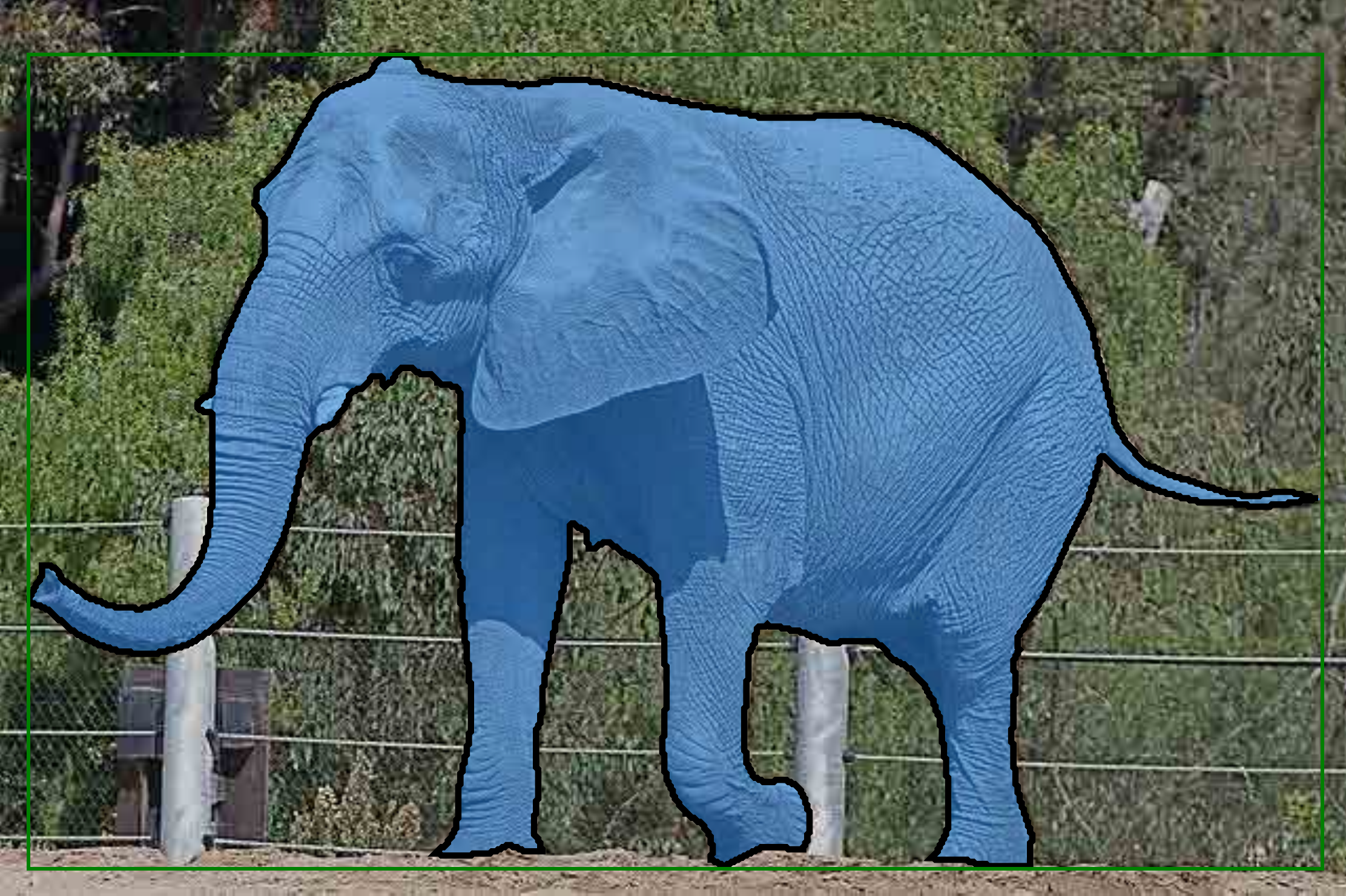} &
\includegraphics[width=0.16\linewidth]{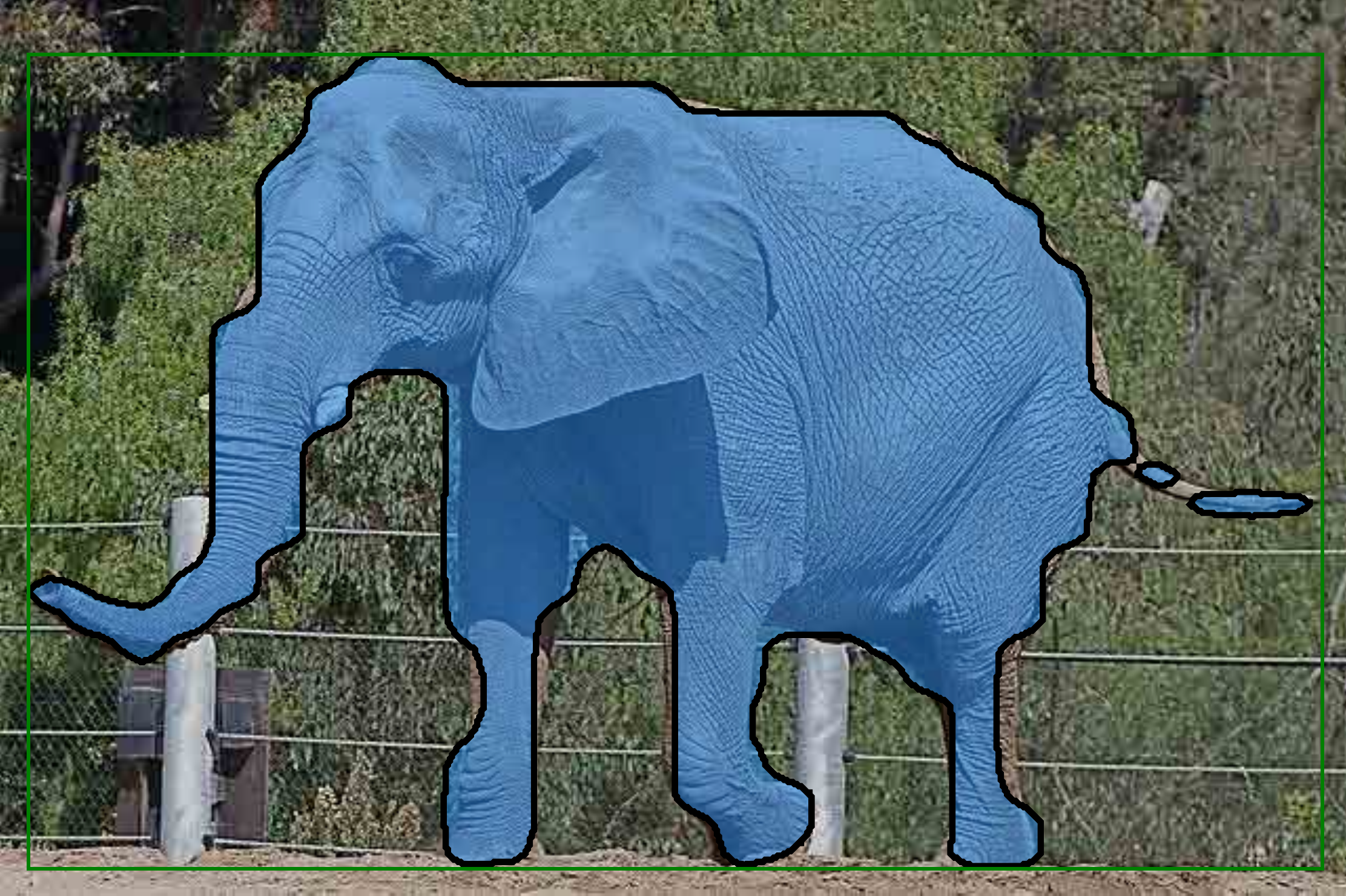} &
\includegraphics[width=0.16\linewidth]{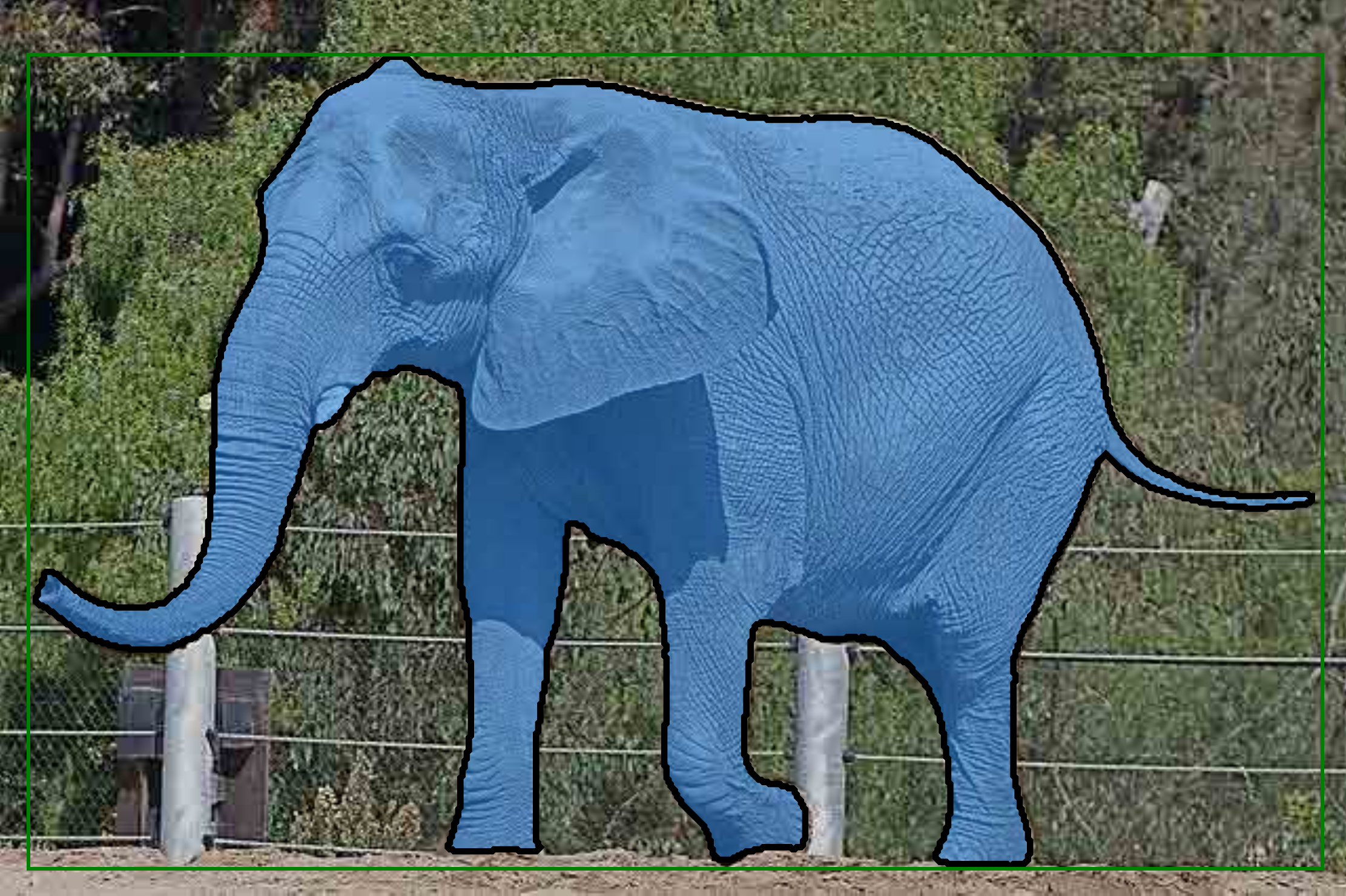}\\
\end{tabular}\vspace{-3mm}
\caption{Zero-shot instance segmentation on LVIS v1. \sam produces higher quality masks than ViTDet. As a zero-shot model, \sam does not have the opportunity to learn specific training data biases; see top-right as an example where \sam makes a modal prediction, whereas the ground truth in LVIS is amodal given that mask annotations in LVIS have no holes.}
\label{fig:instanceseg}\vspace{-3mm}
\end{figure*}
%##################################################################################################

\subsection{Zero-Shot Instance Segmentation}\label{app:instseg}

\paragraph{Method.} For zero-shot instance segmentation, we prompt \sam with the boxes output by a fully-supervised \mbox{ViTDet-H} on COCO and LVIS v1 validation splits. We apply an additional mask refinement iteration by feeding the most confident predicted mask, together with the box prompt, back to the mask decoder to produce the final prediction. We show zero-shot instance segmentations predicted on LVIS in \fig{fig:instanceseg}. Compared to ViTDet, \sam tends to produce higher quality masks with cleaner boundaries. We confirm this observation with human studies in \S\ref{sec:eval:instseg}. Note that as a zero-shot model, \sam is not able to learn annotation biases in a dataset. For instance, we see that \sam makes a valid modal prediction for the plate, whereas LVIS masks cannot contain holes by design so the plate is annotated amodally.

\subsection{Zero-Shot Text-to-Mask}\label{app:text_to_mask}

\paragraph{Model and training.} We use the largest publicly available CLIP model~\cite{Radford2021} ({\tt ViT-L/14@336px}) to compute text and image embeddings, which we $\ell^2$ normalize prior to use. To train \sam, we use masks from the first two stages of our data engine. Moreover, we discard all masks with an area smaller than $\textrm{100}^\textrm{2}$ pixels. We train this model with large-scale jitter~\cite{Ghiasi2021} for 120k iterations with batch size 128. All other training parameters follow our default settings.

\paragraph{Generating training prompts.} To extract an input prompt we first expand the bounding box around each mask by a random factor from 1$\x$ to 2$\x$, square-crop the expanded box to maintain its aspect ratio, and resize it to 336$\x$336 pixels. Before feeding the crop to the CLIP image encoder, with 50\% probability we zero-out pixels outside the mask. To ensure the embedding focuses on the object, we use masked attention in the last layer to restrict attention from the output token to the image positions inside the mask. Finally, our prompt is the output token embedding. For training we supply the CLIP-based prompt first, followed by additional iterative point prompts to refine the prediction.

\paragraph{Inference.} During inference we use the CLIP text encoder without any modifications to create a prompt for \sam. We rely on the fact that text and image embeddings are aligned by CLIP, which allows us to train without any explicit text supervision while using text-based prompts for inference.

%##################################################################################################
\begin{figure}\centering
\includegraphics[width=\linewidth]{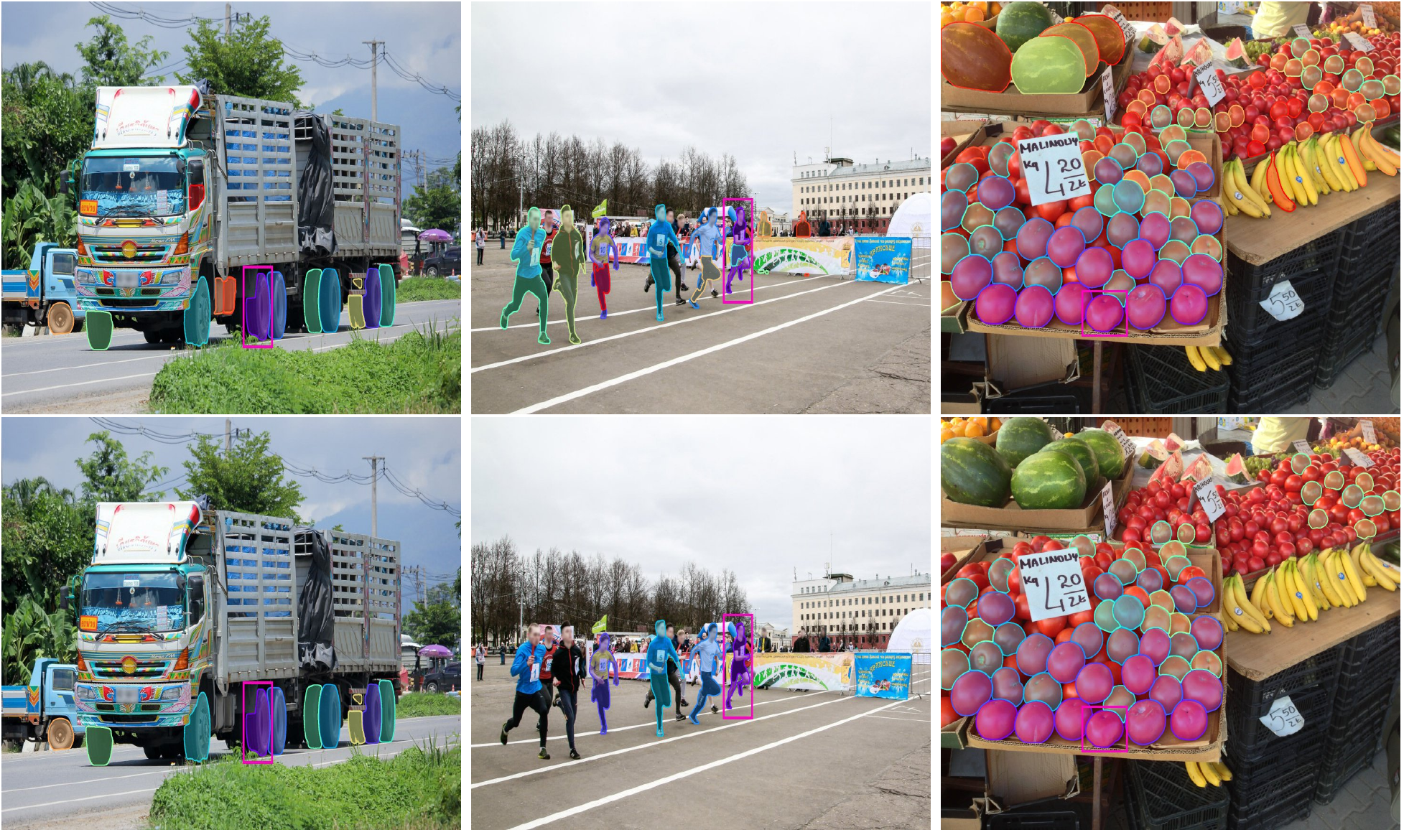}\vspace{-2mm}
\caption{Visualization of thresholding the similarities of mask embeddings from \sam's latent space. A query is indicated by the magenta box; top row shows matches at a low threshold, bottom row at a high threshold. The most similar mask embeddings in the same image can often be semantically similar to the query mask embedding, even though \sam is not trained with explicit semantic supervision.}
\label{fig:latent_visualization}\vspace{-1mm}
\end{figure}
%##################################################################################################

\subsection{Probing the Latent Space of \sam}\label{app:latent}

Finally, we perform an initial investigation to qualitatively probe the latent space learned by \sam. In particular, we are interested in whether \sam is able to capture any semantics in its representation even though is not trained with explicit semantic supervision. To do so, we compute \emph{mask embeddings} by extracting an image embedding from \sam from an image crop around a mask and its horizontally flipped version, multiplying the image embedding by the binary mask, and averaging over spatial locations. In \fig{fig:latent_visualization}, we show 3 examples of a query mask and similar masks (in the latent space) in the same image. We observe that the nearest neighbors for each query show some, albeit imperfect, shape and semantic similarity. Although these results are preliminary, they indicate that the representations from \sam may be useful for a variety of purposes, such as further data labeling, understanding the contents of datasets, or as features for downstream tasks.

%%%%%%%%%%%%%%%%%%%%%%%%%%%%%%%%%%%%%%%%%%%%%%%%%%%%%%%%%%%%%%%%%%%%%%%%%%%%%%%%%%%%%%%%%%%%%%%%%%%
\section{Human Study Experimental Design}\label{app:human_study}

Here we describe details of the human study used to evaluate mask quality in \S\ref{sec:eval:single_point} and \S\ref{sec:eval:instseg}. The purpose of the human study is to address two limitations of using IoU to ground truth as a measure of predicted mask quality. The first limitation is that, for ambiguous inputs such as a single point, the model may be strongly penalized for returning a valid mask of a different object than the ground truth. The second limitation is that ground truth masks may include various biases, such as systematic errors in the edge quality or decisions to modally or amodally segment occluding objects. A model trained in-domain can learn these biases and obtain a higher IoU without necessarily producing better masks. Human review can obtain a measure of mask quality independent of an underlying ground truth mask in order to alleviate these issues.

\paragraph{Models.} For single-point evaluation, we use RITM~\cite{sofiiuk2022reviving}, single-output \sam, and \sam to test two hypotheses. First, we hypothesize that \sam produces visually higher quality masks than baseline interactive segmentation models when given a single point, even when metrics such as IoU with ground truth do not reveal this. Second, we hypothesize that \sam's ability to disambiguate masks improves mask quality for single point inputs, since single output \sam may return masks that average over ambiguous masks.

For instance segmentation experiments, we evaluate cascade \mbox{ViTDet-H}~\cite{li2022exploring} and \sam in order to test the hypothesis that \sam produces visually higher quality masks, even if it obtains a lower AP due to the inability to learn specific annotation biases of the validation dataset.

\paragraph{Datasets.} For single-point experiments, we select 7 datasets from our set of 23 datasets, since the full suite is too large for human review. We choose LVIS v0.5~\cite{chen20223D}, VISOR~\cite{VISOR, EpicKitchens}, DRAM~\cite{DRAM}, IBD~\cite{chen20223D}, NDD20~\cite{ndd20}, OVIS~\cite{ovis}, and iShape~\cite{iShape}, which provide a diverse collection of images, including scene-level, ego-centric, drawn, overhead, underwater, and synthetic imagery. Additionally, this set includes datasets both where \sam outperforms RITM with IoU metrics and vice-versa. For instance segmentation experiments, we use the LVIS v1 validation set, allowing for direct comparison to ViTDet, which was trained on LVIS.

\paragraph{Methodology.} We presented masks generated by the models to professional annotators and asked them to rate each mask using provided guidelines (see \S\ref{app:annotation_guidelines} for the complete guidelines). Annotators were sourced from the same company that collected manually annotated masks for the data engine. An annotator was provided access to an image, the predicted mask of a single model, and the input to the model (either a single point or single box) and asked to judge the mask on three criterion: Does the mask correspond to a valid object? Does the mask have a clean boundary? and Does the mask correspond to the input? They then submitted a rating from 1-10 indicating the overall mask quality.

A score of 1 indicates a mask that corresponds to no object at all; a low score (2-4) indicates that the mask has huge errors, such including huge regions of other objects or having large areas of nonsensical boundaries; a middle score (5-6) indicates masks that are mostly sensible but still have significant semantic or boundary errors; a high score (7-9) indicates masks with only minor boundary errors; and a score of 10 is for masks with no visible errors. Annotators were provided with five different views, each designed to help identify different error types.

For single point experiments, 1000 masks per dataset were selected randomly from the same subsets used for benchmarking zero-shot interactive segmentation (see \S\ref{app:benchmark} for details on these subsets). The model input was the centermost point, calculated as the largest value of the distance transform from the edge of the mask. For instance segmentation experiments, 1000 masks were selected from the LVIS v1 validation set, and the model input was the LVIS ground truth box. In all experiments, masks with a size smaller than $\textrm{24}^\textrm{2}$ pixels were excluded from sampling, to prevent showing raters a mask that was too small to judge accurately. For both memory and display reasons, large images were rescaled to have a max side-length of 2000 before predicting a mask. In all experiments, the same inputs were fed to each model to produce a predicted mask.

For comparison, the ground truth masks from each dataset were also submitted for rating. For single-point experiments, this gave 4000 total rating jobs per dataset (1000 masks each for RITM, \sam single-output, \sam, and ground truth); for instance segmentation experiments, it gave 3000 total jobs (ViTDet, \sam, and ground truth).

%##################################################################################################
\begin{figure*}\centering\vspace{5mm}
\begin{subfigure}{0.48\linewidth}
\includegraphics[width=\textwidth]{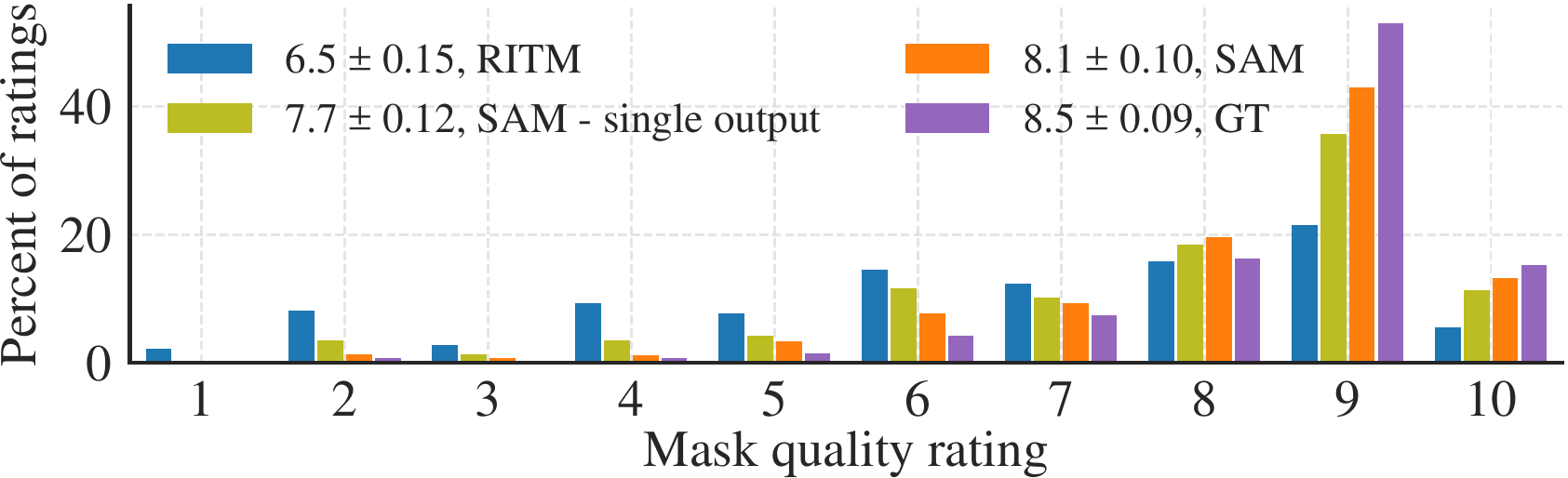} \vspace{-6mm}
\caption{LVIS v0.5~\cite{chen20223D}}
\end{subfigure} ~
\begin{subfigure}{0.48\linewidth}
\includegraphics[width=\textwidth]{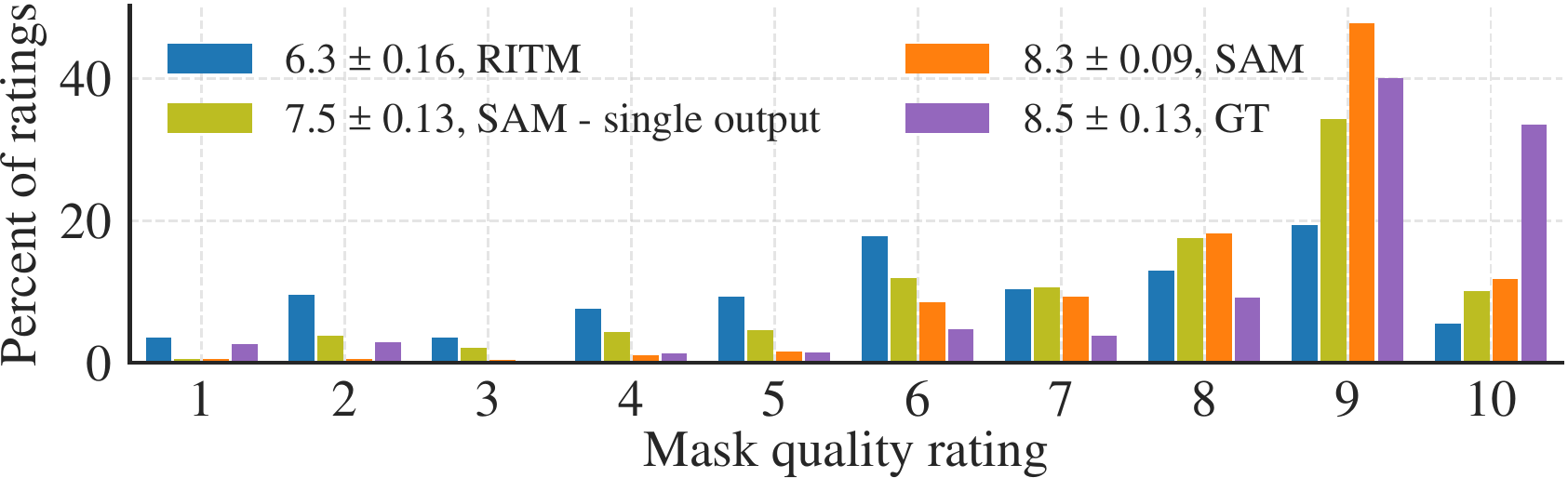} \vspace{-6mm}
\caption{VISOR~\cite{VISOR, EpicKitchens}}
\end{subfigure}\\
\begin{subfigure}{0.48\linewidth}
\includegraphics[width=\textwidth]{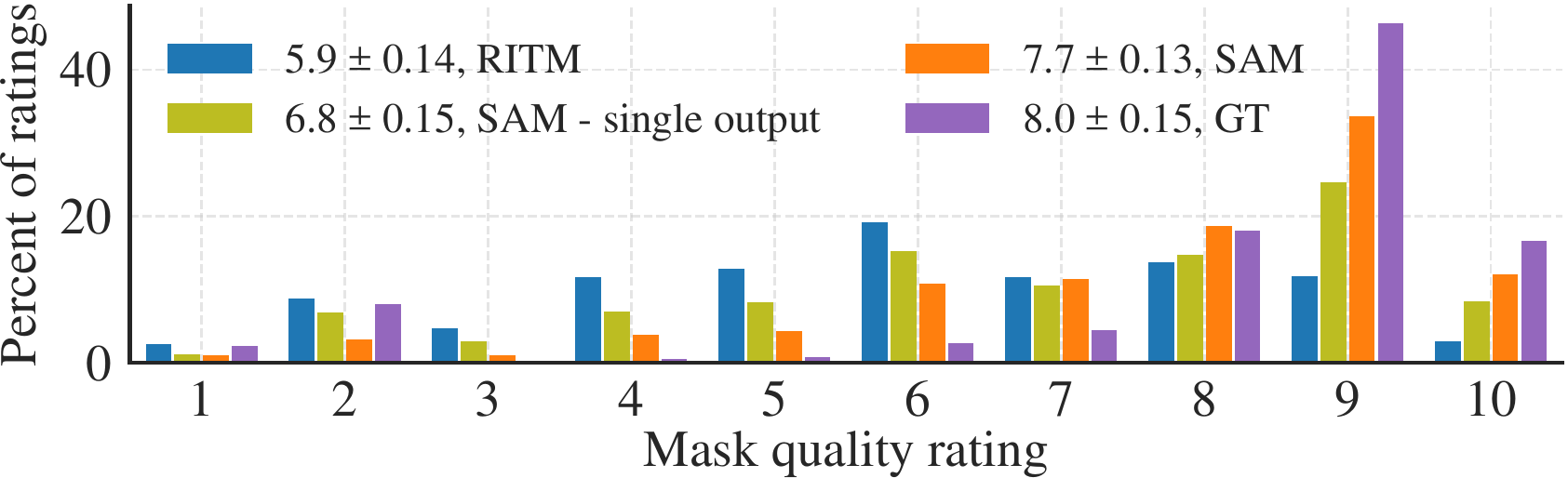} \vspace{-6mm}
\caption{DRAM~\cite{DRAM}}
\end{subfigure} ~
\begin{subfigure}{0.48\linewidth}
\includegraphics[width=\textwidth]{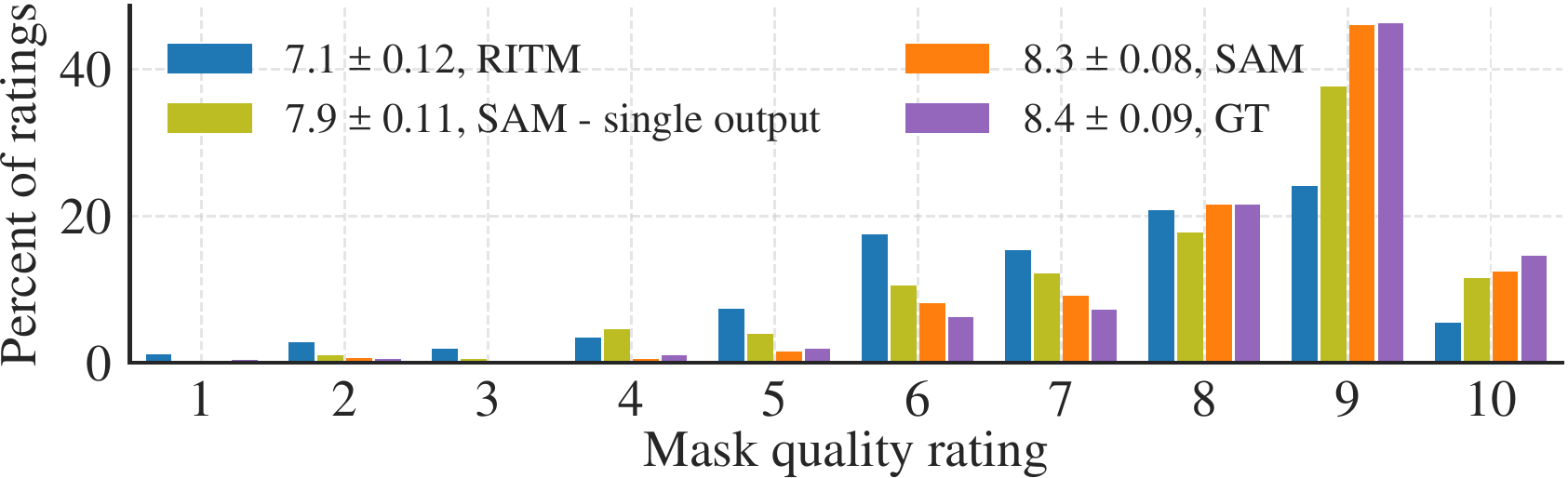} \vspace{-6mm}
\caption{IBD~\cite{chen20223D}}
\end{subfigure}\\
\begin{subfigure}{0.48\linewidth}
\includegraphics[width=\textwidth]{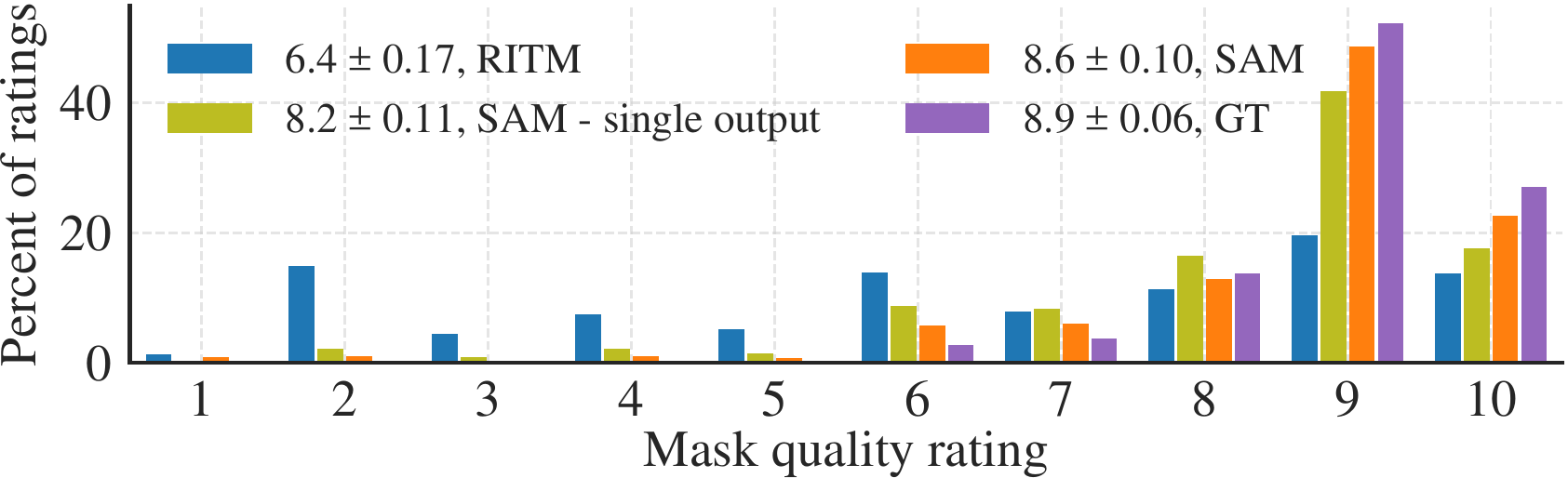} \vspace{-6mm}
\caption{NDD20~\cite{ndd20}}
\end{subfigure} ~
\begin{subfigure}{0.48\linewidth}
\includegraphics[width=\textwidth]{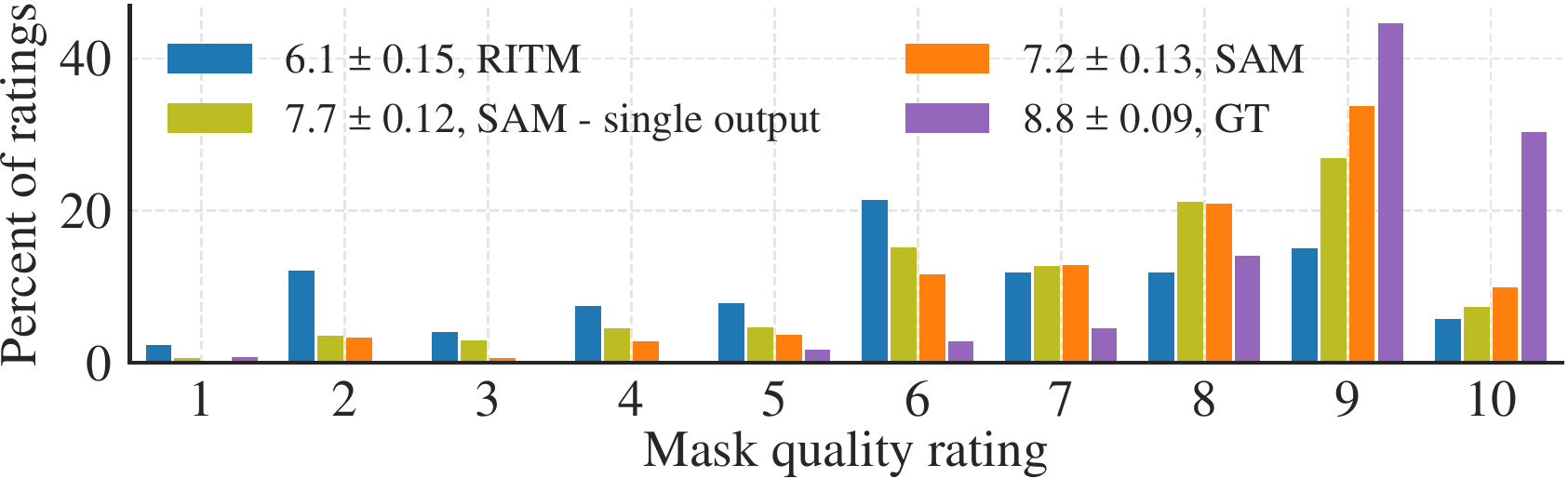} \vspace{-6mm}
\caption{OVIS~\cite{ovis}}
\end{subfigure}\\
\begin{subfigure}{0.48\linewidth}
\includegraphics[width=\textwidth]{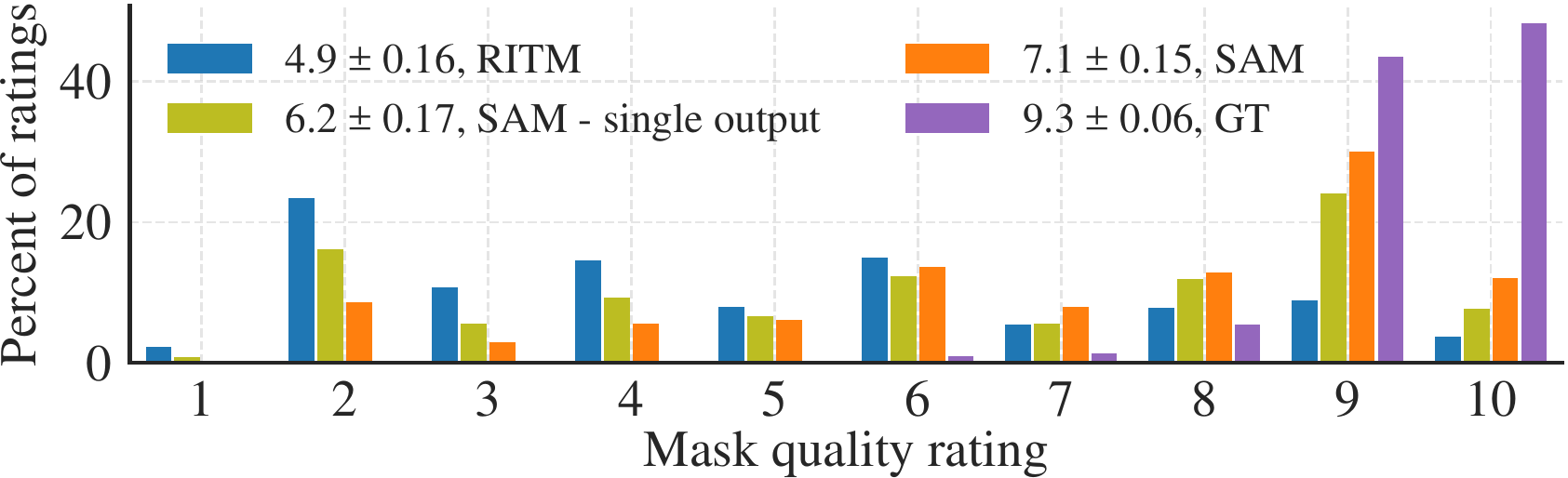} \vspace{-6mm}
\caption{iShape~\cite{iShape}}
\end{subfigure} \hfill \vspace{-1mm}
\caption{Mask quality rating distributions by dataset from our human evaluation study.}
\label{fig:humanstudy:points}
\end{figure*}
%##################################################################################################

For each dataset, these jobs were inserted with random ordering into a queue from which 30 annotators drew jobs. In initial testing of the review study, we provided each job to five different annotators and found reasonable consistency in scores: the average standard deviation in score over the five annotators was 0.83. Additionally, the annotation company deployed quality assurance testers who spot checked a fraction of results for extreme departures from the guidelines. Thus for our experiments each job (\ie, rating one mask in one image) was completed by only a single annotator. Average time spent per annotator per job was 90 seconds, longer than our initial target of 30 seconds, but still sufficiently fast to collect a large number of ratings on each of the 7 selected datasets.

%##################################################################################################
\begin{table}[t]
\centering
\tablestyle{2pt}{1.1}
\footnotesize
\begin{tabular}{@{}l x{40}x{40}| x{40}x{38}}
\multirow{2}{*}{} & \multicolumn{2}{c|}{\sam $>$ baseline} & \multicolumn{2}{c}{\sam $>$ \sam single out.}\\
dataset & p-value & $\mathrm{CI}_{99}(\Delta \mu)$ & p-value & $\mathrm{CI}_{99}(\Delta \mu)$ \\
\hline
\multicolumn{5}{@{}l}{\emph{point input (RITM~\cite{sofiiuk2022reviving} baseline):}} \\
LVIS v0.5~\cite{Gupta2019} & 4e-69 & (1.40, 1.84) & 2e-11 & (0.29, 0.64)\\
VISOR~\cite{VISOR,EpicKitchens}& 7e-98 & (1.81, 2.24) & 7e-26 & (0.58, 0.94) \\
DRAM~\cite{DRAM} &1e-76 & (1.54, 2.00) & 2e-24 & (0.62, 1.03) \\
IBD~\cite{chen20223D} & 2e-57 & (1.03, 1.39) & 1e-15& (0.32, 0.62) \\
NDD20~\cite{ndd20} & 2e-86& (1.88, 2.37) & 5e-08& (0.19, 0.55) \\
OVIS~\cite{ovis} &2e-64 &(1.38, 1.84) &3e-10 &(0.27, 0.63) \\
iShape~\cite{iShape} & 2e-88& (1.97, 2.47) & 7e-23& (0.65, 1.10) \\
\multicolumn{5}{@{}l}{\emph{box input (ViTDet-H~\cite{li2022exploring} baseline):}} \\
LVIS v1 \cite{Gupta2019} & 2e-05 & (0.11, 0.42) & N/A & N/A
\end{tabular}
\vspace{-2mm}
\caption{Statistical tests showing significance that \sam has higher mask quality ratings than baseline and single-output \sam. P-values are calculated by paired t-test, while confidence intervals for the difference in mean scores are calculated by paired bootstrap on 10k samples. All p-values are significant, and all confidence intervals exclude zero.}
\label{tab:humanreviewpvalues}
\end{table}
%##################################################################################################

\paragraph{Results.} \fig{fig:humanstudy:points} shows histograms over ratings for each dataset in the single-point experiments. We run statistical tests for two hypotheses: (1) that \sam gets higher scores than the baseline model (RITM or ViTDet) and (2) that \sam gets higher scores than single-output \sam. P-values are calculated via a paired t-test on the means of the model scores, which we supplement with a paired bootstrap test on 10k samples to find the 99\% confidence interval for the difference of means. Table \ref{tab:humanreviewpvalues} shows p-values and confidence intervals for these tests. All statistical tests are strongly significant, and all confidence intervals exclude zero.

For instance segmentation, \fig{fig:humanstudy:inst} of the main text shows the histogram for ratings. To compare to COCO ground truth, we additionally include 794 ratings of COCO ground truth masks that were collected during our testing of the human review process. These masks were presented to raters using an identical setup as the LVIS results. For fair comparison, results for LVIS in \fig{fig:humanstudy:inst} were subsampled to the same 794 inputs for each model and ground truth. For Table \ref{tab:humanreviewpvalues}, the full 1000 ratings are used to run statistical tests, which show that \sam's mask quality improvement over ViTDet is statistically significant.

%%%%%%%%%%%%%%%%%%%%%%%%%%%%%%%%%%%%%%%%%%%%%%%%%%%%%%%%%%%%%%%%%%%%%%%%%%%%%%%%%%%%%%%%%%%%%%%%%%%
\section{Dataset, Annotation, and Model Cards}\label{app:cards}

In \S\ref{app:datacard} we provide a Dataset Card for \sad, following~\cite{gebru2021datasheets}, in a list of questions and answers. Next, we provide a Data Annotation Card in \S\ref{app:crowdworksheets} for the first two stages of our data engine described in \S\ref{sec:engine}, following CrowdWorkSheets~\cite{diaz2022crowdworksheets}, again as a list of questions and answers. We provide a Model Card following~\cite{mitchell2019model} in Table~\ref{tab:modelcard}.

\subsection{Dataset Card for \sad}\label{app:datacard}
{\fontsize{7.5}{8.4}\selectfont
\paragraph{Motivation}
\begin{enumeratecard}
\item \qtxt{For what purpose was the dataset created? Was there a specific task in mind? Was there a specific gap that needed to be filled? Please provide a description.} The contributions of our dataset to the vision community are fourfold: (1) We release a dataset of 11M images and 1.1B masks, by far the largest segmentation dataset to date. (2) The dataset we release is privacy protecting: we have blurred faces and license plates in all images. (3) The dataset is licensed under a broad set of terms of use which can be found at \href{https://ai.facebook.com/datasets/segment-anything}{https://ai.facebook.com/datasets/segment-anything}. (4) The data is more geographically diverse than its predecessors, and we hope it will bring the community one step closer to creating fairer and more equitable models.
\item \qtxt{Who created the dataset (\eg, which team, research group) and on behalf of which entity (\eg, company, institution, organization)?} The dataset was created by the FAIR team of Meta AI. The underlying images were collected and licensed from a third party photo company.
\item \qtxt{Who funded the creation of the dataset? If there is an associated grant, please provide the name of the grantor and the grant name and number.} Meta AI funded the creation of the dataset.
\item \qtxt{Any other comments?} No.
\end{enumeratecard}
\paragraph{Composition}
\begin{enumeratecard}
\item \qtxt{What do the instances that comprise the dataset represent (\eg, documents, photos, people, countries)? Are there multiple types of instances (\eg, movies, users, and ratings; people and interactions between them; nodes and edges)? Please provide a description.} All of the instances in the dataset are photos. The photos vary in subject matter; common themes of the photo include: locations, objects, scenes. All of the photos are distinct, however there are some sets of photos that were taken of the same subject matter.
\item \qtxt{How many instances are there in total (of each type, if appropriate)?} There are 11 million images.
\item \qtxt{Does the dataset contain all possible instances or is it a sample (not necessarily random) of instances from a larger set? If the dataset is a sample, then what is the larger set? Is the sample representative of the larger set (\eg, geographic coverage)? If so, please describe how this representativeness was validated/verified. If it is not representative of the larger set, please describe why not (\eg, to cover a more diverse range of instances, because instances were withheld or unavailable).} The dataset is composed of images licensed from a photo provider. The dataset contains all instances licensed. The images are photos, \ie not artwork, although there are a few exceptions. The dataset includes all generated masks for each image in the dataset. We withheld \app2k randomly selected images for testing purposes.
\item \qtxt{What data does each instance consist of? ``Raw'' data (\eg, unprocessed text or images) or features? In either case, please provide a description.} Each instance in the dataset is an image. The images were processed to blur faces and license plates to protect the identities of those in the image.
\item \qtxt{Is there a label or target associated with each instance? If so, please provide a description.} Each image is annotated with masks. There are no categories or text associated with the masks. The average image has \app100 masks, and there are \app1.1B masks in total.
\item \qtxt{Is any information missing from individual instances? If so, please provide a description, explaining why this information is missing (\eg, because it was unavailable). This does not include intentionally removed information, but might include, \eg, redacted text.} Yes. Each image is accompanied by a short caption that describes the content and place of the photo in a free form text. Per our agreement with the photo provider we are not allowed to release these captions. However, we use them in our paper to analyze the geographical distribution of the dataset.
\item \qtxt{Are relationships between individual instances made explicit (\eg, users' movie ratings, social network links)? If so, please describe how these relationships are made explicit.} No, there are no known relationships between instances in the dataset.
\item \qtxt{Are there any errors, sources of noise, or redundancies in the dataset? If so, please provide a description.} \textit{Errors:} The masks are generated by a segmentation model, so there may be errors or inconsistencies in the masks. \textit{Redundancies:} While no two images are the same, there are instances of images of the same subject taken close together in time.
\item \qtxt{Is the dataset self-contained, or does it link to or otherwise rely on external resources (\eg, websites, tweets, other datasets)? If it links to or relies on external resources, a) are there guarantees that they will exist, and remain constant, over time; b) are there official archival versions of the complete dataset (\ie, including the external resources as they existed at the time the dataset was created); c) are there any restrictions (\eg, licenses, fees) associated with any of the external resources that might apply to a dataset consumer? Please provide descriptions of all external resources and any restrictions associated with them, as well as links or other access points, as appropriate.} The dataset is self-contained.
\item \qtxt{Does the dataset contain data that might be considered confidential (\eg, data that is protected by legal privilege or by doctor-patient confidentiality, data that includes the content of individuals' non-public communications)? If so, please provide a description.} No.
\item \qtxt{Does the dataset contain data that, if viewed directly, might be offensive, insulting, threatening, or might otherwise cause anxiety? If so, please describe why.} We have two safety measures to prevent objectionable content: (1) Photos are licensed from a photo provider and had to meet the terms of service of the photo provider. We requested that all objectionable content be filtered from the images we licensed. (2) If a user observes objectionable image(s) in the dataset, we invite them to report the image(s) at \href{mailto:segment-anything@meta.com}{segment-anything@meta.com} for removal. Despite the measures taken, we observe that a small portion of images contains scenes of protests or other gatherings that focus on a diverse spectrum of religious beliefs or political opinions that may be offensive. We were not able to produce a filtering strategy that removes all such images and rely on users to report this type of content.
\item \qtxt{Does the dataset identify any subpopulations (\eg, by age, gender)? If so, please describe how these subpopulations are identified and provide a description of their respective distributions within the dataset.} The dataset does not identify any subpopulations of the people in the photos.
\item \qtxt{Is it possible to identify individuals (\ie, one or more natural persons), either directly or indirectly (\ie, in combination with other data) from the dataset? If so, please describe how.} No. Images were subjected to a face blurring model to remove any personally identifiable information. If a user observes any anonymization issue, we invite them to report the issue and the image id(s) at \href{mailto:segment-anything@meta.com}{segment-anything@meta.com}.
\item \qtxt{Does the dataset contain data that might be considered sensitive in any way (\eg, data that reveals race or ethnic origins, sexual orientations, religious beliefs, political opinions or union memberships, or locations; financial or health data; biometric or genetic data; forms of government identification, such as social security numbers; criminal history)? If so, please provide a description.} The dataset contains scenes of protests, or other gatherings that may suggest religious beliefs, political opinions or union memberships. However, the faces of all people in the dataset have been anonymized via facial blurring, so it is not possible to identify any person in the dataset.
\item \qtxt{Any other comments?} No.
\end{enumeratecard}
\paragraph{Collection Process}
\begin{enumeratecard}
\item \qtxt{How was the data associated with each instance acquired? Was the data directly observable (\eg, raw text, movie ratings), reported by subjects (\eg, survey responses), or indirectly inferred/derived from other data (\eg, part-of-speech tags, model-based guesses for age or language)? If the data was reported by subjects or indirectly inferred/derived from other data, was the data validated/verified? If so, please describe how.} The released masks associated with each image were automatically inferred by our segmentation model, \sam. The masks that were collected using model-assisted manual annotation will not be released. Quality was validated as described in \S\ref{sec:dataset}.
\item \qtxt{What mechanisms or procedures were used to collect the data (\eg, hardware apparatuses or sensors, manual human curation, software programs, software APIs)? How were these mechanisms or procedures validated?} The images in the dataset are licensed from an image provider. They are all photos taken by photographers with different cameras.
\item \qtxt{If the dataset is a sample from a larger set, what was the sampling strategy (\eg, deterministic, probabilistic with specific sampling probabilities)?} We withheld \app 2k randomly selected images for testing purposes. The rest of the licensed images are included in the dataset.
\item \qtxt{Who was involved in the data collection process (\eg, students, crowdworkers, contractors) and how were they compensated (\eg, how much were crowdworkers paid)?} The released masks were automatically inferred by \sam. For details on our model-assisted manual annotation process see our Data Annotation Card in \S\ref{app:crowdworksheets}. Note these masks will not be released.
\item \qtxt{Over what timeframe was the data collected? Does this timeframe match the creation timeframe of the data associated with the instances (\eg, recent crawl of old news articles)? If not, please describe the timeframe in which the data associated with the instances was created.} The licensed photos vary in their date taken over a wide range of years up to 2022.
\item \qtxt{Were any ethical review processes conducted (\eg, by an institutional review board)? If so, please provide a description of these review processes, including the outcomes, as well as a link or other access point to any supporting documentation. If the dataset does not relate to people, you may skip the remaining questions in this section.} We underwent an internal privacy review to evaluate and determine how to mitigate any potential risks with respect to the privacy of people in the photos. Blurring faces and license plates protects the privacy of the people in the photos.
\item \qtxt{Did you collect the data from the individuals in question directly, or obtain it via third parties or other sources (\eg, websites)?} We licensed the data from a third party photo provider.
\item \qtxt{Were the individuals in question notified about the data collection? If so, please describe (or show with screenshots or other information) how notice was provided, and provide a link or other access point to, or otherwise reproduce, the exact language of the notification itself.} The images are licensed from a third party who provided appropriate representations regarding the collection of any notices and consents as required from individuals. In addition, all identifiable information (\eg faces, license plates) was blurred. Under the terms of the dataset license it is prohibited to attempt to identify or associate an image with a particular individual.
\item \qtxt{Did the individuals in question consent to the collection and use of their data? If so, please describe (or show with screenshots or other information) how consent was requested and provided, and provide a link or other access point to, or otherwise reproduce, the exact language to which the individuals consented.} The images are licensed from a third party who provided appropriate representations regarding the collection of any notices and consents as required from individuals. In addition, all identifiable information (\eg faces, license plates) was blurred from all images. For avoidance of doubt, under the terms of the dataset license it is prohibited to attempt to identify or associate an image with a particular individual.
\item \qtxt{If consent was obtained, were the consenting individuals provided with a mechanism to revoke their consent in the future or for certain uses? If so, please provide a description, as well as a link or other access point to the mechanism (if appropriate).} We invite users to report at \href{mailto:segment-anything@meta.com}{segment-anything@meta.com} for image(s) removal.
\item \qtxt{Has an analysis of the potential impact of the dataset and its use on data subjects (\eg, a data protection impact analysis) been conducted? If so, please provide a description of this analysis, including the outcomes, as well as a link or other access point to any supporting documentation.} To eliminate any potential impact on people whose photos are included in the dataset, identifiable information (faces, license plates) has been blurred.
\item \qtxt{Any other comments?} No.
\end{enumeratecard}
\paragraph{Preprocessing / Cleaning / Labeling}
\begin{enumeratecard}
\item \qtxt{Was any preprocessing / cleaning / labeling of the data done (\eg, discretization or bucketing, tokenization, part-of-speech tagging, SIFT feature extraction, removal of instances, processing of missing values)? If so, please provide a description. If not, you may skip the remaining questions in this section.} We resized the high-resolution licensed images such that the shorter side is 1500 pixels and only processed the images to remove any identifiable and personal information from the photos (faces, license plates).
\item \qtxt{Was the ``raw'' data saved in addition to the preprocessed/cleaned/labeled data (\eg, to support unanticipated future uses)? If so, please provide a link or other access point to the ``raw'' data.} No, as we removed the data for safety reasons and to respect privacy, we do not release the unaltered photos.
\item \qtxt{Is the software that was used to preprocess/clean/label the data available? If so, please provide a link or other access point.} We used the RetinaFace~\cite{serengil2020lightface, serengil2021lightface} model (\href{https://github.com/serengil/retinaface}{https://github.com/serengil/retinaface}) to detect faces. The model used to blur license plates has not been made public.
\end{enumeratecard}
\paragraph{Uses}
\begin{enumeratecard}
\item \qtxt{Has the dataset been used for any tasks already? If so, please provide a description.} The dataset was used to train our segmentation model, \sam.
\item \qtxt{Is there a repository that links to any or all papers or systems that use the dataset? If so, please provide a link or other access point.} No. However, all users of the dataset must cite it, so its use is trackable via citation explorers.
\item \qtxt{What (other) tasks could the dataset be used for?} We intend the dataset to be a large-scale segmentation dataset. However, we invite the research community to gather additional annotations for the dataset.
\item \qtxt{Is there anything about the composition of the dataset or the way it was collected and preprocessed/cleaned/labeled that might impact future uses? For example, is there anything that a dataset consumer might need to know to avoid uses that could result in unfair treatment of individuals or groups (\eg, stereotyping, quality of service issues) or other risks or harms (\eg, legal risks, financial harms)? If so, please provide a description. Is there anything a dataset consumer could do to mitigate these risks or harms?} We have an analysis of the approximate geographic and income level coverage of our dataset in \S\ref{sec:rai}. While we believe our dataset to be more representative than most of the publicly existing datasets at this time, we acknowledge that we do not have parity across all groups, and we encourage users to be mindful of potential biases their models have learned using this dataset.
\item \qtxt{Are there tasks for which the dataset should not be used? If so, please provide a description.} Full terms of use for the dataset including prohibited use cases can be found at \href{https://ai.facebook.com/datasets/segment-anything}{https://ai.facebook.com/datasets/segment-anything}.
\item \qtxt{Any other comments?} No.
\end{enumeratecard}
\paragraph{Distribution}
\begin{enumeratecard}
\item \qtxt{Will the dataset be distributed to third parties outside of the entity (\eg, company, institution, organization) on behalf of which the dataset was created? If so, please provide a description.} The dataset will be available for the research community.
\item \qtxt{How will the dataset will be distributed (\eg, tarball on website, API, GitHub)? Does the dataset have a digital object identifier (DOI)?} The dataset is available at \href{https://ai.facebook.com/datasets/segment-anything}{https://ai.facebook.com/datasets/segment-anything}.
\item \qtxt{When will the dataset be distributed?} The dataset will be released in 2023.
\item \qtxt{Will the dataset be distributed under a copyright or other intellectual property (IP) license, and/or under applicable terms of use (ToU)? If so, please describe this license and/or ToU, and provide a link or other access point to, or otherwise reproduce, any relevant licensing terms or ToU, as well as any fees associated with these restrictions.} Yes. The license agreement and terms of use for the dataset can be found at \href{https://ai.facebook.com/datasets/segment-anything}{https://ai.facebook.com/datasets/segment-anything}. Users must agree to the terms of use before downloading or using the dataset.
\item \qtxt{Have any third parties imposed IP-based or other restrictions on the data associated with the instances? If so, please describe these restrictions, and provide a link or other access point to, or otherwise reproduce, any relevant licensing terms, as well as any fees associated with these restrictions.} Full terms of use and restrictions on use of the \sad dataset can be found at \href{https://ai.facebook.com/datasets/segment-anything}{https://ai.facebook.com/datasets/segment-anything}.
\item \qtxt{ Do any export controls or other regulatory restrictions apply to the dataset or to individual instances? If so, please describe these restrictions, and provide a link or other access point to, or otherwise reproduce, any supporting documentation.} The license and restrictions on use of the SA-1B dataset can be found at \href{https://ai.facebook.com/datasets/segment-anything}{https://ai.facebook.com/datasets/segment-anything}.
\item \qtxt{Any other comments?} No.
\end{enumeratecard}
\paragraph{Maintenance}
\begin{enumeratecard}
\item \qtxt{Who will be supporting/hosting/maintaining the dataset?} The dataset will be hosted at \href{https://ai.facebook.com/datasets/segment-anything}{https://ai.facebook.com/datasets/segment-anything} and maintained by Meta AI.
\item \qtxt{How can the owner/curator/manager of the dataset be contacted (\eg, email address)?} Please email \href{mailto:segment-anything@meta.com}{segment-anything@meta.com}.
\item \qtxt{Is there an erratum? If so, please provide a link or other access point.} No.
\item \qtxt{Will the dataset be updated (\eg, to correct labeling errors, add new instances, delete instances)? If so, please describe how often, by whom, and how updates will be communicated to dataset consumers (\eg, mailing list, GitHub)?} To aid reproducibility of research using \sad, the only updates will be to remove reported images.
\item \qtxt{If the dataset relates to people, are there applicable limits on the retention of the data associated with the instances (\eg, were the individuals in question told that their data would be retained for a fixed period of time and then deleted)? If so, please describe these limits and explain how they will be enforced.} There are no limits on data retention. We took measures to remove personally identifiable information from any images of people. Users may report content for potential removal here: \href{mailto:segment-anything@meta.com}{segment-anything@meta.com}.
\item \qtxt{Will older versions of the dataset continue to be supported/hosted/maintained? If so, please describe how. If not, please describe how its obsolescence will be communicated to dataset consumers.} No, as the only updates will be to remove potentially harmful content, we will not keep older versions with the content.
\item \qtxt{If others want to extend/augment/build on/contribute to the dataset, is there a mechanism for them to do so? If so, please provide a description. Will these contributions be validated/verified? If so, please describe how. If not, why not? Is there a process for communicating/distributing these contributions to dataset consumers? If so, please provide a description.} We encourage users to gather further annotations for \sad. Any users who generate annotations will be liable for hosting and distributing their annotations.
\item \qtxt{Any other comments?} No.
\end{enumeratecard}}

\subsection{Data Annotation Card}\label{app:crowdworksheets}
{\fontsize{7.5}{8.4}\selectfont
\paragraph{Task Formulation}
\begin{enumeratecard}
\item \qtxt{At a high level, what are the subjective aspects of your task?} Segmenting objects present in an image is inherently a subjective task. For instance, one annotator may segment two boots as one mask, whereas another may segment each boot separately. Depending on annotators's skills, the quality of the mask and the number of masks per image are different between annotators. Despite these subjective aspects of the task, we believed efficient annotation was possible as the data was annotated in a per-mask fashion with the main focus on the diversity of the data rather than completeness.
\item \qtxt{What assumptions do you make about annotators?} Our annotators worked full time on our annotation task with very small attrition rate. This made it possible to train the annotators providing feedback and answering their questions on a regular basis. Specifically: (1) By giving a clear understanding of the goals of this work and providing clear guidelines, including visuals and video recordings of the tasks, annotators had enough context to understand and perform the tasks reasonably. (2) Sharing objectives and key results and meeting weekly with annotators increased the likelihood that annotators improved annotation quality and quantity over time.
\item \qtxt{How did you choose the specific wording of your task instructions? What steps, if any, were taken to verify the clarity of task instructions and wording for annotators?} As our task was annotating images, the annotation guidelines included visual examples. Our research team completed 30 annotation tasks to identify any obvious challenges using the annotation tool, collectively decide how to handle complex cases, and refine the guidelines. The research team met with the annotators weekly for feedback sessions. Videos of the research team performing the task were shared live with the annotators, followed by Q\&A sessions. Annotators were able to give feedback on unclear aspects, both during the feedback session and asynchronously.
\item \qtxt{What, if any, risks did your task pose for annotators and were they informed of the risks prior to engagement with the task?} No identified risks. Images were filtered for objectionable content prior to the annotation phase.
\item \qtxt{What are the precise instructions that were provided to annotators?} We provide only high-level instructions: Given an image, we aim at segmenting every possible object. Annotators generate a mask for every potential object they can identify. An object can be segmented using our interactive segmentation tool either by using corrective foreground/background clicks to add/remove parts of the mask or by drawing a bounding box around the object. Masks can be refined using pixel-precise tools.
\end{enumeratecard}
\paragraph{Selecting Annotations}
\begin{enumeratecard}
\item \qtxt{Are there certain perspectives that should be privileged? If so, how did you seek these perspectives out?} We chose to work with annotators that have worked on other vision annotation tasks before.
\item \qtxt{Are there certain perspectives that would be harmful to include? If so, how did you screen these perspectives out?} No.
\item \qtxt{Were sociodemographic characteristics used to select annotators for your task? If so, please detail the process.} No.
\item \qtxt{If you have any aggregated socio-demographic statistics about your annotator pool, please describe. Do you have reason to believe that sociodemographic characteristics of annotators may have impacted how they annotated the data? Why or why not?} We worked with 130 annotators. The annotators were all based in Kenya. We do not believe sociodemographic characteristics of annotators meaningfully impacted the annotated data.
\item \qtxt{Consider the intended context of use of the dataset and the individuals and communities that may be impacted by a model trained on this dataset. Are these communities represented in your annotator pool?} The Segment Anything 1B (\sad) dataset is to be used for research purposes only. The \sad dataset is one of the most geographically diverse segmentation dataset, as discussed in \S\ref{sec:rai}. In addition, we analyze the responsible AI axes of a model trained on the dataset in \S\ref{sec:rai}.
\end{enumeratecard}
\paragraph{Platform and Infrastructure Choices}
\begin{enumeratecard}
\item \qtxt{What annotation platform did you utilize? At a high level, what considerations informed your decision to choose this platform? Did the chosen platform sufficiently meet the requirements you outlined for annotator pools? Are any aspects not covered?} We used a proprietary annotation platform.
\item \qtxt{What, if any, communication channels did your chosen platform offer to facilitate communication with annotators? How did this channel of communication influence the annotation process and/or resulting annotations?} We manually reviewed annotations and shared feedback with the annotators on a weekly basis. We communicated common mistakes or inconsistencies and the corresponding corrections. In addition, the annotators were given feedback for improvements daily by the annotation QA team. Outside the weekly feedback sessions, annotators had access to a spreadsheet and chat group to facilitate communication with the research team. This process greatly improved the average speed and quality of the annotations.
\item \qtxt{How much were annotators compensated? Did you consider any particular pay standards, when determining their compensation? If so, please describe.} Annotators were compensated with an hourly wage set by the vendor. The vendor is a Certified B Corporation.
\end{enumeratecard}
\paragraph{Dataset Analysis and Evaluation}
\begin{enumeratecard}
\item \qtxt{How do you define the quality of annotations in your context, and how did you assess the quality in the dataset you constructed?} Annotators were first placed into training. They followed a 1-day training session led by the vendor and then were asked to annotate a large number of examples from a training queue. Annotators graduated from training to production after the vendor QA team, in collaboration with the research team, manually spot-checked the annotator’s masks to ensure quality. On average, annotators spent one week in training before graduating. Production quality assessment followed a similar process: the vendor QA team and the research team manually reviewed the annotations weekly, sharing feedback weekly.
\item \qtxt{Have you conducted any analysis on disagreement patterns? If so, what analyses did you use and what were the major findings? Did you analyze potential sources of disagreement?} We pointed out common mistakes during weekly meetings with the annotators.
\item \qtxt{How do the individual annotator responses relate to the final labels released in the dataset?} The annotations were only used to train early versions of the \sam model and we do not currently plan to release them.
\end{enumeratecard}
\paragraph{Dataset Release and Maintenance}
\begin{enumeratecard}
\item \qtxt{Do you have reason to believe the annotations in this dataset may change over time? Do you plan to update your dataset?} No, except to remove objectionable images.
\item \qtxt{Are there any conditions or definitions that, if changed, could impact the utility of your dataset?} We do not believe so.
\item \qtxt{Will you attempt to track, impose limitations on, or otherwise influence how your dataset is used? If so, how?} The SA-1B dataset will be released under a license agreement allowing use for certain research purposes and protections for researchers. Researchers must agree to the terms of the license agreement to access the dataset.
\item \qtxt{Were annotators informed about how the data is externalized? If changes to the dataset are made, will they be informed?} No, we do not plan to release the manual annotations at the moment.
\item \qtxt{Is there a process by which annotators can later choose to withdraw their data from the dataset? If so, please detail.} No.
\end{enumeratecard}}

%##################################################################################################
\begin{table*}\centering\fontsize{7.5}{8.4}\selectfont
\begin{tabular*}{\linewidth}{r|p{12cm}}
\multicolumn{2}{l}{\bf\underline{Model Overview}} \\
Name & \sam or Segment Anything Model \\
Version & 1.0 \\
Date & 2023 \\
Organization & The FAIR team of Meta AI\\
Mode type & Promptable segmentation model \\
Architecture & See \S\ref{sec:model} \\
Repository & \href{https://github.com/facebookresearch/segment-anything}{https://github.com/facebookresearch/segment-anything} \\
Citation & \href{https://research.facebook.com/publications/segment-anything}{https://research.facebook.com/publications/segment-anything} \\
License & Apache 2.0 \\
\multicolumn{2}{c}{} \\
\multicolumn{2}{l}{\bf\underline{Intended Use}} \\
Primary intended uses & \sam is intended to be used for any prompt-based segmentation task. We explored its use in \textit{segmenting objects from a point} (\S\ref{sec:eval:single_point}), \textit{edge detection} (\S\ref{sec:eval:edge}), \textit{segmenting all objects} (\S\ref{subsec:proposals}), and \textit{segmenting detected objects} (\S\ref{sec:eval:instseg}). We explored how \sam can integrate with other vision models to \textit{segment objects from text} (\S\ref{sec:eval:text_to_mask}). \\
Primary intended users & \sam was primarily developed for research. The license for \sam can be found at \href{https://github.com/facebookresearch/segment-anything}{https://github.com/facebookresearch/segment-anything}. \\
Out-of-scope use cases & See terms of use for \sam found at \href{https://github.com/facebookresearch/segment-anything}{https://github.com/facebookresearch/segment-anything}. See \textit{Use Cases} under \textit{Ethical Considerations}. \\
Caveats and recommendations & \sam has impressive zero-shot performance across a wide range of tasks. We note, however, that in the zero-shot setting there may be multiple valid ground truth masks for a given input. We recommend users take this into consideration when using \sam for zero-shot segmentation. \sam can miss fine structures and can hallucinate small disconnected components. See \S\ref{sec:disc} for a discussion of limitations. \\
\multicolumn{2}{c}{} \\
\multicolumn{2}{l}{\bf\underline{Relevant Factors}} \\
Groups & \sam was designed to segment any object. This includes \textit{stuff} and \textit{things}.\\
Instrumentation and environment & We benchmarked \sam on a diverse set of datasets and found that \sam can handle a variety of visual data including \textit{simulations, paintings, underwater images, microscopy images, driving data, stereo images, fish-eye images}. See \S\ref{app:benchmark} and Table~\ref{app:tab:datasets_all} for information on the benchmarks used. \\
\multicolumn{2}{c}{} \\
\multicolumn{2}{l}{\bf\underline{Metrics}} \\
Model performance measures & We evaluated \sam on a variety of metrics based on the downstream task in our experiments.
\begin{itemize}
\itemsep-0.2em
\item{\textit{mIoU}: We used the mean intersection-over-union after a given number of prompts to evaluate the segmentation quality of a mask when prompted with points.}
\item{\textit{Human evaluation}: We performed a human study (detailed in \S\ref{app:human_study}) to evaluate the real world performance of \sam. We compared the masks generated by \sam to a baseline state-of-the-art interactive segmentation model, RITM~\cite{sofiiuk2022reviving}, using a perceptual quality scale from 1 to 10.}
\item{\textit{AP}: We used average precision to evaluate instance segmentation for a given box and edge detection.}
\item{\textit{AR@1000}: We used average recall to evaluate object proposal generation.}
\item{\textit{ODS, OIS, AP, R50}: We used the standard edge detection evaluation metrics from BSDS500~\cite{martin2001database,arbelaez2010contour}.}\vspace{-1.2em}%remove weird extra space after itemize env
\end{itemize}\\
\multicolumn{2}{c}{} \\
\multicolumn{2}{l}{\bf\underline{Evaluation Data}} \\
Data sources & See \S\ref{app:benchmark}. \\
\multicolumn{2}{c}{} \\
\multicolumn{2}{l}{\bf\underline{Training Data}} \\
Data source & See Data Card in \S\ref{app:datacard}. \\
\multicolumn{2}{c}{} \\
\multicolumn{2}{l}{\bf\underline{Ethical Considerations}} \\
Data &
We trained \sam on licensed images. The images were filtered for objectionable content by the provider, but we acknowledge the possibility of false negatives. We performed a geographic analysis of the \sad dataset in \S\ref{sec:rai}. While \sad is more geographically diverse than many of its predecessors, we acknowledge that some geographic regions and economic groups are underrepresented. \\
Cost and impact of compute & \sam was trained on 256 A100 GPUS for 68 hours. We acknowledge the environmental impact and cost of training large scale models. The environmental impact of training the released \sam model is approximately 6963 kWh resulting in an estimated 2.8 metric tons of carbon dioxide given the specific data center used, using the calculation described in~\cite{patterson2021carbon} and the ML CO$_2$ Impact calculator~\cite{lacoste2019quantifying}. This is equivalent to \app7k miles driven by the average gasoline-powered passenger vehicle in the US~\cite{epa2022}. We released the \sam models to both reduce the need for retraining and lower the barrier to entry for large scale vision research. \\
Risks and harms & We evaluated \sam for fairness in \S\ref{sec:rai}. Downstream use cases of \sam will create their own potential for biases and fairness concerns. As such we recommend users run their own fairness evaluation when using \sam for their specific use case. \\
Use cases & We implore users to use their best judgement for downstream use of the model. \\
\end{tabular*}
\caption{Model Card for \sam, following the procedure detailed in~\cite{mitchell2019model}.}
\label{tab:modelcard}\vspace{20mm}
\end{table*}
%##################################################################################################

%##################################################################################################
\newcommand{\incannpanel}[2]{\begin{subfigure}[t]{.24\textwidth}
 \includegraphics[width=\linewidth,height=26mm]{figs/annotator_guidelines/#1.jpg}
 \vspace{-6mm}\caption*{\tiny #2}\end{subfigure}\vspace{1mm}\hspace{1.5mm}}
\newcommand{\incannpaneldouble}[3]{\begin{subfigure}[t]{.24\textwidth}
 \begin{minipage}[b]{\linewidth}
 \includegraphics[width=\linewidth,height=13mm]{figs/annotator_guidelines/#1.jpg}
 \includegraphics[width=\linewidth,height=13mm]{figs/annotator_guidelines/#2.jpg}\end{minipage}
 \vspace{-6mm}\caption*{\tiny #3}\end{subfigure}\vspace{1mm}\hspace{1.5mm}}
\newcommand{\incanntext}[2]{\begin{subfigure}[t]{.24\textwidth}
 \begin{minipage}[b]{\linewidth}{\fontsize{3}{3.6}\selectfont#2}\vspace{2mm}\end{minipage}
 \vspace{-6mm}\caption*{\tiny #1}\end{subfigure}\vspace{1mm}\hspace{1.5mm}}
%##################################################################################################

%##################################################################################################
\begin{figure*}
\incanntext{Objective and Setup}{
We have several models that, when provided with a click or a box as input, output a mask. We would like to compare the quality of these models by rating the quality of their masks on many examples. The interface will be different than for regular mask annotation.
\begin{itemize}[leftmargin=2.5mm, itemsep=-3pt, topsep=-3pt]
\item Each job reviews one mask in one image.
\item On the right, there will be five image thumbnails in two rows. Each thumbnail can be moused-over to show the image at a larger size. Clicking on the thumbnail will make it full screen, and clicking again will return to the original screen.
\begin{itemize}[leftmargin=1.5mm, itemsep=-2pt, topsep=-2pt]
\item The images show the same mask in five different views. On the top row: (left) the image without the mask, (middle) the mask overlaid on the image, and (right) the mask alone. On the bottom row: (left) a zoomed in view of the object without a mask, and (right) a zoomed in view of the mask overlaid on the image. These views are provided to make it easy to see different types of mask errors.
\item The mask will be in red when overlaid on the image.
\item When shown by itself, the mask is yellow, and the background is purple.
\item Each image will include either a blue dot or a blue and white box. This is the input to the model, as if you had clicked at this location or drawn this box.
\end{itemize}
\item On the left, there are buttons labeled 1-10. This is used to rate the quality of the shown mask.
\end{itemize}}
\incannpanel{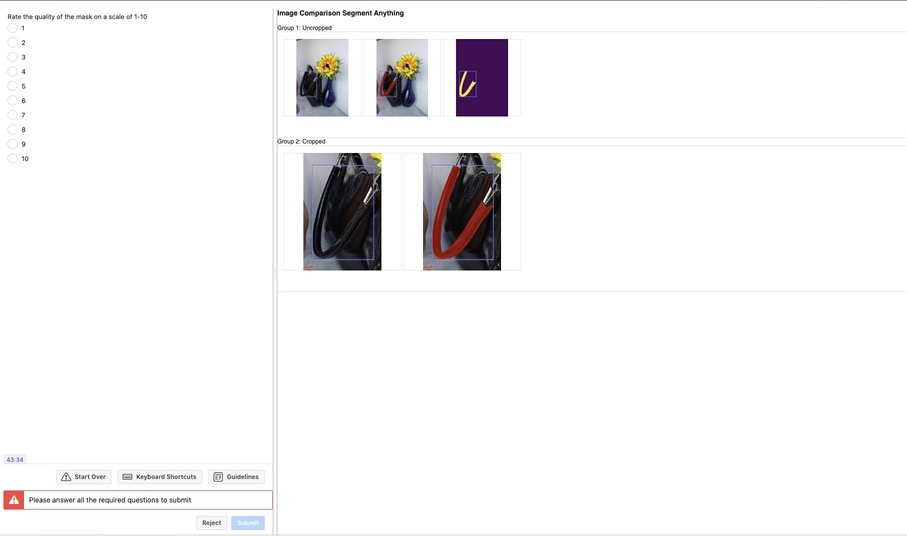}{Example interface page. There will be five images on the right and a question box on the left.}
\incannpanel{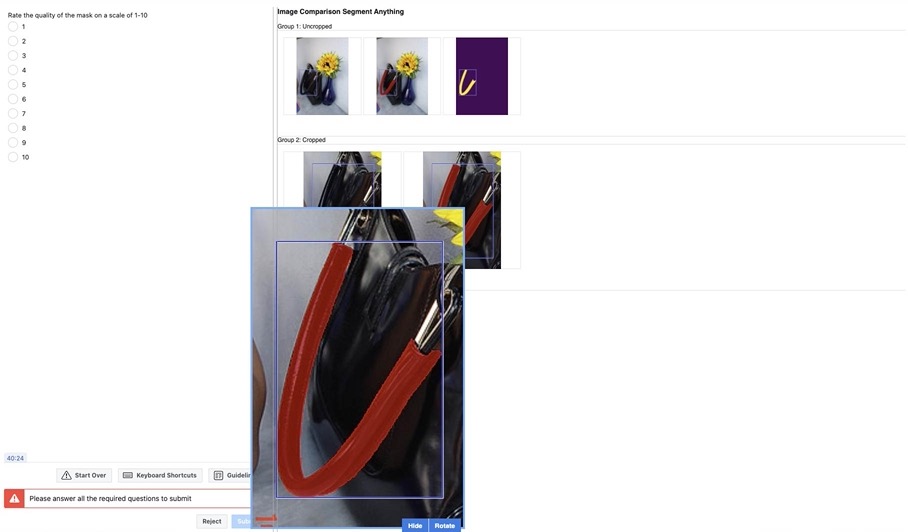}{Mouse over an image to show the full image.}
\incannpanel{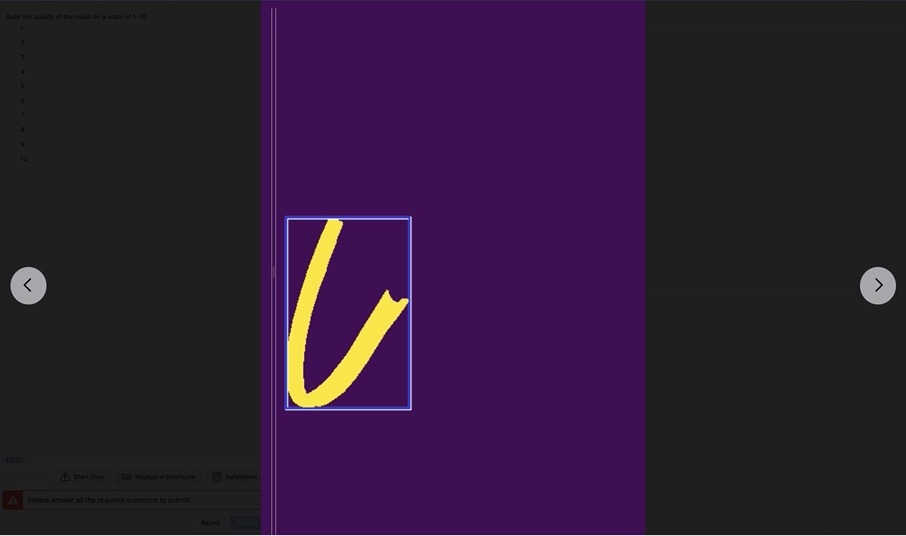}{Click on an image to make it full screen. The arrows will cycle between images. Click again to return to previous view.}\\
\incannpanel{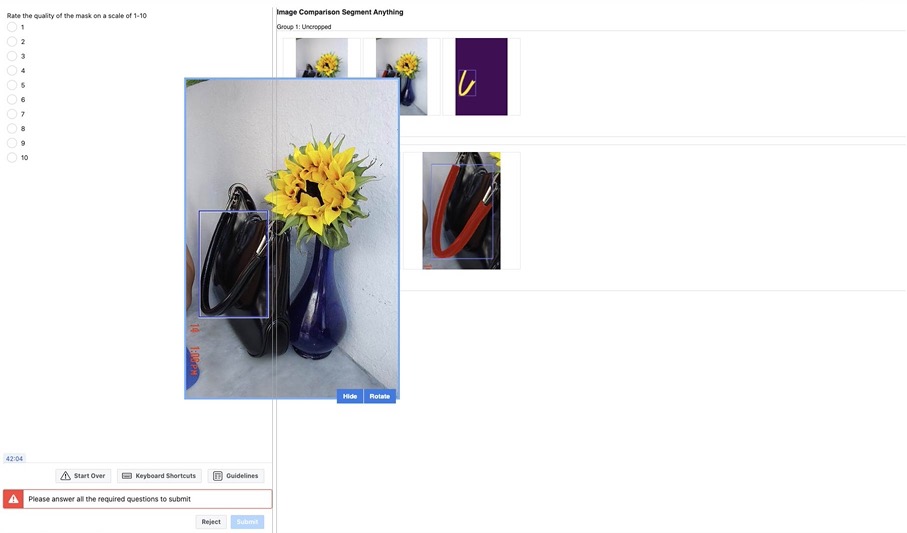}{The first image on the top row shows the image without a mask. A blue point will be on the object of interest, or a blue and white box will surround it.}
\incannpanel{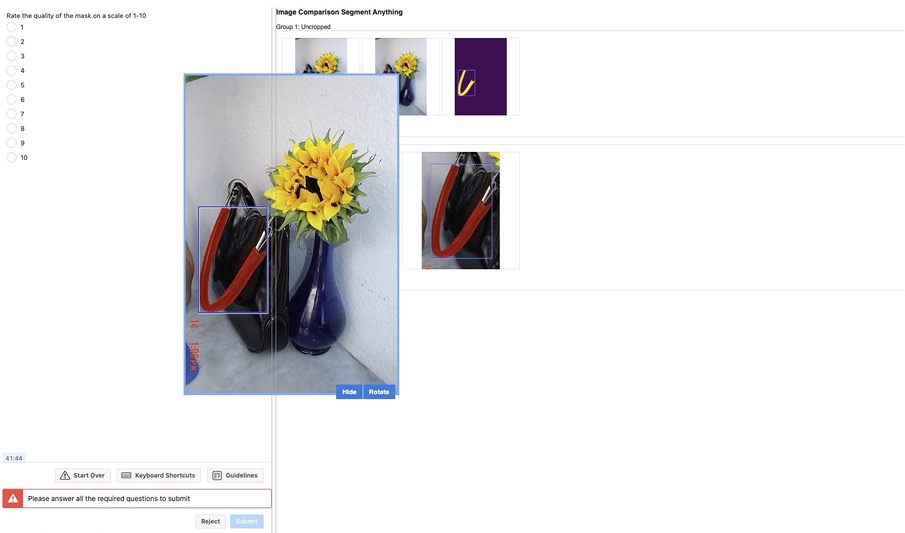}{The second image on the top row shows the mask for the object in red.}
\incannpanel{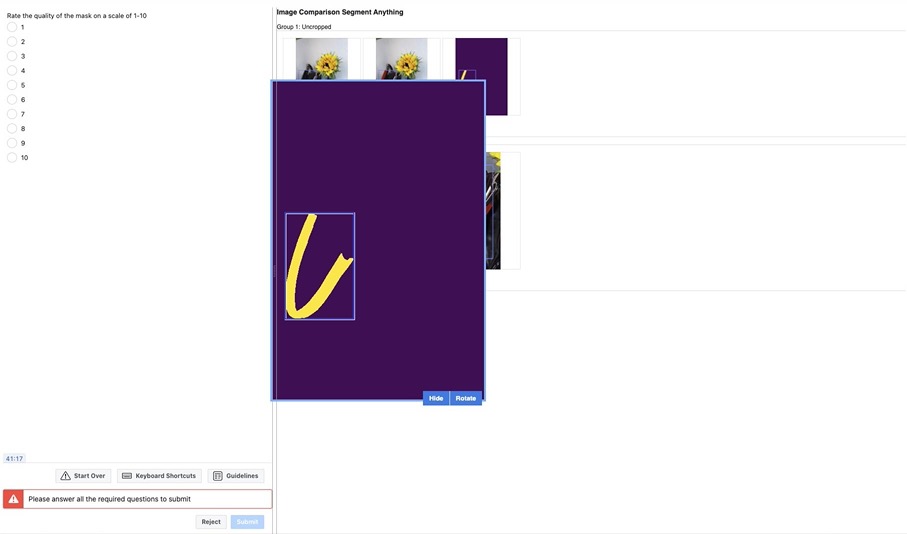}{The third image on the top row shows the mask only. The mask is in yellow and the background is purple.}
\incannpanel{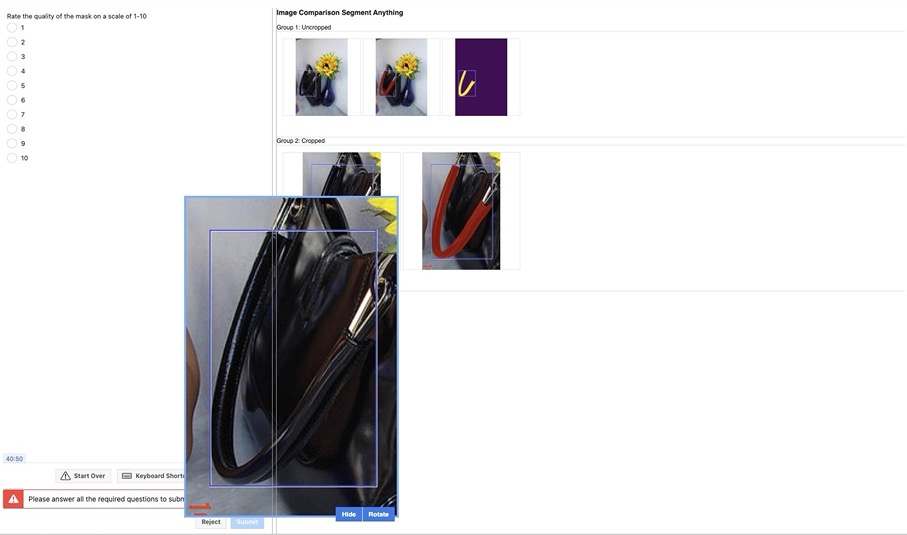}{The first image on the bottom row shows a zoomed in view of the object without a mask.}\\
\incannpanel{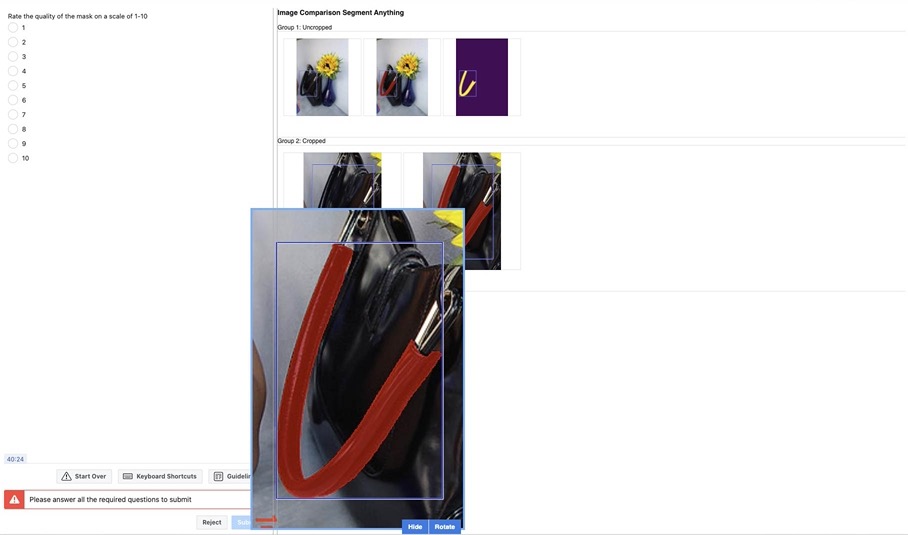}{The second image on the bottom row shows a zoomed in view of the object with a mask. The mask is in red.}
\incannpanel{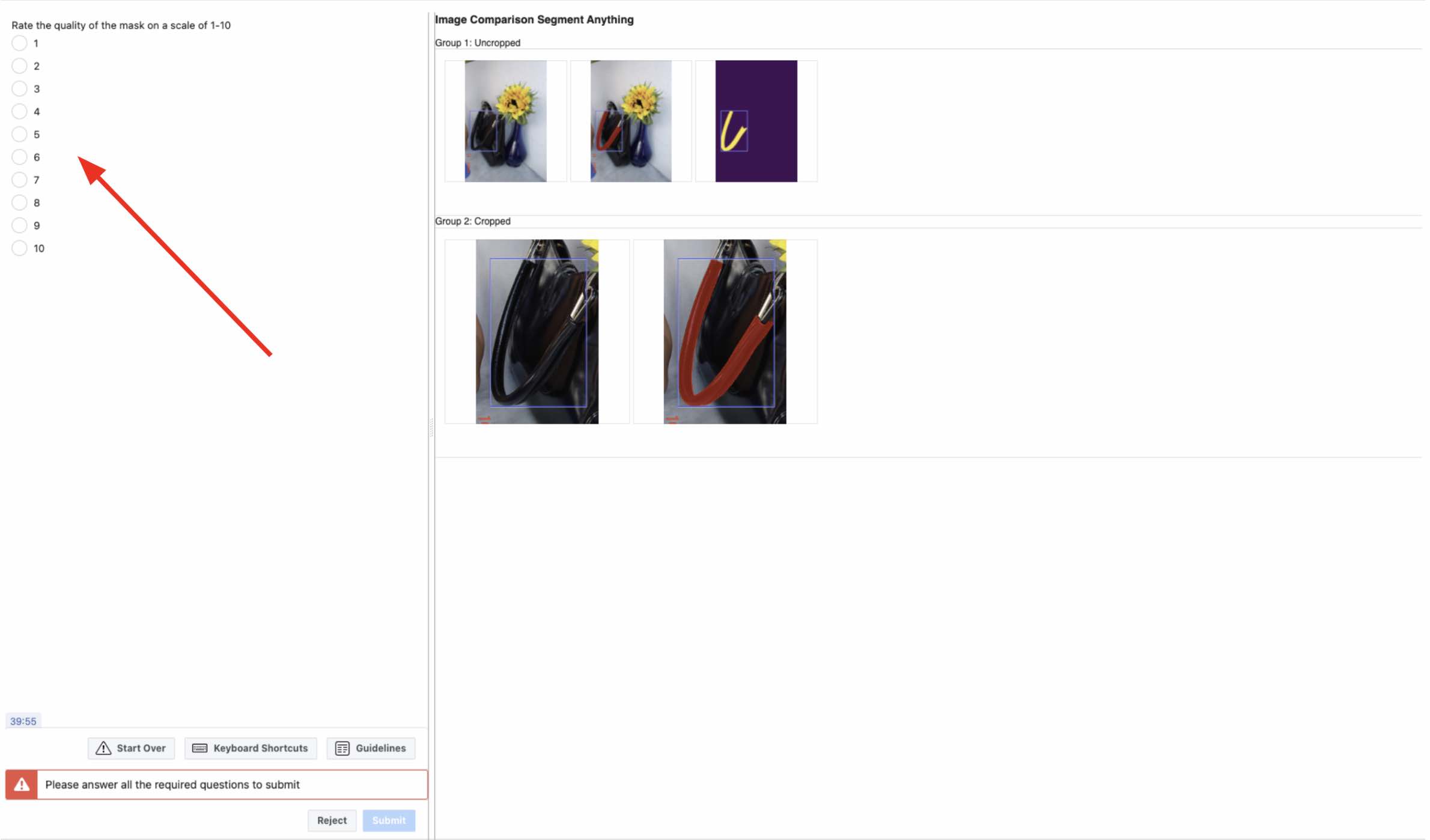}{On the left are buttons to rate the mask quality, with selections 1-10.}
\incanntext{Task}{
What we would like you to do for each job:
\begin{itemize}[leftmargin=2.5mm, itemsep=-3pt, topsep=-3pt]
\item Please aim to spend up to 30 seconds per job.
\item Mouse-over or click each of the three images of the mask on the right to get a sense of the quality of the mask. The thumbnail is too small to judge a mask, do not judge a mask by the thumbnail alone. Each image can provide a different signal on possible mask errors:
\begin{itemize}[leftmargin=1.5mm, itemsep=-2pt, topsep=-2pt]
\item The unzoomed image can give context for the mask: does this mask correspond to an actual object?
\item The mask-only image can show if the mask has small holes or separated, incorrect pixels.
\item The zoomed image can show if the mask boundaries make sense.
\end{itemize}
\item Judge the quality of the mask on three criterion. Examples will follow.
\begin{itemize}[leftmargin=1.5mm, itemsep=-2pt, topsep=-2pt]
\item Does the mask correspond to an actual object?
\item Does the mask have a good boundary?
\item Does the mask correspond to the provided point or box?
\end{itemize}
\item Rate the quality of the mask on a scale of 1-10 using the drop-down box on the left.
\item Next are details and examples for judging mask quality according to the three criterion. These are just examples and other cases may come up, please use your best judgment when determining if something is a good mask.
\end{itemize}}
\incanntext{Judging Mask Quality (1 of 3)}{
Does the mask correspond to an actual object?
\begin{itemize}[leftmargin=2.5mm, itemsep=-3pt, topsep=-3pt]
\item Valid objects can include:
\begin{itemize}[leftmargin=1.5mm, itemsep=-2pt, topsep=-2pt]
\item Entire single objects (such as a person, shirt, or tree)
\item Logical parts of objects (a chair leg, a car door, a tabletop)
\item Collections of objects (a stack of books, a crowd of people)
\item ‘Stuff’ (the ground, the sky).
\end{itemize}
\item Example errors a mask may have. The severity of these errors may be minor or major:
\begin{itemize}[leftmargin=1.5mm, itemsep=-2pt, topsep=-2pt]
\item Include a piece of another object (the mask of a person including the arm of a nearby person)
\item Miss part of an object (the mask covers only one part of a building obscured by a tree in the foreground),
\item Combine two unrelated things (a single mask covers both a mug and a pen on a desk)
\item Include an arbitrary part of a collection for a point input (a point is on one apple, but the mask covers three apples in a pile of many apples). If a box surrounds an arbitrary collection, it is not an error to provide a mask for these objects.
\end{itemize}
\item If you are unsure, a good rule-of-thumb is: can you name the object in question? However, some things that are hard to name may still be good objects (an unusual component of a machine, something at the edge of the image for which it is hard to determine what it is).\\\\
\end{itemize}}\\
\incanntext{Judging Mask Quality (2 of 3)}{
Does the mask have a good boundary?
\begin{itemize}[leftmargin=2.5mm, itemsep=-3pt, topsep=-3pt]
\item Errors in the boundary can include:
\begin{itemize}[leftmargin=1.5mm, itemsep=-2pt, topsep=-2pt]
\item Incorrect holes in the mask
\item Incorrect pixels included separated from the main part of the mask
\item Poor edge quality, where the mask does not exactly match the edge of the object.
\item Failure to consistently handle obscuring foreground objects (a mask that covers obscuring objects is fine, and a mask that doesn’t cover obscuring objects is fine, but one that does some of both has an error)
\item Pixelation of a small mask is not an error, as long as the mask still matches the edges of the object.
\end{itemize}
\end{itemize}}
\incanntext{Judging Mask Quality (3 of 3)}{
Does the mask correspond to the provided point or box?
\begin{itemize}[leftmargin=2.5mm, itemsep=-3pt, topsep=-3pt]
\item For points:
\begin{itemize}[leftmargin=1.5mm, itemsep=-2pt, topsep=-2pt]
\item The point needs to be on the mask.
\item The size or position of the object with respect to the point does not matter (a point on someone’s gloved hand can correspond to the glove or to the entire person, both are valid masks).
\end{itemize}
\item For boxes:
\begin{itemize}[leftmargin=1.5mm, itemsep=-2pt, topsep=-2pt]
\item The object needs to be the best object that is the size of the box (if a box is around someone’s entire head but the mask is of their hair, this is an error: their hair is in the box but is not the correct object).
\item If the box clearly corresponds to a given object but is slightly smaller than it, it is okay if the mask goes slightly outside a box (if a box around a person misses their extended hand, the mask can still include their hand even if the mask goes outside the box).
\end{itemize}
\end{itemize}}
\incannpanel{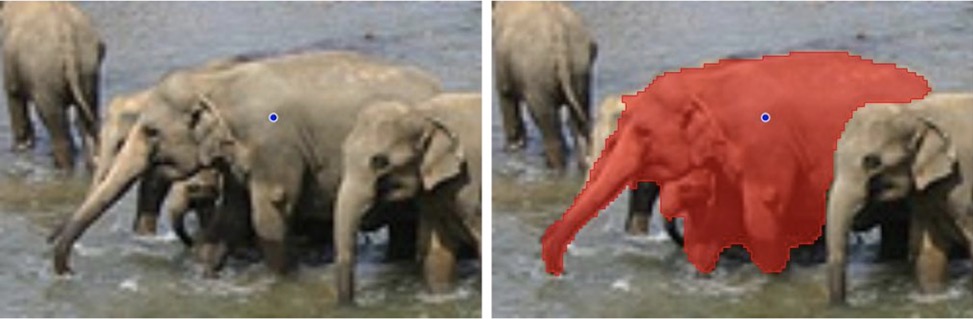}{Example error of ‘Include a piece of another object’: The elephant mask contains a piece of another nearby elephant.}
\incannpaneldouble{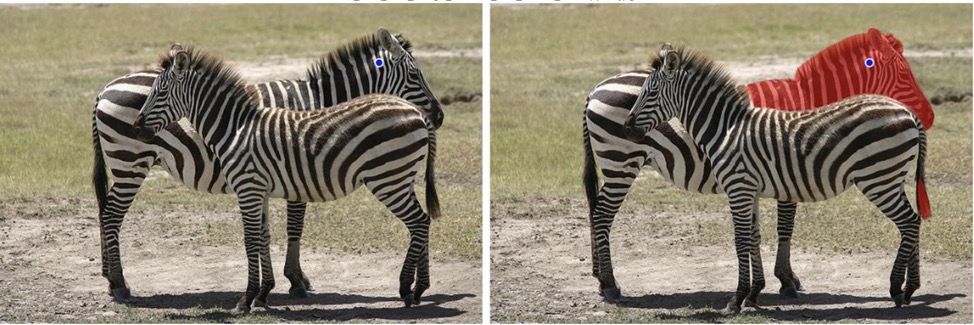}{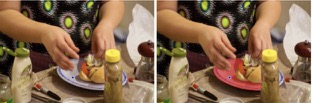}{Example error of ‘Missing a part of an object’: the mask is missing a disconnected part of the object: the back half of the zebra, and the right portion of the plate.}\\
\incannpaneldouble{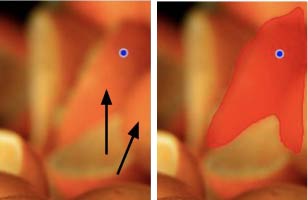}{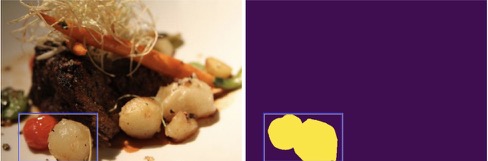}{Example error of ‘Include an arbitrary part of a collection’: In top top image, the point is on one orange rind, but the mask covers two orange rinds. This is a mask error: the mask covers an arbitrary number of objects in the collection, and should either cover one orange rind or all of them. In the bottom image, the box is around both vegetables. Since this is the best match to the box, this is not a mask error.}
\incannpanel{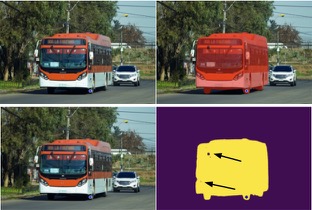}{Example error for ‘Incorrect holes in the mask’: This mask has holes in the upper left and on the left sides (black arrows). These holes are much easier to see on the ‘mask only’ image.}
\incannpanel{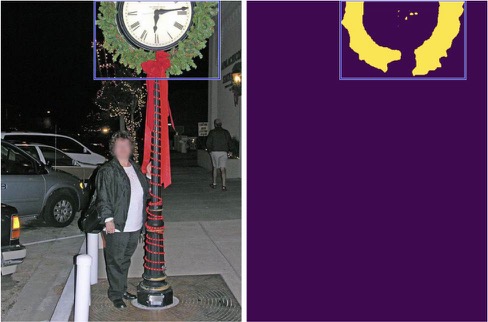}{Example error for ‘Incorrect pixels included separated from the main part of the mask’: The ‘mask only’ view reveals a few stray incorrect pixels on the clock face.}
\incannpanel{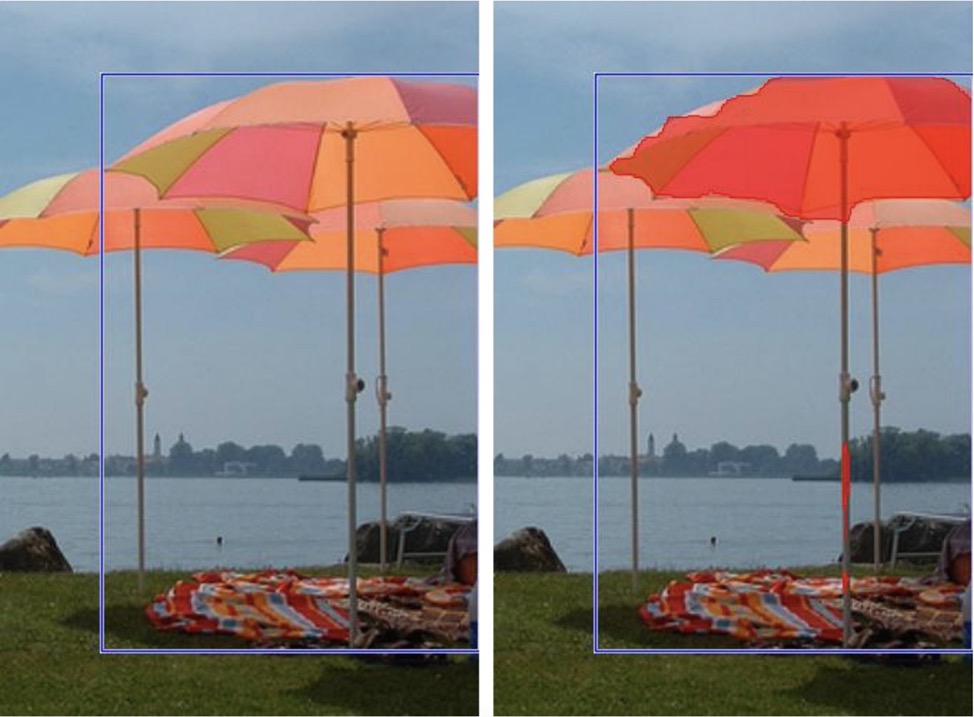}{Example error for ‘Poor edge quality’: The mask has poor edge quality, both along the edge of the umbrella, as well as along the thin pole.}\\
\vspace{-5mm}\caption{Here we provide the complete guidelines given to annotations for the human review of mask quality. Some images been edited slightly and faces have been blurred to enable release. Best viewed with zoom (part 1 of 2).}
\label{fig:guidelines_a}\vspace{-2mm}
\end{figure*}
%##################################################################################################

%%%%%%%%%%%%%%%%%%%%%%%%%%%%%%%%%%%%%%%%%%%%%%%%%%%%%%%%%%%%%%%%%%%%%%%%%%%%%%%%%%%%%%%%%%%%%%%%%%%
\section{Annotation Guidelines}\label{app:annotation_guidelines}

\begin{figure*}
\incannpanel{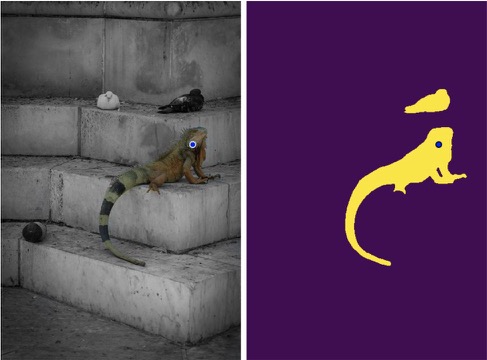}{Example for ‘Combine two unrelated things’: The point indicates the lizard, but the mask covers both the lizard and a bird. This is a mask error.}
\incannpanel{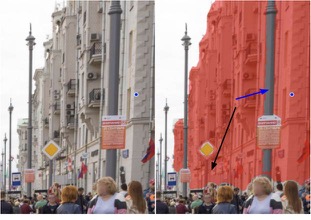}{Example error for ‘Failure to consistently handle obscuring foreground objects’: The pole on the right (blue arrow) is excluded from the mask, while the pole on the left is included in the object (black arrow). The mask should either include or exclude both of these.}
\incannpanel{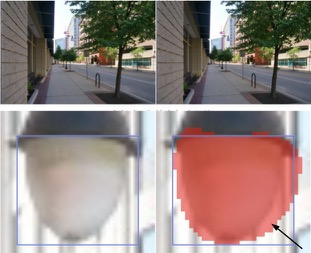}{Example of ‘Pixelation of a small mask’: this mask has an imperfect boundary, since it extends beyond the object at the black arrow. However, the ‘blocky’ pattern of the mask is not an error, since, when zoomed in this much, the image is also blocky the same way.}
\incannpanel{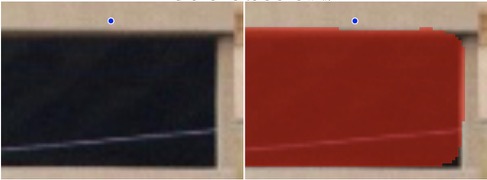}{Example error for consistency with the provided point: The mask does not agree with the blue point, so this is a mask error.}\\
\incannpanel{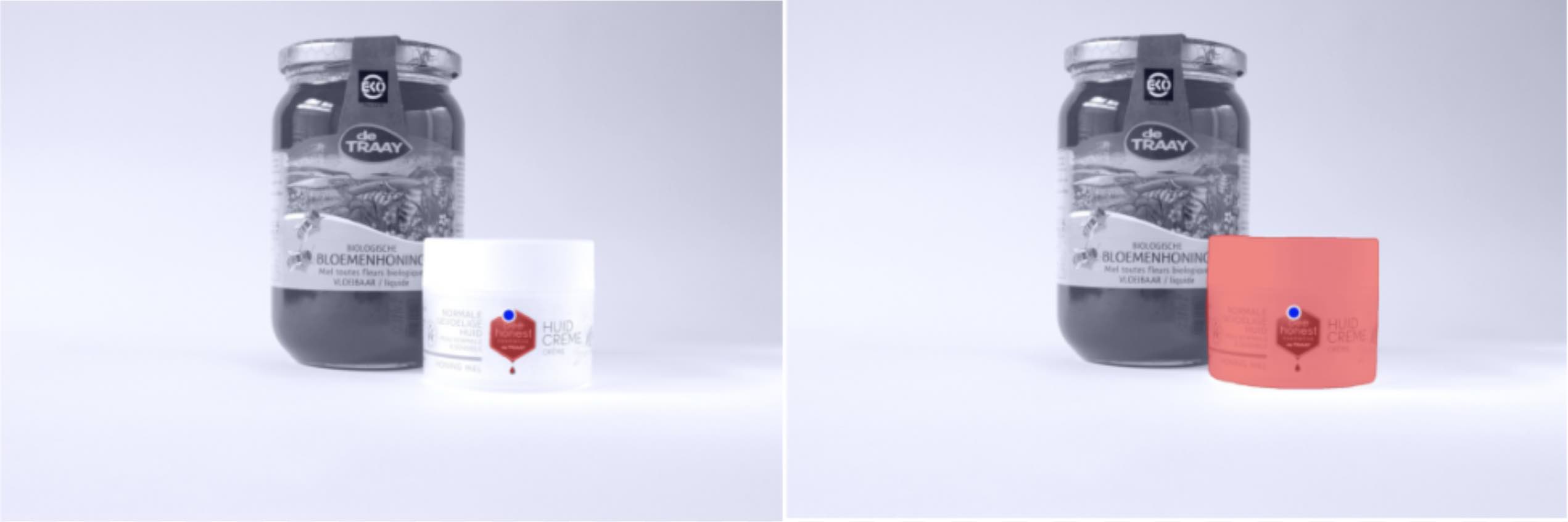}{Example for consistency with the provided point: For this input point, but the logo (left) and the container (right) are valid objects, since the blue point lies on both of them. Neither mask has a mask error.}
\incannpanel{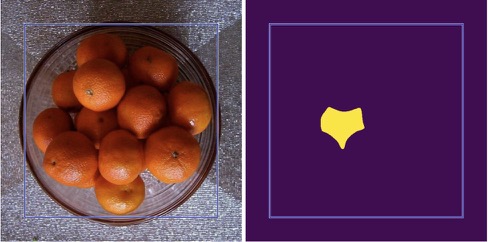}{Example for consistency with a box: The box surrounds the bowl of oranges, but the mask is only of a single orange. This is a mask error.}
\incannpanel{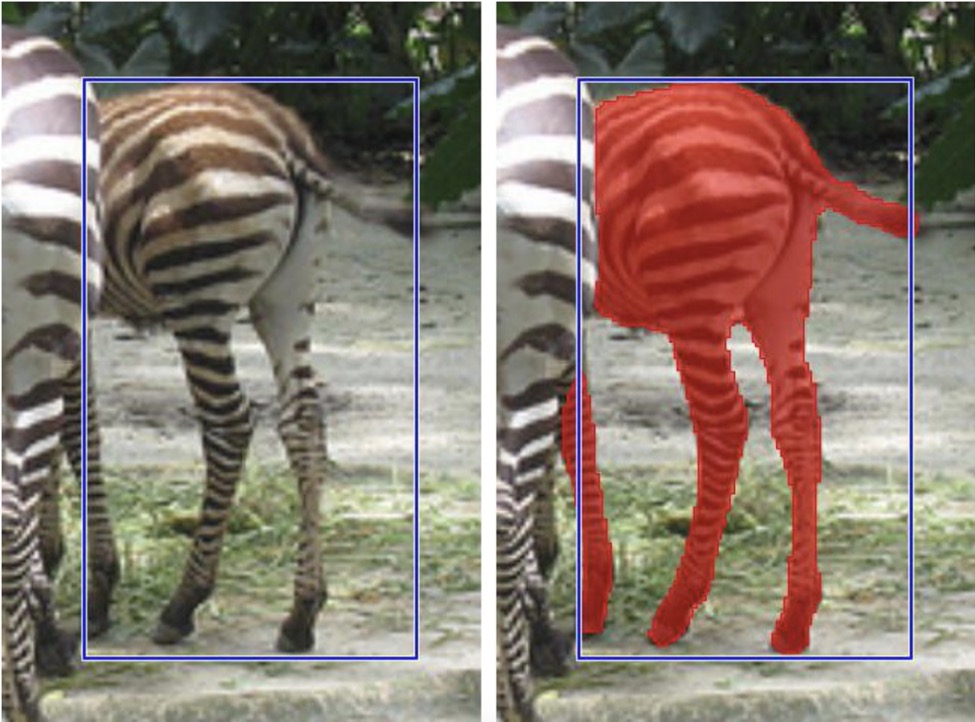}{Example for consistency with a box: The box’s shape fits the zebra. Even though the mask extends slightly outside the box to include the zebra’s left leg, this is not an error.}
\incanntext{Mask Scoring}{
Overall mask quality is subjective, each of the above errors may hurt mask quality only a little or a lot, depending on how large the error is. Please use your best judgment when choosing mask scores, and try to stay consistent from mask-to-mask. Here are some general guidelines for what different scores should correspond to:
\begin{itemize}[leftmargin=2.5mm, itemsep=-3pt, topsep=-3pt]
\item A score of 1: It is not possible to tell what object this mask corresponds to. This includes the case that there is no mask visible at all.
\item A low score (2-4): The object is mostly identifiable, but the mask quality is extremely poor (\eg large regions of the mask cover other objects; large regions of the object missing; extremely splotchy mask boundaries that cut through the middle of the object).
\item A mid score (5-6): The object is identifiable and the boundary is mostly correct, but there are major errors (missing a significant disconnected part of the object; containing a significant part of another object; very poor boundary quality in one area of the object but not the entire object).
\item A high score (7-9): The object is identifiable and errors are small and rare (missing a small, heavily obscured disconnected component, having small regions where the mask boundary does not quite match the object boundary).
\item A score of 10: The mask is pixel-perfect; it has no identifiable errors at all.
\end{itemize}}\\
\incannpanel{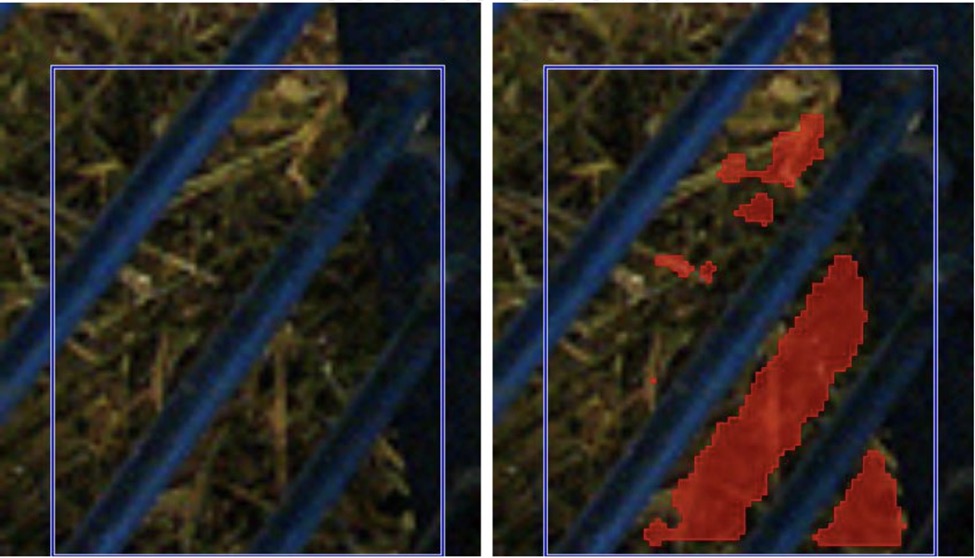}{Example of a mask with a score of 1: It is not clear what object this mask corresponds to.}
\incannpanel{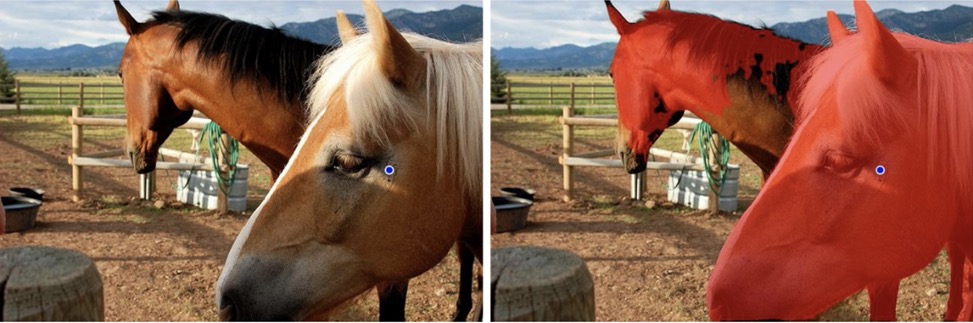}{Example of a mask with a low score (2-4): The main object is identifiable, but the mask includes a large, incorrect portion of another object.}
\incannpanel{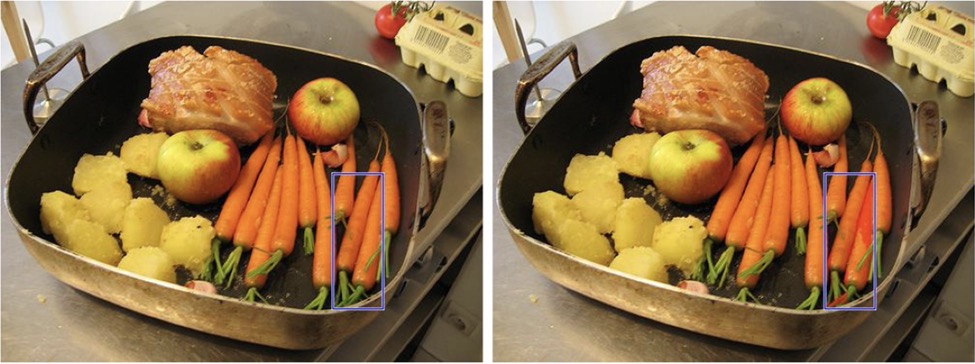}{Example of a mask with a low score (2-4): The main object is identifiable, but a large, random part of the object is missing.}
\incannpanel{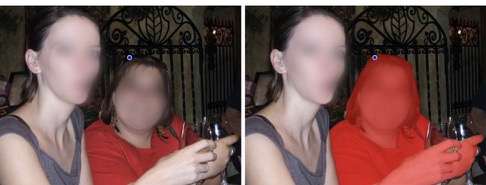}{Example of a mask with a low-to-medium score (4-5): The object is identifiable and the edges are all correct, but the mask incorrectly includes the hand of the person on the left.}\\
\incannpanel{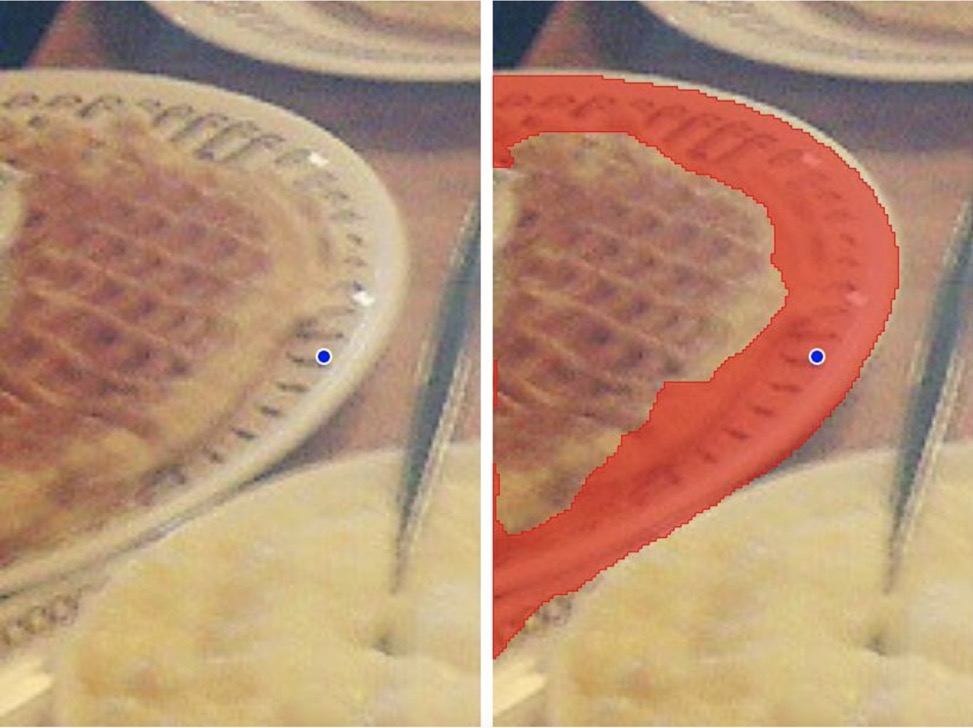}{Example of a mask with a medium score (5-6): The mask clearly corresponds to the plate, but the boundary with the waffle is quite poor.}
\incannpanel{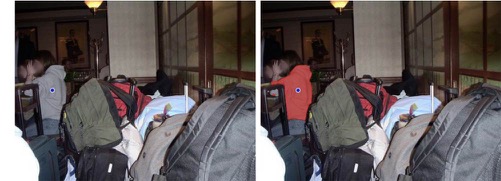}{Example of a mask with a medium score (5-6): the object is easy to identify, and most of the edges make sense. However, there is a significant disconnected part (their arm inside the frame) that is mostly missing, as well as splotchy pixels in this region.}
\incannpanel{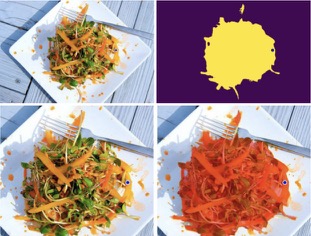}{Example of a mask with a medium-to-high score (6-8): the mask has two small-ish regions of poor boundary, at the top of the mask and on the bottom right.}
\incannpanel{im16}{Example of a mask with a medium-to-high score (6-8): The wreath is a valid object that is the size of the box (the entire wreath + clock would also be a valid object). However, there are incorrect stray mask pixels on the clock.}\\
\incannpanel{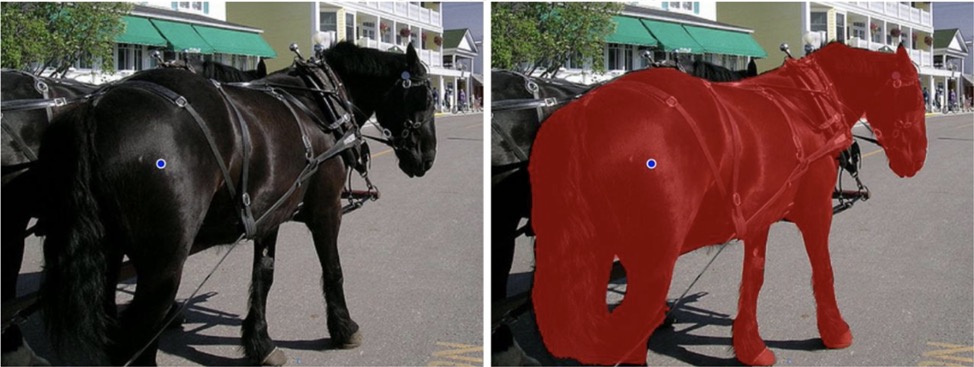}{Example of a mask with a high score (7-9): The boundary of the horse is almost entirely correct, except for the right side of its back leg. The mask consistently includes all of the equipment that horse is wearing, and has logical boundaries.}
\incannpanel{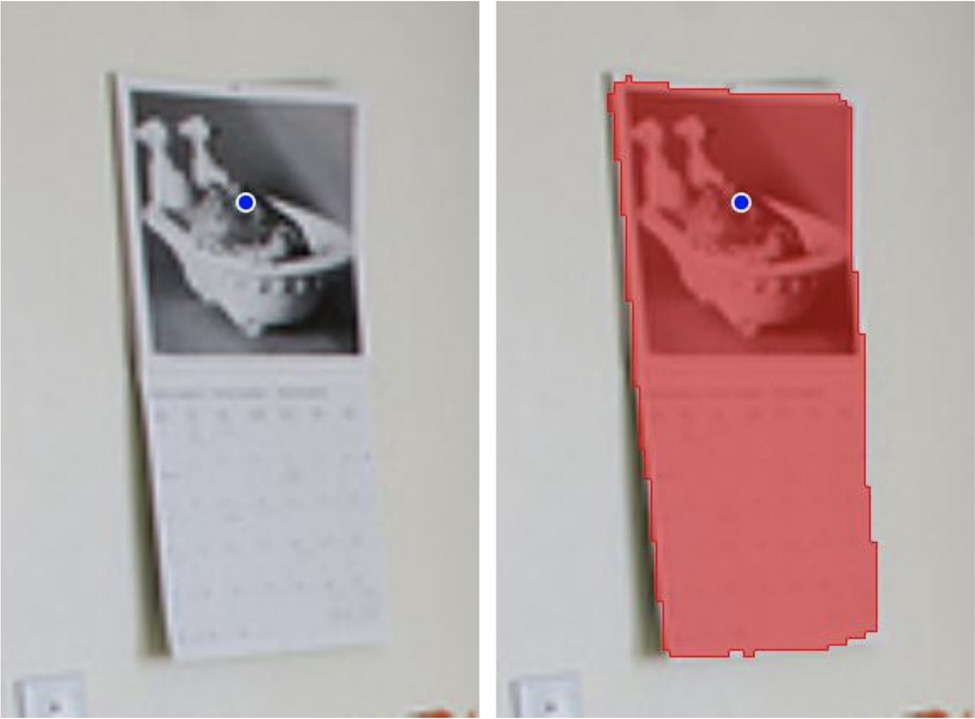}{Example of a mask with a very high score ($\sim$9): There are only minor errors around the edge of the mask. The blocky ‘pixelation’ is not an error, since the image is also blocky at this scale.}
\incannpanel{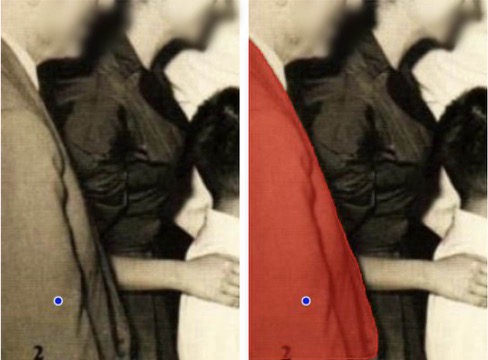}{Example of a mask with a very high score (9-10): the mask has only very minor errors in the edge on the bottom right.}
\incannpanel{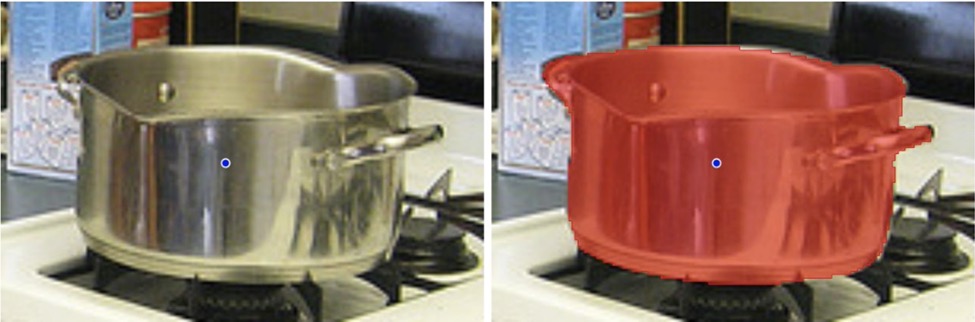}{Example of a mask with a very high score (9-10): There are only minor errors around the edge of the mask.}\\
\vspace{-5mm}\caption{Here we provide the complete guidelines given to annotations for the human review of mask quality. Some images been edited slightly and faces have been blurred to enable release. Best viewed with zoom (part 2 of 2).}
\label{fig:guidelines_b}
\end{figure*}

We provide the complete guidelines given to annotations for the human review of mask quality in \fig{fig:guidelines_a} and \fig{fig:guidelines_b}.

%##################################################################################################
%##################################################################################################

\end{document}